\documentclass[acmsmall]{acmart}
\renewcommand\footnotetextcopyrightpermission[1]{} 
\usepackage{multirow}
\usepackage{subfigure}
\usepackage{changepage}
\usepackage{enumitem}
\usepackage{fancyhdr}

\setcopyright{acmcopyright}
\copyrightyear{2021}
\acmYear{2021}

\AtBeginDocument{%
  \providecommand\BibTeX{{%
    \normalfont B\kern-0.5em{\scshape i\kern-0.25em b}\kern-0.8em\TeX}}}

\begin{document}


\title{The Elements of End-to-end Deep Face Recognition: A Survey of Recent Advances}

\author{Hang Du}
\authornote{Equal contribution. This work was performed at JD AI Research.}
\email{duhang@shu.edu.cn}
\affiliation{%
  \institution{Shanghai University}
  \city{Shanghai}
  \country{China}
 }
\author{Hailin Shi}
\authornotemark[1]
\email{shihialin@jd.com}
\affiliation{%
  \institution{JD AI Research}
  \city{Beijing}
  \country{China}
}

\author{Dan Zeng}
\authornote{Corresponding author.}
\affiliation{%
  \institution{Shanghai University}
  \city{Shanghai}
  \country{China}
  }
\email{dzeng@shu.edu.cn}

\author{Xiao-Ping Zhang}
\email{xzhang@ryerson.ca}
\affiliation{%
 \institution{Ryerson University}
  \city{Toronto}
  \country{Canada}
 }

\author{Tao Mei}
\email{tmei@jd.com}
\affiliation{%
 \institution{JD AI Research}
  \city{Beijing}
  \country{China}
 }

\renewcommand{\shortauthors}{Du et al.}

\begin{abstract}
Face recognition is one of the most popular and long-standing topics in computer vision. With the recent development of deep learning techniques and large-scale datasets, deep face recognition has made remarkable progress and been widely used in many real-world applications. 
Given a natural image or video frame as input, an end-to-end deep face recognition system outputs the face feature for recognition. 
To achieve this, a typical end-to-end system is built with three key elements: face detection, face alignment, and face representation. The face detection locates faces in the image or frame. Then, the face alignment is proceeded to calibrate the faces to the canonical view and crop them with a normalized pixel size. Finally, in the stage of face representation, the discriminative features are extracted from the aligned face for recognition. Nowadays, all of the three elements are fulfilled by the technique of deep convolutional neural network.
In this survey article, we present a comprehensive review about the recent advance of each element of the end-to-end deep face recognition, since the thriving deep learning techniques have greatly improved the capability of them. 
To start with, we present an overview of the end-to-end deep face recognition. 
Then, we review the advance of each element, respectively, covering many aspects such as the to-date algorithm designs, evaluation metrics, datasets, performance comparison, existing challenges, and promising directions for future research. 
Also, we provide a detailed discussion about the effect of each element on its subsequent elements and the holistic system. 
Through this survey, we wish to bring contributions in two aspects: first, readers can conveniently identify the methods which are quite strong-baseline style in the subcategory for further exploration; second, one can also employ suitable methods for establishing a state-of-the-art end-to-end face recognition system from scratch.


\end{abstract}




\maketitle

\section{Introduction}



Face recognition (FR) is an extensively studied topic in computer vision. Among the existing technologies of human biometrics, face recognition is the most widely used one in real-world applications.
With the great advance of deep convolutional neural networks (DCNNs), the deep learning based methods have achieved significant improvements on various computer vision tasks, including face recognition. In this survey, we focus on 2D image based end-to-end deep face recognition which takes the general images or video frames as input, and extracts the deep feature of each face as output. We provide a comprehensive review of the recent advances of the elements of end-to-end deep face recognition. Specifically, an end-to-end deep face recognition system is composed of three key elements: face detection, face alignment, and face representation. In the following, we give a brief introduction of each element.

\begin{figure}[t]
\subfigure[]{\label{publications} \includegraphics[height=2.6cm]{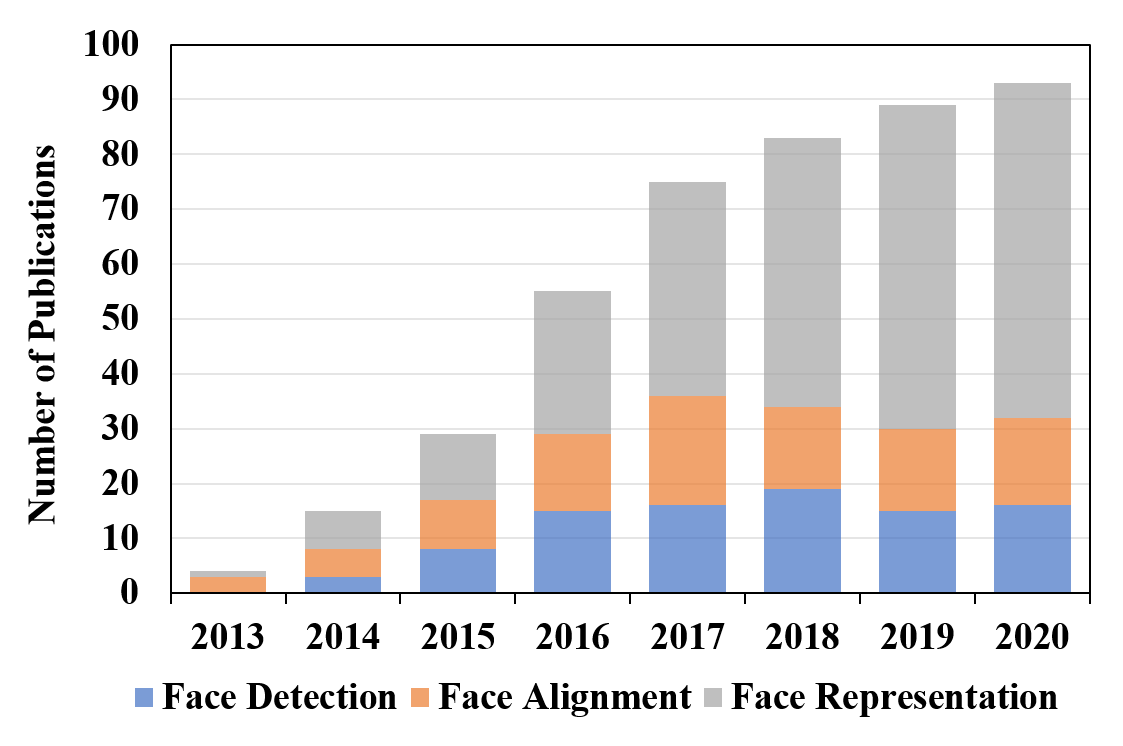}}
\subfigure[]
{\label{Pipeline}
\includegraphics[height=2.6cm]{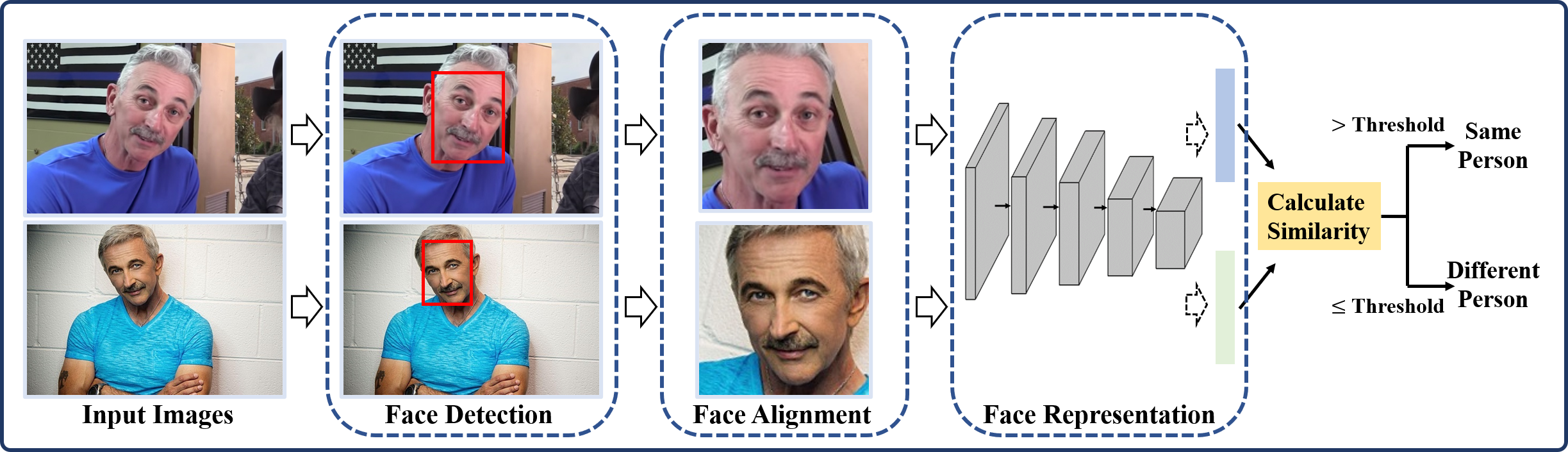}}

\caption{(a) The publication trend of the elements of end-to-end deep face recognition from 2013 to 2020. (b) The standard pipeline of end-to-end deep face recognition system. First, the face detection stage aims to localize the face region on the input image. Then, the face alignment is proceeded to normalize the detected face to the  canonical view. Finally, the face representation devotes to extracting features for recognition.}

\end{figure}

Face detection is the first step of end-to-end face recognition. It aims to locate the face regions in the still images or video frames. 
Before the deep learning era, one of the pioneering works for face detection is Viola-Jones~\cite{viola2001rapid} face detector, which utilizes AdaBoost classifiers with Haar features to build a cascaded structure. 
Later on, the subsequent approaches explore the effective hand-craft features~\cite{Ojala2002Multiresolution,Mita2005Joint,Yang2014acf} and various classifiers~\cite{Li2002Statistical,Pham2007Fast,Brubaker2008On} to improve the detection performance. 
One can refer to~\cite{Yang2002DetectingFI,Stefanos2015A} for a thorough survey of traditional face detection methods. 

Next, face alignment refers to calibrate the detected face to the canonical view and crop it to a normalized pixel size, in order to facilitate the subsequent task of face representation computing. It is an essential intermediate procedure for face recognition system. 
Generally, the facial landmark localization is necessary for face alignment, while some approaches can directly generate aligned face from the input one.
Most traditional works of facial landmark localization focus on either generative methods~\cite{Cootes1992ActiveSM,Cootes2000ViewbasedA} or discriminative methods~\cite{Zhou2007ShapeRM,Martnez2013LocalEA}, and there are several exhaustive surveys about them~\cite{eliktutan2013ACS,Jin2017FaceAI,Wang2018FacialFP}. 

In the face representation stage, the discriminative features are extracted from the aligned face images for recognition.
This is the final and core step of face recognition. 
In early studies, many approaches calculate the face representation by projecting face images into low-dimensional subspace, such as Eigenfaces~\cite{1991Eigenfaces} and Fisherfaces~\cite{Belhumeur1997Eigenfaces}. 
Later on, handcrafted local descriptors based methods~\cite{Liu2002GaborFB,Ahonen2004Face} prevail in this area. For a detailed review of these traditional methods, one can refer to ~\cite{W2003Face,2006fr,2009fr}. 
In the last few years, the face representation benefits from the development of DCNNs and witnesses great improvements for high performance face recognition.

This survey focuses on reviewing and analyzing the recent advances in each element. An important fact is that, the performance of face recognition depends on the contribution of all the elements (\textit{i.e.,} face detection, alignment and representation). In other words, inferiority in any one of the elements will become the bottleneck and harm the final performance. In order to establish high-performance end-to-end face recognition system, it is necessary to understand every element of the holistic framework and their intrinsic connection. A number of face recognition surveys have been published in the past twenty years. The main differences between our survey and the existing ones are summarized as follows.

\begin{itemize}
\item\textbf{The relationship between the elements and whole.}
We provide the thorough discussion about the effect of each element on its subsequent one and the holistic system, which are overlooked in the existing surveys. From the existing experiments and detailed analysis, we can conclude the performance of the holistic system depends on the three elements. Therefore, it is necessary to review them together for helping the readers who aim to establish state-of-the-art face recognition system from scratch.

\item\textbf{More recently published works.}
The publications in the last three years (2018-2020) are much more than all those published before 2018 (as illustrated in Fig.~\ref{publications}). 
In view of the rapid development of face detection, face alignment and face representation in the past few years, this survey covers the recently published articles.
By doing so, we provide the up-to-date review of the elements, and large number of newly presented methods.

\item\textbf{New analysis for future work.}
Based on the up-to-date review, we conclude the promising trends from the newest frontier, and several insightful thoughts of each element as well as the holistic system, to enlighten the future research. 
\end{itemize}

Specifically, there are certain surveys~\cite{W2003Face,2006fr,2009fr} about face recognition who, however, do not cover deep learning based methods since they are published early before the deep learning era; besides, another set of surveys focus on 3D face recognition~\cite{bowyer2006survey,Soltanpour2017ASO} and specific tasks~\cite{Zou2007IlluminationIF,Ding2016ACS}. 
Instead, we focus on the 2D face recognition which is the most needed in practical applications.
For deep learning based 2D face recognition, there are a small number of articles that fulfil relevant survey, which differ from this paper in many ways.
Among them, Ranjan~\textit{et al.}~\cite{Ranjan2018deep} do not include the recent techniques that rapidly evolved in the past few years. 
In fact, the number of published works has been increasing dramatically during these years (as shown in Fig.~\ref{publications}). 
Wang and Deng~\cite{wang2018deep} present a systematic review about deep face representation rather than the end-to-end face recognition. 
More recently, Insaf~\textit{et al.}~\cite{electronics9081188} provide a review of 2D and 3D face recognition from the traditional to deep-learning era, while the scope is still limited in the face representation.
In summary, the end-to-end face recognition, covering all the elements of the pipeline, needs to be systematically reviewed, while seldom of the existing survey articles attach importance to this task.

Therefore, we systematically review the deep learning based approaches of each element in the end-to-end face recognition, respectively. The review of each element covers many aspects: algorithm designs, evaluation metrics, datasets, performance comparisons, remaining challenges, and promising directions for future research. 
We hope this survey could bring helpful thoughts for better understanding of the big picture of end-to-end face recognition and deeper exploration in a systematic way. 
Specifically, the main contributions can be summarized as follows:

\begin{itemize}
\item {We provide a comprehensive survey of the elements of end-to-end deep face recognition. We review the recent advances of each element, respectively, and present elaborated categorizations of them to make the readers understand them in a systematic way.}

\item {We review the three elements from many aspects: algorithm designs, evaluation metrics, datasets, and performance comparison.
Moreover, we point out the effect of each element on its subsequent elements and the holistic system. }

\item {We collect the existing challenges and promising directions for each element and its subcategories to facilitate the future research, and further discuss the major challenges and future trends from the view of the holistic framework. }

\end{itemize}

\begin{figure*}[t]
\centering
\includegraphics[height=10cm]{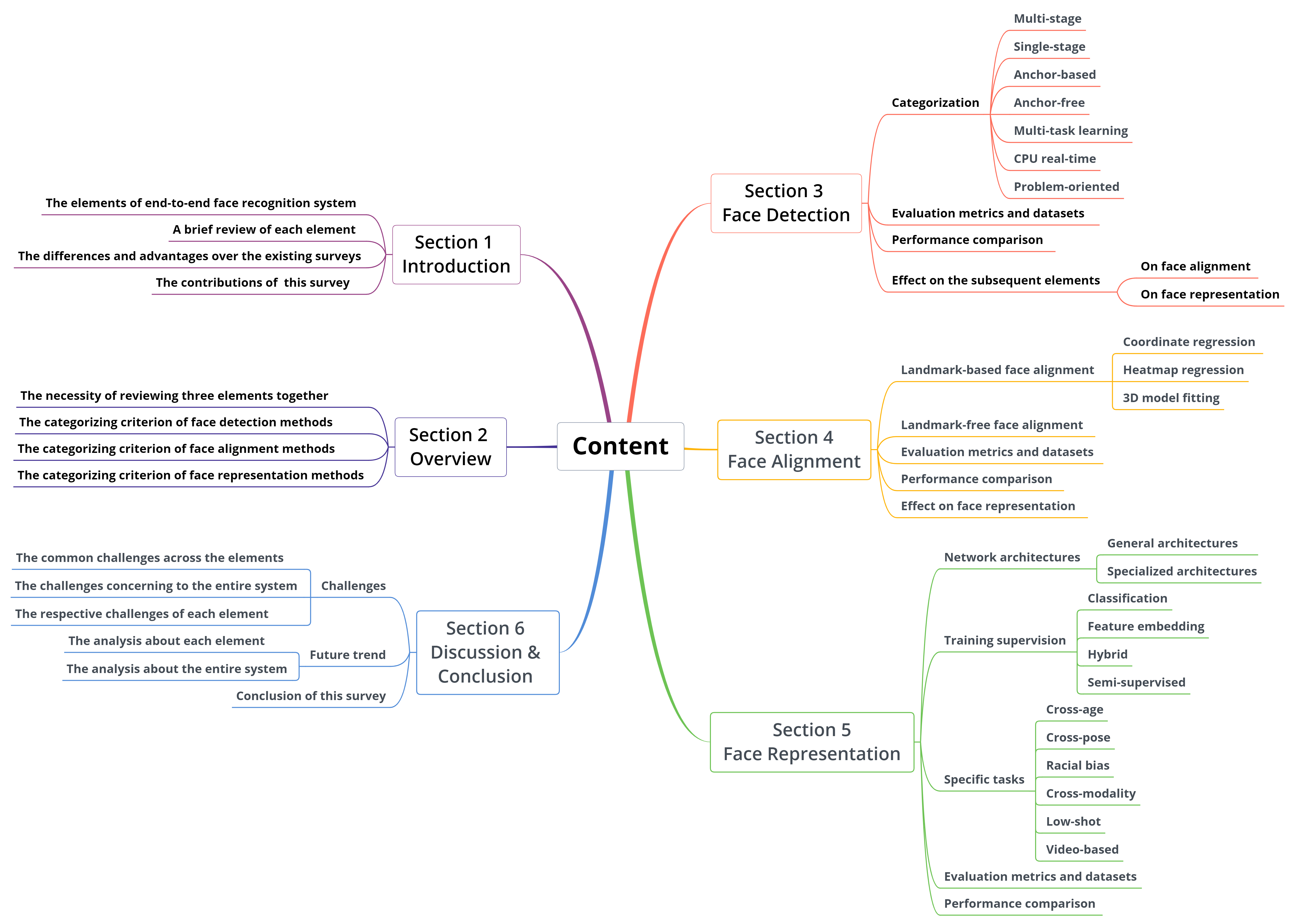}

\caption{
The structure of this survey. The left parts (Section 1, 2, 6) refer to the functional contents that provide overall introduction and discussion. The right parts (Section 3, 4, 5) refer to the technical contents that provide the detailed reviewing of three elements. 
}

\label{Content}
\end{figure*}

\section{OVERVIEW}
A typical end-to-end deep face recognition system includes three basic elements: face detection, face alignment, and face representation, as shown in Fig.~\ref{Pipeline}. 
First, face detection localizes the face region on the input image. Then, face alignment is proceeded to normalize the detected face into the canonical layout. 
Finally, face representation devotes to extracting  discriminative features from the aligned face. The features are used to calculate the similarity between them, in order to make the decision that whether the faces belong to the same identity. 

The structure of this survey is illustrated in Fig.~\ref{Content}. We structure the body sections  (Section~\ref{sec:face_detection},~\ref{sec:face_alignment},~\ref{sec:face_representation}) with respect to the three elements, each of which is a research topic that covers abundant literatures in computer vision. We give an overview of the three elements briefly in this section, and dive into each of them in the following body sections.  

\subsection{Face Detection}
\label{sec:overview:face_detection}
Face detection is the first procedure of the face recognition system. Given an input image, the face detection aims to find all the faces in the image and give the coordinates of bounding box with a confidence score. The major challenges of face detection contain varying resolution, scale, pose, illumination, occlusion, \textit{etc}. 
In Section~\ref{sec:face_detection}, we provide a categorization of the deep learning based face detection methods from multiple dimensions, which includes multi-stage, single-stage, anchor-based, anchor-free, multi-task learning, CPU real-time and problem-oriented methods. 
It is worth noting that there exist overlapping techniques between the categories, because, the categorization is built up from multiple perspectives. 

\textbf{Differences to the existing survey of face detection.}  Minaee~\textit{et al.}~\cite{Minaee2021GoingDI} review face detection methods from the beginning of deep learning era, and categorize them by design of network architecture. 
Compared with them, our categorizing criterion covers poly-aspects. 
Specifically, we provide a multiple-dimension categorization, to discuss the face detection methods from many different perspectives, which will help us to better understand the developing line and conclude the future trend.
Since face detection state of the art is relatively advanced, such comprehensive categorization is necessary for readers.

\subsection{Face Alignment}
\label{sec:overview:face_alignment}
\begin{figure}
     \centering
        \includegraphics[height=1.6cm]{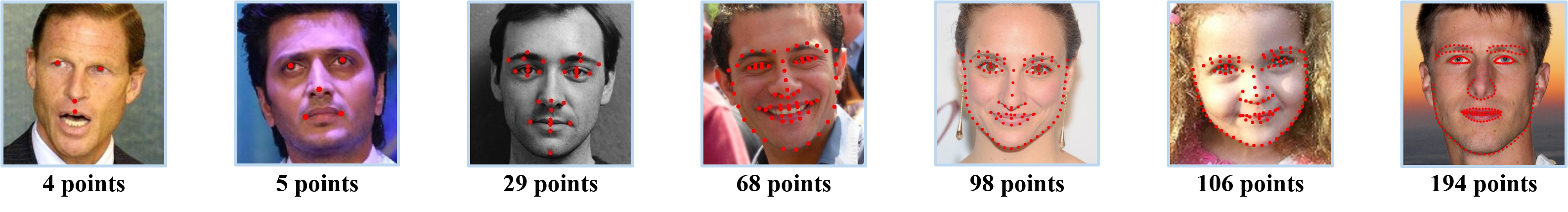}
        
        \caption{Visualization of facial landmarks of different versions. The 4-point and 5-point landmarks are often used for face alignment.}
        \label{fig:lmk_sample}
        
\end{figure}

In the second stage, face alignment aims to calibrate the detected face to the canonical view. 
Since human face appears with the regular structure, in which the facial parts (eyes, nose, mouth, \textit{etc}) have constant arrangement, the alignment of face is of great benefit to the subsequent feature computation for face recognition.
For most existing methods of face alignment, the facial landmarks, or so-called facial keypoints (as shown in Fig.~\ref{fig:lmk_sample}), are indispensable, because they are involved as the reference for similarity transformation or affine transformation. 
So, the facial landmark localization is a prerequisite for face alignment.
The DCNNs based facial landmark localization methods can be divided into three subcategories: coordinate regression, heatmap regression and 3D model fitting based approaches. 
Without relying on the facial landmarks, several approaches can directly output aligned face from the input by learning the transformation parameters. 
We will review these methods in Section~\ref{sec:face_alignment}.

\textbf{Differences to the existing survey of face alignment.} Previous surveys of face alignment~\cite{eliktutan2013ACS,Jin2017FaceAI,Wang2018FacialFP} only focus on reviewing the facial landmark localization methods. Since the landmark-free face alignment is also a kind of methods to generate aligned images for face recognition, we further collect them in this survey. 

\subsection{Face Representation}
\label{sec:overview:face_representation}
As the key step of face recognition system, face representation devotes to learning deep face model and using it to extract features from aligned faces for recognition. The features are used to calculate the similarity of the matched faces. 
In Section~\ref{sec:face_representation}, we provide a review of deep learning based methods for discriminative face features, and retrospect these methods with respect to the network architecture and the training supervision. 
For network architecture, we introduce the general architectures which are designed for a wide range of computer vision tasks, and the special architectures which are specialized for face representation. 
As for training supervision, we mainly introduce four schemes, including the classification, feature embedding, hybrid and semi-supervised schemes. 
Additionally, we present several specific face recognition scenes, including cross domain, low-shot learning and video based scenarios. 

\textbf{Differences to the existing survey of face representation.}
This survey aims to provide the readers with a better understanding of the end-to-end face recognition. 
Recently, Wang and Deng~\cite{wang2018deep} present a systematic review about deep face recognition, in which they mainly focus on deep face representation, and the categorization of training loss is sub-optimal. 
For instance, they sort the supervised learning of deep face representation by Euclidean-distance based loss, angular/cosine-margin-based loss, softmax loss and its variations; while, in fact, almost all the angular/cosine-margin-based losses are implemented as the variation of softmax loss rather than an individual set. In contrast, we suggest a more reasonable categorization with three subcategories, \textit{i.e.,} classification, feature embedding, and hybrid methods (in Section
~\ref{sec:face_representation:supervision}).

\section{Face Detection}
\label{sec:face_detection}

Face detection is the first step of end-to-end face recognition system, which aims to locate the face regions from the input images. 
In this section, first, we categorize and make comparison of the existing deep learning methods for face detection. Next, we introduce several popular datasets of face detection and the common metrics for evaluation. Finally, we provide a performance comparison of state-of-the-art face detection methods and detailed discussion about the effect of face detection on its subsequent elements.

\subsection{Categorization of Face Detection}
\label{sec:face_detection:categorization}

\begin{figure}[t]
\centering\includegraphics[height=2.5cm]{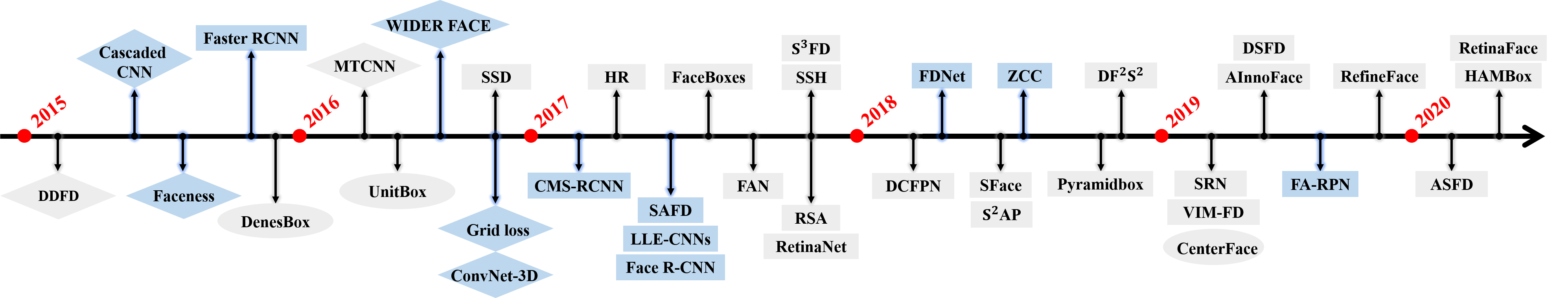}

\caption{The development of representative face detection methods. The blue and gray represent multi-stage and single-stage methods; according to the anchor usage, the rectangle, oval, and diamond denote anchor-based, anchor-free and other methods. One can refer to Table~\ref{fd_class} for the references of these methods.}

\label{Development_fd}
\end{figure}

In order to present the deep face detection methods with a clear categorization, we group them with seven sets, \textit{i.e.,} multi-stage, single-stage, anchor-based, anchor-free, multi-task learning, CPU real-time, and problem-oriented methods (in Table~\ref{fd_class}). These sets are not necessarily exclusive, because we establish the categorization from multiple perspective.
Fig.~\ref{Development_fd} is the development of representative methods for face detection.

\subsubsection{Multi-stage methods} Following the coarse-to-fine manner or the proposal-to-refine strategy, multi-stage based detectors first generate a number of candidate boxes, and then refine the candidates by one or more additional stages. 
The first stage employs sliding window to propose the candidate bounding boxes at a given scale, and the latter stages reject the false positives and refine the remaining boxes. 
In such regime, the cascaded architecture~\cite{Cascade_CNN,mtcnn,pcn,ZENG2019PPN} is naturally an effective solution for the coarse-to-fine face detection.

Face detection can be considered as a specific objective of general object detection. Thus, many works~\cite{HyperFace,Zhu2017CMS-RCNN,Jiang2017650,wang2017facercnn,zhang2018face,SUN201842,Ge_2017_CVPR,2016stn,Li2016FaceDW,Najibi_2019_CVPR} inherit the remarkable achievements from the general object detectors. 
For instance, Faster R-CNN~\cite{ren2015faster} is a classic and effective detection framework which employs a region proposal network  to generate region proposals with a set of dense anchor boxes in the first stage, and then refines the proposals in the second stage. 
Based on the proposal-to-refine scheme, many works have dedicated to improve the modeling of the refinement stage~\cite{Zhu2017CMS-RCNN,Jiang2017650,wang2017facercnn,zhang2018face,SUN201842} and the proposal stage~\cite{2016stn,Li2016FaceDW,Najibi_2019_CVPR,hao2017scale,Song_2018_CVPR}, and achieved great progress for accurate face detection.
Apart from the modeling, how to train the multi-stage detector is another interesting topic. 
To tackle the issue of inferior optimization for multi-stage detectors, a joint training strategy~\cite{Qin2016Joint} is designed for both Cascaded CNN~\cite{Cascade_CNN} and Faster R-CNN to achieve end-to-end optimization and better performance.

\subsubsection{Single-stage methods} The single-stage methods accomplish the candidate classification and bounding box regression from the feature maps directly, without the dependence on proposal stage. 

A classic structure of single stage comes from a general object detector named Single Shot multibox Detector (SSD)~\cite{Liu_2016_ssd}. It runs much faster than the multi-stage ones while maintaining comparable accuracy.
Based on SSD, many studies~\cite{Zhang2017S3FD,Zhang2017Faceboxes,Zhang2018DCFPN,2019fbi,tang2018pyramidbox} develop deep face detectors those are robust to different scales of face.
As for the backbone architecture, many face detectors resort to the feature pyramid network (FPN)~\cite{lin2016feature} which consists of a top-down architecture with skip connections and merges the high-level and low-level features for detection. The high-level feature maps provide rich semantic information, while the low-level layers supplement more local information. The feature fusion preserves the advantages from both sides, and brings great progress in detecting objects with a wide range of scales. 
Therefore, many single-stage face detectors~\cite{2017SSH,wang2017fan,zhang2017fanet,tang2018pyramidbox,tian2018df2s2,li2019pyramidbox,deng2020retinaface,chi2019selective,2019DSFD,zhang2019improved} are developed with the advantage of FPN. Not only handling the scale issue in face detection via FPN, but also these methods attempt to solve the inherent shortcomings of original FPN such like the conflict of receptive field.

\begin{table*}[t]
\centering
\begin{center}
\caption{The categorization of deep face detection methods.}
\label{fd_class}
\resizebox{1\linewidth}{!}{
\begin{tabular}{|p{2cm}|p{8.5cm}|p{12cm}|}
\hline
\multicolumn{1}{|c|}{Category}&\multicolumn{1}{c|}{Description}&\multicolumn{1}{c|}{Method}\\
\hline
    {Multi-stage} & 
    Detectors first generate candidate boxes, then the following one or more stages refine the candidates.
    & Faceness~\cite{2015Faceness}, HyperFace~\cite{HyperFace}, STN~\cite{2016stn}, ConvNet-3D~\cite{Li2016FaceDW}, 
    SAFD~\cite{hao2017scale},  CMS-RCNN~\cite{Zhu2017CMS-RCNN}, Wan~\textit{et al.}~\cite{wan2016bootstrapping}, 
    Jiang~\textit{et al.}~\cite{Jiang2017650}, DeepIR~\cite{SUN201842}, Grid loss~\cite{2016Grid_Loss}, Face R-CNN~\cite{wang2017facercnn}, Face R-FCN~\cite{wang2017facerfcn}, ZCC~\cite{Zhu2018}, FDNet~\cite{zhang2018face}, FA-RPN~\cite{Najibi_2019_CVPR}, Cascaded CNN~\cite{Cascade_CNN}, MTCNN~\cite{mtcnn}, Qin~\textit{et al.}~\cite{Qin2016Joint}, LLE-CNNs~\cite{Ge_2017_CVPR}, PCN~\cite{pcn}, PPN~\cite{ZENG2019PPN}
    \\
    \hline
    Single-stage & 
    Detectors accomplish face classification and bounding box regression from feature maps at once. 
    & DDFD~\cite{farfade2015multiview},     DenseBox~\cite{huang2015densebox}, UnitBox~\cite{UnitBox},
    HR~\cite{2017HR}, Faceboxes~\cite{Zhang2017Faceboxes}, SSH~\cite{2017SSH}, S$^3$FD~\cite{Zhang2017S3FD}, DCFPN~\cite{Zhang2018DCFPN}, FAN~\cite{wang2017fan},
    FANet~\cite{zhang2017fanet}, RSA~\cite{Liu2016rsa},  S$^2$AP~\cite{Song_2018_CVPR},  PyramidBox~\cite{tang2018pyramidbox},  DF$^2$S$^2$~\cite{tian2018df2s2}, SFace~\cite{wang2018sface}, DSFD~\cite{2019DSFD}, RefineFace~\cite{zhang2019refineface}, SRN~\cite{chi2019selective},  PyramidBox++~\cite{li2019pyramidbox},
    CenterFace~\cite{xu2019centerface},    VIM-FD~\cite{zhang2019robust}, ISRN~\cite{zhang2019improved}, AInnoFace~\cite{zhang2019accurate}, ASFD~\cite{Zhang2020ASFDAA},   RetinaFace~\cite{deng2020retinaface},    HAMBox~\cite{Liu_2020_HAMBox}
    \\ 
    \hline
    Anchor-based& Detectors deploy a number of dense anchors on the feature maps, and then proceed the classification and regression on these anchors. & Wan~\textit{et al.}~\cite{wan2016bootstrapping}, Face Faster RCNN~\cite{Jiang2017650}, RSA~\cite{Liu2016rsa}, Face R-CNN~\cite{wang2017facercnn}, FDNet~\cite{zhang2018face}, DeepIR~\cite{SUN201842}, SAFD~\cite{hao2017scale}, 
    SSH~\cite{2017SSH}, S$^3$FD~\cite{Zhang2017S3FD}, DCFPN~\cite{Zhang2018DCFPN},  Faceboxes~\cite{Zhang2017Faceboxes},  FAN~\cite{wang2017fan}, 
    FANet~\cite{zhang2017fanet}, PyramidBox~\cite{tang2018pyramidbox}, ZCC~\cite{Zhu2018}, S$^2$AP~\cite{Song_2018_CVPR}, DF$^2$S$^2$~\cite{tian2018df2s2}, SFace~\cite{wang2018sface}, 
    RetinaFace~\cite{deng2020retinaface}, DSFD~\cite{2019DSFD}, RefineFace~\cite{zhang2019refineface}, SRN~\cite{chi2019selective}, 
    VIM-FD~\cite{zhang2019robust}, PyramidBox++~\cite{li2019pyramidbox}, FA-RPN~\cite{Najibi_2019_CVPR}, ISRN~\cite{zhang2019improved}, AInnoFace~\cite{zhang2019accurate}, Group Sampling~\cite{Ming_2019_Group_Sampling}, HAMBox~\cite{Liu_2020_HAMBox}, ASFD~\cite{Zhang2020ASFDAA},
    \\
    \hline
    Anchor-free & Detectors directly find faces without preset anchors. &DenseBox~\cite{huang2015densebox}, UnitBox~\cite{UnitBox}, CenterFace~\cite{xu2019centerface}
    \\
    \hline
    Multi-task learning  & Detectors jointly learn the classification and bounding box regression with additional  tasks (\textit{e.g.,} landmark localization) in one framework.
    & STN~\cite{2016stn}, ConvNet-3D~\cite{Li2016FaceDW},  HyperFace~\cite{HyperFace}, MTCNN~\cite{mtcnn}, Face R-CNN~\cite{wang2017facercnn}, RetinaFace~\cite{deng2020retinaface}, DF$^2$S$^2$~\cite{tian2018df2s2}, FLDet~\cite{2019fldet}, PyramidBox++~\cite{li2019pyramidbox}, CenterFace~\cite{xu2019centerface}
    \\
    
    \hline
    CPU real-time &
    Detectors can run on a single CPU core in real-time for VGA-resolution images.
    & Cascade CNN~\cite{Cascade_CNN}, STN~\cite{2016stn},  MTCNN~\cite{mtcnn}, DCFPN~\cite{Zhang2018DCFPN}, Faceboxes~\cite{Zhang2017Faceboxes}, PCN~\cite{pcn},  RetinaFace~\cite{deng2020retinaface}, FLDet~\cite{2019fldet}, FBI~\cite{2019fbi}, PPN~\cite{ZENG2019PPN}, CenterFace~\cite{xu2019centerface}
    \\

    \hline
    Problem-oriented & 
    Detectors aim to solve specific challenges in face detection, such as tiny faces, occluded faces, rotated and blurry faces.
    & HR~\cite{2017HR}, SSH~\cite{2017SSH}, S$^3$FD~\cite{Zhang2017S3FD}, Bai~\textit{et al.}~\cite{bai2018finding}, PyramidBox~\cite{tang2018pyramidbox}, Grid loss~\cite{2016Grid_Loss}, FAN~\cite{wang2017fan}, LLE-CNNs~\cite{Ge_2017_CVPR}, PCN~\cite{pcn}, Group Sampling~\cite{Ming_2019_Group_Sampling} 
    \\
\hline
\end{tabular}}
\end{center}
\end{table*}

Although the single-stage methods have the advantage of high efficiency, their detection accuracy is below that of the two-stage methods. It is partially because the imbalance problem of positives and negatives brought by the dense anchors, whereas the proposal-to-refine scheme is able to alleviate this issue.   
Accordingly, RefineDet~\cite{Zhang_2018_refinedet} sets up an anchor refinement module in its network to remove large number of negatives. 
Inspired by RefineDet, SRN~\cite{chi2019selective} presents a selective two-step classification and regression method; the two-step classification is performed at the low-level layers to reduce the search space of classifier, and the two-step regression is performed at high-level layers to obtain accurate location. Later on, many works~\cite{zhang2019robust,zhang2019improved,zhang2019accurate,zhang2019refineface} improve SRN with several effective techniques, such as training data augmentation, improved feature extractor and training supervision, anchor assignment and matching strategy, multi-scale test strategy, \textit{etc}. 

Most aforementioned methods need to preset anchors for face detection, while some representative detectors of single-stage, such as DenseBox~\cite{huang2015densebox}, UnitBox~\cite{UnitBox} and CenterFace~\cite{xu2019centerface}, fulfil the detection without preset anchors. We will present them as anchor-free type in the next subsection.

\subsubsection{Anchor-based and anchor-free methods} As shown in Table~\ref{fd_class}, most current face detectors are anchor-based due to the long-time development and superior performance.
Generally, we preset the anchors on the feature maps, then fulfil the classification and bounding box regression on these anchors one or more times, and finally output the accepted ones as the detection results.
Therefore, the anchor allocation and matching strategy are  crucial to the detection accuracy. 
Most anchor-based methods focus on the algorithms along this direction, such as scale compensation~\cite{Zhang2017S3FD,Liu_2020_HAMBox}, max-out background label~\cite{Zhang2017S3FD}, expected max overlapping score~\cite{Zhu2018}, group sampling by scale~\cite{Ming_2019_Group_Sampling}, \textit{etc}.
However, the settings (\textit{e.g.,} scale, stride, ratio, number) of anchors need to be carefully tuned for each particular dataset, limiting their generalization ability. Besides, the dense anchors increase the computational cost and bring the imbalance problem of positive and negative anchors.

Anchor-free methods~\cite{Law2018CornerNet,zhu2019fs,Tian2019FCOS} attract growing attention in general object detection. As for face detection, certain pioneering works have emerged in recent years. 
DenseBox~\cite{huang2015densebox} and UnitBox~\cite{UnitBox} attempt to predict the pixel-wise bounding box on face. CenterFace~\cite{xu2019centerface} regards face detection as a generalized task of keypoint estimation, which predicts the facial center point and the size of bounding box in feature map. In brief, the anchor-free detectors get rid of the preset anchors and achieve better generalization capacity. 
Regarding to the detection accuracy, it needs further exploration for better robustness to false positives and stability in training process.

\subsubsection{Multi-task learning methods} 
Generally, the multi-task learning methods are designed for solving a problem together with other related tasks by sharing the visual representation. 
Here, we introduce the multi-task learning methods that train the face detector with the associated facial tasks or auxiliary supervision branches to enrich the feature representation and detection robustness. Many approaches~\cite{zhang2014mt,huang2015densebox,2016stn,mtcnn,Li2016FaceDW,2019fldet,xu2019centerface} have explored the joint learning of face detection and facial landmark localization. 
Among them, MTCNN~\cite{mtcnn} is the most representative one, which exploits the inherent correlation between facial bounding boxes and landmarks. 
Subsequently, HyperFace~\cite{HyperFace} fuses the low-level features and high-level features to simultaneously conduct four tasks, including face detection, facial landmark localization, gender classification and pose estimation. Based on RetinaNet~\cite{lin2017focal}, RetinaFace~\cite{deng2020retinaface} integrates face detection, facial landmark localization and dense 3D face regression in one framework. 
From the multi-task routine, we can see that the face detectors can benefit from the associated facial tasks. Moreover, certain methods~\cite{wang2017facercnn,tian2018df2s2,wang2018sface,li2019pyramidbox} exploit auxiliary supervision branches, such as segmentation branch, anchor-free branch, \textit{etc}. These branches are used to boost the training of face detection.

\subsubsection{CPU real-time methods} 
Although state-of-the-art face detectors have achieved great success in accuracy, their efficiency is not enough for real-world applications, especially on non-GPU devices. According to the demand of inference speed on CPU, we collect the CPU real-time face detectors~\cite{2016stn,Zhang2017Faceboxes,Zhang2018DCFPN,pcn,2019fldet,ZENG2019PPN,xu2019centerface,deng2020retinaface} here for convenient retrieval. These detectors are able to run at least 20 frames per second (FPS) on a single CPU with VGA-resolution input images. We provide a table in the supplemental material which shows the running efficiency of them, among which the lightweight backbone~\cite{xu2019centerface,deng2020retinaface}, rapidly digested convolutional layer~\cite{Zhang2017Faceboxes,Zhang2018DCFPN}, knowledge distillation~\cite{2019fbi} and region-of-interest (RoI) convolution~\cite{2016stn} are the common practices.

\subsubsection{Problem-oriented methods} 
We highlight some problem-oriented methods which are designed against a variety of specific challenges in face detection. Detecting faces with a wide range of scale is a long-existing challenge in face detection. A group of methods~\cite{2017HR,2017SSH,Zhang2017S3FD,tang2018pyramidbox,Ming_2019_Group_Sampling} are designed for scale-invariant face detection, including scale selection, multi-scale detection, dense anchor setting, scale balancing strategy, \textit{etc}. 
The partially visible faces (\textit{i.e.,} with occlusion) is another issue that harms the detection recall.
The existing solutions~\cite{2015Faceness,2016Grid_Loss,wang2017fan,Ge_2017_CVPR} resort to the facial part arrangement, anchor-level attention and data augmentation by generation, \textit{etc}.
Likewise, the in-plane rotation is an existing factor that impedes face detection. To tackle this problem, PCN~\cite{pcn} calibrates the candidates against the rotation progressively.  

\subsection{Evaluation Metrics and Datasets}
\label{sec:face_detection:evaluation}

\subsubsection{Metrics}
\label{metrics}
Like the general object detection, average precision (AP) is a widely used metric for evaluating the face detection methods. AP is derived from the precision-recall curve. To obtain precision and recall, Intersection over Union (IoU) is used to measure the overlap of the predicted bounding box ($Box_\text{p}$) and the ground-truth ($Box_\text{gt}$), which can be formulated as
\begin{equation}
\mathrm{IoU}=\frac{area(Box_\text{p}\cap Box_\text{gt})}
{area(Box_\text{p} \cup Box_\text{gt})}. 
\end{equation}

The prediction of face detector includes a predicted bounding box and its confidence score.
The confidence score is used to determine whether to accept this according to the confidence threshold.
Then, an accepted prediction can be regarded as true positive (TP) if the IoU is larger than a preset IoU threshold (usually 0.5 for face detection). Otherwise, it will be regarded as a false positive (FP). After determining the TP and FP, the precision-recall curve can be drawn by varying the confidence threshold. AP is computed as the mean precision at a series of uniformly-spaced discrete recall levels~\cite{Everingham2010The}. 
Apart from AP, the receiver operating characteristic (ROC) curve is also adopted as the metric, such as the evaluation in FDDB~\cite{fddbTech}; frames per second (FPS) is used to measure the runtime efficiency of detectors. 

\begin{table}[t]
\begin{center}
\caption{Statistics of popular datasets for face detection.}
\label{fd_dataset}

\resizebox{0.65\linewidth}{!}{
\begin{tabular}{|c|c|c|c|c|c|}
\hline
{Datasets}&{Year}&{$\#$Image}&{$\#$Face}&{$\#$ of faces per image}&{Description}\\
\hline\hline
\multicolumn{6}{|c|}{Training}\\
\hline
ALFW~\cite{Kostinger2011alfw}&2011&21,997&25, 993&1.18& Training source for face detection.\\ 
\hline
WIDER FACE~\cite{Yang_2016_CVPR}&2016&16K&199K&12.43&The largest face detection dataset.\\
\hline\hline
\multicolumn{6}{|c|}{Test}\\
\hline
FDDB~\cite{fddbTech}&2010&2,845&5,171&1.82& A classic face detection benchmark. \\
\hline
AFW~\cite{Zhu2012Face}&2012&205&473&2.31& Multiple facial annotations.\\
\hline
PASCAL faces~\cite{YAN2014790}&2014&851&1,335&1.57&Large facial variations.\\
\hline
MALF~\cite{faceevaluation15}&2015&5,250&11,931&2.27& Fine-grained evaluation.\\ 
\hline
WIDER FACE~\cite{Yang_2016_CVPR}&2016&16K&194K&12.12&The largest face detection dataset.\\
\hline
MAFA~\cite{Ge_2017_CVPR} &2017&30,811&35,806&1.16&Masked face detection.\\
\hline
\end{tabular}}
\end{center}

\end{table}

\subsubsection{Datasets}
We introduce several widely used datasets for face detection. The statistics of them are given in Table~\ref{fd_dataset}. Among them, FDDB~\cite{fddbTech} is a classic dataset of unconstrained face detection which includes low resolution, occlusion and difficult pose variations. It is noteworthy that FDDB uses ellipse as ground-truth instead of rectangular box. 
The images in PASCAL faces dataset~\cite{YAN2014790} are taken from the Pascal person layout dataset~\cite{everingham2011pascal}. 
WIDER FACE~\cite{Yang_2016_CVPR} provides a large number of training data and a challenging test benchmark with large data  variations. 

\begin{table}[t]
    \centering
    \caption{ The performance of state-of-the-art methods on the WIDER FACE~\cite{Yang_2016_CVPR} validation and test subsets. The evaluation metric is AP.} 
    \label{tab:performance_face_detection}
    
    \resizebox{1\linewidth}{!}{
    \begin{tabular}{|r|c|p{9cm}|c|c|c|c|c|c|}
    \hline
         \multirow{2}{*}{ Method }&\multirow{2}{*}{Publication}&\multirow{2}{*}{Subcategory}&\multicolumn{3}{c|}{WIDER FACE Val.}&\multicolumn{3}{c|}{WIDER FACE Test} \\
    \cline{4-9}&&& Easy & Medium &Hard&Easy&Medium&Hard\\
  \hline \hline
    Faceness-WIDER~\cite{Yang_2016_CVPR}&CVPR’16&{ Multi-stage}&0.713&0.634&0.345&0.716&0.604&0.315\\
    \hline
    MSC-CNN~\cite{Yang_2016_CVPR}&CVPR’16&{Multi-stage}&0.691&0.664&0.424&0.711&0.636&0.400\\
    \hline
    CMS-RCNN~\cite{Zhu2017CMS-RCNN}&DLB'17&{Multi-stage}&\textbf{0.899}&\textbf{0.874}&\textbf{0.624}&\textbf{0.902}&\textbf{0.874}&\textbf{0.643}\\
    \hline
    \hline
    Face R-CNN~\cite{wang2017facercnn}&arXiv'17&{Multi-stage, Anchor-based}&0.937&0.921&0.831&0.932&0.916&0.827\\ 
        \hline
    Face R-FCN~\cite{wang2017facerfcn}&arXiv'17&{Multi-stage, Anchor-based}&0.947&0.935&0.874&0.943&0.931&0.876\\ 
       \hline
    ZCC~\cite{Zhu2018}&CVPR'18&{Multi-stage, Anchor-based} &0.949&0.933&0.861&0.949&0.935&0.865\\ 
        \hline
    FDNet~\cite{zhang2018face}&arXiv'18&{Multi-stage, Anchor-based} &\textbf{0.959}&\textbf{0.945}&\textbf{0.879}&\textbf{0.950}&\textbf{0.939}&0.878\\ 
    \hline
    FA-RPN~\cite{Najibi_2019_CVPR}&CVPR'19&{Multi-stage, Anchor-based} &0.949&0.941&0.894&0.945&0.937&\textbf{0.891}\\ 
     \hline
    \hline
    MTCNN~\cite{mtcnn}&SPL’16&{Multi-stage, CPU real-time, Multi-task learning}&0.848&0.825&0.598&0.851&0.820&0.607\\
    \hline
    \hline
    HR~\cite{2017HR}&CVPR'17&{Single-stage}&0.925&0.910&0.806&0.923&0.910&0.819\\
    \hline
    SSH~\cite{2017SSH}&ICCV'17&{Single-stage, Anchor-based} &0.931&0.921&0.845&0.927&0.915&0.844\\
    \hline
    SF$^3$D~\cite{Zhang2017S3FD}&ICCV'17&{Single-stage, Anchor-based} &0.937&0.925&0.859&0.935&0.921&0.858\\    
     \hline
    FAN~\cite{wang2017fan}&arXiv'17&{Single-stage, Anchor-based} &0.952&0.940&0.900&0.946&0.936&0.885\\ 
    \hline
    PyramidBox~\cite{tang2018pyramidbox}&ECCV'18&{ Single-stage, Anchor-based} &0.961&0.950&0.889&0.956&0.946&0.887\\ 
     \hline
    SRN~\cite{chi2019selective}&AAAI'19&{ Single-stage, Anchor-based} &0.964&0.952&0.901&0.959&0.948&0.896\\  
       \hline
    VIM-FD~\cite{zhang2019robust}&arXiv'19&{Single-stage, Anchor-based} &0.967&0.957&0.907&0.962&0.953&0.902\\   
     \hline
    DSFD~\cite{2019DSFD}&CVPR'19&{ Single-stage, Anchor-based} &0.964&0.957&0.904&0.960&0.953&0.900\\    
     \hline
    ISRN~\cite{zhang2019improved}&arXiv'19&{Single-stage, Anchor-based} &0.967&0.958&0.909&0.963&0.954&0.903\\   
         \hline
    AInnoFace~\cite{zhang2019accurate}&arXiv'19&{ Single-stage, Anchor-based} &0.971&0.961&0.918&0.965&0.957&0.912\\     
     \hline
    RefineFace~\cite{zhang2019refineface}&TPAMI'20&{ Single-stage, Anchor-based} &0.971&0.962&0.920&0.965&0.958&0.914\\  
        \hline
    HAMBox~\cite{Liu_2020_HAMBox}&CVPR'20&{Single-stage, Anchor-based} &0.970&0.964&\textbf{0.933}&0.960&0.955&\textbf{0.923} \\
    \hline
    ASFD~\cite{Zhang2020ASFDAA}&arXiv'20&{Single-stage, Anchor-based}&\textbf{0.972}&\textbf{0.965}&0.925&\textbf{0.967}&\textbf{0.962}&0.921\\  
     \hline
    \hline     
     Faceboxes~\cite{Zhang2017Faceboxes}&IJCB'17&{Single-stage, Anchor-based, CPU real-time} &0.840&0.766&0.395&0.839&0.763&0.396\\
    \hline     
    DF$^2$S$^2$~\cite{tian2018df2s2}&arXiv'18&{Single-stage, Anchor-based, Multi-task learning }&0.969&0.959&0.912&\textbf{0.963}&0.954&0.907\\   
     \hline
    PyramidBox++~\cite{li2019pyramidbox}&arXiv'19&{ Single-stage, Anchor-based, Multi-task learning} &0.965&0.959&0.912&0.956&0.952&0.909\\  
    \hline
        CenterFace~\cite{xu2019centerface}&arXiv'19&{ Single-stage, Anchor-free, CPU real-time, Multi-task learning} &0.935&0.924&0.875&0.932&0.921&0.873\\
     \hline
    RetinaFace~\cite{deng2020retinaface}&CVPR'20&{Single-stage, Anchor-based, CPU real-time, Multi-task learning }&\textbf{0.971}&\textbf{0.961}&\textbf{0.918}&\textbf{0.963}&\textbf{0.958}&\textbf{0.914}\\
    \hline
    \end{tabular}
    }
    
\end{table}

 \subsection{Performance Comparison}
Table~\ref{tab:performance_face_detection} shows the performance  of the existing face detectors on WIDER FACE validation and test subsets. 
From the viewpoint of subcategory, we can observe that the single-stage methods with anchor-based mechanism (\textit{e.g.,} RefineFace~\cite{zhang2019refineface}, HAMBox~\cite{Liu_2020_HAMBox}) dominate the state-of-the-art performance. 
For many real-world applications, MTCNN~\cite{mtcnn}, Faceboxes~\cite{Zhang2017Faceboxes}, and RetinaFace~\cite{deng2020retinaface} are the widely used face detectors for building a face recognition system, since they can achieve good balance between the detection accuracy and efficiency.

\begin{figure}[t]
\subfigure[]{{\label{fd_to_fa} \includegraphics[height=3cm]{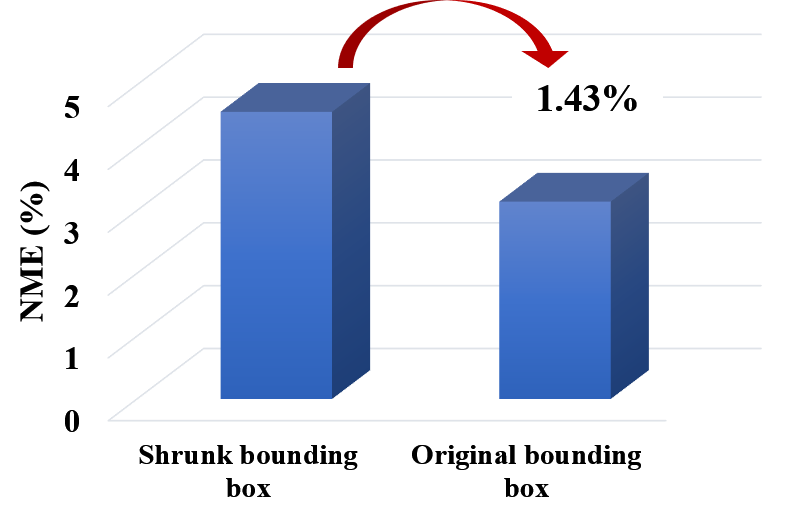}}}
\subfigure[]
{{\label{fd_to_fr}
\includegraphics[height=3cm]{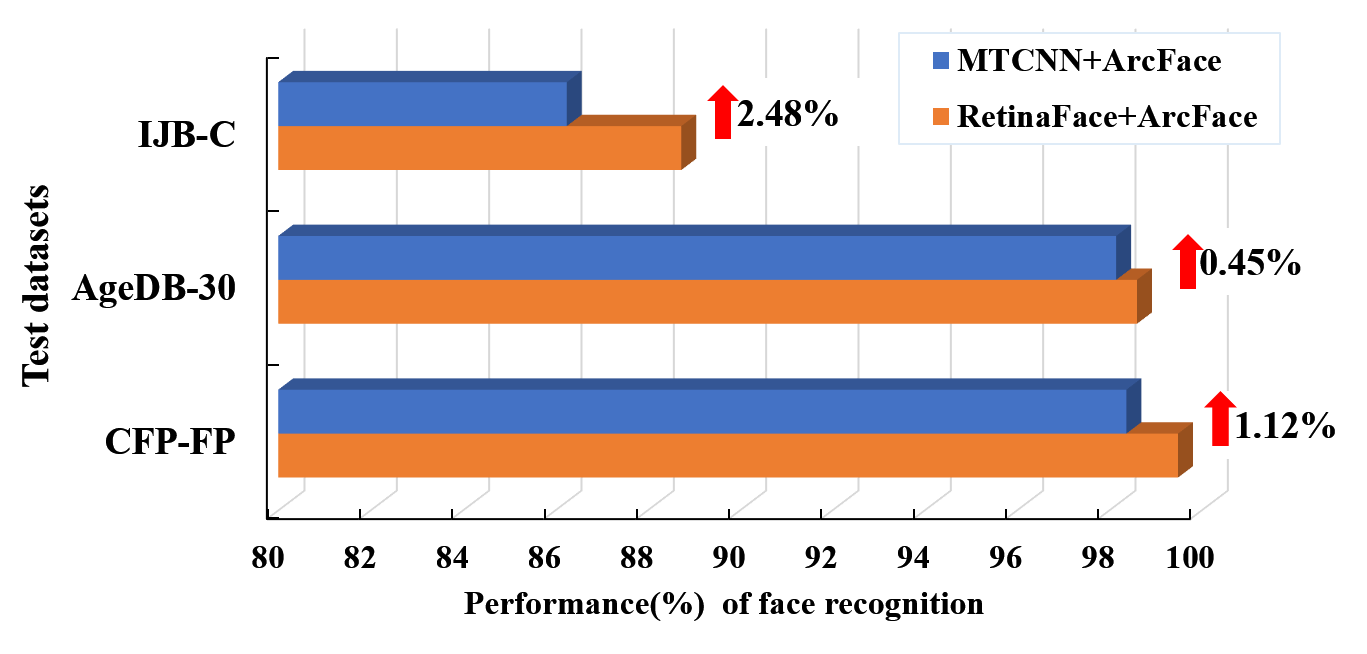}}}

\caption{The accuracy of face detection can influence on the subsequent elements, \textit{i.e.,} face alignment and face representation. (a) Inaccurately detected bounding boxes will bring the performance degradation to facial landmark localization~\cite{Xiong2020GaussianVA}. (b) A more robust face detector can further improve the recognition accuracy~\cite{deng2020retinaface}.}

\end{figure}

\subsection{Effect on the Subsequent Elements}
Face detection is the very first procedure in the end-to-end face recognition system, and thereby plays the role of \textit{input} towards face alignment and face representation. The quality of detection bounding box directly influences on the performance of the subsequent alignment. 
There are two possible cases, \textit{i.e.,} the loss of facial region and the excessively residual context region in the bounding box, both of which are the adverse factors to the subsequent process. 
Certain relevant literature shows solid evidence of that face detection influence on the face alignment and face recognition. Firstly, the quality of detected bounding boxes has a significant impact to facial landmark localization. For example, Xiong~\textit{et al.}~\cite{Xiong2020GaussianVA} compare the performance (Fig.~\ref{fd_to_fa}) of facial landmark localization on the correct face bounding boxes and shrunk face bounding boxes, indicating that the inaccurately detected bounding boxes will bring the performance degradation to the landmark localization.
Moreover, as shown in Fig.~\ref{fd_to_fr}, RetinaFace~\cite{deng2020retinaface} compares the recognition accuracy after using different face detection methods, which proves that a robust face detector can further improve the face recognition accuracy. In summary, face detection has significant impact to both face alignment and face representation. It is indispensable to consider the effect of face detection when establishing  high-performance face recognition system.

\section{face alignment}
\label{sec:face_alignment}
\begin{figure}[t]
\centering
\includegraphics[width=0.95\linewidth]{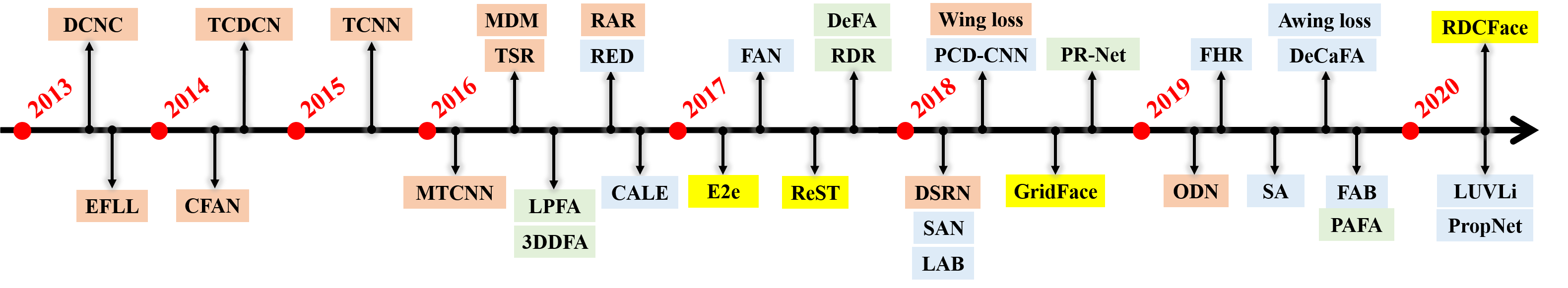}

\caption{The development of representative methods for face alignment. The orange, blue, green, and yellow represent coordinate regression, heatmap regression, 3D model fitting, and landmark-free face alignment methods, respectively. One can refer to Table~\ref{fp_class} for the references of these methods.  }

\label{developmen_fp}
\end{figure}

Given the detected face, face alignment aims to calibrate unconstrained faces to the canonical layout for facilitating the downstream tasks of recognition and analysis. In this section, we review the mainstream routines for face alignment, including landmark-based face alignment, and landmark-free face alignment. Fig.~\ref{developmen_fp} shows the development of representative methods for face alignment.

\subsection{Landmark-based Face Alignment}
\label{sec:face_alignment:lm_based_algin}
Landmark-based face alignment utilizes the spatial transformation to calibrate faces to the predefined canonical layout by involving the facial landmarks as the reference. 
Therefore, the facial landmark localization is the core task of landmark-based alignment.
We sort the existing landmark-based alignment methods into three subcategories, \textit{i.e.,} coordinate regression based methods, heatmap regression based methods and 3D model fitting based methods.

\begin{table}[t]
\begin{center}
\caption{The categorization of face alignment methods.}
\label{fp_class}

\resizebox{1\linewidth}{!}{
\begin{tabular}{|p{2.5cm}|p{1.5cm}|p{6cm}|p{11cm}|}
\hline
\multicolumn{2}{|c|}{Category}&\multicolumn{1}{|c|}{Description}&\multicolumn{1}{|c|}{Method}\\
\hline
{Landmark-based Face Alignment}
& Coordinate regression& Take the landmark coordinates as the target of regression, and learn the nonlinear mapping from the input face image to the landmark coordinates.& DCNC~\cite{sun2013}, EFLL~\cite{Zhou2013ExtensiveFL}, CFAN~\cite{Jie2014Coarse}, TCDCN~\cite{Zhang2014Facial}, RAR~\cite{Xiao2016Robust}, MDM~\cite{Trigeorgis2016MDM}, TSR~\cite{Lv2016TSR}, JFA~\cite{Xu2017JFA}, SIR~\cite{Fan2018SelfReinforcedCR}, TCNN~\cite{Wu2018TCNN}, DSRN~\cite{Miao2018DirectSR}, SBR~\cite{Dong2018SBR}, Wing loss~\cite{Feng2018wingloss}, AAN~\cite{Yue2018AttentionalAN}, ODN~\cite{Zhu_2019_CVPR}, HyperFace~\cite{HyperFace}, MTCNN~\cite{mtcnn}, RetinaFace~\cite{deng2020retinaface}, FLDet~\cite{2019fldet}, CenterFace~\cite{xu2019centerface}, RDN~\cite{Liu2020LearningRN}\\

\cline{2-4}
& Heatmap regression& Output the likelihood response maps of each landmark. & 
CALE~\cite{Bulat2016Convolutional}, RED~\cite{Peng2016RED}, 
Yang~\textit{et al.}~\cite{Jing2017Stacked}, JMFA~\cite{Deng2019JMVFA}, FAN~\cite{Bulat2017HowFA}, LAB~\cite{Wu2018LAB}, SAN~\cite{Dong_2018_CVPR}, FALGCN~\cite{Merget_2018_CVPR}, DU-Net~\cite{Tang2018QuantizedDC},  Guo~\textit{et al.}~\cite{Guo2018StackedDU}, PCD-CNN~\cite{Kumar2018Disentangling3P},  RCN(L+ELT)~\cite{Honari2018ImprovingLL}, HR-Net~\cite{wang2020deep},  Zhang~\textit{et al.}~\cite{Zhang2019StackedHN}, SA~\cite{Liu2019SemanticAF}, FHR~\cite{Tai2019TowardsHA}, Awing loss~\cite{Wang2019adawing}, DeCaFA~\cite{Dapogny2019decafa}, HSLE~\cite{Zou2019LearningRF}, FAB~\cite{Sun2019FABAR}, KDN~\cite{Chen2019FaceAW}, Dong~\textit{et al.}~\cite{Dong2019TeacherSS}, LaplaceKL~\cite{Robinson2019LaplaceLL}, LUVLi~\cite{Kumar2020LUVLiFA},
PropagationNet~\cite{Huang_2020_PropagationNet}\\

\cline{2-4}
&3D model fitting& Infer the 3D face shape from 2D image, and then project it to the image plane to obtain 2D landmarks. & 
LPFA~\cite{Jourabloo2016_D3PF},  
3DDFA~\cite{Zhu2016_3DDFA}, FacePoseNet~\cite{Chang2017FacePoseNet}, 
PIFASCNN~\cite{Jourabloo2017}, 
DeFA~\cite{Liu2017DenseFA}, 
RDR~\cite{Xiao2017RDR},  
Bhagavatula~\textit{et al.}~\cite{Bhagavatula2017FasterTR},
Zhang~\textit{et al.}~\cite{Zhang2018FaceAA}, 
PR-Net~\cite{feng2018joint},
PAFA~\cite{Li2019PoseAwareFA}\\

\hline
\multicolumn{2}{|c|}{Landmark-free Face Alignment} &Directly output aligned faces without the explicit use of landmark.& 
Hayat~\textit{et al.}~\cite{Hayat2017JointRA}, E2e~\cite{Zhong2017e2e}, ReST~\cite{Wu2017ReST}, GridFace~\cite{Zhou2018GridFaceFR}, Wei~\textit{et al.}~\cite{Wei2020BalancedAF}, RDCFace~\cite{Zhao_2020_CVPR}\\ 
\hline
\end{tabular}}
\end{center}

\end{table}

\subsubsection{Coordinate regression} The coordinate regression based methods regard the landmark coordinates as the numerical objective of the regression via neural networks. In other words, they focus on learning the nonlinear mapping from the face image to the landmark coordinate vectors. 

Following the coarse-to-fine manner, most methods employ cascaded regression~\cite{sun2013,Zhou2013ExtensiveFL,Jie2014Coarse,Lv2016TSR} or recurrent neural network (RNN)~\cite{Xiao2016Robust,Trigeorgis2016MDM} to progressively refine the prediction of landmark coordinate. 
Besides, the multi-task learning is also a common routine to facilitate landmark localization with the related facial tasks, such as face detection~\cite{HyperFace, mtcnn,deng2020retinaface,2019fldet,xu2019centerface} and facial attribute recognition~\cite{Zhang2014Facial,Xu2017JFA}. 
Moreover, many regression methods employ the L1, L2, or smoothed L1 loss functions, which are effective but, nonetheless, sensitive to outliers. To handle this problem, Wing loss~\cite{Feng2018wingloss} amplifies the impact of the samples with small or medium range errors. 
The above methods study the facial landmark localization on still images. 
For video face landmark localization, how to leverage the temporal information across frames becomes necessary.
TSTN~\cite{Liu2018TwoStreamTN} develops a two-stream architecture, which locates the landmark from a single frame and captures the temporal consistency for refinement. Besides, SBR~\cite{Dong2018SBR} proposes to exploit the optical flow coherency of detected landmarks when training with video data.

\subsubsection{Heatmap regression} In contrast to the coordinate regression, the heatmap regression based methods output likelihood response maps of each landmark. 
The early exploration~\cite{Bulat2016Convolutional} studies how to aggregate the score maps and refine the prediction with DCNNs.
Later on, Newell~\textit{et al.}~\cite{Newell2016StackedHN} design stacked hourglass (HG) network to generate heatmap for human pose estimation, which has achieved great success. 
As the facial landmark localization task is similar to the human pose estimation, many works~\cite{Jing2017Stacked,Bulat2017HowFA,Deng2019JMVFA,Zhang2019StackedHN,Wang2019adawing,Huang_2020_PropagationNet} adopt the stacked HG network for facial landmark localization and greatly improve the state-of-the-art performance.

The dense pixel-wise classification by the fully convolutional network (FCN) is an effective way for the heatmap regression task.
The HG structure can be regarded as an instance of the fully convolutional network.
Beyond the HG structure, a number of effective network architectures~\cite{Merget_2018_CVPR,Kumar2018Disentangling3P,Dong_2018_CVPR,Dapogny2019decafa,wang2020deep} are newly designed for heatmap regression.
Among them, DeCaFA~\cite{Dapogny2019decafa} utilizes stacked U-nets to preserve the spatial resolution, and landmark-wise attention maps to extract local information around the current estimation. 
High-resolution network (HR-Net)~\cite{wang2020deep} is designed to maintain the high-resolution representation and shows its advantage for landmark-kind tasks.

The above-mentioned wing loss, which is designed for the coordinate regression, however, does not guarantee the convergence for the heatmap regression, due to the imbalance pixel number of foreground and background.
To address this issue, Wang~\textit{et al.}~\cite{Wang2019adawing} propose adaptive wing loss to penalize more on foreground pixels than on background pixels; similarly, PropNet~\cite{Huang_2020_PropagationNet} presents a focal wing loss which adjusts the loss weight of samples in each mini-batch. 

Some facial landmarks have ambiguous definition, such as those on cheek, leading to inconsistent annotations by different annotators. Besides, the landmarks in occluded facial regions also cause imprecise annotations. 
Many methods~\cite{Wu2018LAB,Zou2019LearningRF,Liu2019SemanticAF,Liu2019SemanticAF,Chen2019FaceAW,Kumar2020LUVLiFA} devote to these two issues. 
Facial boundary heatmap~\cite{Wu2018LAB} is a good choice to provide the facial geometric structure for reducing the semantic ambiguities. 
Regarding the semantic ambiguities as noisy annotation, Liu~\textit{et al.}~\cite{Liu2019SemanticAF} provide another path to estimate the real landmark location with a probabilistic model.
More recently, KDN~\cite{Chen2019FaceAW} and LUVLi~\cite{Kumar2020LUVLiFA} propose to estimate the uncertainty of predictions. The uncertainty can be used to identify the images in which the face alignment fails.

\subsubsection{3D model fitting} 
Considering the explicit relationship between 2D facial landmarks and 3D face shape, the 3D model fitting based methods reconstruct the 3D face shape from 2D image, and then project it onto the image plane to obtain the 2D landmarks.
Compared with the regular 2D methods which estimate a set of landmarks, 3D model fitting based methods are able to fit faces with 3D model of thousands of vertexes and align them with large poses. 

Since the cascaded regression is an effective manner to estimate model parameters, 
some methods~\cite{Jourabloo2016_D3PF,Zhu2016_3DDFA,Liu2017DenseFA} combine the cascaded CNN regressor with a dense 3D Morphable Model (3DMM)~\cite{BlanzVolker2003FaceRB} to estimate the 3D face shape.
Despite many advantages, the cascaded CNNs often suffer from the lack of end-to-end training. As a roundabout, Jourabloo~\textit{et al.}~\cite{Jourabloo2017} attempt to fit a 3D face model through a single CNN, which consists of several blocks to adjust the 3D shape and projection matrix according to the features and predictions from the previous blocks. 

Although the above methods take great advantages from 3DMM, the diverse facial shape would lead to inaccurate 2D landmark location, especially when the 3D shape coefficients are sparse.
To tackle this problem, RDR~\cite{Xiao2017RDR} proposes to fit 3D faces by a dynamic expression model and use a recurrent 3D-2D dual learning model to alternatively refine 3D face model and 2D landmarks. 
Beyond regressing the parameters of 3D face shape, Faster-TRFA~\cite{Bhagavatula2017FasterTR} and FacePoseNet~\cite{Chang2017FacePoseNet} estimate the warping parameters of rendering a different view of a general 3D face model. Besides, some methods~\cite{feng2018joint,Zhang2018FaceAA} aim to directly regress the landmarks from the 3D coordinates of face shape.

\subsection{Landmark-free Face Alignment}
\label{sec:face_alignment:lm_free_align}

Landmark-free face alignment methods integrate the alignment transformation processing into DCNNs and output aligned face without relying on facial landmarks. 
This set of methods generally employ the spatial transformer network (Spatial-TN)~\cite{Jaderberg2015SpatialTN} for geometric warping, where the transformation parameters are learned via end-to-end training.
Based on Spatial-TN, Hayat~\textit{et al.}~\cite{Hayat2017JointRA} and Zhong~\textit{et al.}~\cite{Zhong2017e2e} propose to optimize the face alignment with a subsequent module of face representation jointly. 
Since the facial variations are quite complex with various factors, some methods~\cite{Wu2017ReST,Zhou2018GridFaceFR} are designed to improve the deformation ability of Spatial-TN. 
Besides, the radial distortion of face images is another common problem, which is brought by the wide-angle cameras. RDCFace~\cite{Zhao_2020_CVPR} presents a cascaded network which learns the rectification against the radial lens distortion, the face alignment transformation, and the face representation in an end-to-end manner.

\subsection{Evaluation Metrics and Datasets}
\label{sec:face_alignment:evaluation}

We introduce the commonly used evaluation metrics and datasets for face alignment. 
As presented in the following part of this subsection, most landmark-based methods employ the quantitative metrics, such as normalized mean error. Besides, landmark-free methods employ the evaluation oriented to face recognition, and we will describe their metrics in the face representation section.

\begin{table}[t]
\begin{center}
\caption{Statistics of popular facial landmark datasets. ``-'' refers to none official protocol for splitting the training and test set.}

\label{flmk_dataset}
\resizebox{0.75\linewidth}{!}{
\begin{tabular}{|c|c|c|c|c|c|c|}
\hline
{Datasets}&{Year}&{$\#$ Total}&{$\#$ Training}&{$\#$ Test}&{$\#$ Point}&{Description}\\
\hline\hline
Multi-PIE~\cite{Gross2008MultiPIE}&2008&755,370&-&-&68& The largest facial dataset in controlled condition.\\
\hline
LFPW~\cite{Belhumeur2011LocalizingPO}&2010&2,845&-&-&35& Images taken from uncontrolled setting.\\
\hline
ALFW~\cite{Kostinger2011alfw}&2011&24,386&20,000&4,386&21&A large-scale facial landmark dataset.\\
\hline
AFW~\cite{Zhu2012Face}&2012&473&-&-&6& Multiple facial annotations.\\
\hline
HELEN~\cite{Le2012InteractiveFF}&2012&2,330&2,000&330&194&Providing dense landmark annotations.\\
\hline
COFW~\cite{BurgosArtizzu2013RobustFL}&2013&1,852&1,345&507&29& Containing occluded faces.\\
\hline
300-W~\cite{Sagonas2013300FI} &2013&3,837&3,148&689&68&The most frequently used dataset of facial landmark.\\
\hline
300-VW~\cite{Shen2015TheFF}&2015&114&50&64&68& A video facial landmark dataset.\\
\hline
Menpo~\cite{Zafeiriou2017TheMF} &2017&28,273&12,014&16,259&68&Containing both semi-frontal and profile faces.\\
\hline
WFLW~\cite{Wu2018LAB} &2018&10,000&7,500&2,500&98& Multiple annotations and large variations.\\
\hline
JD-landmark~\cite{Liu2019GrandCO} &2019&15,393&13,393&2,000&106& Covering large facial variations.\\
\hline
\end{tabular}}
\end{center}

\end{table}

\subsubsection{Metrics}
The widely used evaluation metric is to measure the point-to-point Euclidean distance by normalized mean error (NME), which can be defined as

\begin{equation}{NME}=\frac{1}{M} \sum_{k=1}^{M} \frac{\left\|p_{k}-{g}_{k}\right\|_{2}}{d},\end{equation}
where $M$ is the number of landmarks, $p_{k}$ and ${g}_{k}$ represent the prediction and ground-truth coordinates of the face landmarks, $k$ denotes the index of landmarks, and $d$ refers to the normalized distance which is used to alleviate the abnormal measurement caused by different face scales and large pose. 
There are four types of normalized distance for computing NME, \textit{i.e.,} the geometric mean of the width and height of the 
face bounding box, the distance between the outer corners of eyes, the distance between the pupils, and the diagonal of the face bounding box.

The cumulative errors distribution (CED) curve is also used as an evaluation criterion. 
CED is a distribution function of NME. The vertical axis of CED represents the proportion of test images that have an error value less than or equal to the error value on the horizontal axis. 
The area under the curve (AUC) also provides a reference of how the algorithm performs at a given error: 
\begin{equation}{AUC}_{\alpha}=\int_{0}^{\alpha} f(e) d e,\end{equation}
where ${\alpha}$ is the given error corresponding to the upper bound of integration calculation, $e$ is the progressive normalized errors and $f(e)$ refers to the CED curve. Larger AUC indicates better performance. Based on CED curve, failure rate can be used to measure the robustness of  a algorithm, which denotes the percentage of samples in the test set whose NME is larger than a threshold.

\begin{table}[t]
    \centering
     \caption{ Performance of facial landmark localization methods on the 300W, WFLW-All, ALFW-Full, and COFW datasets. The evaluation metric is NME ($\%$). For 300W test set, two types of NME normalization (\textit{i.e.,} inter-pupil normalization and inter-ocular normalization) are used. For WFLW-All dataset, the inter-ocular normalization is applied. For ALFW-Full dataset, the diagonal of face bounding box is adopted as the normalization factor. For COFW dataset, the inter-pupil normalization is applied. ``-'' indicates that the authors do not report the performance with the corresponding protocol. }
    \label{tab:performance_aglinment}
    
    \resizebox{1\linewidth}{!}{
    \begin{tabular}{|r|c|c|c|c|c|c|c|c|c|c|c|}
    \hline
    \multirow{2}{*}{Method}&\multirow{2}{*}{Publication}&\multirow{2}{*}{Subcategory}&\multicolumn{3}{c|}{300-W (inter-pupil normalization) }&\multicolumn{3}{c|}{300-W (inter-ocular normalization)}&\multirow{2}{*}{WFLW}& \multirow{2}{*}{ALFW}& \multirow{2}{*}{COFW}\\ 
    \cline{4-9}&&&Com. subset&Chall. subset&Full set&Com. subset&Chall. subset&Full set&&&\\
   \hline \hline 
        CFAN~\cite{Jie2014Coarse} &  ECCV'14 & Coordinate regression & 5.50 & 16.78 & 7.69 &-&-&- &-& 10.94 & 8.38\\
    \hline
        TCDCN~\cite{Zhang2014Facial} & ECCV'14 & Coordinate regression & 4.80 & 8.60 & 5.54  &-&-&- &-& 7.60& 8.05 \\
        \hline
        MDM~\cite{Trigeorgis2016MDM} & CVPR'16  & Coordinate regression & 4.83 & 5.88 & 10.14  &-&-&- &-& -& 6.26\\   
        \hline
        RAR~\cite{Xiao2016Robust} & ECCV'16 & Coordinate regression & 4.12 & 8.35 & 4.94  &-&-&-  &-& 7.23& 6.03\\        
        \hline
         TSR~\cite{Lv2016TSR} &CVPR'17 & Coordinate regression & 4.36 & 7.56 & 4.99 &-&-&- &-& 2.17& - \\
        \hline
        SIR~\cite{Fan2018SelfReinforcedCR} &AAAI'18 & Coordinate regression & 4.29 & 8.14 & 5.04 &-&-&- &-&- & -\\
        \hline
        DSRN~\cite{Miao2018DirectSR} & CVPR'18 & Coordinate regression & 4.12 & 9.68 & 5.21 &-&-&- &-& 1.86&- \\   
        \hline
        Wing loss~\cite{Feng2018wingloss} & CVPR'18  & Coordinate regression & \textbf{3.01} & \textbf{6.01} &\textbf{3.60}  &-&-&- & \textbf{5.11}& \textbf{1.47} & 5.44\\ 
        \hline
        CPM + SBR~\cite{Dong2018SBR}& CVPR'18 & Coordinate regression &-&-&-  & \textbf{3.28} & 7.58 & \textbf{4.10}&- &2.14 & -\\   
        \hline 
         ODN~\cite{Zhu_2019_CVPR} &CVPR'19 &Coordinate regression &-&-&-  & 3.56&\textbf{6.67}&4.17& -&1.63& \textbf{5.30} \\ 
        \hline
        RDN~\cite{Liu2020LearningRN} &TPAMI'20 & Coordinate regression & 3.31 & 7.04 & 4.23 &-&-&- &-& 2.06& 5.82 \\    
        \hline  \hline
        3DDFA~\cite{Zhu2016_3DDFA} & CVPR'16 & 3D model fitting & 6.15 & 10.59 & 7.01  &-&-&- &-& 5.60&- \\
        \hline
        PIFASCNN~\cite{Jourabloo2017} & ICCV'17  & 3D model fitting & 5.43 & 9.88 & 6.30 &-&-&- &-& 4.45&- \\        
        \hline
        DeFA~\cite{Liu2017DenseFA} &ICCVW'17 & 3D model fitting & 5.37 & 9.38 & 6.10 &-&-&- &-& -& -\\      
        \hline
        RDR~\cite{Xiao2017RDR} &ICCV'17 & 3D model fitting & 5.03 & 8.95 & 5.80 &-&-&- &-&4.41 & -\\   
        \hline
        PAFA~\cite{Li2019PoseAwareFA} &BMVC'19 & 3D model fitting & \textbf{3.42} &\textbf{5.73}& \textbf{3.87} &-&-&- &-&\textbf{1.51}& \textbf{3.55}\\    
        \hline \hline
         RCN$^+$~\cite{Honari2018ImprovingLL} &CVPR'18  & Heatmap regression & 4.20 & 7.78 & 4.90 &-&-&- &-& 2.17& -\\    
        \hline
        PCD-CNN~\cite{Kumar2018Disentangling3P}& CVPR'18 & Heatmap regression &-&-&-  & 3.67 & 7.62 & 4.44&-&2.40 & 5.77\\ 
    \hline
        SAN~\cite{Dong_2018_CVPR}&CVPR'18& Heatmap regression &-&-&- &3.34& 6.60 &3.98&-& 1.91& - \\
    \hline
        HR-Net~\cite{wang2020deep} & CVPR'18 & Heatmap regression &-&-&-  & 2.87 & 5.15 & 3.32 &- &1.57 &- \\ 
    \hline
    
        LAB~\cite{Wu2018LAB} & CVPR'18  & Heatmap regression & 3.42 & 6.98 & 4.12 & 2.98 & 5.19 & 3.49 &5.27 & \textbf{1.25}& 3.92\\  
    \hline 
        DU-Net~\cite{Tang2018QuantizedDC}& ECCV'18 & Heatmap regression &-&-&-  & 2.90 & 5.15 & 3.35&-& -& -\\ 
                \hline
         SA~\cite{Liu2019SemanticAF} &CVPR'19  & Heatmap regression & 3.45 & 6.38 & 4.02 &-&-&- &-&1.60 &- \\ 
          \hline
        HG-HSLE~\cite{Zou2019LearningRF} & ICCV'19 & Heatmap regression& 3.94 & 7.24 & 4.59 & 2.85 & 5.03 & 3.28 &-& -& -\\    
        \hline
        Awing loss~\cite{Wang2019adawing} & ICCV'19  & Heatmap regression & 3.77 & 6.52 & 4.31 &2.72 & 4.52 & 3.07 &\textbf{4.36}& -& 4.94\\    
        \hline
        LaplaceKL~\cite{Robinson2019LaplaceLL} &ICCV'19  &Heatmap regression & \textbf{3.28} & 7.01 & \textbf{4.01} &-&-&- &-& 1.97& -\\    
            \hline 
    
        DeCaFA~\cite{Dapogny2019decafa}& ICCV'19 & Heatmap regression &-&-&- & 2.93 & 5.26 & 3.39& 4.62 & -& -\\
    \hline
        LUVLi~\cite{Kumar2020LUVLiFA}& CVPR'20 & Heatmap regression &-&-&-  & 2.76 & 5.16 & 3.23 &4.37 & 1.39& - \\   
     \hline
         PropNet~\cite{Huang_2020_PropagationNet} & CVPR'20 & Heatmap regression & 3.70 &\textbf{5.75}& 4.10 &\textbf{2.67} &\textbf{3.99} & \textbf{2.93}& \textbf{4.05} & -&\textbf{3.71}\\   
    \hline

    \end{tabular}
    }
    
\end{table}

\subsubsection{Datasets} 
The facial landmark datasets can be sorted by the constrained condition and in-the-wild condition. The statistics of these datasets are given in Table~\ref{flmk_dataset}. CMU Multi Pose, Illumination, and Expression (Multi-PIE)~\cite{Gross2008MultiPIE} is the largest facial dataset in constrained condition, which provides 337 subjects with 15 predefined poses, 19 illumination conditions and 6 facial expressions.
The annotated facial landmarks are 68 points for frontal faces and 39 points for profile ones.

In addition, more in-the-wild datasets~\cite{Belhumeur2011LocalizingPO,Kostinger2011alfw,Zhu2012Face,Le2012InteractiveFF,BurgosArtizzu2013RobustFL,Sagonas2013300FI,Shen2015TheFF,Zafeiriou2017TheMF,Wu2018LAB,Liu2019GrandCO} are proposed for facial landmark localization. Among them,  300-W~\cite{Sagonas2013300FI} is the most frequently used dataset, which follows the annotation configuration of Multi-PIE and re-annotates the images in LFPW, AFW, HELEN, and iBug~\cite{Sagonas2016300FI}. 
Besides, Menpo~\cite{Zafeiriou2017TheMF} is a large-scale facial landmark dataset with more difficult cases for facial landmark localization. 
JD-landmark~\cite{Liu2019GrandCO} annotates face images with 106 facial landmarks, providing more structural information of facial components. 300-VW~\cite{Shen2015TheFF} provides 50 video clips for training and 64 for test of landmark localization in video.

\subsection{Performance Comparison}
Table~\ref{tab:performance_aglinment} shows the comparison of state-of-the-art facial landmark localization methods on various test datasets, including 300-W, WLFW-ALL, ALFW-Full, and COFW. 
Among coordinate regression based methods, Wing loss~\cite{Feng2018wingloss} is a simple but effective approach which has been widely used. 
More recently, the heatmap regression based methods attract more attention, since they can obtain the leading performance by maintaining facial structure information throughout the models. 

\begin{figure}[t]
\centering
\includegraphics[height=3cm]{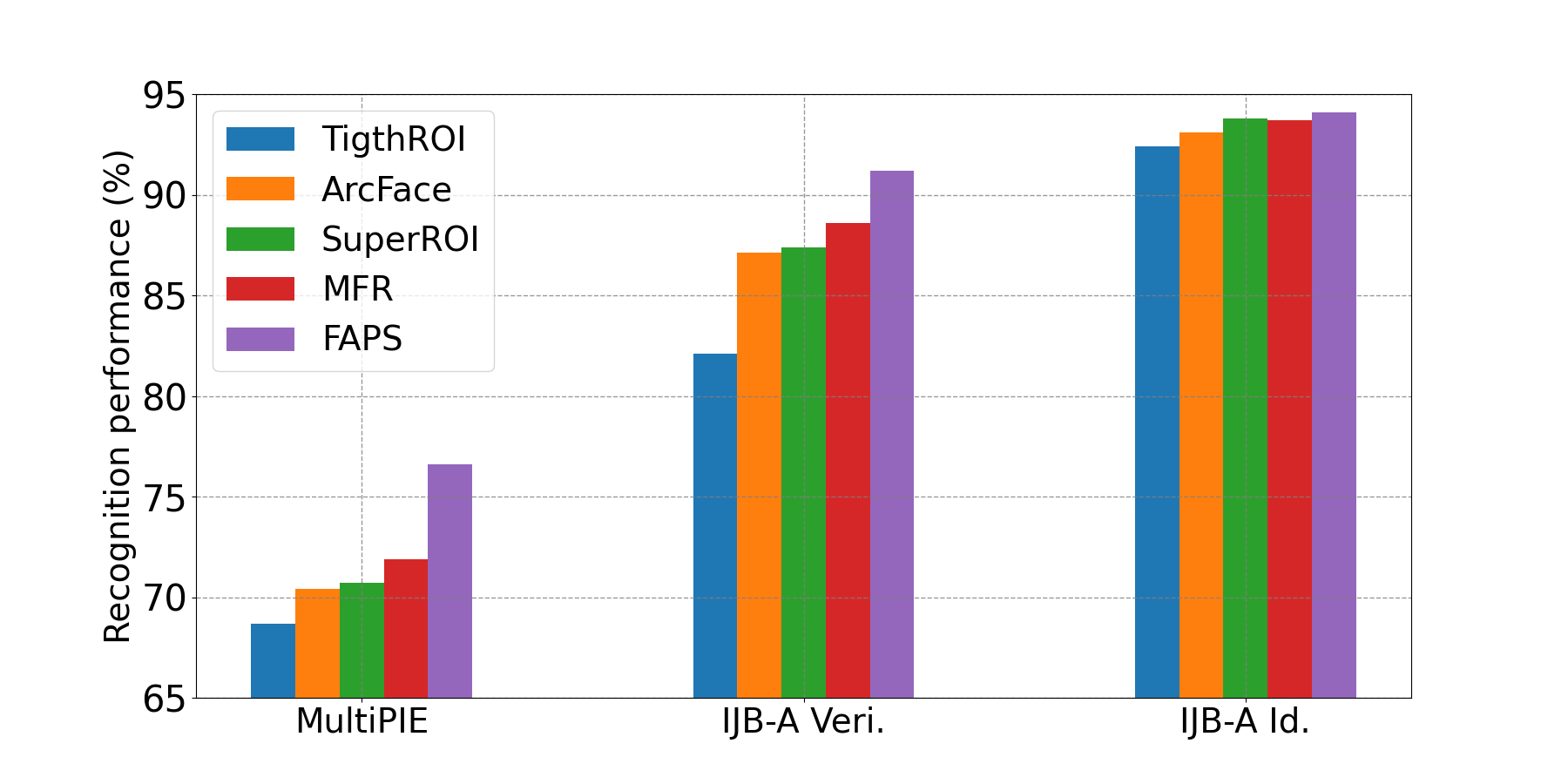}

\caption{Appropriate face alignment policy is beneficial to face recognition in many situations~\cite{Xu2021SearchingFA}. 
The indicated choices of alignment policy are different in number of used facial landmarks, cropping size of face image, and vertical shift. 
Among them, ArcFace~\cite{deng2020retinaface} employs a 5-point alignment template, and MFR~\cite{Cao2020DomainBF} utilizes a 25-point one. TigthROI~\cite{Xu2021SearchingFA} involves few external facial feature (\textit{e.g.,} jaw-line, ears, part of hair), which lacks useful facial features. SuperRoI~\cite{Xu2021SearchingFA} uses large cropping size, which potentially covers irrelevant background. FAPS~\cite{Xu2021SearchingFA} is designed to search an optimal face alignment template. The latter three policies 
use 68 landmarks which provide adequate information for computing the affine transformation matrix.   }
\label{fa_to_fr}

\end{figure}

\subsection{Effect on the Face Representation}
Face alignment is the intermediate procedure.
The study of how face alignment influences on face representation is vital for tuning the recognition system to attain its maximum effect.
For landmark-based face alignment, a set of inaccurate facial landmarks will harm the alignment and then impede the following feature computation as well. Specifically, human faces appear in the images with similar layout, and such layout can be regarded as a template in spatial coordinates. In fact, the alignment is accomplished mostly by warping the face to the predefined coordinates according to the predicted landmarks. Then, the face representation model learns identity feature from facial images with such layout. Once the predicted landmarks are inaccurate, the facial image will drift away from the predefined coordinates, which is unexpected layout for the face representation model.
Guo~\textit{et al.}~\cite{Guo2018StackedDU} and Deng~\textit{et al.}~\cite{deng2020retinaface} both compare the widely used MTCNN~\cite{mtcnn} and their methods, and find that poor landmark localization will bring shift variation, while robust face alignment can boost recognition accuracy, especially for the cross-pose face recognition. 
Besides, as discussed in certain studies~\cite{Parkhi2015DeepFR,Schroff2015FaceNetAU,Xu2021SearchingFA}, the configuration of face alignment process (so-called face alignment policy), including the number of used facial landmarks, the cropping size of face image, and the vertical shift, greatly influences on the performance of face recognition. 
As shown in Fig.~\ref{fa_to_fr}, 
the results from \cite{Xu2021SearchingFA} indicate that the proper face alignment policy is beneficial to face recognition in many situations.
Moreover, a moderate degree of spatial transformation is required in the alignment processing~\cite{Wei2020BalancedAF}. Both limited and excessive transformation will bring disturbance.

\section{Face Representation}
\label{sec:face_representation}
Subsequent to face alignment, the face representation stage aims to map the aligned face images to a feature space, where the features of the same identity are close and those of different identities are far apart. 
In practical applications, there are two major tasks of face recognition, \textit{i.e.,} face verification and face identification. The face verification refers to predict whether a pair of face images belong to the same identity. The face identification can be regarded as an extension of face verification, which aims to determine the specific identity of a face (\textit{i.e.,} probe) among a set of identities (\textit{i.e.,} gallery); moreover, in the case of open-set face identification, a prior task is needed, whose target is predicting whether the face belongs to one of the gallery identities or not. 

For both the face verification and face identification, face representation is used to measure the similarity between face images. Therefore, how to learn discriminative face representation is the core target.  
With the advanced feature learning ability of DCNNs, face representation has made great progress. In the followings, we provide a systematic review of the learning methods of face representation from two major aspects, \textit{i.e.,} network architecture and training supervision.

\begin{table*}[t]
\begin{center}
\caption{The categorization of face representation learning.}

\label{fr_class}
\resizebox{1\linewidth}{!}{
\begin{tabular}{|p{2cm}|p{2.5cm}|p{6cm}|p{13cm}|}
\hline
\multicolumn{2}{|c|}{Category}&\multicolumn{1}{c|}{Description}&\multicolumn{1}{c|}{Method}\\
\hline

{Network Architectures} 
&General & The basic and universal designs for common visual recognition tasks. & AlexNet~\cite{Krizhevsky2012ImageNetCW}, VGGNet~\cite{Simonyan2015VeryDC}, GoogleNet~\cite{Szegedy2015GoingDW}, ResNet~\cite{He2016DeepRL}, 
Xception~\cite{Chollet2017XceptionDL}, 
DenseNet~\cite{Huang2017DenselyCC}
AttentionNet~\cite{Wang2017ResidualAN}, 
SENet~\cite{Hu2018SqueezeandExcitationN}, 
SqueezeNet~\cite{Iandola2017SqueezeNetAA}, 
MobileNet~\cite{Howard2017MobileNetsEC}, 
ShuffleNet~\cite{Zhang2018ShuffleNetAE}, MobileNetV2~\cite{Sandler2018MobileNetV2IR}, 
Shufflenetv2~\cite{Ma2018ShuffleNetVP}
\\

\cline{2-4}
& Specialized & The modified or ensemble designs oriented to face recognition. & HybridDL\cite{Sun2013HybridDL}, 
DeepID series~\cite{sun2014deep,Sun2015DeeplyLF,Sun2015DeepID3FR}, 
MM-DFR~\cite{Ding2015RobustFR}, 
B-CNN~\cite{Chowdhury2016OnetomanyFR}, 
ComparatorNet~\cite{Xie2018ComparatorN}, 
Contrastive CNN~\cite{Han2018FaceRW}, 
PRN~\cite{Kang2018PairwiseRN}, 
AFRN~\cite{Kang2019AttentionalFR}, 
FANFace~\cite{Yang2020FANFaceAS}, 
SparseNet~\cite{sun2015sparsifying}, 
Light-CNN~\cite{Wu2018ALC}, 
MobileFaceNet~\cite{chen2018mobilefacenets}, 
Mobiface~\cite{Duong2018MobiFaceAL}, 
ShuffleFaceNet~\cite{MartnezDaz2019ShuffleFaceNetAL}, Hayat~\textit{et al.}~\cite{Hayat2017JointRA}, 
E2e~\cite{Zhong2017e2e}, 
ReST~\cite{Wu2017ReST}, 
GridFace~\cite{Zhou2018GridFaceFR}, 
RDCFace~\cite{Zhao_2020_CVPR}, 
Wei~\textit{et al.}~\cite{Wei2020BalancedAF}, 
Co-Mining~\cite{wang2019co}, 
GroupFace~\cite{Kim2020GroupFaceLL}, 
MFR~\cite{Cao2020DomainBF}
\\

\hline
{Training Supervision} 
& Classification& Considering the face representation learning as a classification task.&
DeepFace~\cite{taigman2014deepface}, DeepID~\cite{Sun2014DeepID}, 
MM-DFR~\cite{Ding2015RobustFR}, 
L-softmax~\cite{liu2016large}, NormFace~\cite{wang2017normface}, L2-softmax~\cite{ranjan2017l2},
COCO loss~\cite{Liu2017RethinkingFD},
SphereFace~\cite{liu2017sphereface}, Ring loss~\cite{zheng2018ring}, AM-softmax~\cite{wang2018additive}, CosFace~\cite{wang2018cosface}, ArcFace~\cite{deng2019arcface}, AdaptiveFace~\cite{liu2019adaptiveface}, 
Fair loss~\cite{Fair_Loss}, MV-softmax~\cite{Wang2019MisclassifiedVG}, ArcNeg~\cite{liu2019towards},  AdaCos~\cite{zhang2019adacos}, P2SGrad~\cite{Zhang2019P2SGradRG}, NTP~\cite{Hu2019NoiseTolerantPF},  Co-Mining~\cite{wang2019co},
PFE~\cite{Shi2019ProbabilisticFE}, 
CurricularFace~\cite{Huang2020CurricularFaceAC},
Shi~\textit{et al.}~\cite{shi2020universal}, GroupFace~\cite{Kim2020GroupFaceLL}, 
MFR~\cite{Cao2020DomainBF}, RCM loss~\cite{Wu_2020_CVPR},  DUL~\cite{Chang2020DataUL}
\\
\cline{2-4}
& Feature embedding&Optimizing the feature distance according to the label of sample pair. &
DeepID2~\cite{sun2014deep}, FaceNet~\cite{Schroff2015FaceNetAU}, 
VGG Face~\cite{Parkhi2015DeepFR}, 
Lifted structured~\cite{oh2016deep}, 
N-pair loss~\cite{sohn2016improved}, 
Multibatch~\cite{Tadmor2016LearningAM},  
TPE~\cite{Sankaranarayanan2016TripletPE}, 
Smart mining~\cite{Manmatha2017SamplingMI}, 
Contrastive CNN~\cite{Han2018FaceRW}
\\ 
\cline{2-4}
&Hybrid & Applying classification and feature embedding together as the supervisory signals.&
DeepID2~\cite{sun2014deep}, DeepID2+~\cite{Sun2015DeeplyLF}, DeepID3~\cite{Sun2015DeepID3FR}, TUA~\cite{liu2015targeting},
 Center loss~\cite{wen2016discriminative},
Marginal loss~\cite{Deng2017MarginalLF}, 
Range loss~\cite{Zhang2017RangeLF}, 
DM~\cite{smirnov2017doppelganger}, 
PRN~\cite{Kang2018PairwiseRN}, 
UniformFace~\cite{UniformFace},  RegularFace~\cite{zhao2019regularface}, 
UT~\cite{Zhong_2019_CVPR}, 
LBL~\cite{zhu2019large}, 
AFRN~\cite{Kang2019AttentionalFR}, 
Circle loss~\cite{sun2020circle} 
\\
\cline{2-4}
&Semi-supervised & Exploiting labeled and unlabeled faces for representation learning. &
CDP~\cite{Zhan2018ConsensusDrivenPI},
GCN-DS~\cite{yang2019learning}, 
GCN-VE~\cite{Yang2020LearningTC}, 
UIR~\cite{Yu2019UnknownIR}, 
RoyChowdhury~\textit{et al.}~\cite{RoyChowdhury2020ImprovingFR}
\\
\hline
{Specific Tasks} 
& Cross-age& Identifying faces across a wide range of ages. & 
LF-CNNs~\cite{Wen2016LatentFG}, CAN~\cite{Xu2017AgeIF}, AFRN~\cite{Du2019AgeFR}, DAL~\cite{Wang2019DecorrelatedAL}, 
AE-CNN~\cite{Zheng2017AgeEG}, OE-CNN~\cite{Wang2018OrthogonalDF}, IPCGANs~\cite{Wang2018FaceAW}, LMA~\cite{Antipov2017BoostingCF}, Dual cGANs~\cite{song2018dual}, AIM~\cite{Zhao2019LookAE} 
\\

\cline{2-4}
& Cross-pose& Identifying faces across a wide range of poses. &
TP-GAN~\cite{Huang2017tpgan}, PIM~\cite{Zhao2018pim}, DREAM~\cite{Cao2018PoseRobustFR}, DA-GAN~\cite{Zhao2017DualAgentGF}, DR-GAN~\cite{Tran2017DisentangledRL}, UV-GAN~\cite{Deng2018UVGANAF}, 
CAPG-GAN~\cite{Hu2018CAPG}, 
PAMs~\cite{Masi2016PoseAwareFR}, 
MPRs~\cite{AbdAlmageed2016FaceRU}, 
MvDN~\cite{kan2016multi}
\\

\cline{2-4}
&Racial bias& Addressing the imbalance race distribution of training datasets. &  
IMAN~\cite{Wang2019RacialFI}, 
RL-RBN~\cite{Wang_2020_CVPR}\\

\cline{2-4}
&Cross-modality& Performing face recognition on a pair of images captured by different sensing modalities. & 
Reale~\textit{et al.}~\cite{Reale2016SeeingTF}, HFR-CNNs~\cite{Saxena2016HeterogeneousFR}, 
TRIVET~\cite{Liu2016TransferringDR}, 
IDR~\cite{He2017LearningID}, 
DVR~\cite{Wu2018DisentangledVR}, 
MC-CNN~\cite{Deng2019MutualCC}, 
WCNN~\cite{He2019WassersteinCL}, 
NAD~\cite{Lezama2017NotAO}, ADHFR~\cite{Song2018AdversarialDH}, CFC~\cite{He2020AdversarialCF}, Mittal~\textit{et al.}~\cite{Mittal2015CompositeSR},  ForensicFR~\cite{Galea2017ForensicFP}, TDFL~\cite{Wan2019TransferDF}, E2EPG~\cite{Zhang2015EndtoEndPG}, CASPG~\cite{Zhang2017ContentAdaptiveSP}, DualGAN~\cite{Yi2017DualGANUD}, 
PS2-MAN~\cite{Wang2018HighQualityFP},
DTFS~\cite{Zhang2019DualTransferFS},
Cascaded-FS~\cite{Zhang2020CascadedFS}, 
PTFS~\cite{Zhang2019SynthesisOH} 
\\

\cline{2-4}
&Low-shot & Training and test with the data that has a small number of samples per identity.
& 
SSPP-DAN~\cite{Hong2017SSPPDANDD}, 
Guo~\textit{et al.}~\cite{guo2017one}, 
Choe~\textit{et al.}~\cite{Choe2017FaceGF}, 
Hybrid Classifiers~\cite{Wu2017LowShotFR}, 
Cheng~\textit{et al.}~\cite{cheng2017know}, 
DM~\cite{smirnov2017doppelganger}, Yin~\textit{et al.}~\cite{yin2019feature}, 
\\
\cline{2-4}
&Video-based & Performing face recognition with video sequences.
&
TBE-CNN~\cite{Ding2018TrunkBranchEC}, NAN~\cite{Yang2017NeuralAN}, C-FAN~\cite{Gong2019VideoFR}, FANVFR~\cite{Liu2019FeatureAN},  MARN~\cite{Gong2019LowQV}, Rao~\textit{et al.}~\cite{Rao2017LearningDA}, CFR-CNN~\cite{Parchami2017UsingDA}, ADRL~\cite{Rao2017AttentionAwareDR}, DAC~\cite{Liu2018DependencyAwareAC} 
\\
\hline
\end{tabular}}
\end{center}

\end{table*}

\subsection{Network Architectures} 
\label{sec:face_representation:architecture}
The recent improvement of face representation partly benefits from the advance of deep architecture design. We first review the literature of network architecture for face representation learning. According to the designing purpose, we divide them into general architectures and specialized architectures. The general architectures are the basic and universal designs for common visual recognition tasks in the first place, and applied to face representation learning afterward.
The specialized architectures include the modified or ensemble designs oriented to face recognition. 

\subsubsection{General architectures}
With the advanced feature learning ability of DCNNs~\cite{Krizhevsky2012ImageNetCW,Simonyan2015VeryDC,Szegedy2015GoingDW,He2016DeepRL,Chollet2017XceptionDL,Huang2017DenselyCC,Wang2017ResidualAN,Hu2018SqueezeandExcitationN}, face representation has made great progress. 
Among them, AlexNet~\cite{Krizhevsky2012ImageNetCW} obtains the first place in ImageNet~\cite{deng2009imagenet} competition (ILSVRC) 2012 and achieves significant improvement compared with the traditional methods. 
Then, VGGNet~\cite{Simonyan2015VeryDC} presents a more generic network, which replaces the large convolutional kernels by the stacked 3$\times$3 ones, enabling the network to grow in depth. 
In order to enlarge the network without the extra increase of computational budget, GoogleNet~\cite{Szegedy2015GoingDW} develops an inception architecture to concatenate the feature maps that are generated by the convolutions of different receptive field. Soon, GoogleNet is applied to face representation learning, namely FaceNet~\cite{Schroff2015FaceNetAU}.
More recently, ResNet~\cite{He2016DeepRL} proposes a residual structure to make it possible for training deep networks that have hundreds of layers. ResNet is a modern network that has been widely used on many visual tasks, including face recognition. 
Additionally, several lightweight neural networks~\cite{Iandola2017SqueezeNetAA,Zhang2018ShuffleNetAE,Howard2017MobileNetsEC,Sandler2018MobileNetV2IR,Ma2018ShuffleNetVP} are proposed to achieve the trade-off between speed and accuracy. All of them have been employed as backbone network for representation learning in the face recognition literature after being designed.

\subsubsection{Specialized architectures}
The aforementioned architectures are initially proposed for general visual tasks. Besides, many works develop specialized architectures for face representation learning. 
At first, many works~\cite{Sun2013HybridDL,Sun2014DeepID,sun2014deep,Ding2015RobustFR} attempt to assemble multiple convolution networks together for learning multiple local features from a set of facial patches. Given the human face appearing with regular arrangement of facial parts (eyes, nose, mouth, \textit{etc}), such combination of multiple networks with respect to facial part can be more reliable than a single network.
Besides, Xie~\textit{et al.}~\cite{Xie2018ComparatorN} design an end-to-end architecture, namely Comparator Network, to measure the similarity of two sets of a variable number of face images.
Certain approaches~\cite{Kang2018PairwiseRN,Kang2019AttentionalFR} develop feature-pair relational network to capture the relations between a pair of local appearance patches. 
More recently, FANFace~\cite{Yang2020FANFaceAS} integrates the face representation network and facial landmark localization network, so that the heatmap of landmarks will boost the features for recognition.

In addition, many studies~\cite{sun2015sparsifying,Wu2015ALC,Wu2018ALC,chen2018mobilefacenets,Duong2018MobiFaceAL,MartnezDaz2019ShuffleFaceNetAL} focus on developing the lightweight architecture. 
To reduce the parameters of deep networks, SparseNet~\cite{sun2015sparsifying} proposes to iteratively learn sparse structures from the previously learned dense models. 
Light-CNN~\cite{Wu2018ALC} introduces a max-feature-map (MFM) activation function to gain better generalization ability than ReLU for face recognition; based on MFM, a lightweight architecture is developed that achieves the advantages in terms of runtime efficiency and model size.
MobileFaceNet~\cite{chen2018mobilefacenets} replaces the global average pooling layer in the original MobileNet~\cite{Sandler2018MobileNetV2IR} with a global depth-wise convolution layer so the output feature can be improved by the spatial importance in the last layer.

It is worth noting that, in some landmark-free face alignment methods~\cite{Hayat2017JointRA,Zhong2017e2e,Wu2017ReST,Zhou2018GridFaceFR,Zhao_2020_CVPR,Wei2020BalancedAF} which have been presented in Section~\ref{sec:face_alignment:lm_free_align}, the network can be optimized with respect to the objective of face representation learning and face alignment jointly.

\begin{figure}[t]
\centering
\includegraphics[height=2.7cm]{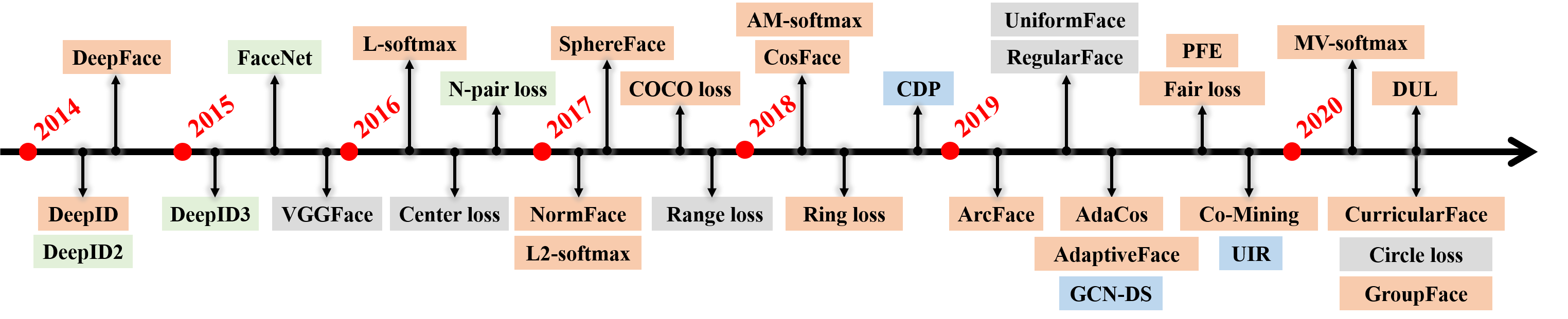}

\caption{The development of training supervision for face representation learning. The orange,  green, gray and blue represent classification, feature embedding, hybrid, and semi-supervised methods, respectively. One can refer to Table~\ref{fr_class} for the detailed references.}
\label{Development_re}

\end{figure}

\subsection{Training Supervision}
\label{sec:face_representation:supervision}

Besides network architectures, the training supervision also plays a key role for learning face representation. 
The objective of supervision for face representation learning is to encourage the faces of same identity to be close and those of different identities to be far apart in the feature space.

Following the convention of representation learning, we categorize the existing methods of training supervision for face representation into supervised scheme, semi-supervised scheme, and unsupervised scheme. 
Although there are certain deep unsupervised learning methods~\cite{shi2018face,lin2018deep,wang2019linkage,GuoDensityAwareFE} for face clustering, in this review, we focus on the supervised and semi-supervised ones which comprise the major literature of state-of-the-art face recognition.
Fig.\ref{Development_re} shows the development of training supervision for face representation learning. 
In the supervised scheme, we can further categorize the existing works into three subsets, \textit{i.e.,} classification, feature embedding and hybrid methods. The classification methods accomplish face representation learning with a $N$-way classification objective, regarding each of the $N$ classes as an identity.
The feature embedding methods aim to optimize the feature distance between samples with respect to the identity label, which means maximizing the inter-person distance and minimizing the intra-person distance. 
Besides, several works employ both classification and feature embedding routine to jointly train the network, namely hybrid methods.
As for the semi-supervised scheme, several works exploit the labeled and unlabeled faces for representation learning.

\subsubsection{Classification scheme}
The classification based deep face representation learning is derived from the general object classification task. Each class corresponds to an identity that contains a number of faces of the same person.
The softmax loss function is the most widely used supervision for classification task, which consists of a fully-connected (FC) layer, the softmax function and the cross-entropy loss.
For face representation learning, DeepFace~\cite{taigman2014deepface} and DeepID~\cite{Sun2014DeepID} are the pioneers of utilizing softmax to predict the probability over a large number of identities of training data. Their training loss function can be formulated as follows:

\begin{equation} \mathcal{L}= -\frac{1}{N} \sum_{i=1}^{N} \log \frac{e^{W_{y_{i}}^{T} {x}_{i}+b_{y_{i}}}}{\sum_{j=1}^{c} e^{W_{j}^{T} {x}_{i} + b_{j}}},
\end{equation}
where $N$ is the batch size, $c$ is the number of classes (identities), $y_i$ is the ground-truth label of sample $x_i$, $W_{y_{i}}$ is the ground-truth weight vector of sample $x_{i}$ in the FC layer, and $b_{j}$ is the bias term. The term inside the logarithm is the predicted probability on the ground-truth class. The training objective is to maximize this probability. Based on the softmax loss function, NormFace~\cite{wang2017normface} and COCO loss~\cite{Liu2017RethinkingFD} study the necessity of the normalization operation and apply $L_{2}$ normalization constraint on both features and weights with omitting the bias term $b_{j}$. 
To effectively train with the normalized features, a scale factor is adopted to re-scale the cosine similarity between the features and the weights. 
Specifically, the normalized softmax loss function can be reformulated as

\begin{equation}\mathcal{L}=-\frac{1}{N} \sum_{i=1}^{N} \log \frac{e^{s\cos(\theta_{y_{i}})}}{e^{s\cos(\theta_{y_{i}})}+\sum_{j=1, j \neq y_{i}}^{c} e^{s \cos \theta_{j}}},\end{equation}
where $\cos(\theta_{j})$ derives from the inner product ${W_{j}^{T}x_{i}}$ with the $L_{2}$ normalization on weights $W_j = \frac{W_j}{\|{W_{j}}\|_{2}}$ and features $x_i = \frac{x_i}{\|{x_i}\|_{2}}$, and $s$ is the scale parameter.

To further improve the intra-class compactness and inter-class separateness, 
L-softmax~\cite{liu2016large} replaces the ground-truth logit $\cos \left( \theta_{y_{i}}\right)$ with $ (-1)^{k} \cos (m \theta_{y_{i}})-2k , \theta_{y_{i}} \in\left[\frac{k \pi}{m}, \frac{(k+1) \pi}{m}\right]$, 
where $m$ is the angular margin that being a positive integer, and $k$ is also an integer that $k \in[0, m-1]$.
Similar to L-softmax, SphereFace~\cite{liu2017sphereface} applies an angular margin in the ground-truth logit $\cos \left( \theta_{y_{i}}\right)$ to make the learned face representation to be more discriminative on a hypersphere manifold. 
However, the multiplicative angular margin in $\cos \left( m \theta_{y_{i}}\right)$ leads to potentially unstable convergence during the training. To overcome the problem, AM-softmax~\cite{wang2018additive} and CosFace~\cite{wang2018cosface} present an additive margin penalty to the logit, $\cos \left( \theta_{y_{i}}\right) + m_{1} $, which brings more stable convergence. 
Subsequently, ArcFace~\cite{deng2019arcface} introduces an  additive angular margin inside the cosine, $\cos \left( \theta_{y_{i}} + m_{2}\right)$, which corresponds to the geodesic distance margin penalty on a hypersphere manifold. The following is a unified formulation of AM-softmax, CosFace, and ArcFace: 
\begin{equation}\mathcal{L}=-\frac{1}{N} \sum_{i=1}^{N} \log \frac{e^{s\left(\cos \left(\theta_{y_{i}}+m_{2}\right)+m_{1}\right)}}{e^{s\left(\cos \left( \theta_{y_{i}}+m_{2}\right)+m_{1}\right)}+\sum_{j=1, j \neq y_{i}}^{c} e ^{s \cos \theta_{j}}},\end{equation}
where $m_{1} < 0$ represents the additive cosine margin of AM-softmax and CosFace, and $m_{2} > 0$ denotes to the additive angular margin of ArcFace. They are easy to be implemented and can achieve better performance than the original softmax loss. 
Going further with the margin based supervision, 
AdaptiveFace~\cite{liu2019adaptiveface} and Fair loss~\cite{Fair_Loss} propose the adaptive margin that being class-wise in the training data. The purpose is to address the imbalance distribution problem in the training dataset. 

Resorting to the advantage of hard sample mining strategy~\cite{shrivastava2016training,lin2017focal}, some approaches~\cite{Wang2019MisclassifiedVG,liu2019towards,Huang2020CurricularFaceAC} reformulate the negative (non-ground-truth) logit in softmax loss function. For example, MV-softmax~\cite{Wang2019MisclassifiedVG} proposes to re-weight the negative logit to emphasize the supervision on the mis-classified samples, and thus to improve the representation learning from the negative view. 
In addition, certain studies~\cite{zhang2019adacos,Zhang2019P2SGradRG} deeply analyze the formulation of margin-based softmax loss function from the perspective of classification probability, and propose hyperparameter-free approaches for face representation leanring.

More recently, many  methods~\cite{Hu2019NoiseTolerantPF,Zhong_2019_CVPR,wang2019co,Luo2016FaceMC,shi2020universal,Kim2020GroupFaceLL,Cao2020DomainBF,Wu_2020_CVPR,Shi2019ProbabilisticFE,Chang2020DataUL} go further with the classification supervision for face representation learning. Some of them~\cite{Hu2019NoiseTolerantPF,Zhong_2019_CVPR,wang2019co} focus on the noise-robust face representation learning, and some of the others~\cite{Luo2016FaceMC,Wu_2020_CVPR} tackle the issue of performance degradation of low-bit quantified model.  Wu~\textit{et al.}~\cite{Wu_2020_CVPR} regard the quantization error as the combination of class error and individual error, and propose a rotation-consistent margin loss to reduce the latter error which is more critical.
Besides, PFE~\cite{Shi2019ProbabilisticFE} and DUL~\cite{Chang2020DataUL} propose to take into account the data uncertainty for modeling deep face representation, preventing from the uncertainty issue  caused by low quality face images.

\subsubsection{Feature embedding scheme}
Feature embedding scheme aims to optimize the feature distance according to the label of sample pair. If the pair belong to the same identity, \textit{i.e.,} positive pair, the objective is to minimize the distance or to maximize the similarity; otherwise, \textit{i.e.,} negative pair, to maximize the distance or to minimize the similarity.
For instance, contrastive loss~\cite{yi2014learning,sun2014deep,Sun2015DeeplyLF,Sun2015DeepID3FR} direct optimizes the pair-wise distance with a margin that to encourage positive pairs to be close together and negative pairs to be far apart. The loss function to be minimized is written as
\begin{equation}
    \mathcal{L}_{c} = 
    \begin{cases}
     \frac{1}{2} \|f(x_{i})-f(x_{j})\| _{2}^{2} & \text{if } y_i = y_j, \\
     \frac{1}{2} \max (0, m_\text{d}- \|f(x_{i})-f(x_{j})\| _{2})^{2} & \text{if } y_i \neq y_j,
    \end{cases}
\end{equation}
where $y_{i}=y_{j}$ denotes $x_{i}$ and $x_{j}$ are positive pair, $y_{i}\not=y_{j}$ denotes negative pair, $f(\cdot)$ is the embedding function, and $m_\text{d}$ is the non-negative distance margin. The contrastive loss drives the supervision on all the positive pairs and those negative pairs whose distance is smaller than the margin. 

FaceNet~\cite{Schroff2015FaceNetAU} first applies the triplet loss~\cite{Schultz2003LearningAD,Weinberger2005DistanceML} to deep face representation learning. Different from contrastive loss, the triplet loss encourages the positive pairs to have smaller distance than the negative pairs with respect to a margin, 
\begin{equation}\mathcal{L}_{t}=\sum_{i}^{N}\left[\left\|f\left(x_{i}^{a}\right)-f\left(x_{i}^{p}\right)\right\|_{2}^{2}-\left\|f\left(x_{i}^{a}\right)-f\left(x_{i}^{n}\right)\right\|_{2}^{2}+m_\text{d} \right]_{+},
\end{equation}
where $m_\text{d}$ is the distance margin, $x_{i}^{a}$ denotes the anchor sample, $x_{i}^{p}$ and $x_{i}^{n}$ refer to the positive sample and negative sample, respectively. The contrastive loss and triplet loss take into account only one negative sample each time, while negative pairs are abundant in training data and deserve thorough involvement in training supervision. Therefore, N-pair loss~\cite{sohn2016improved} generalizes the triplet loss to the form with multiple negative pairs, and gained further improvement on face recognition.

Compared with the supervision of classification, feature embedding can save the parameters of the FC layer in softmax, especially when the training dataset is in large scale. But the batch size of training samples limits the effectiveness of feature embedding. To alleviate this problem, some approaches~\cite{oh2016deep,Manmatha2017SamplingMI,smirnov2017doppelganger} propose the hard sample mining strategy to exploit the effective information in each batch, which is crucial to promote the performance of feature embedding.

\subsubsection{Hybrid methods}
The hybrid methods refer to those which apply classification and feature embedding together as the supervisory signals. 
DeepID series~\cite{sun2014deep,Sun2015DeepID3FR,Sun2015DeepID3FR} utilize softmax loss and contrastive loss jointly for learning face representation.
Later, several methods~\cite{wen2016discriminative,Deng2017MarginalLF,UniformFace,zhao2019regularface} improve the feature embedding portion within the hybrid scheme, by utilizing either the intra-class or the inter-class constraints. 
Some methods~\cite{Zhang2017RangeLF,Zhong_2019_CVPR,zhu2019large} show the advantage for handling the long-tail distributed data which is a widely-existing problem in FR.  
Generally, the classification scheme works well on the head data but poorly on the tail data.
Compared with classification scheme, the feature embedding scheme is able to provide the complementary supervision on the tail data. Thus, the combination of classification and feature embedding can improve the training on long-tail distributed data. More recently, Sun~\textit{et al.}~\cite{sun2020circle} propose a circle loss from a unified perspective of the classification and embedding learning, which integrates the triplet loss with the cross-entropy loss to simultaneously learn deep features with pair-wise labels and class-wise labels.

\subsubsection{Semi-supervised scheme}
The aforementioned methods focus on supervised learning. 
Constructing labeled dataset requires much of annotation effort, while large amount of unlabeled data is easily available. Therefore, it is an attractive direction that to exploit the labeled and unlabeled data together for training deep models. For semi-supervised face representation learning, assuming the identities of unlabeled data being disjoint with the labeled data, several existing works~\cite{Zhan2018ConsensusDrivenPI,yang2019learning,Yang2020LearningTC} focus on generating the pseudo labels for unlabeled data. 
However, these methods assume non-overlapping identities between unlabeled and labeled data, which is generally impractical in real-world scenarios. 
Consequently, the unlabeled samples of overlapping identity will be incorrectly clustered as a new class by the pseudo-labeling methods. The intra-class label noise in pseudo-labeled data is another problem.
To address these issues, RoyChowdhury~\textit{et al.}~\cite{RoyChowdhury2020ImprovingFR} separates unlabeled data into samples of disjoint and overlapping classes via an out-of-distribution detection algorithm. Besides, they design an improved training loss based
on uncertainty to alleviate the label noise of pseudo-labeled data.

\subsection{Specific Face Recognition Tasks}
\label{sec:face_representation:specific_scene}

\subsubsection{Cross-domain face recognition} 
Here, the term of cross-domain refers to a generalized definition that includes various factors, such like cross-age and cross-pose FR.
As deep learning is a data-driven technique, the deep network usually works well on the training domains but poorly on the unseen ones. In real-world applications of face recognition, it is essential to improve the generalization ability of face representation across various domain factors.
In the following, we discuss certain aspects of cross-domain FR; also, the current solutions are presented.

\textbf{Cross-age}: As the facial appearance has large intra-class variation along with the growing age, identifying faces across wide range of age is a challenging task. For such cross-age FR, there are two directions. The first direction~\cite{Wen2016LatentFG,Xu2017AgeIF,Du2019AgeFR,Wang2019DecorrelatedAL,Zheng2017AgeEG,Wang2018OrthogonalDF} aims to learn age-invariant face representation by decomposing deep face features into age-related and identity-related components. The second direction is based on generative mechanism. In this way, several methods~\cite{KemelmacherShlizerman2014IlluminationAwareAP,Wang2016RecurrentFA,Antipov2017FaceAW} attempt to synthesize faces of target age, but they present imperfect preservation of the original identities in aged faces. Thus, supplementary  methods~\cite{Wang2018FaceAW,Antipov2017BoostingCF,song2018dual,Zhao2019LookAE} are designed to improve the identity-preserving ability during the face aging.

\textbf{Cross-pose}: In unconstrained conditions, such as surveillance video, the cameras cannot always capture the frontal face image for every appeared subject. Thus, the captured faces have large pose variation from frontal to profile view. 
However, generating the frontal faces will increase the burden of face recognition system. Cao~\textit{et al.}~\cite{Cao2018PoseRobustFR} alleviate this issue by transforming the representation of a profile face to the frontal view in the feature space. Another problem is that the number of profile faces are much fewer than frontal ones in the training data. Thus, some generative approaches~\cite{Zhao2017DualAgentGF,Tran2017DisentangledRL,Deng2018UVGANAF,Hu2018CAPG} propose to synthesize identity-preserving faces of arbitrary poses to enrich the training data. Moreover, certain methods~\cite{Masi2016PoseAwareFR,AbdAlmageed2016FaceRU,kan2016multi} develop multiple pose-specific deep models to compute the multi-view face representations. 

\textbf{Racial bias}: Due to the imbalance distribution of different races in training data, the deep face feature shows favorable recognition performance to the races of large proportion in training data than those of small proportion.
Recently, Wang~\textit{et al.}~\cite{Wang2019RacialFI} construct an in-the-wild face dataset (RFW) with both identity and race annotation, which consists of four racial subsets, \textit{i.e.,} Caucasian, Asian, Indian, and African. Besides, they propose a domain adaptation method to alleviate the racial bias. Later on, RL-RBN~\cite{Wang_2020_CVPR} sets a fixed margin for the large-proportion races and automatically select an optimal margin for the small-proportion races, in order to achieve balanced performance.

\textbf{Cross-modality}: Cross-modality face recognition generally refers to the heterogeneous face recognition, which performs with a pair of input face images captured by different sensing modalities, such as infrared vs. visible, or sketch vs. photo. 
How to alleviate the domain gaps between different modalities is the major challenge. Besides, 
the available infrared or sketch images are of very limited number. The existing works mainly handle these two issues. 
Many methods~\cite{Reale2016SeeingTF,Saxena2016HeterogeneousFR,Liu2016TransferringDR,Mittal2015CompositeSR,Galea2017ForensicFP,Wan2019TransferDF} exploit the transfer learning, \textit{i.e.,} pretraining on the visible-light (VIS) images and finetuning with the infrared or sketch data, to reduce the domain discrepancy. 
Another set of methods~\cite{He2017LearningID,Wu2018DisentangledVR,Deng2019MutualCC,He2019WassersteinCL} decompose the cross-modality features to the modality-specific and modality-invariant components, and use the latter one for the recognition task. Moreover, recent  methods~\cite{Lezama2017NotAO,Song2018AdversarialDH,He2020AdversarialCF,Yi2017DualGANUD,Zhu2017UnpairedIT,Wang2018HighQualityFP,Zhang2019DualTransferFS,Zhang2019SynthesisOH,Zhang2020CascadedFS} aim to synthesize the common VIS image from infrared or sketch input, and then perform the regular FR in the VIS domain.

\subsubsection{Low-shot face recognition}
Low-shot learning in face recognition focuses on the condition of identification of low-shot face IDs, each of which has a small number of samples. MS-Celeb-1M low-shot learning benchmark~\cite{guo2017one} is most used, which has about 50 to 100 training samples per ID in the base set and only one training sample per ID in the novel set. The target is to recognize the IDs in both base and novel sets. The key challenge is to correctly recognize the subjects in the novel set which has only one training sample per ID. To tackle this problem, many methods~\cite{guo2017one,Wu2017LowShotFR,smirnov2017doppelganger,cheng2017know,yin2019feature} improve the low-shot face recognition with better training supervision or strategy.  
Besides, face generation ~\cite{Choe2017FaceGF,Hong2017SSPPDANDD,Song2019UnsupervisedPI} is another effective routine for low-shot issue.

\subsubsection{Video face recognition}
The above methods focus on still image-based face recognition. For video face recognition, a common way~\cite{Chen2017UnconstrainedSF,Ding2018TrunkBranchEC} is to equally consider the importance of each frame and simply average a set of deep features as the template. However, this routine does not consider the different quality of frames and the temporal information across frames. How to obtain an optimal template feature in video is the major challenge of video face recognition. Several methods~\cite{Yang2017NeuralAN,Gong2019VideoFR,Liu2019FeatureAN,Gong2019LowQV} aggregate the frame-level features with the attention weights or quality scores. 
Synthesizing representative or high-quality face image from a video sequence is another possibility~\cite{Rao2017LearningDA,Parchami2017UsingDA}. Additionally, certain methods~\cite{Rao2017AttentionAwareDR,Mei2008VideoCP,Liu2018DependencyAwareAC} model the temporal-spatial information with the attention mechanism and find the focus of video frames.

\subsection{Evaluation Metrics and Datasets}
\label{sec:face_representation:evaluation}

\subsubsection{Metrics}
The performance of face recognition is usually evaluated on two tasks: verification and identification, each of which has its corresponding evaluation metrics. Specifically, two sets of samples, \textit{i.e.,} gallery and probe, are required for the evaluation. The gallery refers to a set of faces registered in the face recognition system with known identities, while the probe denotes a set of faces need to be recognized in verification or identification. 
Before discussing the commonly used evaluation metrics, we first introduce some basic concepts. A face recognition system determines whether to accept the matching of a probe face and a gallery face by comparing their similarity, computed by some measurement between their features, with a given threshold. Specifically, when a probe face and a gallery face are the same identity, a true acceptance (TA) means their similarity is above the threshold, and a false rejection (FR) represents their similarity is below the threshold; if they are different identities, a true rejection (TR) means their similarity is below the threshold, and a false acceptance (FA) means their similarity is above the threshold. 
These are the basic concepts to build the evaluation metrics in the followings. 
One can refer to~\cite{Grother2003FaceRV,Grother2014Face} for more details.

\textbf{Verification task:}
Face verification is often applied in identity authentication system, which measures the similarity of face pairs. One presents his or her face and claims the enrolled identity in the gallery. Then, the system determines whether it accepts the person being the same one of the claimed identity by calculating the similarity between the presented face and the claimed face. 
Thus, the verification task can be regarded as a one-to-one face matching process. The false accept rate (FAR) and true accept rate (TAR) are used to evaluate the verification performance. 
FAR is the fraction of impostor pairs with the similarity above the threshold, which can be calculated by $\frac{FA}{FA+TR}$; TAR represents the fraction of genuine pairs with the similarity above the threshold, which can be calculated by $\frac{TA}{TA+FR}$. Then, by varying the threshold, the ROC curve can be drawn by many operating points, each of which is determined by a pair of TAR vs. FAR.
The ROC curve (with TAR value at selected FAR) and its AUC (\textit{i.e.,} area under curve) are widely used to evaluate the performance for the face verification task.

\textbf{Identification task:}
Face identification task determines whether a probe face belongs to a enrolled identity in the gallery set. To this end, the probe face needs to be compared with every person in the gallery set. Thus, the identification task can be also referred as one-to-$N$ face matching. 

Generally, face identification includes two tasks, \textit{i.e.,} the open-set and closed-set identification. 
The open-set identification task refers to that the probe face is not necessarily the very identity contained in the gallery set, which is the most general case in practice. 
The true positive identification rate (TPIR) and false positive identification rate (FPIR) are the most used metrics for the following two situations. 
The first situation refers to that the probe corresponds to an enrolled identity in the gallery set. This situation is called mate searching, and the probe is called mate probe.
The succeeded mate searching represents that the rank of true matching is higher than the target rank, and meanwhile its similarity is above the threshold.
In such case, the mate probe is correctly identified as its true identity, and the mate searching is measured by the TPIR which represents the proportion of succeeded trials of mate searching. 
The second is non-mate searching, in which the probe does not correspond to any enrolled identity (\textit{i.e.,} non-mate probe). 
The non-mate searching is measured by the FPIR which reports the proportion of non-mate probes wrongly identified as enrolled identity. 
By fixing the rank and varying the threshold, the ROC curve can be drawn by many operating points, each of which is determined by a pair of TPIR vs. FPIR. The ROC curve (TPIR value at a given FPIR) is used to evaluate performance in the open-set face identification task. 

In the closed-set scenario, the identity of each probe face is included in the gallery set. 
The cumulative match characteristic (CMC) curve is used for evaluating the closed-set face identification. The CMC curve is drawn by the operating points that are determined by a pair of identification rate vs. rank. 
The identification rate refers to the fraction of probe faces that are correctly identified as the true identities, thus the CMC curve reports the fraction of the true matching with a given rank, and the identification rate at rank one is the most commonly used indicator of performance. It is noteworthy that the CMC is a special case of the TPIR when we relax the threshold. 

\begin{table}[t]
\begin{center}
\caption{The commonly used public datasets for training and testing deep face recognition.}

\label{fr_data}
\resizebox{1\linewidth}{!}{
\begin{tabular}{|c|c|c|c|c|c|}
\hline
{Dataset}&{Year}&{$\#$ Subject}&{$\#$ Image/Video}&{$\#$ of Img/Vid per Subj}&{Description}\\
\hline\hline
\multicolumn{6}{|c|}{Training}\\
\hline
CASIA-WebFace~\cite{yi2014learning}&2014&10,575&494,414/-&47&The first public large-scale face dataset\\
\hline
VGGFace~\cite{Parkhi2015DeepFR}&2015&2,622&2.6M/-&1,000&Containing large number of images in each subject \\
\hline
CelebA~\cite{liu2015faceattributes}&2015&10,177 &202,599/-&20& Rich annotations of attributes and identities \\
\hline
UMDFaces~\cite{Bansal2017UMDFacesAA}&2015&8,277&367K/-&45& Abundant variation of facial pose\\
\hline
MS-Celeb-1M~\cite{guo2016ms}&2016&100K&10M/-&100&A large-scale public dataset of celebrity faces\\
\hline
MegaFace~\cite{kemelmacher2016megaface,Nech2017LevelPF}&2016&672,057&4.7M/-&7&A long-tail dataset of non-celebrity \\
\hline
VGGFace2~\cite{Cao2018VGGFace2AD}&2017&9,131&3.31M/-&363&A high-quality dataset with a wide range of variation\\
\hline
UMDFaces-Videos~\cite{Bansal2017TheDA}&2017&3,107&-/22,075&7&A video  training dataset collected from YouTube\\
\hline
MS-Celeb-1M Low-shot~\cite{guo2017one}&2017&20K,1K&1M,1K/-&58,1&Low-shot face recognition\\
\hline
IMDb-Face~\cite{Wang2018TheDO}&2018&57K&1.7M/-&29&A large-scale noise-controlled dataset\\
\hline
QMUL-SurvFace~\cite{Wang2018TheDO}&2018&5,319&220,890/-&41&A low-resolution surveillance dataset\\
\hline
Glint360K~\cite{an2020partical_fc}&2021&360K&17M/-&47&A  large-scale and cleaned dataset\\
\hline 
WebFace260M~\cite{Zhu2021WebFace260MAB}&2021&4M&260M/-&65&The largest public dataset of celebrity faces\\

\hline\hline
\multicolumn{6}{|c|}{Test}\\
\hline
LFW~\cite{LFWTech}&2007&5,749&13,233/-&2.3& A classic benchmark in unconstrained conditions\\
\hline
YouTube Faces (YTF)~\cite{Wolf2011FaceRI}&2011&1,595&-/3,425&2.1&Face recognition in unconstrained videos\\
\hline
CUFSF~\cite{Zhang2011CoupledIE}&2011&1,194&2,388/-&2&Photo-sketch face recognition\\
\hline
CASIA NIR-VIS v2.0~\cite{Li2013TheCN}&2013&725&17,580/-&24.2&Near-infrared vs. RGB face recognition\\
\hline
IJB-A~\cite{Klare2015PushingTF}&2015&500&5,712/2,085&11.4/4.2& Set-based face recognition with large variation\\
\hline
CFP~\cite{sengupta2016frontal}&2016&500&7,000/-&14&Frontal to profile cross-pose face verification\\
\hline
MS-Celeb-1M Low-shot~\cite{guo2017one}&2016&20K,1K&100K,20K/-&5,20&Low-shot face recognition\\
\hline
MegaFace~\cite{kemelmacher2016megaface,Nech2017LevelPF}&2016&690,572&1M/-&1.4&A large-scale benchmark with one million faces\\
\hline
IJB-B~\cite{Whitelam2017IARPAJB}&2017&1,845&11,754/7,011&6.37/3.8&Set-based face recognition with full pose variation \\
\hline
CALFW~\cite{zheng2017cross}&2017&4,025&12,174/-&3& Cross-age face verification\\
\hline 
AgeDB~\cite{Moschoglou2017AgeDBTF}&2017&570&16,516/-&29& Cross-age face verification\\
\hline
SLLFW~\cite{deng2017fine}&2017&5,749&13,233/-&2.3& Improving the difficulty of negative pairs in LFW\\
\hline
CPLFW~\cite{zheng2018cross}&2017&3,968&11,652/-&2.9&Cross-poss face verification\\
\hline
Trillion Pairs~\cite{trillionpairs.org}&2018&1M&1.58M/-&1.6&A large-scale benchmark with massive distractors\\
\hline
IJB-C~\cite{Maze2018IARPAJB}&2018&3,531&31,334/11,779&6/3&Set-based face recognition with large variation\\
\hline
IJB-S~\cite{Kalka2018IJBSIJ}&2018&202&5,656/552&28/12&Real-world surveillance videos \\
\hline
RFW~\cite{Wang2019RacialFI}&2018&11,429&40,607/-&3.6& For reducing racial bias in face recognition\\
\hline
DFW~\cite{Kushwaha2018DisguisedFI}&2018&600&7,771/-&13&Disguised face recognition\\
\hline
QMUL-SurvFace~\cite{Wang2018TheDO}&2018&10,254&242,617/-&23.7&Low-resolution surveillance videos\\
\hline

\end{tabular}}
\end{center}

\end{table}

\subsubsection{Datasets}
With the development of deep face recognition, another key role to promote face representation learning is the growing datasets for training and test. In the past few years, the face datasets have become large scale and diverse, and the testing scene has been approaching to the real-world unconstrained condition. The statistics of them are presented in Table~\ref{fr_data}.

\textbf{Training data:}
Large-scale training datasets are essential for learning deep face representation. The early works often employ the private face datasets, such as Deepface~\cite{taigman2014deepface}, FaceNet~\cite{Schroff2015FaceNetAU}, DeepID~\cite{sun2014deep}. 
To make it possible for fair comparison, Yi~\textit{et al.}~\cite{yi2014learning} release the CASIA-WebFace dataset, which has been one of the most widely-used training datasets. 
Afterward, more public training datasets are published to provide abundant face images for training deep face model. Among them, VGGFace~\cite{Parkhi2015DeepFR} and VGGFace2~\cite{Cao2018VGGFace2AD} contain many training samples for each subject.
In contrast, MS-Celeb-1M~\cite{guo2016ms}, MegaFace~\cite{kemelmacher2016megaface},  IMDb-Face~\cite{Wang2018TheDO} and WebFace260M~\cite{Zhu2021WebFace260MAB} provide a large number of subjects with relatively less training samples per subject. 

\begin{table}[t]
\begin{center}
\caption{Performance ($\%$) comparison of face recognition on various test datasets. ``Training Data'' denotes the number of training face images used by the methods. For the evaluation on MegaFace, ``Id.'' refers to the rank-1 face identification accuracy with 1M distractors, and ``Veri.'' refers to the face verification TAR at $10^{-6}$ FAR. For the evaluation on IJB-B and IJB-C, we report the 1:1 verification TAR (@FAR=$10^{-4}$). The performance with ``$^*$'' refers to the evaluation on the refined version of MegaFace~\cite{deng2019arcface}. ``-'' indicates that the authors do not report the performance with the corresponding protocol. }
\label{performance_fr}
    
    \resizebox{1\linewidth}{!}
    {
    \begin{tabular}{|r|c|c|c|c|c|c|c|c|c|c|c|c|c|c|}
    \hline
    \multirow{2}{*}{Method}&\multirow{2}{*}{Publication}&\multirow{2}{*}{Subcategory}&\multirow{2}{*}{Training Data}&\multirow{2}{*}{Backbone}&\multirow{2}{*}{LFW}&\multicolumn{2}{c|}{MegaFace}&\multirow{2}{*}{IJB-B}&\multirow{2}{*}{IJB-C}&\multirow{2}{*}{YTF}&\multirow{2}{*}{CALFW}&\multirow{2}{*}{CPLFW}&\multirow{2}{*}{CFP-FP}&\multirow{2}{*}{AgeDB30} \\
    \cline{7-8}&&&&&&Id.&Veri.&&&&&&&\\
  \hline\hline
    DeepFace~\cite{taigman2014deepface}&CVPR'14&Classification&4M&CNN-8&97.35&-&-& -& -& 91.4& -& -& -& -\\
    \hline
    DeepID~\cite{taigman2014deepface}&CVPR'14&Classification&0.3M&CNN-8&97.45&-&-& -& -& -& -& -& -& -\\
    \hline
    L-Softmax~\cite{liu2016large}&ICML'16&Classification&0.5M&VGGNet-18&99.10&67.12&80.42& -& -& -& -& -& -& -\\   
    \hline
    NormFace~\cite{wang2017normface}&ACMMM'17&Classification&0.5M&ResNet-28&99.16& -& -& -& -& -& -& -& -& \\   
    \hline
    SphereFace~\cite{liu2017sphereface}&CVPR'17&Classification&0.5M&ResNet-64&99.42&72.72&85.56& -& -& 95.0& -& -& -& -\\ 
    \hline
    ReST~\cite{liu2017sphereface}&CVPR'17&Classification&0.5M&CNN-9&99.03&65.16& -& -& -& 95.4& -& -& -& -\\ 
    \hline
    E2e~\cite{Zhong2017e2e}&SPL'17&Classification&0.7M&ResNet-27&99.33&-& -& -& -& 95.0& -& -& -& -\\
    \hline
    AM-softmax~\cite{wang2018additive}&SPL'18&Classification&0.5M&ResNet-20&98.98&72.47&84.44& -& -& -& -& -& -& -\\  
        \hline
    CosFace~\cite{wang2018cosface}&CVPR'18&Classification&5M&ResNet-64&99.73&82.72&96.65& -& -& 97.6& -& -& -& -\\  
    \hline
     ComparatorNet~\cite{Xie2018ComparatorN}&ECCV'18&Classification&3.3M&ResNet-50&-&-&-&84.1&88.0&-&-&-&-&-\\
     \hline
    ArcFace~\cite{deng2019arcface}&CVPR'19&Classification&0.5M&ResNet-50&99.53&77.50&92.34& -& -& -& -& -& 95.56& 95.15\\  
    \hline
    Fair loss~\cite{Fair_Loss}&ICCV'19&Classification&0.5M&ResNet-50&99.57&77.45&92.87& -& -& 96.2& -& -& -& -\\
    \hline
    PFE~\cite{Shi2019ProbabilisticFE}&ICCV'19&Classification&4.4M&ResNet-64& 99.82&\textbf{78.95}&92.51& -& 93.25& -& -& -& 93.34& -\\ 
    \hline  
    FANFace~\cite{Yang2020FANFaceAS}&AAAI'20&Classification &0.5M&ResNet-50&99.56&78.32&92.83& -& -& 96.72& -& -& -& -\\
    \hline
    TURL~\cite{shi2020universal}&CVPR'20&Classification &4.8M&ResNet-100&99.78&78.60&\textbf{95.04}& -& 96.6& -& -& -& 98.64& -\\
    \hline
      RDCFace~\cite{Zhao_2020_CVPR}&CVPR'20&Classification &  1.7M& ResNet-50&  99.80&-&-& -& -& 97.10& -& -& 96.62& -\\
     \hline 
    AdaCos~\cite{zhang2019adacos}&CVPR'19&Classification &2.35M&ResNet-50&99.73&97.41$^*$&-& -& 92.4& -& -& -& -& -\\
    \hline
    P2SGrad~\cite{Zhang2019P2SGradRG}&CVPR'19&Classification&2.35M&ResNet-50&99.82&97.25$^*$&-& -& 92.3& -& -& -& -& -\\
    \hline
    AdaptiveFace~\cite{liu2019adaptiveface}&CVPR'19&Classification&5M&ResNet-50&99.62&95.02$^*$&95.61$^*$& -& -& -& -& -& -& -\\
    \hline
    ArcFace~\cite{deng2019arcface}&CVPR'19&Classification&5.8M &ResNet-100&99.82 &98.35$^*$&98.48$^*$& 94.2& 95.6& 97.7& 95.45& 92.08& 98.27& 98.15\\
        \hline
    MV-AM-softmax~\cite{Wang2019MisclassifiedVG}&AAAI'20&Classification&3.2M&Attention-56&99.79&98.00$^*$&98.31$^*$& -& -& -& 95.63& 89.19& 95.30& 98.00\\
    \hline
    DUL~\cite{Chang2020DataUL}&CVPR'20&Classification&3.6M&ResNet-64&99.83&98.12$^*$&-& -& 94.21& 96.84& -& -& \textbf{98.78}& -\\
    \hline
    DB~\cite{Cao2020DomainBF}&CVPR'20&Classification&5.8M&ResNet-50&99.78&96.35$^*$&96.56$^*$&-& -& -& 96.08& 92.63& -& 97.90\\
    \hline
    CurricularFace~\cite{Huang2020CurricularFaceAC}&CVPR'20&Classification&5.8M&ResNet-100&99.80 &98.71$^*$&98.64$^*$& 94.8& 96.1& -& \textbf{96.20}& 93.13& 98.37& \textbf{98.32}\\
    \hline
    GroupFace~\cite{Kim2020GroupFaceLL}&CVPR'20&Classification& 5.8M &ResNet-100&\textbf{99.85} &\textbf{98.74}$^*$&\textbf{98.79}$^*$& \textbf{94.93}& \textbf{96.26}& \textbf{97.8}& \textbf{96.20}& \textbf{93.17}& 98.63& 98.28\\
  
    \hline \hline
    FaceNet~\cite{Schroff2015FaceNetAU}&CVPR'15&Embedding&400M&GoogleNet-22&99.63&-&-& -& -& 95.1& -& -& -& -\\
    \hline
    VGG Face~\cite{Parkhi2015DeepFR}&BMVC'15&Embedding&2.6M&CNN-36&98.95&\textbf{64.79}&\textbf{78.32}&-&-& \textbf{97.3}& -& -& -& -\\
         \hline
    {N-pair loss~\cite{sohn2016improved}}&NIPS'16&Embedding&0.5M&CNN-10&98.50&-&-& -& -& -& -& -& -& -\\
    \hline
    GridFace~\cite{Zhou2018GridFaceFR}&ECCV'18&Embedding&10M&GoogLeNet-22&\textbf{99.70}&-&-& -& -& 95.6& -& -& -& -\\
    \hline
    \hline
    DeepID2~\cite{sun2014deep}&NeurIPS'14&Hybrid&0.3M&CNN-8&99.15&65.21&78.86&-&-& -& -& -& -& -\\
    \hline
     SparseNet~\cite{sun2015sparsifying}&CVPR'15&Hybrid& 0.3M& CNN-15& 99.30& -& -& -& -& 92.7& -& -& -& -\\
         \hline
    Center loss~\cite{wen2016discriminative}&ECCV'16&Hybrid&0.7M&CNN-11&99.28&65.49&80.14& -& -& 94.9& -& -& -& -\\  
         \hline
    Ring loss~\cite{zheng2018ring}&CVPR'18&Hybrid&3.5M&ResNet-64&99.50&74.93&-& -& -& 93.7& -& -& -& -\\
        \hline
    PRN~\cite{Kang2018PairwiseRN}&ECCV'18&Hybrid&2.8M&ResNet-101&99.76&-&-& 84.5& -& 96.3& -& -& -& -\\
     \hline
        RegularFace~\cite{zhao2019regularface}&CVPR'19&Hybrid&3.1M&ResNet-20&99.61&75.61&91.13& -& -& 96.7& -& -& -& -\\  
    \hline
    UniformFace~\cite{UniformFace}&CVPR'19&Hybrid&3.8M&ResNet-34&99.8&\textbf{79.98}&\textbf{95.36}& -& -& \textbf{97.7}& -& -& -& -\\
        \hline
    AFRN~\cite{Kang2019AttentionalFR}&ICCV'19&Hybrid&2.8M&ResNet-101&\textbf{99.85}&-&-& \textbf{88.5}& 93.0& 97.1& \textbf{96.30}& \textbf{93.48}& 95.56& \textbf{96.35}\\
     \hline 
    Circle loss~\cite{sun2020circle}&CVPR'20&Hybrid&3.6M&ResNet-34&99.73&\textbf{97.81}
    $^*$& -& -& \textbf{93.44}& 96.38& -& -& \textbf{96.02}& -\\
    \hline
    \end{tabular}}
\end{center}

\end{table}

\textbf{Test data:}
As for testing, Labeled Faces in the Wild (LFW)~\cite{LFWTech} is classic and the most widely used benchmark for face recognition in unconstrained environments. The original protocol of LFW contains 3,000 genuine and 3,000 impostor face pairs, and evaluates the mean accuracy of verification on these 6,000 pairs. So far, the state-of-the-art accuracy has been saturated on LFW, whereas the total samples in LFW are more than those in the original protocol. Based on this, BLUFR~\cite{Liao2014ABS} exploits all the face images in LFW for a large-scale unconstrained face recognition evaluation; SLLFW~\cite{deng2017fine} replaces the negative pairs of LFW with more challenging ones. In addition, CFP~\cite{sengupta2016frontal}, CPLFW~\cite{zheng2018cross}, CALFW~\cite{zheng2017cross}, AgeDB~\cite{Moschoglou2017AgeDBTF} and RFW~\cite{Wang2019RacialFI} utilize the similar evaluation metric of LFW to test face recognition with various challenges, such as cross pose, cross age and multiple races. 
MegaFace~\cite{kemelmacher2016megaface,Nech2017LevelPF} and Trillion Pairs~\cite{trillionpairs.org} focus on the performance at the strict false accept rates (\textit{i.e.,} $10^{-6}$ and $10^{-9}$) on face verification and identification with million-scale distractors. 
The above datasets focus on image-to-image face recognition, whereas YTF~\cite{Wolf2011FaceRI}, IJB series~\cite{Klare2015PushingTF,Whitelam2017IARPAJB,Maze2018IARPAJB,Kalka2018IJBSIJ}, and QMUL-SurvFace~\cite{Wang2018TheDO} serve as the evaluation benchmark of video-based face recognition. Especially, IJB-S and QMUL-SurvFace are constructed from real-world surveillance videos, which are much more difficult and realistic than the tasks on still images. 

\subsection{Performance Comparison}
Table~\ref{performance_fr} shows the performance of face representation methods on various test datasets.
Among them, CosFace~\cite{wang2018cosface} and ArcFace~\cite{deng2019arcface} are the two commonly used methods in many applications of face recognition. 
In addition, with the growing datasets for training and test, the closed-set classification training on the large-scale datasets enables to approach the open-set face recognition scenario. 
This could be the reason why the classification based training methods have been widely studied and dominated the state-of-the-art performance in recent years.
One can find the publication trend of three supervised training schemes with the increasing scale of public face datasets in the supplementary material.

\section{Discussion and Conclusion}

Deep face recognition still remains a number of issues for each element. In the following, we first analyze the major challenges towards end-to-end deep face recognition and the subcategories of each element. Then, we provide a detailed discussion about the promising future trends for each element and the entire system. Finally, the conclusion of this survey is presented.

\subsection{Challenge}
\label{sec:discussion}

\begin{table}[t] 

\begin{center}
\caption{{Summary of the major challenges towards end-to-end deep face recognition.}}
\label{conclusion_challenges}

\resizebox{1\linewidth}{!}{
\begin{tabular}{|p{1cm}|p{4.5cm}|p{8.5cm}|}
\hline
\multicolumn{2}{|c|}{Challenges}
&\multicolumn{1}{|c|}{Description}
\\
\hline

\multicolumn{1}{|l|}{ The issues of each element.} 

& \vspace{-3pt}  Face detection \vspace{-6pt}
& 
\vspace{-7pt}
\begin{itemize}[leftmargin=*]
 \item  Trade-off between detection accuracy and efficiency.
 \item Accuracy of the bounding box location. 
 \item Detecting faces with a wide range of scale. 
\end{itemize}
\vspace{-12pt}
\\

\cline{2-3}
& \vspace{-3pt}  Face alignment  \vspace{-6pt}
& 
\vspace{-7pt}
\begin{itemize}[leftmargin=*] 
 \item  Annotation ambiguity and granularity.
\end{itemize}
\vspace{-12pt}
\\

\cline{2-3}
& \vspace{-3pt}  Face representation  \vspace{-6pt}
& 
\vspace{-7pt}
\begin{itemize}[leftmargin=*]
 \item limited training data and computational budget.
\item Surveillance video face recognition.
 \item Noisy label and imbalance data.

\end{itemize}
\vspace{-12pt}
\\
\hline
\multicolumn{1}{|l|}{The common issues across the elements.} 
& Facial / image variations 
& 
\vspace{-7pt}
\begin{itemize}[leftmargin=*]
\item Large pose, extreme expression, occlusion,  facial scale. 
\item Motion blur, low illumination, low resolution. 
\end{itemize}
\vspace{-12pt}
\\

\cline{2-3}
& Data / label distribution 
& 
\vspace{-7pt}
\begin{itemize}[leftmargin=*]
\item Limited labeled data, label noise.
\item Usage of unlabeled data.
\item Imbalance over scale, identity, race, domain, modality.
\end{itemize}
\vspace{-12pt}
\\

\cline{2-3}
& Computational efficiency 
& 
\vspace{-7pt}
\begin{itemize}[leftmargin=*]
\item Inference on non-GPU device and edge computing.
\item Fast training and convergence. 
\end{itemize}
\vspace{-12pt}
\\

\hline

\multicolumn{1}{|l|}{The issues concerning to the entire system.} 
& Interpretability
& 
\vspace{-7pt}
\begin{itemize}[leftmargin=*]
\item Explainable learning and inference. 
\end{itemize}
\vspace{-12pt}
\\
\cline{2-3}

& Joint modeling and optimization 
& 
\vspace{-7pt}
\begin{itemize}[leftmargin=*]
\item End-to-end training and inference.
\item Unified learning objective.
\item Mutual promotion. 
\end{itemize}
\vspace{-12pt}
\\
\cline{2-3}
&\vspace{-3pt}  Universal pretraining
& 
\vspace{-7pt}
\begin{itemize}[leftmargin=*]
 \item Universal pretrained facial representation.
\end{itemize}
\vspace{-12pt}
\\
\cline{2-3}
& \vspace{-3pt}  Trustworthiness
& 
\vspace{-7pt}
\begin{itemize}[leftmargin=*]
 \item  Robustness, fairness, explainability, security, and privacy.
\end{itemize}
\vspace{-12pt}
\\
\hline
\end{tabular}}
\end{center}

\end{table}

The top rows of Table~\ref{conclusion_challenges} elaborate the issues of each element. For face detection, the state-of-the-art methods are eager for  trade-off between detection accuracy and efficiency. 
For example, in many applications, resizing the input image is a common practice of acceleration for detectors, while it harms the recall of tiny faces as well. 
In the unconstrained condition, human faces with large variation tend to be missed by detectors, whereas the diverse image background often leads to false positives. 
Besides, detecting faces with a wide range of scale is also a great challenge.
As for the face alignment procedure, the facial landmark localization methods are still not robust enough when working with extreme variations, such as severe occlusion, large pose, low illumination. 
In addition, the annotation ambiguity, such as the landmarks on cheek, is a common problem in datasets. Besides, most of the existing facial landmark datasets provide the annotation of 68 or 106 points. More landmark points enable to depict the abundant facial structure. 
For the face representation learning, although existing methods achieve high accuracy on various benchmarks, it is still challenging when training data and computational budget are very limited. In addition, surveillance face recognition is a common scenario, where the challenges include various facial variations, such large poses, motion blur, occlusion, low illumination and resolution, \textit{etc}. Imbalance distribution of training data also brings issues to the face representation learning, such as long-tail distribution over face identities or domains.

The middle rows of Table~\ref{conclusion_challenges} elaborate the common issues shared between face detection, alignment and representation. 
We can find that the issues mainly include three aspects, \textit{i.e.,} facial and image variations, data and label distribution, and computational efficiency. For example, in the first aspect, the facial variations include large facial pose, extreme expression, occlusion and facial scale, while the image variations include the objective factors such as motion blur, low illumination and resolution which occur frequently in video face recognition. 
Another example indicates the need of training efficiency, including fast training and convergence, both of which devote to accelerating the learning of large face representation network (hundreds of layers normally) from weeks to hours; the former generally focuses on the mixed precision training or the distributed framework for large-scale training (over millions of identities), while the latter focuses on improving the supervision, initialization, updating manner, activation, architectures, \textit{etc}. 
Here, rather than replaying every detail, we leave Table~\ref{conclusion_challenges} to readers for exploring the common challenges and further improvement. 
It is worth mentioning that all the elements will benefit from the solutions against these issues, since they are the common issues across the elements.

The bottom of Table~\ref{conclusion_challenges} indicates the major challenges from the perspective of entire system. 
For instance, ideally, the three elements should be jointly modeled and optimized with respect to the end-to-end accuracy.
On the one hand, such integration provides a possibility to search global optimal solution for the holistic system; on the other hand, the individual elements of the system can benefit from the upstream ones. 
However, the elements have different learning objectives regarding to their own tasks. 
How to unify these learning objectives is a challenging and critical issue for the joint optimization.
One can find a group of works~\cite{mtcnn,deng2020retinaface,Wu2017ReST,Zhao_2020_CVPR,Hayat2017JointRA,Zhong2017e2e,Zhou2018GridFaceFR,Zhao2018pim,Wei2020BalancedAF} attempting to integrate face detection and alignment, or face alignment and representation for a joint boost.
But face detection is still difficult to be integrated with face representation because they have quite different objectives and implementation mechanisms. 

\begin{table*}[t] 
\begin{center}
\caption{{Summary of the major challenges towards the subcategories for each element.}}
\label{conclusion_challenges1}

\resizebox{1\linewidth}{!}{
\begin{tabular}{|p{2cm}|p{6cm}|p{10cm}|}
\hline
\multicolumn{1}{|c|}{Element}
&\multicolumn{1}{c|}{Subcategory}
&\multicolumn{1}{c|}{Challenges Description}
\\
\hline

\multicolumn{1}{|l|}{Face detection} 

&\vspace{-3pt} {Multi-stage}
& 
\vspace{-7pt}
\begin{itemize}[leftmargin=*]
 \item Runtime efficiency.
\end{itemize}
\vspace{-12pt}
\\
\cline{2-3}
& \vspace{-3pt}  Single-stage
& 
\vspace{-7pt}
\begin{itemize}[leftmargin=*]
 \item Detecting tiny faces. 
\end{itemize}
\vspace{-12pt}
\\

\cline{2-3}
&\vspace{-3pt}  Anchor-based
& 
\vspace{-7pt}
\begin{itemize}[leftmargin=*]
 \item Well-tuned anchors.
\end{itemize}
\vspace{-12pt}
\\
\cline{2-3}
& \vspace{-3pt}  Anchor-free
& 
\vspace{-7pt}
\begin{itemize}[leftmargin=*]
 \item Training stability and robustness to false positives.   
\end{itemize}
\vspace{-12pt}
\\
\cline{2-3}
&\vspace{-3pt}  CPU real-time
& 
\vspace{-7pt}
\begin{itemize}[leftmargin=*]
 \item Trade-off between accuracy and efficiency. 
\end{itemize}
\vspace{-12pt}
\\

\cline{2-3}
&\vspace{-3pt}  Multi-task learning
& 
\vspace{-7pt}
\begin{itemize}[leftmargin=*]
 \item Balance of multi-task training supervision.
\end{itemize}
\vspace{-12pt}
\\

\cline{2-3}
& \vspace{-3pt}  Problem-oriented
& 
\vspace{-7pt}
\begin{itemize}[leftmargin=*]
 \item Low-illumination and low-resolution.
\end{itemize}
\vspace{-12pt}
\\

\hline
\multicolumn{1}{|l|}{ Face alignment} 
&\vspace{-3pt}  Landmark-based --- Coordinate regression 
& 
\vspace{-7pt}
\begin{itemize}[leftmargin=*]
 \item Prediction bias due to poor initialization. 
\end{itemize}
\vspace{-12pt}
\\

\cline{2-3}
& \vspace{-3pt}  Landmark-based --- Heatmap regression 
& 
\vspace{-7pt}
\begin{itemize}[leftmargin=*]
 \item High computational cost. 
\end{itemize}
\vspace{-12pt}
\\

\cline{2-3}
&\vspace{-3pt}  Landmark-based --- 3D model fitting 
& 
\vspace{-7pt}
\begin{itemize}[leftmargin=*]
 \item Runtime efficiency.
\end{itemize}
\vspace{-12pt}
\\

\cline{2-3}
& \vspace{-3pt}  Landmark-free 
& 
\vspace{-7pt}
\begin{itemize}[leftmargin=*]
 \item Loss of identity discriminative information. 
\end{itemize}
\vspace{-12pt}
\\

\hline

\multicolumn{1}{|l|}{  Face representation} 

& \vspace{-3pt}  Training supervision --- Classification
& 
\vspace{-7pt}
\begin{itemize}[leftmargin=*]
 \item Training on imbalance data. 
\end{itemize}
\vspace{-12pt}
\\

\cline{2-3}
& \vspace{-3pt}  Training supervision --- Feature embedding
& 
\vspace{-7pt}
\begin{itemize}[leftmargin=*]
 \item Efficient training on large-scale datasets. 

\end{itemize}
\vspace{-12pt}
\\

\cline{2-3}
&\vspace{-3pt}   Training supervision --- Hybrid
& 
\vspace{-7pt}
\begin{itemize}[leftmargin=*]
 \item Unified training supervision of classification and feature embedding.
\end{itemize}
\vspace{-12pt}
\\

\cline{2-3}
& \vspace{-3pt}  Training supervision --- Semi-supervised
& 
\vspace{-7pt}
\begin{itemize}[leftmargin=*]
 \item Open-set identities setting. 
\end{itemize}
\vspace{-12pt}
\\

\cline{2-3}
& \vspace{-3pt}  Specific Tasks --- Cross-age
& 
\vspace{-7pt}
\begin{itemize}[leftmargin=*]
 \item Recognizing identities across a wide range of age. 
\end{itemize}
\vspace{-12pt}
\\

\cline{2-3}
& \vspace{-3pt}  Specific Tasks --- Cross-pose
& 
\vspace{-7pt}
\begin{itemize}[leftmargin=*]
 \item Large pose variation. 
\end{itemize}
\vspace{-12pt}
\\

\cline{2-3}
& \vspace{-3pt}  Specific Tasks --- Racial bias
& 
\vspace{-7pt}
\begin{itemize}[leftmargin=*]
 \item Bias reduction. 
\end{itemize}
\vspace{-12pt}
\\

\cline{2-3}
& \vspace{-3pt}  Specific Tasks --- Cross-modality
& 
\vspace{-7pt}
\begin{itemize}[leftmargin=*]
 \item Domain generalization.
\end{itemize}
\vspace{-12pt}
\\

\cline{2-3}
& \vspace{-3pt}  Specific Tasks --- Low-shot
& 
\vspace{-7pt}
\begin{itemize}[leftmargin=*]
 \item One-shot learning.
\end{itemize}
\vspace{-12pt}

\\

\cline{2-3}
& \vspace{-3pt}  Specific Tasks --- Video-based
\vspace{-6pt} & \vspace{-6pt}
\begin{itemize}[leftmargin=*]
 \item Low quality of frames. 
\end{itemize}
\vspace{-12pt}
\\
\hline
\end{tabular}}
\end{center}

\end{table*}

In addition, we are going deeper with Table~\ref{conclusion_challenges1} about the major challenges towards the subcategories of each element. 
For instance, since the anchor-based face detector needs to pre-define a large number of anchors, the settings of preset anchors need to be carefully tuned for each particular dataset, which limits the generalization ability of face detectors. In contrast, anchor-free face detector needs further exploration for better robustness to false positives and stability in training process. 

%
\subsection{Future Trend}

To address the above challenges, a number of worthwhile research directions need to be explored in the future.

\subsubsection{Face detection}
 
\begin{itemize}
\item~\textbf{Generalized anchor settings}. The existing anchor-based methods design the anchor setting from many aspects, such as assignment and matching   strategy~\cite{Zhang2017S3FD,tang2018pyramidbox,2019DSFD,li2019pyramidbox,Liu_2020_HAMBox}, attributes tuning~\cite{Zhang2017Faceboxes,Zhu2018,chi2019selective}, and sampling strategy~\cite{Ming_2019_Group_Sampling}. The well-tuned anchors may limit the generalization ability of face detectors. Hence, it is worth to explore a generalized anchor setting that can be used for different application demand. 

\item~\textbf{Anchor-free face detection framework}.
Anchor-free detectors~\cite{Law2018CornerNet,zhu2019fs,Tian2019FCOS} show flexible designs and more potential in generalization ability for object detection. However, a small number of works~\cite{huang2015densebox,UnitBox,xu2019centerface} have explored the anchor-free mechanism and its advantages for face detection.

\end{itemize}

\subsubsection{Face alignment}

\begin{itemize}
\item~\textbf{High robustness and efficiency}.
There is a large amount of facial variations in real-world conditions, which requires the alignment methods being robust to various input faces while keeping efficiency as an intermediate step of the system. 
\item~\textbf{Dense landmark localization}.
The most existing datasets employ 68 or 106 keypoints as annotation configuration. 
They are enough for face alignment (usually 5 keypoints needed), but insufficient to the complex face analysis tasks, such as facial motion capture. Besides, the dense landmarks will help to locate more accurate alignment-needed keypoints.
\item~\textbf{Video-based landmark localization}.
How to make better use of the temporal information is a major challenge for video-based landmark localization. This topic will enable to address the problems in video, such as large poses, motion blur, low illumination and resolution, \textit{etc}. 
\item~\textbf{Semi-supervised landmark localization}:
The extensive research on landmark localization belongs to the regime of supervised learning, which needs the precise annotated landmarks. However, it is expensive and inefficient to obtain large-scale dataset with the precise annotations. As explored by the pioneering works~\cite{Dong2018SBR,Honari2018ImprovingLL,Robinson2019LaplaceLL,Dong2019TeacherSS}, the semi-supervised routine is a feasible and valuable solution for facial landmark localization. 
 
\end{itemize}

\subsubsection{Face representation}


\begin{itemize}
\item~\textbf{Lightweight face recognition}:
The large memory and computational cost often makes it impractical to employ heavy-weight networks on mobile or embedded devices. Although many works ~\cite{Wu2015ALC,Wu2018ALC,chen2018mobilefacenets,Duong2018MobiFaceAL,MartnezDaz2019ShuffleFaceNetAL,Wu_2020_CVPR} have studied lightweight face recognition, there is still large room to improve the lightweight models with high efficiency and accuracy. 

\item~\textbf{Robustness to variations in video}:
It highly requires robust face representation models against varying conditions in surveillance video. The robustness to low image quality and large facial pose is the core demand in many practical applications.

\item~\textbf{Noisy label learning}:
Label noise is an inevitable problem when collecting large-scale face dataset. Certain works~\cite{Wang2018TheDO,deng2019arcface,trillionpairs.org,Zhang_2020_FaceGraph} study how to remove the noisy data, and some others~\cite{Hu2019NoiseTolerantPF,wang2019co,Zhong_2019_CVPR} aim at learning noise-robust face representation. But most of them are susceptible to the ability of the initial model, and need to be more flexible in real-world scenarios. It is still an open issue for noisy label learning in face recognition. 

\item~\textbf{Cross domain face recognition}:
There are many different domain factors in face data, such as facial age, pose, race, imaging modality, and some works~\cite{Wen2016LatentFG,Du2019AgeFR,Tran2017DisentangledRL,Cao2018PoseRobustFR,Wang2019RacialFI,Wang_2020_CVPR,Zhang2020CascadedFS} have studied the face recognition across a small fraction of them.
How to obtain a universal representation for cross domain face recognition is a challenging research topic.

\item~\textbf{Learning with imbalance data}:
Representation learning on the long-tail data is long-standing topic in many datasets. 
With the scarcity of intra-class variations, the subjects with limited training samples are usually neglected. The domain bias caused by imbalance data scale is another problem. 
It is worth to handle these problems in a unified framework.   

\item~\textbf{Learning with unlabeled faces}:
There are a large amount of unlabeled face data in practical applications. However, it is excessively expensive to manually annotate them when the dataset keeps growing. Recently, semi-supervised learning and face clustering methods attract increasing attention. How to effectively employ unlabeled data for boosting face recognition is a promising direction.

\end{itemize}

\subsubsection{Towards the entire system}
There is very little work to solve the major challenges from the perspective of entire system. We present several promising directions of this area in the following. 

\begin{itemize}

\item~\textbf{Interpretable deep models}: 
Although the explainable artificial intelligence, so-called XAI, has been studied for a long time, the explainable deep face recognition is in its infancy~\cite{zhong2018deep,zee2019enhancing,yin2019towards,williford2020explainable}. 
There are two ways to access the interpretability for deep face recognition, \textit{i.e.,} the top-down and bottom-up, respectively. The top-down way resorts to the human prior knowledge for algorithm exploration, since human shows superior ability of face recognition than deep models in many tough conditions. The bottom-up way denotes the exploration from the perspective of face data itself, such as modeling the explainable deep face recognition in spatial and scale dimension.

\item~\textbf{Joint modeling for the holistic system}: Despite the three elements having different optimized objective, it is still worth to exploit the end-to-end trainable deep face recognition, and study how they can be further improved through the jointly learning.
Furthermore, beyond the topic of this survey, there is also an open question that how should we develop a single network to perform the end-to-end face recognition. 

\item~\textbf{Universal face representation pretraining}: Most studies of face recognition focus on the specific tasks, but overlooking how to learn a pre-trained universal face representation that can be used to facilitate the downstream facial analysis tasks. There is only one work~\cite{Bulat2021PretrainingSA} that studies this topic. The findings show that it is promising to obtain significant performance improvement for related facial tasks by employing unsupervised pretraining.

\item~\textbf{Trustworthy face recognition system}: With the wide application, it is important to evaluate and boost the trustworthiness of the recognition system~\cite{Jain2021BiometricsTB}. The pursuit for  trustworthy face recognition system is becoming a necessity, which mainly involves several aspects, \textit{i.e.,} robustness, fairness, interpretability, security, and privacy. 
Further research on these aspects is essential.

\end{itemize}

\subsection{Conclusion}

\label{sec:conclusion}

In this survey, we review the recent advances of the elements of end-to-end deep face recognition, which consist of face detection, face alignment and face representation. 
Although there are many surveys about face recognition, they mostly focus on face representation without considering the intrinsic connection from other elements in the pipeline;  whereas, this survey is the first one which provides a comprehensive review of the elements of end-to-end deep face recognition. 
We present a detailed discussion and comparison of many approaches in each element from poly-aspects.  
Also, we discuss the relationship between the elements and the holistic framework.
According to these elaborated contents, we can not only find the suitable methods to establish state-of-the-art face recognition system, but also know which method is quite strong-baseline style for comparison in experiment. 
Additionally, we analyze the existing challenges and collect certain promising future research directions. We hope this survey could bring helpful thoughts for better understanding of end-to-end face recognition and deeper exploration in a systematic way.


\bibliographystyle{ACM-Reference-Format}
\bibliography{sample-acmsmall}


\begin{thebibliography}{358}


\ifx \showCODEN    \undefined \def \showCODEN     #1{\unskip}     \fi
\ifx \showDOI      \undefined \def \showDOI       #1{#1}\fi
\ifx \showISBNx    \undefined \def \showISBNx     #1{\unskip}     \fi
\ifx \showISBNxiii \undefined \def \showISBNxiii  #1{\unskip}     \fi
\ifx \showISSN     \undefined \def \showISSN      #1{\unskip}     \fi
\ifx \showLCCN     \undefined \def \showLCCN      #1{\unskip}     \fi
\ifx \shownote     \undefined \def \shownote      #1{#1}          \fi
\ifx \showarticletitle \undefined \def \showarticletitle #1{#1}   \fi
\ifx \showURL      \undefined \def \showURL       {\relax}        \fi
\providecommand\bibfield[2]{#2}
\providecommand\bibinfo[2]{#2}
\providecommand\natexlab[1]{#1}
\providecommand\showeprint[2][]{arXiv:#2}

\bibitem[\protect\citeauthoryear{Abd-Almageed, Wu, Rawls, Harel, Hassner, Masi,
  Choi, Leksut, Kim, Natarajan, Nevatia, and Medioni}{Abd-Almageed
  et~al\mbox{.}}{2016}]%
        {AbdAlmageed2016FaceRU}
\bibfield{author}{\bibinfo{person}{W. Abd-Almageed}, \bibinfo{person}{Y. Wu},
  \bibinfo{person}{S. Rawls}, \bibinfo{person}{S. Harel}, \bibinfo{person}{T.
  Hassner}, \bibinfo{person}{I. Masi}, \bibinfo{person}{J. Choi},
  \bibinfo{person}{J.~T. Leksut}, \bibinfo{person}{J. Kim}, \bibinfo{person}{P.
  Natarajan}, \bibinfo{person}{R. Nevatia}, {and} \bibinfo{person}{G.~G.
  Medioni}.} \bibinfo{year}{2016}\natexlab{}.
\newblock \showarticletitle{Face recognition using deep multi-pose
  representations}. In \bibinfo{booktitle}{\emph{Proceedings of the IEEE Winter
  Conference on Applications of Computer Vision}}. \bibinfo{pages}{1--9}.
\newblock


\bibitem[\protect\citeauthoryear{Adjabi, Ouahabi, Benzaoui, and
  Taleb-Ahmed}{Adjabi et~al\mbox{.}}{2020}]%
        {electronics9081188}
\bibfield{author}{\bibinfo{person}{I. Adjabi}, \bibinfo{person}{A. Ouahabi},
  \bibinfo{person}{A. Benzaoui}, {and} \bibinfo{person}{A. Taleb-Ahmed}.}
  \bibinfo{year}{2020}\natexlab{}.
\newblock \showarticletitle{Past, Present, and Future of Face Recognition: A
  Review}.
\newblock \bibinfo{journal}{\emph{Electronics}} \bibinfo{volume}{9},
  \bibinfo{number}{8} (\bibinfo{year}{2020}).
\newblock


\bibitem[\protect\citeauthoryear{Ahonen, Hadid, and Pietikinen}{Ahonen
  et~al\mbox{.}}{2004}]%
        {Ahonen2004Face}
\bibfield{author}{\bibinfo{person}{T. Ahonen}, \bibinfo{person}{A. Hadid},
  {and} \bibinfo{person}{M. Pietikinen}.} \bibinfo{year}{2004}\natexlab{}.
\newblock \showarticletitle{Face recognition with local binary patterns}. In
  \bibinfo{booktitle}{\emph{Proceedings of the European Conference on Computer
  Vision}}. \bibinfo{pages}{469--481}.
\newblock


\bibitem[\protect\citeauthoryear{An, Zhu, Xiao, Wu, Zhang, Gao, Qin, Zhang, and
  Fu}{An et~al\mbox{.}}{2021}]%
        {an2020partical_fc}
\bibfield{author}{\bibinfo{person}{Xiang An}, \bibinfo{person}{Xuhan Zhu},
  \bibinfo{person}{Yanghua Xiao}, \bibinfo{person}{Lan Wu},
  \bibinfo{person}{Ming Zhang}, \bibinfo{person}{Yuan Gao},
  \bibinfo{person}{Bin Qin}, \bibinfo{person}{Debing Zhang}, {and}
  \bibinfo{person}{Yingnan Fu}.} \bibinfo{year}{2021}\natexlab{}.
\newblock \showarticletitle{Partial FC: Training 10 Million Identities on a
  Single Machine}.
\newblock \bibinfo{journal}{\emph{Proceedings of the IEEE/CVF International
  Conference on Computer Vision Workshops}} (\bibinfo{year}{2021}),
  \bibinfo{pages}{1445--1449}.
\newblock


\bibitem[\protect\citeauthoryear{Antipov, Baccouche, and Dugelay}{Antipov
  et~al\mbox{.}}{2017a}]%
        {Antipov2017BoostingCF}
\bibfield{author}{\bibinfo{person}{G. Antipov}, \bibinfo{person}{M. Baccouche},
  {and} \bibinfo{person}{J. Dugelay}.} \bibinfo{year}{2017}\natexlab{a}.
\newblock \showarticletitle{Boosting cross-age face verification via generative
  age normalization}. In \bibinfo{booktitle}{\emph{Proceedings of the IEEE
  International Joint Conference on Biometrics}}. \bibinfo{pages}{191--199}.
\newblock


\bibitem[\protect\citeauthoryear{Antipov, Baccouche, and Dugelay}{Antipov
  et~al\mbox{.}}{2017b}]%
        {Antipov2017FaceAW}
\bibfield{author}{\bibinfo{person}{G. Antipov}, \bibinfo{person}{M. Baccouche},
  {and} \bibinfo{person}{J. Dugelay}.} \bibinfo{year}{2017}\natexlab{b}.
\newblock \showarticletitle{Face aging with conditional generative adversarial
  networks}. In \bibinfo{booktitle}{\emph{Proceedings of the IEEE International
  Conference on Image Processing}}. \bibinfo{pages}{2089--2093}.
\newblock


\bibitem[\protect\citeauthoryear{Arabnia}{Arabnia}{2009}]%
        {2009fr}
\bibfield{author}{\bibinfo{person}{H.~R. Arabnia}.}
  \bibinfo{year}{2009}\natexlab{}.
\newblock \showarticletitle{A Survey of Face Recognition Techniques}.
\newblock \bibinfo{journal}{\emph{Journal of Information Processing Systems}}
  \bibinfo{volume}{5}, \bibinfo{number}{2} (\bibinfo{year}{2009}),
  \bibinfo{pages}{41--68}.
\newblock


\bibitem[\protect\citeauthoryear{{B. Yang}, {Yan}, {Lei}, and {Li}}{{B. Yang}
  et~al\mbox{.}}{2014}]%
        {Yang2014acf}
\bibfield{author}{\bibinfo{person}{{B. Yang}}, \bibinfo{person}{J. {Yan}},
  \bibinfo{person}{Z. {Lei}}, {and} \bibinfo{person}{S.~Z. {Li}}.}
  \bibinfo{year}{2014}\natexlab{}.
\newblock \showarticletitle{Aggregate channel features for multi-view face
  detection}. In \bibinfo{booktitle}{\emph{Proceedings of the IEEE
  International Joint Conference on Biometrics}}. \bibinfo{pages}{1--8}.
\newblock


\bibitem[\protect\citeauthoryear{Bai, Zhang, Ding, and Ghanem}{Bai
  et~al\mbox{.}}{2018}]%
        {bai2018finding}
\bibfield{author}{\bibinfo{person}{Y. Bai}, \bibinfo{person}{Y. Zhang},
  \bibinfo{person}{M. Ding}, {and} \bibinfo{person}{B. Ghanem}.}
  \bibinfo{year}{2018}\natexlab{}.
\newblock \showarticletitle{Finding tiny faces in the wild with generative
  adversarial network}. In \bibinfo{booktitle}{\emph{Proceedings of the
  IEEE/CVF Conference on Computer Vision and Pattern Recognition}}.
  \bibinfo{pages}{21--30}.
\newblock


\bibitem[\protect\citeauthoryear{Bansal, Castillo, Ranjan, and
  Chellappa}{Bansal et~al\mbox{.}}{2017a}]%
        {Bansal2017TheDA}
\bibfield{author}{\bibinfo{person}{A. Bansal}, \bibinfo{person}{C.~D.
  Castillo}, \bibinfo{person}{R. Ranjan}, {and} \bibinfo{person}{R.
  Chellappa}.} \bibinfo{year}{2017}\natexlab{a}.
\newblock \showarticletitle{The Do’s and Don’ts for CNN-Based Face
  Verification}. In \bibinfo{booktitle}{\emph{Proceedings of the IEEE
  International Conference on Computer Vision Workshops}}.
  \bibinfo{pages}{2545--2554}.
\newblock


\bibitem[\protect\citeauthoryear{Bansal, Nanduri, Castillo, Ranjan, and
  Chellappa}{Bansal et~al\mbox{.}}{2017b}]%
        {Bansal2017UMDFacesAA}
\bibfield{author}{\bibinfo{person}{A. Bansal}, \bibinfo{person}{A. Nanduri},
  \bibinfo{person}{C.~D. Castillo}, \bibinfo{person}{R. Ranjan}, {and}
  \bibinfo{person}{R. Chellappa}.} \bibinfo{year}{2017}\natexlab{b}.
\newblock \showarticletitle{UMDFaces: An annotated face dataset for training
  deep networks}. In \bibinfo{booktitle}{\emph{Proceedings of the IEEE
  International Joint Conference on Biometrics}}. \bibinfo{pages}{464--473}.
\newblock


\bibitem[\protect\citeauthoryear{Belhumeur, Jacobs, Kriegman, and
  Kumar}{Belhumeur et~al\mbox{.}}{2013}]%
        {Belhumeur2011LocalizingPO}
\bibfield{author}{\bibinfo{person}{P.~N. Belhumeur}, \bibinfo{person}{D.~W.
  Jacobs}, \bibinfo{person}{D.~J. Kriegman}, {and} \bibinfo{person}{N. Kumar}.}
  \bibinfo{year}{2013}\natexlab{}.
\newblock \showarticletitle{Localizing parts of faces using a consensus of
  exemplars}.
\newblock \bibinfo{journal}{\emph{IEEE Trans. Pattern Anal. Mach. Intell.}}
  \bibinfo{volume}{35}, \bibinfo{number}{12} (\bibinfo{year}{2013}),
  \bibinfo{pages}{2930--2940}.
\newblock


\bibitem[\protect\citeauthoryear{Belhumeur, Joo, and Kriegman}{Belhumeur
  et~al\mbox{.}}{1997}]%
        {Belhumeur1997Eigenfaces}
\bibfield{author}{\bibinfo{person}{P.~N. Belhumeur}, \bibinfo{person}{P.~H.
  Joo}, {and} \bibinfo{person}{D.~J. Kriegman}.}
  \bibinfo{year}{1997}\natexlab{}.
\newblock \showarticletitle{Eigenfaces vs. Fisherfaces: Recognition Using Class
  Specific Linear Projection}.
\newblock \bibinfo{journal}{\emph{IEEE Trans. Pattern Anal. Mach. Intell.}}
  \bibinfo{volume}{19}, \bibinfo{number}{7} (\bibinfo{year}{1997}),
  \bibinfo{pages}{711--720}.
\newblock


\bibitem[\protect\citeauthoryear{Bhagavatula, Zhu, Luu, and
  Savvides}{Bhagavatula et~al\mbox{.}}{2017}]%
        {Bhagavatula2017FasterTR}
\bibfield{author}{\bibinfo{person}{C. Bhagavatula}, \bibinfo{person}{C. Zhu},
  \bibinfo{person}{K. Luu}, {and} \bibinfo{person}{M. Savvides}.}
  \bibinfo{year}{2017}\natexlab{}.
\newblock \showarticletitle{Faster than Real-Time Facial Alignment: A 3D
  Spatial Transformer Network Approach in Unconstrained Poses}. In
  \bibinfo{booktitle}{\emph{Proceedings of the IEEE International Conference on
  Computer Vision}}. \bibinfo{pages}{4000--4009}.
\newblock


\bibitem[\protect\citeauthoryear{Blanz and Vetter}{Blanz and Vetter}{2003}]%
        {BlanzVolker2003FaceRB}
\bibfield{author}{\bibinfo{person}{V. Blanz} {and} \bibinfo{person}{T.
  Vetter}.} \bibinfo{year}{2003}\natexlab{}.
\newblock \showarticletitle{Face recognition based on fitting a 3d morphable
  model}.
\newblock \bibinfo{journal}{\emph{IEEE Trans. Pattern Anal. Mach. Intell.}}
  \bibinfo{volume}{25}, \bibinfo{number}{9} (\bibinfo{year}{2003}),
  \bibinfo{pages}{1063--1074}.
\newblock


\bibitem[\protect\citeauthoryear{Bowyer, Chang, and Flynn}{Bowyer
  et~al\mbox{.}}{2006}]%
        {bowyer2006survey}
\bibfield{author}{\bibinfo{person}{K.~W. Bowyer}, \bibinfo{person}{K. Chang},
  {and} \bibinfo{person}{P. Flynn}.} \bibinfo{year}{2006}\natexlab{}.
\newblock \showarticletitle{A survey of approaches and challenges in 3D and
  multi-modal 3D+ 2D face recognition}.
\newblock \bibinfo{journal}{\emph{Computer vision and image understanding}}
  \bibinfo{volume}{101}, \bibinfo{number}{1} (\bibinfo{year}{2006}),
  \bibinfo{pages}{1--15}.
\newblock


\bibitem[\protect\citeauthoryear{Brubaker, Wu, Sun, Mullin, and Rehg}{Brubaker
  et~al\mbox{.}}{2008}]%
        {Brubaker2008On}
\bibfield{author}{\bibinfo{person}{S.~C. Brubaker}, \bibinfo{person}{J. Wu},
  \bibinfo{person}{J. Sun}, \bibinfo{person}{M.~D. Mullin}, {and}
  \bibinfo{person}{J.~M. Rehg}.} \bibinfo{year}{2008}\natexlab{}.
\newblock \showarticletitle{On the Design of Cascades of Boosted Ensembles for
  Face Detection}.
\newblock \bibinfo{journal}{\emph{International Journal of Computer Vision}}
  \bibinfo{volume}{77}, \bibinfo{number}{1-3} (\bibinfo{year}{2008}),
  \bibinfo{pages}{65--86}.
\newblock


\bibitem[\protect\citeauthoryear{Bulat, Cheng, Yang, Garbett, Sanchez, and
  Tzimiropoulos}{Bulat et~al\mbox{.}}{2021}]%
        {Bulat2021PretrainingSA}
\bibfield{author}{\bibinfo{person}{Adrian Bulat}, \bibinfo{person}{Shiyang
  Cheng}, \bibinfo{person}{Jing Yang}, \bibinfo{person}{A. Garbett},
  \bibinfo{person}{Enrique Sanchez}, {and} \bibinfo{person}{Georgios
  Tzimiropoulos}.} \bibinfo{year}{2021}\natexlab{}.
\newblock \showarticletitle{Pre-training strategies and datasets for facial
  representation learning}.
\newblock \bibinfo{journal}{\emph{ArXiv}}  \bibinfo{volume}{abs/2103.16554}
  (\bibinfo{year}{2021}).
\newblock


\bibitem[\protect\citeauthoryear{Bulat and Tzimiropoulos}{Bulat and
  Tzimiropoulos}{2016}]%
        {Bulat2016Convolutional}
\bibfield{author}{\bibinfo{person}{A. Bulat} {and} \bibinfo{person}{G.
  Tzimiropoulos}.} \bibinfo{year}{2016}\natexlab{}.
\newblock \showarticletitle{Convolutional aggregation of local evidence for
  large pose face alignment}. In \bibinfo{booktitle}{\emph{Proceedings of the
  British Machine Vision Conference}}. \bibinfo{pages}{86.1--86.12}.
\newblock


\bibitem[\protect\citeauthoryear{Bulat and Tzimiropoulos}{Bulat and
  Tzimiropoulos}{2017}]%
        {Bulat2017HowFA}
\bibfield{author}{\bibinfo{person}{A. Bulat} {and} \bibinfo{person}{G.
  Tzimiropoulos}.} \bibinfo{year}{2017}\natexlab{}.
\newblock \showarticletitle{How Far are We from Solving the 2D $\&$ 3D Face
  Alignment Problem? (and a Dataset of 230,000 3D Facial Landmarks)}. In
  \bibinfo{booktitle}{\emph{Proceedings of the IEEE International Conference on
  Computer Vision}}. \bibinfo{pages}{1021--1030}.
\newblock


\bibitem[\protect\citeauthoryear{Burgos-Artizzu, Perona, and
  Doll{\'a}r}{Burgos-Artizzu et~al\mbox{.}}{2013}]%
        {BurgosArtizzu2013RobustFL}
\bibfield{author}{\bibinfo{person}{X.~P. Burgos-Artizzu}, \bibinfo{person}{P.
  Perona}, {and} \bibinfo{person}{P. Doll{\'a}r}.}
  \bibinfo{year}{2013}\natexlab{}.
\newblock \showarticletitle{Robust face landmark estimation under occlusion}.
  In \bibinfo{booktitle}{\emph{Proceedings of the IEEE International Conference
  on Computer Vision}}. \bibinfo{pages}{1513--1520}.
\newblock


\bibitem[\protect\citeauthoryear{Cao, Zhu, Huang, Guo, and Lei}{Cao
  et~al\mbox{.}}{2020}]%
        {Cao2020DomainBF}
\bibfield{author}{\bibinfo{person}{D. Cao}, \bibinfo{person}{X. Zhu},
  \bibinfo{person}{X. Huang}, \bibinfo{person}{J. Guo}, {and}
  \bibinfo{person}{Z. Lei}.} \bibinfo{year}{2020}\natexlab{}.
\newblock \showarticletitle{Domain Balancing: Face Recognition on Long-Tailed
  Domains}. In \bibinfo{booktitle}{\emph{Proceedings of the IEEE/CVF Conference
  on Computer Vision and Pattern Recognition}}. \bibinfo{pages}{5671--5679}.
\newblock


\bibitem[\protect\citeauthoryear{Cao, Rong, Li, Tang, and Loy}{Cao
  et~al\mbox{.}}{2018a}]%
        {Cao2018PoseRobustFR}
\bibfield{author}{\bibinfo{person}{K. Cao}, \bibinfo{person}{Y. Rong},
  \bibinfo{person}{C. Li}, \bibinfo{person}{X. Tang}, {and}
  \bibinfo{person}{C.~C. Loy}.} \bibinfo{year}{2018}\natexlab{a}.
\newblock \showarticletitle{Pose-Robust Face Recognition via Deep Residual
  Equivariant Mapping}. In \bibinfo{booktitle}{\emph{Proceedings of the
  IEEE/CVF Conference on Computer Vision and Pattern Recognition}}.
  \bibinfo{pages}{5187--5196}.
\newblock


\bibitem[\protect\citeauthoryear{Cao, Shen, Xie, Parkhi, and Zisserman}{Cao
  et~al\mbox{.}}{2018b}]%
        {Cao2018VGGFace2AD}
\bibfield{author}{\bibinfo{person}{Q. Cao}, \bibinfo{person}{L. Shen},
  \bibinfo{person}{W. Xie}, \bibinfo{person}{O.~M. Parkhi}, {and}
  \bibinfo{person}{A. Zisserman}.} \bibinfo{year}{2018}\natexlab{b}.
\newblock \showarticletitle{VGGFace2: A Dataset for Recognising Faces across
  Pose and Age}. In \bibinfo{booktitle}{\emph{Proceedings of the IEEE
  International Conference on Automatic Face $\&$ Gesture Recognition}}.
  \bibinfo{pages}{67--74}.
\newblock


\bibitem[\protect\citeauthoryear{{Chang}, {Tran}, {Hassner}, {Masi}, {Nevatia},
  and {Medioni}}{{Chang} et~al\mbox{.}}{2017}]%
        {Chang2017FacePoseNet}
\bibfield{author}{\bibinfo{person}{F. {Chang}}, \bibinfo{person}{A.~T. {Tran}},
  \bibinfo{person}{T. {Hassner}}, \bibinfo{person}{I. {Masi}},
  \bibinfo{person}{R. {Nevatia}}, {and} \bibinfo{person}{G. {Medioni}}.}
  \bibinfo{year}{2017}\natexlab{}.
\newblock \showarticletitle{FacePoseNet: Making a Case for Landmark-Free Face
  Alignment}. In \bibinfo{booktitle}{\emph{Proceedings of the IEEE
  International Conference on Computer Vision Workshops}}.
  \bibinfo{pages}{1599--1608}.
\newblock


\bibitem[\protect\citeauthoryear{Chang, Lan, Cheng, and Wei}{Chang
  et~al\mbox{.}}{2020}]%
        {Chang2020DataUL}
\bibfield{author}{\bibinfo{person}{J. Chang}, \bibinfo{person}{Z. Lan},
  \bibinfo{person}{C. Cheng}, {and} \bibinfo{person}{Y. Wei}.}
  \bibinfo{year}{2020}\natexlab{}.
\newblock \showarticletitle{Data Uncertainty Learning in Face Recognition}. In
  \bibinfo{booktitle}{\emph{Proceedings of the IEEE/CVF Conference on Computer
  Vision and Pattern Recognition}}. \bibinfo{pages}{5710--5719}.
\newblock


\bibitem[\protect\citeauthoryear{Chen, Hua, Wen, and Sun}{Chen
  et~al\mbox{.}}{2016}]%
        {2016stn}
\bibfield{author}{\bibinfo{person}{D. Chen}, \bibinfo{person}{G. Hua},
  \bibinfo{person}{F. Wen}, {and} \bibinfo{person}{J. Sun}.}
  \bibinfo{year}{2016}\natexlab{}.
\newblock \showarticletitle{Supervised transformer network for efficient face
  detection}. In \bibinfo{booktitle}{\emph{Proceedings of the European
  Conference on Computer Vision}}, Vol.~\bibinfo{volume}{9909}.
  \bibinfo{pages}{122--138}.
\newblock


\bibitem[\protect\citeauthoryear{Chen, Ranjan, Sankaranarayanan, Kumar, Chen,
  Patel, Castillo, and Chellappa}{Chen et~al\mbox{.}}{2017}]%
        {Chen2017UnconstrainedSF}
\bibfield{author}{\bibinfo{person}{J. Chen}, \bibinfo{person}{R. Ranjan},
  \bibinfo{person}{S. Sankaranarayanan}, \bibinfo{person}{A. Kumar},
  \bibinfo{person}{C. Chen}, \bibinfo{person}{V. Patel}, \bibinfo{person}{C.~D.
  Castillo}, {and} \bibinfo{person}{R. Chellappa}.}
  \bibinfo{year}{2017}\natexlab{}.
\newblock \showarticletitle{Unconstrained Still/Video-Based Face Verification
  with Deep Convolutional Neural Networks}.
\newblock \bibinfo{journal}{\emph{International Journal of Computer Vision}}
  \bibinfo{volume}{126} (\bibinfo{year}{2017}), \bibinfo{pages}{272--291}.
\newblock


\bibitem[\protect\citeauthoryear{Chen, Su, and Ji}{Chen et~al\mbox{.}}{2019}]%
        {Chen2019FaceAW}
\bibfield{author}{\bibinfo{person}{L. Chen}, \bibinfo{person}{H. Su}, {and}
  \bibinfo{person}{Q. Ji}.} \bibinfo{year}{2019}\natexlab{}.
\newblock \showarticletitle{Face Alignment With Kernel Density Deep Neural
  Network}. In \bibinfo{booktitle}{\emph{Proceedings of the IEEE International
  Conference on Computer Vision}}. \bibinfo{pages}{6991--7001}.
\newblock


\bibitem[\protect\citeauthoryear{Chen, Liu, Gao, and Han}{Chen
  et~al\mbox{.}}{2018}]%
        {chen2018mobilefacenets}
\bibfield{author}{\bibinfo{person}{S. Chen}, \bibinfo{person}{Y. Liu},
  \bibinfo{person}{X. Gao}, {and} \bibinfo{person}{Z. Han}.}
  \bibinfo{year}{2018}\natexlab{}.
\newblock \showarticletitle{Mobilefacenets: Efficient cnns for accurate
  real-time face verification on mobile devices}. In
  \bibinfo{booktitle}{\emph{Chinese Conference on Biometric Recognition}}.
  \bibinfo{pages}{428--438}.
\newblock


\bibitem[\protect\citeauthoryear{Cheng, Zhao, Wang, Xu, Jayashree, Shen, and
  Feng}{Cheng et~al\mbox{.}}{2017}]%
        {cheng2017know}
\bibfield{author}{\bibinfo{person}{Y. Cheng}, \bibinfo{person}{J. Zhao},
  \bibinfo{person}{Z. Wang}, \bibinfo{person}{Y. Xu}, \bibinfo{person}{K.
  Jayashree}, \bibinfo{person}{S. Shen}, {and} \bibinfo{person}{J. Feng}.}
  \bibinfo{year}{2017}\natexlab{}.
\newblock \showarticletitle{Know you at one glance: A compact vector
  representation for low-shot learning}. In
  \bibinfo{booktitle}{\emph{Proceedings of the IEEE International Conference on
  Computer Vision Workshops}}. \bibinfo{pages}{1924--1932}.
\newblock


\bibitem[\protect\citeauthoryear{Chi, Zhang, Xing, Lei, Li, and Zou}{Chi
  et~al\mbox{.}}{2019}]%
        {chi2019selective}
\bibfield{author}{\bibinfo{person}{C. Chi}, \bibinfo{person}{S. Zhang},
  \bibinfo{person}{J. Xing}, \bibinfo{person}{Z. Lei}, \bibinfo{person}{S.~Z.
  Li}, {and} \bibinfo{person}{X. Zou}.} \bibinfo{year}{2019}\natexlab{}.
\newblock \showarticletitle{Selective refinement network for high performance
  face detection}. In \bibinfo{booktitle}{\emph{Proceedings of the AAAI
  Conference on Artificial Intelligence}}, Vol.~\bibinfo{volume}{33}.
  \bibinfo{pages}{8231--8238}.
\newblock


\bibitem[\protect\citeauthoryear{Choe, Park, Kim, Park, Kim, and Shim}{Choe
  et~al\mbox{.}}{2017}]%
        {Choe2017FaceGF}
\bibfield{author}{\bibinfo{person}{J. Choe}, \bibinfo{person}{S. Park},
  \bibinfo{person}{K. Kim}, \bibinfo{person}{J. Park}, \bibinfo{person}{D.
  Kim}, {and} \bibinfo{person}{H. Shim}.} \bibinfo{year}{2017}\natexlab{}.
\newblock \showarticletitle{Face Generation for Low-Shot Learning Using
  Generative Adversarial Networks}. In \bibinfo{booktitle}{\emph{Proceedings of
  the IEEE International Conference on Computer Vision Workshops}}.
  \bibinfo{pages}{1940--1948}.
\newblock


\bibitem[\protect\citeauthoryear{Chollet}{Chollet}{2017}]%
        {Chollet2017XceptionDL}
\bibfield{author}{\bibinfo{person}{F. Chollet}.}
  \bibinfo{year}{2017}\natexlab{}.
\newblock \showarticletitle{Xception: Deep Learning with Depthwise Separable
  Convolutions}. In \bibinfo{booktitle}{\emph{Proceedings of the IEEE/CVF
  Conference on Computer Vision and Pattern Recognition}}.
  \bibinfo{pages}{1800--1807}.
\newblock


\bibitem[\protect\citeauthoryear{Chowdhury, Lin, Maji, and
  Learned-Miller}{Chowdhury et~al\mbox{.}}{2016}]%
        {Chowdhury2016OnetomanyFR}
\bibfield{author}{\bibinfo{person}{A.~R. Chowdhury}, \bibinfo{person}{T. Lin},
  \bibinfo{person}{S. Maji}, {and} \bibinfo{person}{E.~G. Learned-Miller}.}
  \bibinfo{year}{2016}\natexlab{}.
\newblock \showarticletitle{One-to-many face recognition with bilinear CNNs}.
  In \bibinfo{booktitle}{\emph{Proceedings of the IEEE Winter Conference on
  Applications of Computer Vision}}. \bibinfo{pages}{1--9}.
\newblock


\bibitem[\protect\citeauthoryear{Cootes and Taylor}{Cootes and Taylor}{1992}]%
        {Cootes1992ActiveSM}
\bibfield{author}{\bibinfo{person}{T.~F. Cootes} {and} \bibinfo{person}{C.~J.
  Taylor}.} \bibinfo{year}{1992}\natexlab{}.
\newblock \showarticletitle{Active shape models—‘smart snakes’}. In
  \bibinfo{booktitle}{\emph{Proceedings of the British Machine Vision
  Conference}}. \bibinfo{pages}{266--275}.
\newblock


\bibitem[\protect\citeauthoryear{Cootes, Walker, and Taylor}{Cootes
  et~al\mbox{.}}{2000}]%
        {Cootes2000ViewbasedA}
\bibfield{author}{\bibinfo{person}{T.~F. Cootes}, \bibinfo{person}{K. Walker},
  {and} \bibinfo{person}{C.~J. Taylor}.} \bibinfo{year}{2000}\natexlab{}.
\newblock \showarticletitle{View-based active appearance models}. In
  \bibinfo{booktitle}{\emph{Proceedings of the IEEE International Conference on
  Automatic Face $\&$ Gesture Recognition}}. \bibinfo{pages}{227--232}.
\newblock


\bibitem[\protect\citeauthoryear{{Dapogny}, {Cord}, and {Bailly}}{{Dapogny}
  et~al\mbox{.}}{2019}]%
        {Dapogny2019decafa}
\bibfield{author}{\bibinfo{person}{A. {Dapogny}}, \bibinfo{person}{M. {Cord}},
  {and} \bibinfo{person}{K. {Bailly}}.} \bibinfo{year}{2019}\natexlab{}.
\newblock \showarticletitle{DeCaFA: Deep Convolutional Cascade for Face
  Alignment in the Wild}. In \bibinfo{booktitle}{\emph{Proceedings of the IEEE
  International Conference on Computer Vision}}. \bibinfo{pages}{6892--6900}.
\newblock


\bibitem[\protect\citeauthoryear{Deepglint}{Deepglint}{2020}]%
        {trillionpairs.org}
\bibfield{author}{\bibinfo{person}{Deepglint}.}
  \bibinfo{year}{2020}\natexlab{}.
\newblock \bibinfo{title}{Trillion Pairs}.
\newblock
  \bibinfo{howpublished}{\url{http://trillionpairs.deepglint.com/overview}}.
\newblock
\newblock
\shownote{(Accessed September 15, 2020).}


\bibitem[\protect\citeauthoryear{Deng, Dong, Socher, Li, Li, and Fei-Fei}{Deng
  et~al\mbox{.}}{2009}]%
        {deng2009imagenet}
\bibfield{author}{\bibinfo{person}{J. Deng}, \bibinfo{person}{W. Dong},
  \bibinfo{person}{R. Socher}, \bibinfo{person}{L. Li}, \bibinfo{person}{K.
  Li}, {and} \bibinfo{person}{L. Fei-Fei}.} \bibinfo{year}{2009}\natexlab{}.
\newblock \showarticletitle{Imagenet: A large-scale hierarchical image
  database}. In \bibinfo{booktitle}{\emph{Proceedings of the IEEE/CVF
  Conference on Computer Vision and Pattern Recognition}}.
  \bibinfo{pages}{248--255}.
\newblock


\bibitem[\protect\citeauthoryear{Deng, Guo, Ververas, Kotsia, and
  Zafeiriou}{Deng et~al\mbox{.}}{2020}]%
        {deng2020retinaface}
\bibfield{author}{\bibinfo{person}{Jiankang Deng}, \bibinfo{person}{Jia Guo},
  \bibinfo{person}{Evangelos Ververas}, \bibinfo{person}{Irene Kotsia}, {and}
  \bibinfo{person}{Stefanos Zafeiriou}.} \bibinfo{year}{2020}\natexlab{}.
\newblock \showarticletitle{RetinaFace: single-shot multi-Level face
  localisation in the wild}. In \bibinfo{booktitle}{\emph{Proceedings of the
  IEEE/CVF Conference on Computer Vision and Pattern Recognition}}.
  \bibinfo{pages}{5203--5212}.
\newblock


\bibitem[\protect\citeauthoryear{Deng, Guo, Xue, and Zafeiriou}{Deng
  et~al\mbox{.}}{2019a}]%
        {deng2019arcface}
\bibfield{author}{\bibinfo{person}{J. Deng}, \bibinfo{person}{J. Guo},
  \bibinfo{person}{N. Xue}, {and} \bibinfo{person}{S. Zafeiriou}.}
  \bibinfo{year}{2019}\natexlab{a}.
\newblock \showarticletitle{Arcface: Additive angular margin loss for deep face
  recognition}. In \bibinfo{booktitle}{\emph{Proceedings of the IEEE/CVF
  Conference on Computer Vision and Pattern Recognition}}.
  \bibinfo{pages}{4690--4699}.
\newblock


\bibitem[\protect\citeauthoryear{Deng, S.Cheng, N.Xue, Zhou, and
  Zafeiriou}{Deng et~al\mbox{.}}{2018}]%
        {Deng2018UVGANAF}
\bibfield{author}{\bibinfo{person}{J. Deng}, \bibinfo{person}{S.Cheng},
  \bibinfo{person}{N.Xue}, \bibinfo{person}{Y. Zhou}, {and} \bibinfo{person}{S.
  Zafeiriou}.} \bibinfo{year}{2018}\natexlab{}.
\newblock \showarticletitle{UV-GAN: Adversarial Facial UV Map Completion for
  Pose-Invariant Face Recognition}. In \bibinfo{booktitle}{\emph{Proceedings of
  the IEEE/CVF Conference on Computer Vision and Pattern Recognition}}.
  \bibinfo{pages}{7093--7102}.
\newblock


\bibitem[\protect\citeauthoryear{{Deng}, {Trigeorgis}, {Zhou}, and
  {Zafeiriou}}{{Deng} et~al\mbox{.}}{2019}]%
        {Deng2019JMVFA}
\bibfield{author}{\bibinfo{person}{J. {Deng}}, \bibinfo{person}{G.
  {Trigeorgis}}, \bibinfo{person}{Y. {Zhou}}, {and} \bibinfo{person}{S.
  {Zafeiriou}}.} \bibinfo{year}{2019}\natexlab{}.
\newblock \showarticletitle{Joint Multi-View Face Alignment in the Wild}.
\newblock \bibinfo{journal}{\emph{Trans. Image Process.}} \bibinfo{volume}{28},
  \bibinfo{number}{7} (\bibinfo{year}{2019}), \bibinfo{pages}{3636--3648}.
\newblock


\bibitem[\protect\citeauthoryear{Deng, Zhou, and Zafeiriou}{Deng
  et~al\mbox{.}}{2017b}]%
        {Deng2017MarginalLF}
\bibfield{author}{\bibinfo{person}{Jiankang Deng}, \bibinfo{person}{Yuxiang
  Zhou}, {and} \bibinfo{person}{Stefanos Zafeiriou}.}
  \bibinfo{year}{2017}\natexlab{b}.
\newblock \showarticletitle{Marginal Loss for Deep Face Recognition}. In
  \bibinfo{booktitle}{\emph{Proceedings of the IEEE Conference on Computer
  Vision and Pattern Recognition Workshops}}. \bibinfo{pages}{2006--2014}.
\newblock


\bibitem[\protect\citeauthoryear{Deng, Hu, Zhang, Chen, and Guo}{Deng
  et~al\mbox{.}}{2017a}]%
        {deng2017fine}
\bibfield{author}{\bibinfo{person}{W. Deng}, \bibinfo{person}{J. Hu},
  \bibinfo{person}{N. Zhang}, \bibinfo{person}{B. Chen}, {and}
  \bibinfo{person}{J. Guo}.} \bibinfo{year}{2017}\natexlab{a}.
\newblock \showarticletitle{Fine-grained face verification: FGLFW database,
  baselines, and human-DCMN partnership}.
\newblock \bibinfo{journal}{\emph{Pattern Recognition}}  \bibinfo{volume}{66}
  (\bibinfo{year}{2017}), \bibinfo{pages}{63--73}.
\newblock


\bibitem[\protect\citeauthoryear{Deng, Peng, Li, and Qiao}{Deng
  et~al\mbox{.}}{2019b}]%
        {Deng2019MutualCC}
\bibfield{author}{\bibinfo{person}{Z. Deng}, \bibinfo{person}{X. Peng},
  \bibinfo{person}{Z. Li}, {and} \bibinfo{person}{Y. Qiao}.}
  \bibinfo{year}{2019}\natexlab{b}.
\newblock \showarticletitle{Mutual Component Convolutional Neural Networks for
  Heterogeneous Face Recognition}.
\newblock \bibinfo{journal}{\emph{Trans. Image Process.}}  \bibinfo{volume}{28}
  (\bibinfo{year}{2019}), \bibinfo{pages}{3102--3114}.
\newblock


\bibitem[\protect\citeauthoryear{Ding and Tao}{Ding and Tao}{2015}]%
        {Ding2015RobustFR}
\bibfield{author}{\bibinfo{person}{C. Ding} {and} \bibinfo{person}{D. Tao}.}
  \bibinfo{year}{2015}\natexlab{}.
\newblock \showarticletitle{Robust Face Recognition via Multimodal Deep Face
  Representation}.
\newblock \bibinfo{journal}{\emph{IEEE Trans. Multimedia}}
  \bibinfo{volume}{17} (\bibinfo{year}{2015}), \bibinfo{pages}{2049--2058}.
\newblock


\bibitem[\protect\citeauthoryear{Ding and Tao}{Ding and Tao}{2016}]%
        {Ding2016ACS}
\bibfield{author}{\bibinfo{person}{C. Ding} {and} \bibinfo{person}{D. Tao}.}
  \bibinfo{year}{2016}\natexlab{}.
\newblock \showarticletitle{A Comprehensive Survey on Pose-Invariant Face
  Recognition}.
\newblock \bibinfo{journal}{\emph{ACM Trans. Intell. Syst. Technol.}}
  \bibinfo{volume}{7} (\bibinfo{year}{2016}), \bibinfo{pages}{37:1--37:42}.
\newblock


\bibitem[\protect\citeauthoryear{Ding and Tao}{Ding and Tao}{2018}]%
        {Ding2018TrunkBranchEC}
\bibfield{author}{\bibinfo{person}{C. Ding} {and} \bibinfo{person}{D. Tao}.}
  \bibinfo{year}{2018}\natexlab{}.
\newblock \showarticletitle{Trunk-Branch Ensemble Convolutional Neural Networks
  for Video-Based Face Recognition}.
\newblock \bibinfo{journal}{\emph{IEEE Trans. Pattern Anal. Mach. Intell.}}
  \bibinfo{volume}{40} (\bibinfo{year}{2018}), \bibinfo{pages}{1002--1014}.
\newblock


\bibitem[\protect\citeauthoryear{Dong, Yan, Ouyang, and Yang}{Dong
  et~al\mbox{.}}{2018}]%
        {Dong_2018_CVPR}
\bibfield{author}{\bibinfo{person}{X. Dong}, \bibinfo{person}{Y. Yan},
  \bibinfo{person}{W. Ouyang}, {and} \bibinfo{person}{Y. Yang}.}
  \bibinfo{year}{2018}\natexlab{}.
\newblock \showarticletitle{Style aggregated network for facial landmark
  detection}. In \bibinfo{booktitle}{\emph{Proceedings of the IEEE/CVF
  Conference on Computer Vision and Pattern Recognition}}.
  \bibinfo{pages}{379--388}.
\newblock


\bibitem[\protect\citeauthoryear{Dong and Yang}{Dong and Yang}{2019}]%
        {Dong2019TeacherSS}
\bibfield{author}{\bibinfo{person}{X. Dong} {and} \bibinfo{person}{Y. Yang}.}
  \bibinfo{year}{2019}\natexlab{}.
\newblock \showarticletitle{Teacher Supervises Students How to Learn From
  Partially Labeled Images for Facial Landmark Detection}. In
  \bibinfo{booktitle}{\emph{Proceedings of the IEEE International Conference on
  Computer Vision}}. \bibinfo{pages}{783--792}.
\newblock


\bibitem[\protect\citeauthoryear{{Dong}, {Yu}, {Weng}, {Wei}, {Yang}, and
  {Sheikh}}{{Dong} et~al\mbox{.}}{2018}]%
        {Dong2018SBR}
\bibfield{author}{\bibinfo{person}{X. {Dong}}, \bibinfo{person}{S. {Yu}},
  \bibinfo{person}{X. {Weng}}, \bibinfo{person}{S. {Wei}}, \bibinfo{person}{Y.
  {Yang}}, {and} \bibinfo{person}{Y. {Sheikh}}.}
  \bibinfo{year}{2018}\natexlab{}.
\newblock \showarticletitle{Supervision-by-Registration: An Unsupervised
  Approach to Improve the Precision of Facial Landmark Detectors}. In
  \bibinfo{booktitle}{\emph{Proceedings of the IEEE/CVF Conference on Computer
  Vision and Pattern Recognition}}. \bibinfo{pages}{360--368}.
\newblock


\bibitem[\protect\citeauthoryear{Du, Hu, and Wu}{Du et~al\mbox{.}}{2019}]%
        {Du2019AgeFR}
\bibfield{author}{\bibinfo{person}{L. Du}, \bibinfo{person}{H. Hu}, {and}
  \bibinfo{person}{Y. Wu}.} \bibinfo{year}{2019}\natexlab{}.
\newblock \showarticletitle{Age Factor Removal Network Based on Transfer
  Learning and Adversarial Learning for Cross-Age Face Recognition}.
\newblock \bibinfo{journal}{\emph{IEEE Transactions on Circuits and Systems for
  Video Technology}} \bibinfo{volume}{30}, \bibinfo{number}{9}
  (\bibinfo{year}{2019}), \bibinfo{pages}{2830 -- 2842}.
\newblock


\bibitem[\protect\citeauthoryear{{Duan}, {Lu}, and {Zhou}}{{Duan}
  et~al\mbox{.}}{2019}]%
        {UniformFace}
\bibfield{author}{\bibinfo{person}{Y. {Duan}}, \bibinfo{person}{J. {Lu}}, {and}
  \bibinfo{person}{J. {Zhou}}.} \bibinfo{year}{2019}\natexlab{}.
\newblock \showarticletitle{UniformFace: Learning Deep Equidistributed
  Representation for Face Recognition}. In
  \bibinfo{booktitle}{\emph{Proceedings of the IEEE/CVF Conference on Computer
  Vision and Pattern Recognition}}. \bibinfo{pages}{3410--3419}.
\newblock


\bibitem[\protect\citeauthoryear{Duong, Quach, Le, Nguyen, and Luu}{Duong
  et~al\mbox{.}}{2019}]%
        {Duong2018MobiFaceAL}
\bibfield{author}{\bibinfo{person}{C.~N. Duong}, \bibinfo{person}{K.~G. Quach},
  \bibinfo{person}{N. Le}, \bibinfo{person}{N. Nguyen}, {and}
  \bibinfo{person}{K. Luu}.} \bibinfo{year}{2019}\natexlab{}.
\newblock \showarticletitle{Mobiface: A lightweight deep learning face
  recognition on mobile devices}. In \bibinfo{booktitle}{\emph{Proceedings of
  the IEEE International Conference on Biometrics Theory, Applications and
  Systems}}. \bibinfo{pages}{1--6}.
\newblock


\bibitem[\protect\citeauthoryear{Everingham, Gool, Williams, Winn, and
  Zisserman}{Everingham et~al\mbox{.}}{2010}]%
        {Everingham2010The}
\bibfield{author}{\bibinfo{person}{M. Everingham}, \bibinfo{person}{L.~V.
  Gool}, \bibinfo{person}{C.~K.~I. Williams}, \bibinfo{person}{J. Winn}, {and}
  \bibinfo{person}{A. Zisserman}.} \bibinfo{year}{2010}\natexlab{}.
\newblock \showarticletitle{The Pascal Visual Object Classes (VOC) Challenge}.
\newblock \bibinfo{journal}{\emph{International Journal of Computer Vision}}
  \bibinfo{volume}{88}, \bibinfo{number}{2} (\bibinfo{year}{2010}),
  \bibinfo{pages}{p.303--338}.
\newblock


\bibitem[\protect\citeauthoryear{Everingham and Winn}{Everingham and
  Winn}{2011}]%
        {everingham2011pascal}
\bibfield{author}{\bibinfo{person}{M. Everingham} {and} \bibinfo{person}{J.
  Winn}.} \bibinfo{year}{2011}\natexlab{}.
\newblock \showarticletitle{The pascal visual object classes challenge 2012
  (voc2012) development kit}.
\newblock \bibinfo{journal}{\emph{Pattern Analysis, Statistical Modelling and
  Computational Learning, Tech. Rep}}  \bibinfo{volume}{8}
  (\bibinfo{year}{2011}).
\newblock


\bibitem[\protect\citeauthoryear{Farfade, Saberian, and Li}{Farfade
  et~al\mbox{.}}{2015}]%
        {farfade2015multiview}
\bibfield{author}{\bibinfo{person}{S.~S. Farfade}, \bibinfo{person}{M.~J.
  Saberian}, {and} \bibinfo{person}{L.~J. Li}.}
  \bibinfo{year}{2015}\natexlab{}.
\newblock \showarticletitle{Multi-view face detection using deep convolutional
  neural networks}. In \bibinfo{booktitle}{\emph{Proceedings of the 5th ACM on
  International Conference on Multimedia Retrieval}}.
  \bibinfo{pages}{643--650}.
\newblock


\bibitem[\protect\citeauthoryear{Feng, Wu, Shao, Wang, and Zhou}{Feng
  et~al\mbox{.}}{2018}]%
        {feng2018joint}
\bibfield{author}{\bibinfo{person}{Y. Feng}, \bibinfo{person}{F. Wu},
  \bibinfo{person}{X. Shao}, \bibinfo{person}{Y. Wang}, {and}
  \bibinfo{person}{X. Zhou}.} \bibinfo{year}{2018}\natexlab{}.
\newblock \showarticletitle{Joint 3d face reconstruction and dense alignment
  with position map regression network}. In
  \bibinfo{booktitle}{\emph{Proceedings of the European Conference on Computer
  Vision}}. \bibinfo{pages}{534--551}.
\newblock


\bibitem[\protect\citeauthoryear{{Feng}, {Kittler}, {Awais}, {Huber}, and
  {Wu}}{{Feng} et~al\mbox{.}}{2018}]%
        {Feng2018wingloss}
\bibfield{author}{\bibinfo{person}{Z. {Feng}}, \bibinfo{person}{J. {Kittler}},
  \bibinfo{person}{M. {Awais}}, \bibinfo{person}{P. {Huber}}, {and}
  \bibinfo{person}{X. {Wu}}.} \bibinfo{year}{2018}\natexlab{}.
\newblock \showarticletitle{Wing Loss for Robust Facial Landmark Localisation
  with Convolutional Neural Networks}. In \bibinfo{booktitle}{\emph{Proceedings
  of the IEEE/CVF Conference on Computer Vision and Pattern Recognition}}.
  \bibinfo{pages}{2235--2245}.
\newblock


\bibitem[\protect\citeauthoryear{Galea and Farrugia}{Galea and
  Farrugia}{2017}]%
        {Galea2017ForensicFP}
\bibfield{author}{\bibinfo{person}{C. Galea} {and} \bibinfo{person}{R.~A.
  Farrugia}.} \bibinfo{year}{2017}\natexlab{}.
\newblock \showarticletitle{Forensic Face Photo-Sketch Recognition Using a Deep
  Learning-Based Architecture}.
\newblock \bibinfo{journal}{\emph{IEEE Singal processing letters}}
  \bibinfo{volume}{24} (\bibinfo{year}{2017}), \bibinfo{pages}{1586--1590}.
\newblock


\bibitem[\protect\citeauthoryear{Ge, Li, Ye, and Luo}{Ge et~al\mbox{.}}{2017}]%
        {Ge_2017_CVPR}
\bibfield{author}{\bibinfo{person}{S. Ge}, \bibinfo{person}{J. Li},
  \bibinfo{person}{Q. Ye}, {and} \bibinfo{person}{Z. Luo}.}
  \bibinfo{year}{2017}\natexlab{}.
\newblock \showarticletitle{Detecting masked faces in the wild with lle-cnns}.
  In \bibinfo{booktitle}{\emph{Proceedings of the IEEE/CVF Conference on
  Computer Vision and Pattern Recognition}}. \bibinfo{pages}{2682--2690}.
\newblock


\bibitem[\protect\citeauthoryear{G.Huang, Liu, and Weinberger}{G.Huang
  et~al\mbox{.}}{2017}]%
        {Huang2017DenselyCC}
\bibfield{author}{\bibinfo{person}{G.Huang}, \bibinfo{person}{Z. Liu}, {and}
  \bibinfo{person}{K.~Q. Weinberger}.} \bibinfo{year}{2017}\natexlab{}.
\newblock \showarticletitle{Densely Connected Convolutional Networks}. In
  \bibinfo{booktitle}{\emph{Proceedings of the IEEE/CVF Conference on Computer
  Vision and Pattern Recognition}}. \bibinfo{pages}{2261--2269}.
\newblock


\bibitem[\protect\citeauthoryear{Gong, Shi, and Jain}{Gong
  et~al\mbox{.}}{2019a}]%
        {Gong2019LowQV}
\bibfield{author}{\bibinfo{person}{S. Gong}, \bibinfo{person}{Y. Shi}, {and}
  \bibinfo{person}{A.~K. Jain}.} \bibinfo{year}{2019}\natexlab{a}.
\newblock \showarticletitle{Low Quality Video Face Recognition: Multi-Mode
  Aggregation Recurrent Network (MARN)}. In
  \bibinfo{booktitle}{\emph{Proceedings of the IEEE International Conference on
  Computer Vision Workshops}}. \bibinfo{pages}{1027--1035}.
\newblock


\bibitem[\protect\citeauthoryear{Gong, Shi, Kalka, and Jain}{Gong
  et~al\mbox{.}}{2019b}]%
        {Gong2019VideoFR}
\bibfield{author}{\bibinfo{person}{S. Gong}, \bibinfo{person}{Y. Shi},
  \bibinfo{person}{N.~D. Kalka}, {and} \bibinfo{person}{A.~K. Jain}.}
  \bibinfo{year}{2019}\natexlab{b}.
\newblock \showarticletitle{Video Face Recognition: Component-wise Feature
  Aggregation Network (C-FAN)}. In \bibinfo{booktitle}{\emph{Proceedings of the
  International Conference on Biometrics}}. \bibinfo{pages}{1--8}.
\newblock


\bibitem[\protect\citeauthoryear{Gross, Matthews, Cohn, Kanade, and
  Baker}{Gross et~al\mbox{.}}{2008}]%
        {Gross2008MultiPIE}
\bibfield{author}{\bibinfo{person}{R. Gross}, \bibinfo{person}{I.~A. Matthews},
  \bibinfo{person}{J.~F. Cohn}, \bibinfo{person}{T. Kanade}, {and}
  \bibinfo{person}{S. Baker}.} \bibinfo{year}{2008}\natexlab{}.
\newblock \showarticletitle{Multi-PIE}. In
  \bibinfo{booktitle}{\emph{Proceedings of the IEEE International Conference on
  Automatic Face $\&$ Gesture Recognition}}. \bibinfo{pages}{1--8}.
\newblock


\bibitem[\protect\citeauthoryear{Grother, Micheals, and Phillips}{Grother
  et~al\mbox{.}}{2003}]%
        {Grother2003FaceRV}
\bibfield{author}{\bibinfo{person}{P. Grother}, \bibinfo{person}{R.~J.
  Micheals}, {and} \bibinfo{person}{P.~J. Phillips}.}
  \bibinfo{year}{2003}\natexlab{}.
\newblock \showarticletitle{Face recognition vendor test 2002 performance
  metrics}. In \bibinfo{booktitle}{\emph{International Conference on Audio-and
  Video-based Biometric Person Authentication}}. \bibinfo{pages}{937--945}.
\newblock


\bibitem[\protect\citeauthoryear{Grother and Ngan}{Grother and Ngan}{2014}]%
        {Grother2014Face}
\bibfield{author}{\bibinfo{person}{P. Grother} {and} \bibinfo{person}{M.
  Ngan}.} \bibinfo{year}{2014}\natexlab{}.
\newblock \showarticletitle{Face recognition vendor test (FRVT):Performance of
  face identification algorithms}.
\newblock \bibinfo{journal}{\emph{NIST Interagency report}}
  \bibinfo{volume}{8009}, \bibinfo{number}{5} (\bibinfo{year}{2014}),
  \bibinfo{pages}{14}.
\newblock


\bibitem[\protect\citeauthoryear{Guo, Deng, Xue, and Zafeiriou}{Guo
  et~al\mbox{.}}{2018}]%
        {Guo2018StackedDU}
\bibfield{author}{\bibinfo{person}{J. Guo}, \bibinfo{person}{Jiankang Deng},
  \bibinfo{person}{Niannan Xue}, {and} \bibinfo{person}{S. Zafeiriou}.}
  \bibinfo{year}{2018}\natexlab{}.
\newblock \showarticletitle{Stacked Dense U-Nets with Dual Transformers for
  Robust Face Alignment}. In \bibinfo{booktitle}{\emph{Proceedings of the
  British Machine Vision Conference}}.
\newblock


\bibitem[\protect\citeauthoryear{Guo, Xu, Chen, Zhang, Wang, and Zhao}{Guo
  et~al\mbox{.}}{2020}]%
        {GuoDensityAwareFE}
\bibfield{author}{\bibinfo{person}{S. Guo}, \bibinfo{person}{J. Xu},
  \bibinfo{person}{D. Chen}, \bibinfo{person}{C. Zhang}, \bibinfo{person}{X.
  Wang}, {and} \bibinfo{person}{R. Zhao}.} \bibinfo{year}{2020}\natexlab{}.
\newblock \showarticletitle{Density-Aware Feature Embedding for Face
  Clustering}. In \bibinfo{booktitle}{\emph{Proceedings of the IEEE/CVF
  Conference on Computer Vision and Pattern Recognition}}.
  \bibinfo{pages}{6698--6706}.
\newblock


\bibitem[\protect\citeauthoryear{Guo and Zhang}{Guo and Zhang}{2017}]%
        {guo2017one}
\bibfield{author}{\bibinfo{person}{Y. Guo} {and} \bibinfo{person}{L. Zhang}.}
  \bibinfo{year}{2017}\natexlab{}.
\newblock \showarticletitle{One-shot face recognition by promoting
  underrepresented classes}.
\newblock  (\bibinfo{year}{2017}).
\newblock
\showeprint{1707.05574}


\bibitem[\protect\citeauthoryear{Guo, Zhang, Hu, He, and Gao}{Guo
  et~al\mbox{.}}{2016}]%
        {guo2016ms}
\bibfield{author}{\bibinfo{person}{Y. Guo}, \bibinfo{person}{L. Zhang},
  \bibinfo{person}{Y. Hu}, \bibinfo{person}{X. He}, {and} \bibinfo{person}{J.
  Gao}.} \bibinfo{year}{2016}\natexlab{}.
\newblock \showarticletitle{Ms-celeb-1m: A dataset and benchmark for
  large-scale face recognition}. In \bibinfo{booktitle}{\emph{Proceedings of
  the European Conference on Computer Vision}}. \bibinfo{pages}{87--102}.
\newblock


\bibitem[\protect\citeauthoryear{Han, Shan, Kan, Wu, and Chen}{Han
  et~al\mbox{.}}{2018}]%
        {Han2018FaceRW}
\bibfield{author}{\bibinfo{person}{C. Han}, \bibinfo{person}{S. Shan},
  \bibinfo{person}{M. Kan}, \bibinfo{person}{S. Wu}, {and} \bibinfo{person}{X.
  Chen}.} \bibinfo{year}{2018}\natexlab{}.
\newblock \showarticletitle{Face recognition with contrastive convolution}. In
  \bibinfo{booktitle}{\emph{Proceedings of the European Conference on Computer
  Vision}}. \bibinfo{pages}{118--134}.
\newblock


\bibitem[\protect\citeauthoryear{Hao, Liu, Qin, Yan, Li, and Hu}{Hao
  et~al\mbox{.}}{2017}]%
        {hao2017scale}
\bibfield{author}{\bibinfo{person}{Z. Hao}, \bibinfo{person}{Y. Liu},
  \bibinfo{person}{H. Qin}, \bibinfo{person}{J. Yan}, \bibinfo{person}{X. Li},
  {and} \bibinfo{person}{X. Hu}.} \bibinfo{year}{2017}\natexlab{}.
\newblock \showarticletitle{Scale-aware face detection}. In
  \bibinfo{booktitle}{\emph{Proceedings of the IEEE/CVF Conference on Computer
  Vision and Pattern Recognition}}. \bibinfo{pages}{6186--6195}.
\newblock


\bibitem[\protect\citeauthoryear{Hayat, Khan, Werghi, and Goecke}{Hayat
  et~al\mbox{.}}{2017}]%
        {Hayat2017JointRA}
\bibfield{author}{\bibinfo{person}{M. Hayat}, \bibinfo{person}{S.~H. Khan},
  \bibinfo{person}{N. Werghi}, {and} \bibinfo{person}{R. Goecke}.}
  \bibinfo{year}{2017}\natexlab{}.
\newblock \showarticletitle{Joint Registration and Representation Learning for
  Unconstrained Face Identification}. In \bibinfo{booktitle}{\emph{Proceedings
  of the IEEE/CVF Conference on Computer Vision and Pattern Recognition}}.
  \bibinfo{pages}{1551--1560}.
\newblock


\bibitem[\protect\citeauthoryear{He, Zhang, Ren, and Sun}{He
  et~al\mbox{.}}{2016}]%
        {He2016DeepRL}
\bibfield{author}{\bibinfo{person}{K. He}, \bibinfo{person}{X. Zhang},
  \bibinfo{person}{S. Ren}, {and} \bibinfo{person}{J. Sun}.}
  \bibinfo{year}{2016}\natexlab{}.
\newblock \showarticletitle{Deep Residual Learning for Image Recognition}. In
  \bibinfo{booktitle}{\emph{Proceedings of the IEEE/CVF Conference on Computer
  Vision and Pattern Recognition}}. \bibinfo{pages}{770--778}.
\newblock


\bibitem[\protect\citeauthoryear{He, Cao, Song, Sun, and Tan}{He
  et~al\mbox{.}}{2020}]%
        {He2020AdversarialCF}
\bibfield{author}{\bibinfo{person}{R. He}, \bibinfo{person}{J. Cao},
  \bibinfo{person}{L. Song}, \bibinfo{person}{Z. Sun}, {and}
  \bibinfo{person}{T. Tan}.} \bibinfo{year}{2020}\natexlab{}.
\newblock \showarticletitle{Adversarial Cross-Spectral Face Completion for
  NIR-VIS Face Recognition}.
\newblock \bibinfo{journal}{\emph{IEEE Trans. Pattern Anal. Mach. Intell.}}
  \bibinfo{volume}{42} (\bibinfo{year}{2020}), \bibinfo{pages}{1025--1037}.
\newblock


\bibitem[\protect\citeauthoryear{He, Wu, Sun, and Tan}{He
  et~al\mbox{.}}{2017}]%
        {He2017LearningID}
\bibfield{author}{\bibinfo{person}{R. He}, \bibinfo{person}{X. Wu},
  \bibinfo{person}{Z. Sun}, {and} \bibinfo{person}{T. Tan}.}
  \bibinfo{year}{2017}\natexlab{}.
\newblock \showarticletitle{Learning Invariant Deep Representation for NIR-VIS
  Face Recognition}. In \bibinfo{booktitle}{\emph{Proceedings of the AAAI
  Conference on Artificial Intelligence}}, Vol.~\bibinfo{volume}{33}.
  \bibinfo{pages}{9005--9012}.
\newblock


\bibitem[\protect\citeauthoryear{He, Wu, Sun, and Tan}{He
  et~al\mbox{.}}{2019}]%
        {He2019WassersteinCL}
\bibfield{author}{\bibinfo{person}{R. He}, \bibinfo{person}{X. Wu},
  \bibinfo{person}{Z. Sun}, {and} \bibinfo{person}{T. Tan}.}
  \bibinfo{year}{2019}\natexlab{}.
\newblock \showarticletitle{Wasserstein CNN: Learning Invariant Features for
  NIR-VIS Face Recognition}.
\newblock \bibinfo{journal}{\emph{IEEE Trans. Pattern Anal. Mach. Intell.}}
  \bibinfo{volume}{41} (\bibinfo{year}{2019}), \bibinfo{pages}{1761--1773}.
\newblock


\bibitem[\protect\citeauthoryear{Honari, Molchanov, Tyree, Vincent, Pal, and
  Kautz}{Honari et~al\mbox{.}}{2018}]%
        {Honari2018ImprovingLL}
\bibfield{author}{\bibinfo{person}{S. Honari}, \bibinfo{person}{P. Molchanov},
  \bibinfo{person}{S. Tyree}, \bibinfo{person}{P. Vincent},
  \bibinfo{person}{C.~J. Pal}, {and} \bibinfo{person}{J. Kautz}.}
  \bibinfo{year}{2018}\natexlab{}.
\newblock \showarticletitle{Improving Landmark Localization with
  Semi-Supervised Learning}. In \bibinfo{booktitle}{\emph{Proceedings of the
  IEEE/CVF Conference on Computer Vision and Pattern Recognition}}.
  \bibinfo{pages}{1546--1555}.
\newblock


\bibitem[\protect\citeauthoryear{Hong, Im, Ryu, and Yang}{Hong
  et~al\mbox{.}}{2017}]%
        {Hong2017SSPPDANDD}
\bibfield{author}{\bibinfo{person}{S. Hong}, \bibinfo{person}{W. Im},
  \bibinfo{person}{J.~B. Ryu}, {and} \bibinfo{person}{H. Yang}.}
  \bibinfo{year}{2017}\natexlab{}.
\newblock \showarticletitle{SSPP-DAN: Deep domain adaptation network for face
  recognition with single sample per person}. In
  \bibinfo{booktitle}{\emph{Proceedings of the IEEE International Conference on
  Image Processing}}. \bibinfo{pages}{825--829}.
\newblock


\bibitem[\protect\citeauthoryear{Howard, Zhu, Chen, Kalenichenko, Wang, Weyand,
  Andreetto, and Adam}{Howard et~al\mbox{.}}{2017}]%
        {Howard2017MobileNetsEC}
\bibfield{author}{\bibinfo{person}{A.~G. Howard}, \bibinfo{person}{M. Zhu},
  \bibinfo{person}{B. Chen}, \bibinfo{person}{D. Kalenichenko},
  \bibinfo{person}{W. Wang}, \bibinfo{person}{T. Weyand}, \bibinfo{person}{M.
  Andreetto}, {and} \bibinfo{person}{H. Adam}.}
  \bibinfo{year}{2017}\natexlab{}.
\newblock \showarticletitle{MobileNets: Efficient Convolutional Neural Networks
  for Mobile Vision Applications}.
\newblock  (\bibinfo{year}{2017}).
\newblock
\showeprint{abs/1704.04861}


\bibitem[\protect\citeauthoryear{Hu, Shen, and Sun}{Hu et~al\mbox{.}}{2018}]%
        {Hu2018SqueezeandExcitationN}
\bibfield{author}{\bibinfo{person}{J. Hu}, \bibinfo{person}{L. Shen}, {and}
  \bibinfo{person}{G. Sun}.} \bibinfo{year}{2018}\natexlab{}.
\newblock \showarticletitle{Squeeze-and-Excitation Networks}. In
  \bibinfo{booktitle}{\emph{Proceedings of the IEEE/CVF Conference on Computer
  Vision and Pattern Recognition}}. \bibinfo{pages}{7132--7141}.
\newblock


\bibitem[\protect\citeauthoryear{{Hu} and {Ramanan}}{{Hu} and
  {Ramanan}}{2017}]%
        {2017HR}
\bibfield{author}{\bibinfo{person}{P. {Hu}} {and} \bibinfo{person}{D.
  {Ramanan}}.} \bibinfo{year}{2017}\natexlab{}.
\newblock \showarticletitle{Finding tiny faces}. In
  \bibinfo{booktitle}{\emph{Proceedings of the IEEE/CVF Conference on Computer
  Vision and Pattern Recognition}}. \bibinfo{pages}{1522--1530}.
\newblock


\bibitem[\protect\citeauthoryear{Hu, Huang, Zhang, and Li}{Hu
  et~al\mbox{.}}{2019}]%
        {Hu2019NoiseTolerantPF}
\bibfield{author}{\bibinfo{person}{W. Hu}, \bibinfo{person}{Y. Huang},
  \bibinfo{person}{F. Zhang}, {and} \bibinfo{person}{R. Li}.}
  \bibinfo{year}{2019}\natexlab{}.
\newblock \showarticletitle{Noise-Tolerant Paradigm for Training Face
  Recognition CNNs}. In \bibinfo{booktitle}{\emph{Proceedings of the IEEE/CVF
  Conference on Computer Vision and Pattern Recognition}}.
  \bibinfo{pages}{11879--11888}.
\newblock


\bibitem[\protect\citeauthoryear{{Hu}, {Wu}, {Yu}, {He}, and {Sun}}{{Hu}
  et~al\mbox{.}}{2018}]%
        {Hu2018CAPG}
\bibfield{author}{\bibinfo{person}{Y. {Hu}}, \bibinfo{person}{X. {Wu}},
  \bibinfo{person}{B. {Yu}}, \bibinfo{person}{R. {He}}, {and}
  \bibinfo{person}{Z. {Sun}}.} \bibinfo{year}{2018}\natexlab{}.
\newblock \showarticletitle{Pose-Guided Photorealistic Face Rotation}. In
  \bibinfo{booktitle}{\emph{Proceedings of the IEEE/CVF Conference on Computer
  Vision and Pattern Recognition}}. \bibinfo{pages}{8398--8406}.
\newblock


\bibitem[\protect\citeauthoryear{Huang, Ramesh, Berg, and Learned-Miller}{Huang
  et~al\mbox{.}}{2007}]%
        {LFWTech}
\bibfield{author}{\bibinfo{person}{G.~B. Huang}, \bibinfo{person}{M. Ramesh},
  \bibinfo{person}{T. Berg}, {and} \bibinfo{person}{E. Learned-Miller}.}
  \bibinfo{year}{2007}\natexlab{}.
\newblock \bibinfo{booktitle}{\emph{Labeled Faces in the Wild: A Database for
  Studying Face Recognition in Unconstrained Environments}}.
\newblock \bibinfo{type}{{T}echnical {R}eport} 07-49.
  \bibinfo{institution}{University of Massachusetts, Amherst}.
\newblock


\bibitem[\protect\citeauthoryear{Huang, Yang, Deng, and Yu}{Huang
  et~al\mbox{.}}{2015}]%
        {huang2015densebox}
\bibfield{author}{\bibinfo{person}{L. Huang}, \bibinfo{person}{Y. Yang},
  \bibinfo{person}{Y. Deng}, {and} \bibinfo{person}{Y. Yu}.}
  \bibinfo{year}{2015}\natexlab{}.
\newblock \showarticletitle{Densebox: Unifying landmark localization with end
  to end object detection}.
\newblock  (\bibinfo{year}{2015}).
\newblock
\showeprint{1509.04874}


\bibitem[\protect\citeauthoryear{{Huang}, {Zhang}, {Li}, and {He}}{{Huang}
  et~al\mbox{.}}{2017}]%
        {Huang2017tpgan}
\bibfield{author}{\bibinfo{person}{R. {Huang}}, \bibinfo{person}{S. {Zhang}},
  \bibinfo{person}{T. {Li}}, {and} \bibinfo{person}{R. {He}}.}
  \bibinfo{year}{2017}\natexlab{}.
\newblock \showarticletitle{Beyond Face Rotation: Global and Local Perception
  GAN for Photorealistic and Identity Preserving Frontal View Synthesis}. In
  \bibinfo{booktitle}{\emph{Proceedings of the IEEE International Conference on
  Computer Vision}}. \bibinfo{pages}{2458--2467}.
\newblock


\bibitem[\protect\citeauthoryear{Huang, Deng, Shen, Zhang, and Ye}{Huang
  et~al\mbox{.}}{2020a}]%
        {Huang_2020_PropagationNet}
\bibfield{author}{\bibinfo{person}{X. Huang}, \bibinfo{person}{W. Deng},
  \bibinfo{person}{H. Shen}, \bibinfo{person}{X. Zhang}, {and}
  \bibinfo{person}{J. Ye}.} \bibinfo{year}{2020}\natexlab{a}.
\newblock \showarticletitle{PropagationNet: Propagate Points to Curve to Learn
  Structure Information}. In \bibinfo{booktitle}{\emph{Proceedings of the
  IEEE/CVF Conference on Computer Vision and Pattern Recognition}}.
  \bibinfo{pages}{7265--7274}.
\newblock


\bibitem[\protect\citeauthoryear{Huang, Wang, Tai, Liu, Shen, Li, Li, and
  Huang}{Huang et~al\mbox{.}}{2020b}]%
        {Huang2020CurricularFaceAC}
\bibfield{author}{\bibinfo{person}{Y. Huang}, \bibinfo{person}{Y. Wang},
  \bibinfo{person}{Y. Tai}, \bibinfo{person}{X. Liu}, \bibinfo{person}{P.
  Shen}, \bibinfo{person}{S. Li}, \bibinfo{person}{J. Li}, {and}
  \bibinfo{person}{F. Huang}.} \bibinfo{year}{2020}\natexlab{b}.
\newblock \showarticletitle{Curricularface: adaptive curriculum learning loss
  for deep face recognition}. In \bibinfo{booktitle}{\emph{Proceedings of the
  IEEE/CVF Conference on Computer Vision and Pattern Recognition}}.
  \bibinfo{pages}{5901--5910}.
\newblock


\bibitem[\protect\citeauthoryear{H.Wang, Li, Ji, and Wang}{H.Wang
  et~al\mbox{.}}{2017}]%
        {wang2017facercnn}
\bibfield{author}{\bibinfo{person}{H.Wang}, \bibinfo{person}{Z. Li},
  \bibinfo{person}{X. Ji}, {and} \bibinfo{person}{Y. Wang}.}
  \bibinfo{year}{2017}\natexlab{}.
\newblock \showarticletitle{Face r-cnn}.
\newblock  (\bibinfo{year}{2017}).
\newblock
\showeprint{1706.01061}


\bibitem[\protect\citeauthoryear{Iandola, Moskewicz, Ashraf, Han, Dally, and
  Keutzer}{Iandola et~al\mbox{.}}{2017}]%
        {Iandola2017SqueezeNetAA}
\bibfield{author}{\bibinfo{person}{F.~N. Iandola}, \bibinfo{person}{M.~W.
  Moskewicz}, \bibinfo{person}{K. Ashraf}, \bibinfo{person}{S. Han},
  \bibinfo{person}{W.~J. Dally}, {and} \bibinfo{person}{K. Keutzer}.}
  \bibinfo{year}{2017}\natexlab{}.
\newblock \showarticletitle{SqueezeNet: AlexNet-level accuracy with 50x fewer
  parameters and <1MB model size}.
\newblock  (\bibinfo{year}{2017}).
\newblock
\showeprint{1602.07360}


\bibitem[\protect\citeauthoryear{Jaderberg, Simonyan, Zisserman, and
  Kavukcuoglu}{Jaderberg et~al\mbox{.}}{2015}]%
        {Jaderberg2015SpatialTN}
\bibfield{author}{\bibinfo{person}{M. Jaderberg}, \bibinfo{person}{K.
  Simonyan}, \bibinfo{person}{A. Zisserman}, {and} \bibinfo{person}{K.
  Kavukcuoglu}.} \bibinfo{year}{2015}\natexlab{}.
\newblock \showarticletitle{Spatial transformer networks}. In
  \bibinfo{booktitle}{\emph{Advances in neural information processing
  systems}}. \bibinfo{pages}{2017--2025}.
\newblock


\bibitem[\protect\citeauthoryear{Jain, Deb, and Engelsma}{Jain
  et~al\mbox{.}}{2021}]%
        {Jain2021BiometricsTB}
\bibfield{author}{\bibinfo{person}{Anil~K. Jain}, \bibinfo{person}{Debayan
  Deb}, {and} \bibinfo{person}{Joshua~J. Engelsma}.}
  \bibinfo{year}{2021}\natexlab{}.
\newblock \showarticletitle{Biometrics: Trust, but Verify}.
\newblock \bibinfo{journal}{\emph{ArXiv}}  \bibinfo{volume}{abs/2105.06625}
  (\bibinfo{year}{2021}).
\newblock


\bibitem[\protect\citeauthoryear{Jiang and Learned-Miller}{Jiang and
  Learned-Miller}{2017}]%
        {Jiang2017650}
\bibfield{author}{\bibinfo{person}{H. Jiang} {and} \bibinfo{person}{E.
  Learned-Miller}.} \bibinfo{year}{2017}\natexlab{}.
\newblock \showarticletitle{Face detection with the faster r-cnn}. In
  \bibinfo{booktitle}{\emph{Proceedings of the IEEE International Conference on
  Automatic Face $\&$ Gesture Recognition}}. \bibinfo{pages}{650--657}.
\newblock


\bibitem[\protect\citeauthoryear{Jin, Zhang, Zhu, Tang, Lei, and Li}{Jin
  et~al\mbox{.}}{2019}]%
        {2019fbi}
\bibfield{author}{\bibinfo{person}{H. Jin}, \bibinfo{person}{S. Zhang},
  \bibinfo{person}{X.u Zhu}, \bibinfo{person}{Y. Tang}, \bibinfo{person}{Z.
  Lei}, {and} \bibinfo{person}{S.~Z. Li}.} \bibinfo{year}{2019}\natexlab{}.
\newblock \showarticletitle{Learning Lightweight Face Detector with Knowledge
  Distillation}. In \bibinfo{booktitle}{\emph{Proceedings of the International
  Conference on Biometrics}}. \bibinfo{pages}{1--7}.
\newblock


\bibitem[\protect\citeauthoryear{Jin and Tan}{Jin and Tan}{2017}]%
        {Jin2017FaceAI}
\bibfield{author}{\bibinfo{person}{X. Jin} {and} \bibinfo{person}{X. Tan}.}
  \bibinfo{year}{2017}\natexlab{}.
\newblock \showarticletitle{Face alignment in-the-wild: A survey}.
\newblock \bibinfo{journal}{\emph{Computer Vision and Image Understanding}}
  \bibinfo{volume}{162} (\bibinfo{year}{2017}), \bibinfo{pages}{1--22}.
\newblock


\bibitem[\protect\citeauthoryear{Jing, Liu, and Zhang}{Jing
  et~al\mbox{.}}{2017}]%
        {Jing2017Stacked}
\bibfield{author}{\bibinfo{person}{Y. Jing}, \bibinfo{person}{Q. Liu}, {and}
  \bibinfo{person}{K. Zhang}.} \bibinfo{year}{2017}\natexlab{}.
\newblock \showarticletitle{Stacked Hourglass Network for Robust Facial
  Landmark Localisation}. In \bibinfo{booktitle}{\emph{Proceedings of the
  IEEE/CVF Conference on Computer Vision and Pattern Recognition Workshops}}.
  \bibinfo{pages}{79--87}.
\newblock


\bibitem[\protect\citeauthoryear{{Jourabloo} and {Liu}}{{Jourabloo} and
  {Liu}}{2016}]%
        {Jourabloo2016_D3PF}
\bibfield{author}{\bibinfo{person}{A. {Jourabloo}} {and} \bibinfo{person}{X.
  {Liu}}.} \bibinfo{year}{2016}\natexlab{}.
\newblock \showarticletitle{Large-Pose Face Alignment via CNN-Based Dense 3D
  Model Fitting}. In \bibinfo{booktitle}{\emph{Proceedings of the IEEE/CVF
  Conference on Computer Vision and Pattern Recognition}}.
  \bibinfo{pages}{4188--4196}.
\newblock


\bibitem[\protect\citeauthoryear{{Jourabloo}, {Ye}, {Liu}, and
  {Ren}}{{Jourabloo} et~al\mbox{.}}{2017}]%
        {Jourabloo2017}
\bibfield{author}{\bibinfo{person}{A. {Jourabloo}}, \bibinfo{person}{M. {Ye}},
  \bibinfo{person}{X. {Liu}}, {and} \bibinfo{person}{L. {Ren}}.}
  \bibinfo{year}{2017}\natexlab{}.
\newblock \showarticletitle{Pose-Invariant Face Alignment with a Single CNN}.
  In \bibinfo{booktitle}{\emph{Proceedings of the IEEE International Conference
  on Computer Vision}}. \bibinfo{pages}{3219--3228}.
\newblock


\bibitem[\protect\citeauthoryear{Kalka, Maze, Duncan, OrConnor, Elliott,
  Hebert, Bryan, and Jain}{Kalka et~al\mbox{.}}{2018}]%
        {Kalka2018IJBSIJ}
\bibfield{author}{\bibinfo{person}{N.~D. Kalka}, \bibinfo{person}{B. Maze},
  \bibinfo{person}{J.~A. Duncan}, \bibinfo{person}{K. OrConnor},
  \bibinfo{person}{S. Elliott}, \bibinfo{person}{K. Hebert},
  \bibinfo{person}{J. Bryan}, {and} \bibinfo{person}{A.~K. Jain}.}
  \bibinfo{year}{2018}\natexlab{}.
\newblock \showarticletitle{IJB–S: IARPA Janus Surveillance Video Benchmark}.
  In \bibinfo{booktitle}{\emph{Proceedings of the IEEE International Conference
  on Biometrics Theory, Applications and Systems (BTAS)}}.
  \bibinfo{pages}{1--9}.
\newblock


\bibitem[\protect\citeauthoryear{Kan, Shan, and Chen}{Kan
  et~al\mbox{.}}{2016}]%
        {kan2016multi}
\bibfield{author}{\bibinfo{person}{M. Kan}, \bibinfo{person}{S. Shan}, {and}
  \bibinfo{person}{X. Chen}.} \bibinfo{year}{2016}\natexlab{}.
\newblock \showarticletitle{Multi-view deep network for cross-view
  classification}. In \bibinfo{booktitle}{\emph{Proceedings of the IEEE/CVF
  Conference on Computer Vision and Pattern Recognition}}.
  \bibinfo{pages}{4847--4855}.
\newblock


\bibitem[\protect\citeauthoryear{Kang, Kim, Jun, and Kim}{Kang
  et~al\mbox{.}}{2019}]%
        {Kang2019AttentionalFR}
\bibfield{author}{\bibinfo{person}{B. Kang}, \bibinfo{person}{Y. Kim},
  \bibinfo{person}{B. Jun}, {and} \bibinfo{person}{D. Kim}.}
  \bibinfo{year}{2019}\natexlab{}.
\newblock \showarticletitle{Attentional Feature-Pair Relation Networks for
  Accurate Face Recognition}. In \bibinfo{booktitle}{\emph{Proceedings of the
  IEEE International Conference on Computer Vision}}.
  \bibinfo{pages}{5471--5480}.
\newblock


\bibitem[\protect\citeauthoryear{Kang, Kim, and Kim}{Kang
  et~al\mbox{.}}{2018}]%
        {Kang2018PairwiseRN}
\bibfield{author}{\bibinfo{person}{B. Kang}, \bibinfo{person}{Y. Kim}, {and}
  \bibinfo{person}{D. Kim}.} \bibinfo{year}{2018}\natexlab{}.
\newblock \showarticletitle{Pairwise relational networks for face recognition}.
  In \bibinfo{booktitle}{\emph{Proceedings of the European Conference on
  Computer Vision}}. \bibinfo{pages}{628--645}.
\newblock


\bibitem[\protect\citeauthoryear{Kemelmacher-Shlizerman, Seitz, Miller, and
  Brossard}{Kemelmacher-Shlizerman et~al\mbox{.}}{2016}]%
        {kemelmacher2016megaface}
\bibfield{author}{\bibinfo{person}{I. Kemelmacher-Shlizerman},
  \bibinfo{person}{S.~M. Seitz}, \bibinfo{person}{D. Miller}, {and}
  \bibinfo{person}{E. Brossard}.} \bibinfo{year}{2016}\natexlab{}.
\newblock \showarticletitle{The megaface benchmark: 1 million faces for
  recognition at scale}. In \bibinfo{booktitle}{\emph{Proceedings of the
  IEEE/CVF Conference on Computer Vision and Pattern Recognition}}.
  \bibinfo{pages}{4873--4882}.
\newblock


\bibitem[\protect\citeauthoryear{Kemelmacher-Shlizerman, Suwajanakorn, and
  Seitz}{Kemelmacher-Shlizerman et~al\mbox{.}}{2014}]%
        {KemelmacherShlizerman2014IlluminationAwareAP}
\bibfield{author}{\bibinfo{person}{I. Kemelmacher-Shlizerman},
  \bibinfo{person}{S. Suwajanakorn}, {and} \bibinfo{person}{S. Seitz}.}
  \bibinfo{year}{2014}\natexlab{}.
\newblock \showarticletitle{Illumination-Aware Age Progression}. In
  \bibinfo{booktitle}{\emph{Proceedings of the IEEE/CVF Conference on Computer
  Vision and Pattern Recognition}}. \bibinfo{pages}{3334--3341}.
\newblock


\bibitem[\protect\citeauthoryear{Kim, Park, Roh, and Shin}{Kim
  et~al\mbox{.}}{2020}]%
        {Kim2020GroupFaceLL}
\bibfield{author}{\bibinfo{person}{Y. Kim}, \bibinfo{person}{W. Park},
  \bibinfo{person}{M.l Roh}, {and} \bibinfo{person}{J. Shin}.}
  \bibinfo{year}{2020}\natexlab{}.
\newblock \showarticletitle{GroupFace: Learning Latent Groups and Constructing
  Group-based Representations for Face Recognition}. In
  \bibinfo{booktitle}{\emph{Proceedings of the IEEE/CVF Conference on Computer
  Vision and Pattern Recognition}}. \bibinfo{pages}{5621--5630}.
\newblock


\bibitem[\protect\citeauthoryear{Klare, Klein, Taborsky, Blanton, Cheney,
  Allen, Grother, Mah, Burge, and Jain}{Klare et~al\mbox{.}}{2015}]%
        {Klare2015PushingTF}
\bibfield{author}{\bibinfo{person}{B. Klare}, \bibinfo{person}{B. Klein},
  \bibinfo{person}{E. Taborsky}, \bibinfo{person}{A. Blanton},
  \bibinfo{person}{J. Cheney}, \bibinfo{person}{K.~E. Allen},
  \bibinfo{person}{P. Grother}, \bibinfo{person}{A. Mah}, \bibinfo{person}{M.
  Burge}, {and} \bibinfo{person}{A.~K. Jain}.} \bibinfo{year}{2015}\natexlab{}.
\newblock \showarticletitle{Pushing the frontiers of unconstrained face
  detection and recognition: IARPA Janus Benchmark A}. In
  \bibinfo{booktitle}{\emph{Proceedings of the IEEE/CVF Conference on Computer
  Vision and Pattern Recognition}}. \bibinfo{pages}{1931--1939}.
\newblock


\bibitem[\protect\citeauthoryear{Kostinger, Wohlhart, Roth, and
  Bischof}{Kostinger et~al\mbox{.}}{2011}]%
        {Kostinger2011alfw}
\bibfield{author}{\bibinfo{person}{M. Kostinger}, \bibinfo{person}{P.
  Wohlhart}, \bibinfo{person}{P.~M. Roth}, {and} \bibinfo{person}{H. Bischof}.}
  \bibinfo{year}{2011}\natexlab{}.
\newblock \showarticletitle{Annotated Facial Landmarks in the Wild: A
  large-scale, real-world database for facial landmark localization}. In
  \bibinfo{booktitle}{\emph{Proceedings of the IEEE International Conference on
  Computer Vision Workshops}}. \bibinfo{pages}{2144--2151}.
\newblock


\bibitem[\protect\citeauthoryear{Krizhevsky, Sutskever, and Hinton}{Krizhevsky
  et~al\mbox{.}}{2012}]%
        {Krizhevsky2012ImageNetCW}
\bibfield{author}{\bibinfo{person}{A. Krizhevsky}, \bibinfo{person}{I.
  Sutskever}, {and} \bibinfo{person}{G.~E. Hinton}.}
  \bibinfo{year}{2012}\natexlab{}.
\newblock \showarticletitle{ImageNet Classification with Deep Convolutional
  Neural Networks}. In \bibinfo{booktitle}{\emph{Advances in Neural Information
  Processing Systems}}. \bibinfo{pages}{1097--1105}.
\newblock


\bibitem[\protect\citeauthoryear{Kumar and Chellappa}{Kumar and
  Chellappa}{2018}]%
        {Kumar2018Disentangling3P}
\bibfield{author}{\bibinfo{person}{A. Kumar} {and} \bibinfo{person}{R.
  Chellappa}.} \bibinfo{year}{2018}\natexlab{}.
\newblock \showarticletitle{Disentangling 3D Pose in a Dendritic CNN for
  Unconstrained 2D Face Alignment}. In \bibinfo{booktitle}{\emph{Proceedings of
  the IEEE/CVF Conference on Computer Vision and Pattern Recognition}}.
  \bibinfo{pages}{430--439}.
\newblock


\bibitem[\protect\citeauthoryear{Kumar, Marks, Mou, Wang, Jones, Cherian,
  Koike-Akino, Liu, and Feng}{Kumar et~al\mbox{.}}{2020}]%
        {Kumar2020LUVLiFA}
\bibfield{author}{\bibinfo{person}{A.n Kumar}, \bibinfo{person}{T.K. Marks},
  \bibinfo{person}{W. Mou}, \bibinfo{person}{Y. Wang}, \bibinfo{person}{M.
  Jones}, \bibinfo{person}{A. Cherian}, \bibinfo{person}{T. Koike-Akino},
  \bibinfo{person}{X. Liu}, {and} \bibinfo{person}{C. Feng}.}
  \bibinfo{year}{2020}\natexlab{}.
\newblock \showarticletitle{LUVLi face alignment: estimating landmarks'
  location, uncertainty, and visibility likelihood}. In
  \bibinfo{booktitle}{\emph{Proceedings of the IEEE/CVF Conference on Computer
  Vision and Pattern Recognition}}. \bibinfo{pages}{8236--8246}.
\newblock


\bibitem[\protect\citeauthoryear{Kushwaha, Singh, Singh, Vatsa, Ratha, and
  Chellappa}{Kushwaha et~al\mbox{.}}{2018}]%
        {Kushwaha2018DisguisedFI}
\bibfield{author}{\bibinfo{person}{V. Kushwaha}, \bibinfo{person}{M. Singh},
  \bibinfo{person}{R. Singh}, \bibinfo{person}{M. Vatsa},
  \bibinfo{person}{N.~K. Ratha}, {and} \bibinfo{person}{R. Chellappa}.}
  \bibinfo{year}{2018}\natexlab{}.
\newblock \showarticletitle{Disguised Faces in the Wild}. In
  \bibinfo{booktitle}{\emph{Proceedings of the IEEE/CVF Conference on Computer
  Vision and Pattern Recognition Workshops}}. \bibinfo{pages}{1--18}.
\newblock


\bibitem[\protect\citeauthoryear{Law and Deng}{Law and Deng}{2018}]%
        {Law2018CornerNet}
\bibfield{author}{\bibinfo{person}{H. Law} {and} \bibinfo{person}{J. Deng}.}
  \bibinfo{year}{2018}\natexlab{}.
\newblock \showarticletitle{Cornernet: Detecting objects as paired keypoints}.
  In \bibinfo{booktitle}{\emph{Proceedings of the European Conference on
  Computer Vision}}. \bibinfo{pages}{734--750}.
\newblock


\bibitem[\protect\citeauthoryear{Le, Brandt, Lin, Bourdev, and Huang}{Le
  et~al\mbox{.}}{2012}]%
        {Le2012InteractiveFF}
\bibfield{author}{\bibinfo{person}{V. Le}, \bibinfo{person}{J. Brandt},
  \bibinfo{person}{Z.~L. Lin}, \bibinfo{person}{L.~D. Bourdev}, {and}
  \bibinfo{person}{T.~S. Huang}.} \bibinfo{year}{2012}\natexlab{}.
\newblock \showarticletitle{Interactive Facial Feature Localization}. In
  \bibinfo{booktitle}{\emph{Proceedings of the European Conference on Computer
  Vision}}. \bibinfo{pages}{679--692}.
\newblock


\bibitem[\protect\citeauthoryear{Lezama, Qiu, and Sapiro}{Lezama
  et~al\mbox{.}}{2017}]%
        {Lezama2017NotAO}
\bibfield{author}{\bibinfo{person}{J. Lezama}, \bibinfo{person}{Q. Qiu}, {and}
  \bibinfo{person}{G. Sapiro}.} \bibinfo{year}{2017}\natexlab{}.
\newblock \showarticletitle{Not Afraid of the Dark: NIR-VIS Face Recognition
  via Cross-Spectral Hallucination and Low-Rank Embedding}. In
  \bibinfo{booktitle}{\emph{Proceedings of the IEEE/CVF Conference on Computer
  Vision and Pattern Recognition}}. \bibinfo{pages}{6807--6816}.
\newblock


\bibitem[\protect\citeauthoryear{{Li}, {Lin}, {Shen}, {Brandt}, and {Hua}}{{Li}
  et~al\mbox{.}}{2015}]%
        {Cascade_CNN}
\bibfield{author}{\bibinfo{person}{H. {Li}}, \bibinfo{person}{Z. {Lin}},
  \bibinfo{person}{X. {Shen}}, \bibinfo{person}{J. {Brandt}}, {and}
  \bibinfo{person}{G. {Hua}}.} \bibinfo{year}{2015}\natexlab{}.
\newblock \showarticletitle{A convolutional neural network cascade for face
  detection}. In \bibinfo{booktitle}{\emph{Proceedings of the IEEE/CVF
  Conference on Computer Vision and Pattern Recognition}}.
  \bibinfo{pages}{5325--5334}.
\newblock


\bibitem[\protect\citeauthoryear{Li, Wang, Wang, Tai, Qian, Yang, Wang, Li, and
  Huang}{Li et~al\mbox{.}}{2019c}]%
        {2019DSFD}
\bibfield{author}{\bibinfo{person}{J. Li}, \bibinfo{person}{Y. Wang},
  \bibinfo{person}{C. Wang}, \bibinfo{person}{Y. Tai}, \bibinfo{person}{J.
  Qian}, \bibinfo{person}{J. Yang}, \bibinfo{person}{C. Wang},
  \bibinfo{person}{J. Li}, {and} \bibinfo{person}{F. Huang}.}
  \bibinfo{year}{2019}\natexlab{c}.
\newblock \showarticletitle{DSFD: dual shot face detector}. In
  \bibinfo{booktitle}{\emph{Proceedings of the IEEE/CVF Conference on Computer
  Vision and Pattern Recognition}}. \bibinfo{pages}{5060--5069}.
\newblock


\bibitem[\protect\citeauthoryear{Li, Li, Cui, and Zha}{Li
  et~al\mbox{.}}{2019a}]%
        {Li2019PoseAwareFA}
\bibfield{author}{\bibinfo{person}{S. Li}, \bibinfo{person}{H. Li},
  \bibinfo{person}{J. Cui}, {and} \bibinfo{person}{H. Zha}.}
  \bibinfo{year}{2019}\natexlab{a}.
\newblock \showarticletitle{Pose-Aware Face Alignment based on CNN and 3DMM}.
  In \bibinfo{booktitle}{\emph{Proceedings of the British Machine Vision
  Conference}}. \bibinfo{pages}{106}.
\newblock


\bibitem[\protect\citeauthoryear{Li, Yi, Lei, and Liao}{Li
  et~al\mbox{.}}{2013}]%
        {Li2013TheCN}
\bibfield{author}{\bibinfo{person}{S.~Z. Li}, \bibinfo{person}{D. Yi},
  \bibinfo{person}{Z. Lei}, {and} \bibinfo{person}{S. Liao}.}
  \bibinfo{year}{2013}\natexlab{}.
\newblock \showarticletitle{The CASIA NIR-VIS 2.0 Face Database}. In
  \bibinfo{booktitle}{\emph{Proceedings of IEEE Conference on Computer Vision
  and Pattern RecognitionWorkshops}}. \bibinfo{pages}{348--353}.
\newblock


\bibitem[\protect\citeauthoryear{Li, Zhu, Zhang, Blake, Zhang, and Shum}{Li
  et~al\mbox{.}}{2002}]%
        {Li2002Statistical}
\bibfield{author}{\bibinfo{person}{S.~Z. Li}, \bibinfo{person}{L. Zhu},
  \bibinfo{person}{Z. Zhang}, \bibinfo{person}{A. Blake}, \bibinfo{person}{H.
  Zhang}, {and} \bibinfo{person}{H. Shum}.} \bibinfo{year}{2002}\natexlab{}.
\newblock \showarticletitle{Statistical Learning of Multi-view Face Detection}.
  In \bibinfo{booktitle}{\emph{Proceedings of the European Conference on
  Computer Vision}}. \bibinfo{pages}{67--81}.
\newblock


\bibitem[\protect\citeauthoryear{Li, Sun, Wu, and Wang}{Li
  et~al\mbox{.}}{2016}]%
        {Li2016FaceDW}
\bibfield{author}{\bibinfo{person}{Y. Li}, \bibinfo{person}{B. Sun},
  \bibinfo{person}{T. Wu}, {and} \bibinfo{person}{Y. Wang}.}
  \bibinfo{year}{{2016}}\natexlab{}.
\newblock \showarticletitle{Face detection with end-to-end integration of a
  convNet and a 3d model}. In \bibinfo{booktitle}{\emph{Proceedings of the
  European Conference on Computer Vision}}. \bibinfo{pages}{{420--436}}.
\newblock


\bibitem[\protect\citeauthoryear{Li, Tang, Han, Liu, and He}{Li
  et~al\mbox{.}}{2019b}]%
        {li2019pyramidbox}
\bibfield{author}{\bibinfo{person}{Z. Li}, \bibinfo{person}{X. Tang},
  \bibinfo{person}{J. Han}, \bibinfo{person}{J. Liu}, {and} \bibinfo{person}{R.
  He}.} \bibinfo{year}{2019}\natexlab{b}.
\newblock \showarticletitle{PyramidBox++: High Performance Detector for Finding
  Tiny Face}.
\newblock  (\bibinfo{year}{2019}).
\newblock
\showeprint{1904.00386}


\bibitem[\protect\citeauthoryear{Liao, Lei, Yi, and Li}{Liao
  et~al\mbox{.}}{2014}]%
        {Liao2014ABS}
\bibfield{author}{\bibinfo{person}{S. Liao}, \bibinfo{person}{Z. Lei},
  \bibinfo{person}{D. Yi}, {and} \bibinfo{person}{S.~Z. Li}.}
  \bibinfo{year}{2014}\natexlab{}.
\newblock \showarticletitle{A benchmark study of large-scale unconstrained face
  recognition}. In \bibinfo{booktitle}{\emph{Proceedings of the IEEE
  International Joint Conference on Biometrics}}. \bibinfo{pages}{1--8}.
\newblock


\bibitem[\protect\citeauthoryear{Lin, Dollár, Girshick, He, Hariharan, and
  Belongie}{Lin et~al\mbox{.}}{2017a}]%
        {lin2016feature}
\bibfield{author}{\bibinfo{person}{T. Lin}, \bibinfo{person}{P. Dollár},
  \bibinfo{person}{R. Girshick}, \bibinfo{person}{K. He}, \bibinfo{person}{B.
  Hariharan}, {and} \bibinfo{person}{S. Belongie}.}
  \bibinfo{year}{2017}\natexlab{a}.
\newblock \showarticletitle{Feature pyramid networks for object detection}. In
  \bibinfo{booktitle}{\emph{Proceedings of the IEEE/CVF Conference on Computer
  Vision and Pattern Recognition}}. \bibinfo{pages}{2117--2125}.
\newblock


\bibitem[\protect\citeauthoryear{Lin, Goyal, Girshick, He, and Doll{\'a}r}{Lin
  et~al\mbox{.}}{2017b}]%
        {lin2017focal}
\bibfield{author}{\bibinfo{person}{T. Lin}, \bibinfo{person}{P. Goyal},
  \bibinfo{person}{R. Girshick}, \bibinfo{person}{K. He}, {and}
  \bibinfo{person}{P. Doll{\'a}r}.} \bibinfo{year}{2017}\natexlab{b}.
\newblock \showarticletitle{Focal loss for dense object detection}. In
  \bibinfo{booktitle}{\emph{Proceedings of the IEEE International Conference on
  Computer Vision}}. \bibinfo{pages}{2980--2988}.
\newblock


\bibitem[\protect\citeauthoryear{Lin, Chen, Castillo, and Chellappa}{Lin
  et~al\mbox{.}}{2018}]%
        {lin2018deep}
\bibfield{author}{\bibinfo{person}{W. Lin}, \bibinfo{person}{J. Chen},
  \bibinfo{person}{C.~D. Castillo}, {and} \bibinfo{person}{R. Chellappa}.}
  \bibinfo{year}{2018}\natexlab{}.
\newblock \showarticletitle{Deep Density Clustering of Unconstrained Faces}. In
  \bibinfo{booktitle}{\emph{Proceedings of the IEEE/CVF Conference on Computer
  Vision and Pattern Recognition}}. \bibinfo{pages}{8128--8137}.
\newblock


\bibitem[\protect\citeauthoryear{{Liu}, {Deng}, {Zhong}, {Wang}, {Hu}, {Tao},
  and {Huang}}{{Liu} et~al\mbox{.}}{2019}]%
        {Fair_Loss}
\bibfield{author}{\bibinfo{person}{B. {Liu}}, \bibinfo{person}{W. {Deng}},
  \bibinfo{person}{Y. {Zhong}}, \bibinfo{person}{M. {Wang}},
  \bibinfo{person}{J. {Hu}}, \bibinfo{person}{X. {Tao}}, {and}
  \bibinfo{person}{Y. {Huang}}.} \bibinfo{year}{2019}\natexlab{}.
\newblock \showarticletitle{Fair Loss: Margin-Aware Reinforcement Learning for
  Deep Face Recognition}. In \bibinfo{booktitle}{\emph{Proceedings of the IEEE
  International Conference on Computer Vision}}. \bibinfo{pages}{10051--10060}.
\newblock


\bibitem[\protect\citeauthoryear{Liu and Wechsler}{Liu and Wechsler}{2002}]%
        {Liu2002GaborFB}
\bibfield{author}{\bibinfo{person}{C. Liu} {and} \bibinfo{person}{H.
  Wechsler}.} \bibinfo{year}{2002}\natexlab{}.
\newblock \showarticletitle{Gabor feature based classification using the
  enhanced fisher linear discriminant model for face recognition}.
\newblock \bibinfo{journal}{\emph{Trans. Image Process.}}  \bibinfo{volume}{11
  4} (\bibinfo{year}{2002}), \bibinfo{pages}{467--76}.
\newblock


\bibitem[\protect\citeauthoryear{Liu, Lu, Feng, and Zhou}{Liu
  et~al\mbox{.}}{2018b}]%
        {Liu2018TwoStreamTN}
\bibfield{author}{\bibinfo{person}{H. Liu}, \bibinfo{person}{J. Lu},
  \bibinfo{person}{J. Feng}, {and} \bibinfo{person}{J. Zhou}.}
  \bibinfo{year}{2018}\natexlab{b}.
\newblock \showarticletitle{Two-Stream Transformer Networks for Video-Based
  Face Alignment}.
\newblock \bibinfo{journal}{\emph{IEEE Trans. Pattern Anal. Mach. Intell.}}
  \bibinfo{volume}{40} (\bibinfo{year}{2018}), \bibinfo{pages}{2546--2554}.
\newblock


\bibitem[\protect\citeauthoryear{Liu, Lu, Guo, Wu, and Zhou}{Liu
  et~al\mbox{.}}{2020a}]%
        {Liu2020LearningRN}
\bibfield{author}{\bibinfo{person}{H. Liu}, \bibinfo{person}{J. Lu},
  \bibinfo{person}{M. Guo}, \bibinfo{person}{S. Wu}, {and} \bibinfo{person}{J.
  Zhou}.} \bibinfo{year}{2020}\natexlab{a}.
\newblock \showarticletitle{Learning Reasoning-Decision Networks for Robust
  Face Alignment}.
\newblock \bibinfo{journal}{\emph{IEEE Trans. Pattern Anal. Mach. Intell.}}
  \bibinfo{volume}{42} (\bibinfo{year}{2020}), \bibinfo{pages}{679--693}.
\newblock


\bibitem[\protect\citeauthoryear{Liu, Zhu, Lei, and Li}{Liu
  et~al\mbox{.}}{2019c}]%
        {liu2019adaptiveface}
\bibfield{author}{\bibinfo{person}{H. Liu}, \bibinfo{person}{X. Zhu},
  \bibinfo{person}{Z. Lei}, {and} \bibinfo{person}{S.~Z. Li}.}
  \bibinfo{year}{2019}\natexlab{c}.
\newblock \showarticletitle{AdaptiveFace: Adaptive Margin and Sampling for Face
  Recognition}. In \bibinfo{booktitle}{\emph{Proceedings of the IEEE/CVF
  Conference on Computer Vision and Pattern Recognition}}.
  \bibinfo{pages}{11947--11956}.
\newblock


\bibitem[\protect\citeauthoryear{Liu, Deng, Bai, Wei, and Huang}{Liu
  et~al\mbox{.}}{2015}]%
        {liu2015targeting}
\bibfield{author}{\bibinfo{person}{J. Liu}, \bibinfo{person}{Y. Deng},
  \bibinfo{person}{T. Bai}, \bibinfo{person}{Z. Wei}, {and} \bibinfo{person}{C.
  Huang}.} \bibinfo{year}{2015}\natexlab{}.
\newblock \showarticletitle{Targeting Ultimate Accuracy: Face Recognition via
  Deep Embedding}.
\newblock  (\bibinfo{year}{2015}).
\newblock
\showeprint{1506.07310}


\bibitem[\protect\citeauthoryear{Liu, Anguelov, Erhan, Szegedy, Reed, Fu, and
  Berg}{Liu et~al\mbox{.}}{2016a}]%
        {Liu_2016_ssd}
\bibfield{author}{\bibinfo{person}{W. Liu}, \bibinfo{person}{D. Anguelov},
  \bibinfo{person}{D. Erhan}, \bibinfo{person}{Christian Szegedy},
  \bibinfo{person}{Scott Reed}, \bibinfo{person}{Cheng-Yang Fu}, {and}
  \bibinfo{person}{Alexander~C. Berg}.} \bibinfo{year}{2016}\natexlab{a}.
\newblock \showarticletitle{SSD: single shot multiBox detector}. In
  \bibinfo{booktitle}{\emph{Proceedings of the European Conference on Computer
  Vision}}. \bibinfo{pages}{21–37}.
\newblock


\bibitem[\protect\citeauthoryear{Liu, Wen, Yu, Li, Raj, and Song}{Liu
  et~al\mbox{.}}{2017}]%
        {liu2017sphereface}
\bibfield{author}{\bibinfo{person}{W. Liu}, \bibinfo{person}{Y. Wen},
  \bibinfo{person}{Z. Yu}, \bibinfo{person}{M. Li}, \bibinfo{person}{B. Raj},
  {and} \bibinfo{person}{L. Song}.} \bibinfo{year}{2017}\natexlab{}.
\newblock \showarticletitle{Sphereface: Deep hypersphere embedding for face
  recognition}. In \bibinfo{booktitle}{\emph{Proceedings of the IEEE/CVF
  Conference on Computer Vision and Pattern Recognition}}.
  \bibinfo{pages}{212--220}.
\newblock


\bibitem[\protect\citeauthoryear{Liu, Wen, Yu, and Yang}{Liu
  et~al\mbox{.}}{2016c}]%
        {liu2016large}
\bibfield{author}{\bibinfo{person}{W. Liu}, \bibinfo{person}{Y. Wen},
  \bibinfo{person}{Z. Yu}, {and} \bibinfo{person}{M. Yang}.}
  \bibinfo{year}{2016}\natexlab{c}.
\newblock \showarticletitle{Large-margin softmax loss for convolutional neural
  networks.}. In \bibinfo{booktitle}{\emph{ICML}}, Vol.~\bibinfo{volume}{2}.
  \bibinfo{pages}{7}.
\newblock


\bibitem[\protect\citeauthoryear{Liu, Kumar, Yang, Tang, and You}{Liu
  et~al\mbox{.}}{2018a}]%
        {Liu2018DependencyAwareAC}
\bibfield{author}{\bibinfo{person}{X. Liu}, \bibinfo{person}{B.~V. K.~V.
  Kumar}, \bibinfo{person}{C. Yang}, \bibinfo{person}{Q. Tang}, {and}
  \bibinfo{person}{J. You}.} \bibinfo{year}{2018}\natexlab{a}.
\newblock \showarticletitle{Dependency-Aware Attention Control for
  Unconstrained Face Recognition with Image Sets}. In
  \bibinfo{booktitle}{\emph{Proceedings of the European Conference on Computer
  Vision}}. \bibinfo{pages}{548--565}.
\newblock


\bibitem[\protect\citeauthoryear{Liu, Song, Wu, and Tan}{Liu
  et~al\mbox{.}}{2016b}]%
        {Liu2016TransferringDR}
\bibfield{author}{\bibinfo{person}{X. Liu}, \bibinfo{person}{L. Song},
  \bibinfo{person}{X. Wu}, {and} \bibinfo{person}{T. Tan}.}
  \bibinfo{year}{2016}\natexlab{b}.
\newblock \showarticletitle{Transferring deep representation for NIR-VIS
  heterogeneous face recognition}.
\newblock \bibinfo{journal}{\emph{Proceedings of the International Conference
  on Biometrics}} (\bibinfo{year}{2016}), \bibinfo{pages}{1--8}.
\newblock


\bibitem[\protect\citeauthoryear{Liu, Jourabloo, Ren, and Liu}{Liu
  et~al\mbox{.}}{2017a}]%
        {Liu2017DenseFA}
\bibfield{author}{\bibinfo{person}{Y. Liu}, \bibinfo{person}{A. Jourabloo},
  \bibinfo{person}{W. Ren}, {and} \bibinfo{person}{X. Liu}.}
  \bibinfo{year}{2017}\natexlab{a}.
\newblock \showarticletitle{Dense Face Alignment}. In
  \bibinfo{booktitle}{\emph{Proceedings of the IEEE International Conference on
  Computer Vision Workshops}}. \bibinfo{pages}{1619--1628}.
\newblock


\bibitem[\protect\citeauthoryear{Liu, Li, and Wang}{Liu et~al\mbox{.}}{2017b}]%
        {Liu2017RethinkingFD}
\bibfield{author}{\bibinfo{person}{Y. Liu}, \bibinfo{person}{H. Li}, {and}
  \bibinfo{person}{X. Wang}.} \bibinfo{year}{2017}\natexlab{b}.
\newblock \showarticletitle{Rethinking Feature Discrimination and
  Polymerization for Large-scale Recognition}.
\newblock  (\bibinfo{year}{2017}).
\newblock
\showeprint{1710.00870}


\bibitem[\protect\citeauthoryear{{Liu}, {Li}, {Yan}, {Wei}, {Wang}, and
  {Tang}}{{Liu} et~al\mbox{.}}{2017}]%
        {Liu2016rsa}
\bibfield{author}{\bibinfo{person}{Y. {Liu}}, \bibinfo{person}{H. {Li}},
  \bibinfo{person}{J. {Yan}}, \bibinfo{person}{F. {Wei}}, \bibinfo{person}{X.
  {Wang}}, {and} \bibinfo{person}{X. {Tang}}.} \bibinfo{year}{2017}\natexlab{}.
\newblock \showarticletitle{Recurrent Scale Approximation for Object Detection
  in CNN}. In \bibinfo{booktitle}{\emph{Proceedings of the IEEE International
  Conference on Computer Vision}}. \bibinfo{pages}{571--579}.
\newblock


\bibitem[\protect\citeauthoryear{Liu, Shen, Si, Wang, Zhu, Shi, Hong, Guo, Guo,
  Chen, Li, Xi, Yu, Xie, Xie, Li, Lu, Wang, Lai, Chai, and Wei}{Liu
  et~al\mbox{.}}{2019a}]%
        {Liu2019GrandCO}
\bibfield{author}{\bibinfo{person}{Y. Liu}, \bibinfo{person}{H. Shen},
  \bibinfo{person}{Y. Si}, \bibinfo{person}{X. Wang}, \bibinfo{person}{X. Zhu},
  \bibinfo{person}{H. Shi}, \bibinfo{person}{Z. Hong}, \bibinfo{person}{H.
  Guo}, \bibinfo{person}{Z. Guo}, \bibinfo{person}{Y. Chen},
  \bibinfo{person}{B. Li}, \bibinfo{person}{T. Xi}, \bibinfo{person}{J. Yu},
  \bibinfo{person}{H. Xie}, \bibinfo{person}{G. Xie}, \bibinfo{person}{M. Li},
  \bibinfo{person}{Q. Lu}, \bibinfo{person}{Z. Wang}, \bibinfo{person}{S. Lai},
  \bibinfo{person}{Z. Chai}, {and} \bibinfo{person}{X. Wei}.}
  \bibinfo{year}{2019}\natexlab{a}.
\newblock \showarticletitle{Grand Challenge of 106-Point Facial Landmark
  Localization}. In \bibinfo{booktitle}{\emph{Proceedings of the IEEE ICME
  Workshop}}. \bibinfo{pages}{613--616}.
\newblock


\bibitem[\protect\citeauthoryear{Liu, Tang, Han, Liu, Rui, and Wu}{Liu
  et~al\mbox{.}}{2020b}]%
        {Liu_2020_HAMBox}
\bibfield{author}{\bibinfo{person}{Y. Liu}, \bibinfo{person}{X. Tang},
  \bibinfo{person}{J. Han}, \bibinfo{person}{J. Liu}, \bibinfo{person}{D. Rui},
  {and} \bibinfo{person}{X. Wu}.} \bibinfo{year}{2020}\natexlab{b}.
\newblock \showarticletitle{HAMBox: Delving Into Mining High-Quality Anchors on
  Face Detection}. In \bibinfo{booktitle}{\emph{Proceedings of the IEEE/CVF
  Conference on Computer Vision and Pattern Recognition}}.
  \bibinfo{pages}{13043--13051}.
\newblock


\bibitem[\protect\citeauthoryear{Liu, CloudMinds, Bai, Li, and CloudMinds}{Liu
  et~al\mbox{.}}{2019}]%
        {Liu2019FeatureAN}
\bibfield{author}{\bibinfo{person}{Z. Liu}, \bibinfo{person}{H.~H. CloudMinds},
  \bibinfo{person}{J. Bai}, \bibinfo{person}{S. Li}, {and}
  \bibinfo{person}{S.~L. CloudMinds}.} \bibinfo{year}{2019}\natexlab{}.
\newblock \showarticletitle{Feature Aggregation Network for Video Face
  Recognition}. In \bibinfo{booktitle}{\emph{Proceedings of the IEEE
  International Conference on Computer Vision Workshops}}.
  \bibinfo{pages}{990--998}.
\newblock


\bibitem[\protect\citeauthoryear{Liu, Zhu, Hu, Guo, Tang, Lei, Robertson, and
  Wang}{Liu et~al\mbox{.}}{2019b}]%
        {Liu2019SemanticAF}
\bibfield{author}{\bibinfo{person}{Z. Liu}, \bibinfo{person}{X. Zhu},
  \bibinfo{person}{G. Hu}, \bibinfo{person}{H. Guo}, \bibinfo{person}{M. Tang},
  \bibinfo{person}{Z. Lei}, \bibinfo{person}{N.~M. Robertson}, {and}
  \bibinfo{person}{J. Wang}.} \bibinfo{year}{2019}\natexlab{b}.
\newblock \showarticletitle{Semantic Alignment: Finding Semantically Consistent
  Ground-Truth for Facial Landmark Detection}. In
  \bibinfo{booktitle}{\emph{Proceedings of the IEEE/CVF Conference on Computer
  Vision and Pattern Recognition}}. \bibinfo{pages}{3462--3471}.
\newblock


\bibitem[\protect\citeauthoryear{Luo, Zhu, Liu, Wang, and Tang}{Luo
  et~al\mbox{.}}{2016}]%
        {Luo2016FaceMC}
\bibfield{author}{\bibinfo{person}{Ping Luo}, \bibinfo{person}{Zhenyao Zhu},
  \bibinfo{person}{Ziwei Liu}, \bibinfo{person}{Xiaogang Wang}, {and}
  \bibinfo{person}{Xiaoou Tang}.} \bibinfo{year}{2016}\natexlab{}.
\newblock \showarticletitle{Face Model Compression by Distilling Knowledge from
  Neurons}. In \bibinfo{booktitle}{\emph{Proceedings of the AAAI Conference on
  Artificial Intelligence}}. \bibinfo{pages}{3560–3566}.
\newblock


\bibitem[\protect\citeauthoryear{{Lv}, {Shao}, {Xing}, {Cheng}, and
  {Zhou}}{{Lv} et~al\mbox{.}}{2017}]%
        {Lv2016TSR}
\bibfield{author}{\bibinfo{person}{J. {Lv}}, \bibinfo{person}{X. {Shao}},
  \bibinfo{person}{J. {Xing}}, \bibinfo{person}{C. {Cheng}}, {and}
  \bibinfo{person}{X. {Zhou}}.} \bibinfo{year}{2017}\natexlab{}.
\newblock \showarticletitle{A Deep Regression Architecture with Two-Stage
  Re-initialization for High Performance Facial Landmark Detection}. In
  \bibinfo{booktitle}{\emph{Proceedings of the IEEE/CVF Conference on Computer
  Vision and Pattern Recognition}}. \bibinfo{pages}{3691--3700}.
\newblock


\bibitem[\protect\citeauthoryear{m.~t. Pham and Cham}{m.~t. Pham and
  Cham}{2007}]%
        {Pham2007Fast}
\bibfield{author}{\bibinfo{person}{m.~t. Pham} {and} \bibinfo{person}{T.J.
  Cham}.} \bibinfo{year}{2007}\natexlab{}.
\newblock \showarticletitle{Fast training and selection of Haar features using
  statistics in boosting-based face detection}. In
  \bibinfo{booktitle}{\emph{Proceedings of the IEEE International Conference on
  Computer Vision}}. \bibinfo{pages}{1--7}.
\newblock


\bibitem[\protect\citeauthoryear{Ma, Zhang, Zheng, and Sun}{Ma
  et~al\mbox{.}}{2018}]%
        {Ma2018ShuffleNetVP}
\bibfield{author}{\bibinfo{person}{N. Ma}, \bibinfo{person}{X. Zhang},
  \bibinfo{person}{H. Zheng}, {and} \bibinfo{person}{J. Sun}.}
  \bibinfo{year}{2018}\natexlab{}.
\newblock \showarticletitle{Shufflenet v2: Practical guidelines for efficient
  cnn architecture design}. In \bibinfo{booktitle}{\emph{Proceedings of the
  European Conference on Computer Vision}}. \bibinfo{pages}{116--131}.
\newblock


\bibitem[\protect\citeauthoryear{Manmatha, Wu, Smola, and
  Kr{\"a}henb{\"u}hl}{Manmatha et~al\mbox{.}}{2017}]%
        {Manmatha2017SamplingMI}
\bibfield{author}{\bibinfo{person}{R. Manmatha}, \bibinfo{person}{C.n Wu},
  \bibinfo{person}{A. Smola}, {and} \bibinfo{person}{P. Kr{\"a}henb{\"u}hl}.}
  \bibinfo{year}{2017}\natexlab{}.
\newblock \showarticletitle{Sampling Matters in Deep Embedding Learning}. In
  \bibinfo{booktitle}{\emph{Proceedings of the IEEE International Conference on
  Computer Vision}}. \bibinfo{pages}{2859--2867}.
\newblock


\bibitem[\protect\citeauthoryear{Mart{\'i}nez, Valstar, Binefa, and
  Pantic}{Mart{\'i}nez et~al\mbox{.}}{2013}]%
        {Martnez2013LocalEA}
\bibfield{author}{\bibinfo{person}{B. Mart{\'i}nez}, \bibinfo{person}{M.~F.
  Valstar}, \bibinfo{person}{X. Binefa}, {and} \bibinfo{person}{M. Pantic}.}
  \bibinfo{year}{2013}\natexlab{}.
\newblock \showarticletitle{Local Evidence Aggregation for Regression-Based
  Facial Point Detection}.
\newblock \bibinfo{journal}{\emph{IEEE Trans. Pattern Anal. Mach. Intell.}}
  \bibinfo{volume}{35} (\bibinfo{year}{2013}), \bibinfo{pages}{1149--1163}.
\newblock


\bibitem[\protect\citeauthoryear{Mart{\'i}nez-D{\'i}az, Luevano, Vazquez,
  Nicol{\'a}s-D{\'i}az, Chang, and Gonz{\'a}lez-Mendoza}{Mart{\'i}nez-D{\'i}az
  et~al\mbox{.}}{2019}]%
        {MartnezDaz2019ShuffleFaceNetAL}
\bibfield{author}{\bibinfo{person}{Y. Mart{\'i}nez-D{\'i}az},
  \bibinfo{person}{L.~S. Luevano}, \bibinfo{person}{H.~M. Vazquez},
  \bibinfo{person}{M. Nicol{\'a}s-D{\'i}az}, \bibinfo{person}{L. Chang}, {and}
  \bibinfo{person}{M. Gonz{\'a}lez-Mendoza}.} \bibinfo{year}{2019}\natexlab{}.
\newblock \showarticletitle{ShuffleFaceNet: A Lightweight Face Architecture for
  Efficient and Highly-Accurate Face Recognition}. In
  \bibinfo{booktitle}{\emph{Proceedings of the IEEE International Conference on
  Computer Vision Workshops}}. \bibinfo{pages}{2721--2728}.
\newblock


\bibitem[\protect\citeauthoryear{Masi, Rawls, Medioni, and Natarajan}{Masi
  et~al\mbox{.}}{2016}]%
        {Masi2016PoseAwareFR}
\bibfield{author}{\bibinfo{person}{I. Masi}, \bibinfo{person}{S. Rawls},
  \bibinfo{person}{G.~G. Medioni}, {and} \bibinfo{person}{P. Natarajan}.}
  \bibinfo{year}{2016}\natexlab{}.
\newblock \showarticletitle{Pose-Aware Face Recognition in the Wild}. In
  \bibinfo{booktitle}{\emph{Proceedings of the IEEE/CVF Conference on Computer
  Vision and Pattern Recognition}}. \bibinfo{pages}{4838--4846}.
\newblock


\bibitem[\protect\citeauthoryear{Maze, Adams, Duncan, Kalka, Miller, Otto,
  Jain, Niggel, Anderson, Cheney, and Grother}{Maze et~al\mbox{.}}{2018}]%
        {Maze2018IARPAJB}
\bibfield{author}{\bibinfo{person}{B. Maze}, \bibinfo{person}{J.~C. Adams},
  \bibinfo{person}{J.s~A. Duncan}, \bibinfo{person}{N.~D. Kalka},
  \bibinfo{person}{T. Miller}, \bibinfo{person}{C. Otto},
  \bibinfo{person}{A.~K. Jain}, \bibinfo{person}{W.~T. Niggel},
  \bibinfo{person}{J. Anderson}, \bibinfo{person}{J. Cheney}, {and}
  \bibinfo{person}{Patrick Grother}.} \bibinfo{year}{2018}\natexlab{}.
\newblock \showarticletitle{IARPA Janus Benchmark - C: Face Dataset and
  Protocol}. In \bibinfo{booktitle}{\emph{Proceedings of the International
  Conference on Biometrics}}. \bibinfo{pages}{158--165}.
\newblock


\bibitem[\protect\citeauthoryear{Mei, Yang, Yang, and Hua}{Mei
  et~al\mbox{.}}{2008}]%
        {Mei2008VideoCP}
\bibfield{author}{\bibinfo{person}{Tao Mei}, \bibinfo{person}{Bo Yang},
  \bibinfo{person}{Shiqiang Yang}, {and} \bibinfo{person}{Xiansheng Hua}.}
  \bibinfo{year}{2008}\natexlab{}.
\newblock \showarticletitle{Video collage: presenting a video sequence using a
  single image}.
\newblock \bibinfo{journal}{\emph{The Visual Computer}}  \bibinfo{volume}{25}
  (\bibinfo{year}{2008}), \bibinfo{pages}{39--51}.
\newblock


\bibitem[\protect\citeauthoryear{Merget, Rock, and Rigoll}{Merget
  et~al\mbox{.}}{2018}]%
        {Merget_2018_CVPR}
\bibfield{author}{\bibinfo{person}{D. Merget}, \bibinfo{person}{M. Rock}, {and}
  \bibinfo{person}{G. Rigoll}.} \bibinfo{year}{2018}\natexlab{}.
\newblock \showarticletitle{Robust Facial Landmark Detection via a
  Fully-Convolutional Local-Global Context Network}. In
  \bibinfo{booktitle}{\emph{Proceedings of the IEEE/CVF Conference on Computer
  Vision and Pattern Recognition}}. \bibinfo{pages}{781--790}.
\newblock


\bibitem[\protect\citeauthoryear{Miao, Zhen, Liu, Deng, Athitsos, and
  Huang}{Miao et~al\mbox{.}}{2018}]%
        {Miao2018DirectSR}
\bibfield{author}{\bibinfo{person}{X. Miao}, \bibinfo{person}{X. Zhen},
  \bibinfo{person}{X. Liu}, \bibinfo{person}{C. Deng}, \bibinfo{person}{V.
  Athitsos}, {and} \bibinfo{person}{H. Huang}.}
  \bibinfo{year}{2018}\natexlab{}.
\newblock \showarticletitle{Direct Shape Regression Networks for End-to-End
  Face Alignment}. In \bibinfo{booktitle}{\emph{Proceedings of the IEEE/CVF
  Conference on Computer Vision and Pattern Recognition}}.
  \bibinfo{pages}{5040--5049}.
\newblock


\bibitem[\protect\citeauthoryear{Minaee, Luo, Lin, and Bowyer}{Minaee
  et~al\mbox{.}}{2021}]%
        {Minaee2021GoingDI}
\bibfield{author}{\bibinfo{person}{Shervin Minaee}, \bibinfo{person}{Ping Luo},
  \bibinfo{person}{Zhe Lin}, {and} \bibinfo{person}{K. Bowyer}.}
  \bibinfo{year}{2021}\natexlab{}.
\newblock \showarticletitle{Going Deeper Into Face Detection: A Survey}.
\newblock \bibinfo{journal}{\emph{ArXiv}}  \bibinfo{volume}{abs/2103.14983}
  (\bibinfo{year}{2021}).
\newblock


\bibitem[\protect\citeauthoryear{Ming, Wei, Zhang, Chen, and Wen}{Ming
  et~al\mbox{.}}{2019}]%
        {Ming_2019_Group_Sampling}
\bibfield{author}{\bibinfo{person}{X. Ming}, \bibinfo{person}{F. Wei},
  \bibinfo{person}{T. Zhang}, \bibinfo{person}{D. Chen}, {and}
  \bibinfo{person}{F. Wen}.} \bibinfo{year}{2019}\natexlab{}.
\newblock \showarticletitle{Group Sampling for Scale Invariant Face Detection}.
  In \bibinfo{booktitle}{\emph{Proceedings of the IEEE/CVF Conference on
  Computer Vision and Pattern Recognition}}. \bibinfo{pages}{3446--3456}.
\newblock


\bibitem[\protect\citeauthoryear{Mita, Kaneko, and Hori}{Mita
  et~al\mbox{.}}{2005}]%
        {Mita2005Joint}
\bibfield{author}{\bibinfo{person}{T. Mita}, \bibinfo{person}{T. Kaneko}, {and}
  \bibinfo{person}{O. Hori}.} \bibinfo{year}{2005}\natexlab{}.
\newblock \showarticletitle{Joint Haar-like features for face detection}. In
  \bibinfo{booktitle}{\emph{Proceedings of the IEEE International Conference on
  Computer Vision}}. \bibinfo{pages}{1619--1626}.
\newblock


\bibitem[\protect\citeauthoryear{Mittal, Vatsa, and Singh}{Mittal
  et~al\mbox{.}}{2015}]%
        {Mittal2015CompositeSR}
\bibfield{author}{\bibinfo{person}{P. Mittal}, \bibinfo{person}{M. Vatsa},
  {and} \bibinfo{person}{R. Singh}.} \bibinfo{year}{2015}\natexlab{}.
\newblock \showarticletitle{Composite sketch recognition via deep network - a
  transfer learning approach}. In \bibinfo{booktitle}{\emph{Proceedings of
  International Conference on Biometrics}}. \bibinfo{pages}{251--256}.
\newblock


\bibitem[\protect\citeauthoryear{Moschoglou, Papaioannou, Sagonas, Deng,
  Kotsia, and Zafeiriou}{Moschoglou et~al\mbox{.}}{2017}]%
        {Moschoglou2017AgeDBTF}
\bibfield{author}{\bibinfo{person}{S. Moschoglou}, \bibinfo{person}{A.
  Papaioannou}, \bibinfo{person}{C. Sagonas}, \bibinfo{person}{J. Deng},
  \bibinfo{person}{I. Kotsia}, {and} \bibinfo{person}{S. Zafeiriou}.}
  \bibinfo{year}{2017}\natexlab{}.
\newblock \showarticletitle{AgeDB: The First Manually Collected, In-the-Wild
  Age Database}. In \bibinfo{booktitle}{\emph{Proceedings of the IEEE/CVF
  Conference on Computer Vision and Pattern Recognition Workshops}}.
  \bibinfo{pages}{1997--2005}.
\newblock


\bibitem[\protect\citeauthoryear{{Najibi}, {Samangouei}, {Chellappa}, and
  {Davis}}{{Najibi} et~al\mbox{.}}{2017}]%
        {2017SSH}
\bibfield{author}{\bibinfo{person}{M. {Najibi}}, \bibinfo{person}{P.
  {Samangouei}}, \bibinfo{person}{R. {Chellappa}}, {and} \bibinfo{person}{L.~S.
  {Davis}}.} \bibinfo{year}{2017}\natexlab{}.
\newblock \showarticletitle{SSH: single stage headless face detector}. In
  \bibinfo{booktitle}{\emph{Proceedings of the IEEE International Conference on
  Computer Vision}}. \bibinfo{pages}{4885--4894}.
\newblock


\bibitem[\protect\citeauthoryear{Najibi, Singh, and Davis}{Najibi
  et~al\mbox{.}}{2019}]%
        {Najibi_2019_CVPR}
\bibfield{author}{\bibinfo{person}{M. Najibi}, \bibinfo{person}{B. Singh},
  {and} \bibinfo{person}{L.~S. Davis}.} \bibinfo{year}{2019}\natexlab{}.
\newblock \showarticletitle{FA-RPN: Floating Region Proposals for Face
  Detection}. In \bibinfo{booktitle}{\emph{Proceedings of the IEEE/CVF
  Conference on Computer Vision and Pattern Recognition}}.
  \bibinfo{pages}{7723--7732}.
\newblock


\bibitem[\protect\citeauthoryear{Nech and Kemelmacher-Shlizerman}{Nech and
  Kemelmacher-Shlizerman}{2017}]%
        {Nech2017LevelPF}
\bibfield{author}{\bibinfo{person}{A. Nech} {and} \bibinfo{person}{I.
  Kemelmacher-Shlizerman}.} \bibinfo{year}{2017}\natexlab{}.
\newblock \showarticletitle{Level Playing Field for Million Scale Face
  Recognition}. In \bibinfo{booktitle}{\emph{Proceedings of the IEEE/CVF
  Conference on Computer Vision and Pattern Recognition}}.
  \bibinfo{pages}{3406--3415}.
\newblock


\bibitem[\protect\citeauthoryear{Newell, Yang, and Deng}{Newell
  et~al\mbox{.}}{2016}]%
        {Newell2016StackedHN}
\bibfield{author}{\bibinfo{person}{A. Newell}, \bibinfo{person}{K. Yang}, {and}
  \bibinfo{person}{J. Deng}.} \bibinfo{year}{2016}\natexlab{}.
\newblock \showarticletitle{Stacked Hourglass Networks for Human Pose
  Estimation}. In \bibinfo{booktitle}{\emph{Proceedings of the European
  Conference on Computer Vision}}. \bibinfo{pages}{483--499}.
\newblock


\bibitem[\protect\citeauthoryear{Ojala, Pietikainen, and Maenpaa}{Ojala
  et~al\mbox{.}}{2002}]%
        {Ojala2002Multiresolution}
\bibfield{author}{\bibinfo{person}{T. Ojala}, \bibinfo{person}{M. Pietikainen},
  {and} \bibinfo{person}{T. Maenpaa}.} \bibinfo{year}{2002}\natexlab{}.
\newblock \showarticletitle{Multiresolution gray-scale and rotation invariant
  texture classification with local binary patterns}.
\newblock \bibinfo{journal}{\emph{IEEE Trans. Pattern Anal. Mach. Intell.}}
  \bibinfo{volume}{24}, \bibinfo{number}{7} (\bibinfo{year}{2002}),
  \bibinfo{pages}{971--987}.
\newblock


\bibitem[\protect\citeauthoryear{Opitz, Waltner, Poier, Possegger, and
  Bischof}{Opitz et~al\mbox{.}}{2016}]%
        {2016Grid_Loss}
\bibfield{author}{\bibinfo{person}{M. Opitz}, \bibinfo{person}{G. Waltner},
  \bibinfo{person}{G. Poier}, \bibinfo{person}{H. Possegger}, {and}
  \bibinfo{person}{H. Bischof}.} \bibinfo{year}{2016}\natexlab{}.
\newblock \showarticletitle{Grid loss: detecting occluded faces}. In
  \bibinfo{booktitle}{\emph{Proceedings of the European Conference on Computer
  Vision}}, Vol.~\bibinfo{volume}{9907}. \bibinfo{pages}{386--402}.
\newblock


\bibitem[\protect\citeauthoryear{Parchami, Bashbaghi, Granger, and
  Sayed}{Parchami et~al\mbox{.}}{2017}]%
        {Parchami2017UsingDA}
\bibfield{author}{\bibinfo{person}{M. Parchami}, \bibinfo{person}{S.
  Bashbaghi}, \bibinfo{person}{E. Granger}, {and} \bibinfo{person}{S. Sayed}.}
  \bibinfo{year}{2017}\natexlab{}.
\newblock \showarticletitle{Using deep autoencoders to learn robust
  domain-invariant representations for still-to-video face recognition}. In
  \bibinfo{booktitle}{\emph{Proceedings of the IEEE International Conference on
  Advanced Video and Signal Based Surveillance (AVSS)}}. \bibinfo{pages}{1--6}.
\newblock


\bibitem[\protect\citeauthoryear{Parkhi, Vedaldi, and Zisserman}{Parkhi
  et~al\mbox{.}}{2015}]%
        {Parkhi2015DeepFR}
\bibfield{author}{\bibinfo{person}{O.~M. Parkhi}, \bibinfo{person}{A. Vedaldi},
  {and} \bibinfo{person}{A. Zisserman}.} \bibinfo{year}{2015}\natexlab{}.
\newblock \showarticletitle{Deep Face Recognition}. In
  \bibinfo{booktitle}{\emph{Proceedings of the British Machine Vision
  Conference}}. \bibinfo{pages}{41.1--41.12}.
\newblock


\bibitem[\protect\citeauthoryear{Peng, Feris, Wang, and Metaxas}{Peng
  et~al\mbox{.}}{2016}]%
        {Peng2016RED}
\bibfield{author}{\bibinfo{person}{X. Peng}, \bibinfo{person}{R.~S. Feris},
  \bibinfo{person}{X. Wang}, {and} \bibinfo{person}{D.~N. Metaxas}.}
  \bibinfo{year}{2016}\natexlab{}.
\newblock \showarticletitle{A Recurrent Encoder-Decoder Network for Sequential
  Face Alignment}. In \bibinfo{booktitle}{\emph{Proceedings of the European
  Conference on Computer Vision}}. \bibinfo{pages}{38--56}.
\newblock


\bibitem[\protect\citeauthoryear{Qin, Yan, Xiu, and Hu}{Qin
  et~al\mbox{.}}{2016}]%
        {Qin2016Joint}
\bibfield{author}{\bibinfo{person}{H. Qin}, \bibinfo{person}{J. Yan},
  \bibinfo{person}{L. Xiu}, {and} \bibinfo{person}{X. Hu}.}
  \bibinfo{year}{2016}\natexlab{}.
\newblock \showarticletitle{Joint Training of Cascaded CNN for Face Detection}.
  In \bibinfo{booktitle}{\emph{Proceedings of the IEEE/CVF Conference on
  Computer Vision and Pattern Recognition}}. \bibinfo{pages}{3456--3465}.
\newblock


\bibitem[\protect\citeauthoryear{Ranjan, Castillo, and Chellappa}{Ranjan
  et~al\mbox{.}}{2017}]%
        {ranjan2017l2}
\bibfield{author}{\bibinfo{person}{R. Ranjan}, \bibinfo{person}{C.~D.
  Castillo}, {and} \bibinfo{person}{R. Chellappa}.}
  \bibinfo{year}{2017}\natexlab{}.
\newblock \showarticletitle{L2-constrained softmax loss for discriminative face
  verification}.
\newblock  (\bibinfo{year}{2017}).
\newblock
\showeprint{1703.09507}


\bibitem[\protect\citeauthoryear{{Ranjan}, {Patel}, and {Chellappa}}{{Ranjan}
  et~al\mbox{.}}{2019}]%
        {HyperFace}
\bibfield{author}{\bibinfo{person}{R. {Ranjan}}, \bibinfo{person}{V.~M.
  {Patel}}, {and} \bibinfo{person}{R. {Chellappa}}.}
  \bibinfo{year}{2019}\natexlab{}.
\newblock \showarticletitle{HyperFace: A Deep Multi-Task Learning Framework for
  Face Detection, Landmark Localization, Pose Estimation, and Gender
  Recognition}.
\newblock \bibinfo{journal}{\emph{IEEE Trans. Pattern Anal. Mach. Intell.}}
  \bibinfo{volume}{41}, \bibinfo{number}{1} (\bibinfo{year}{2019}),
  \bibinfo{pages}{121--135}.
\newblock


\bibitem[\protect\citeauthoryear{{Ranjan}, {Sankaranarayanan}, {Bansal},
  {Bodla}, {Chen}, {Patel}, {Castillo}, and {Chellappa}}{{Ranjan}
  et~al\mbox{.}}{2018}]%
        {Ranjan2018deep}
\bibfield{author}{\bibinfo{person}{R. {Ranjan}}, \bibinfo{person}{S.
  {Sankaranarayanan}}, \bibinfo{person}{A. {Bansal}}, \bibinfo{person}{N.
  {Bodla}}, \bibinfo{person}{J. {Chen}}, \bibinfo{person}{V.~M. {Patel}},
  \bibinfo{person}{C.~D. {Castillo}}, {and} \bibinfo{person}{R. {Chellappa}}.}
  \bibinfo{year}{2018}\natexlab{}.
\newblock \showarticletitle{Deep Learning for Understanding Faces: Machines May
  Be Just as Good, or Better, than Humans}.
\newblock \bibinfo{journal}{\emph{IEEE Signal Processing Magazine}}
  \bibinfo{volume}{35}, \bibinfo{number}{1} (\bibinfo{year}{2018}),
  \bibinfo{pages}{66--83}.
\newblock


\bibitem[\protect\citeauthoryear{Rao, Lin, Lu, and Zhou}{Rao
  et~al\mbox{.}}{2017a}]%
        {Rao2017LearningDA}
\bibfield{author}{\bibinfo{person}{Y. Rao}, \bibinfo{person}{J. Lin},
  \bibinfo{person}{J. Lu}, {and} \bibinfo{person}{J. Zhou}.}
  \bibinfo{year}{2017}\natexlab{a}.
\newblock \showarticletitle{Learning Discriminative Aggregation Network for
  Video-Based Face Recognition}. In \bibinfo{booktitle}{\emph{Proceedings of
  the IEEE International Conference on Computer Vision}}.
  \bibinfo{pages}{3801--3810}.
\newblock


\bibitem[\protect\citeauthoryear{Rao, Lu, and Zhou}{Rao et~al\mbox{.}}{2017b}]%
        {Rao2017AttentionAwareDR}
\bibfield{author}{\bibinfo{person}{Y. Rao}, \bibinfo{person}{J. Lu}, {and}
  \bibinfo{person}{J. Zhou}.} \bibinfo{year}{2017}\natexlab{b}.
\newblock \showarticletitle{Attention-Aware Deep Reinforcement Learning for
  Video Face Recognition}. In \bibinfo{booktitle}{\emph{Proceedings of the IEEE
  International Conference on Computer Vision}}. \bibinfo{pages}{3951--3960}.
\newblock


\bibitem[\protect\citeauthoryear{Reale, Nasrabadi, Kwon, and Chellappa}{Reale
  et~al\mbox{.}}{2016}]%
        {Reale2016SeeingTF}
\bibfield{author}{\bibinfo{person}{C. Reale}, \bibinfo{person}{N.~M.
  Nasrabadi}, \bibinfo{person}{H. Kwon}, {and} \bibinfo{person}{R. Chellappa}.}
  \bibinfo{year}{2016}\natexlab{}.
\newblock \showarticletitle{Seeing the Forest from the Trees: A Holistic
  Approach to Near-Infrared Heterogeneous Face Recognition}. In
  \bibinfo{booktitle}{\emph{Proceedings of the IEEE/CVF Conference on Computer
  Vision and Pattern RecognitionW}}. \bibinfo{pages}{320--328}.
\newblock


\bibitem[\protect\citeauthoryear{Ren, He, Girshick, and Sun}{Ren
  et~al\mbox{.}}{2015}]%
        {ren2015faster}
\bibfield{author}{\bibinfo{person}{S. Ren}, \bibinfo{person}{K. He},
  \bibinfo{person}{R. Girshick}, {and} \bibinfo{person}{J. Sun}.}
  \bibinfo{year}{2015}\natexlab{}.
\newblock \showarticletitle{Faster r-cnn: Towards real-time object detection
  with region proposal networks}. In \bibinfo{booktitle}{\emph{Advances in
  neural information processing systems}}. \bibinfo{pages}{91--99}.
\newblock


\bibitem[\protect\citeauthoryear{Robinson, Li, Zhang, Fu, and
  S.Tulyakov}{Robinson et~al\mbox{.}}{2019}]%
        {Robinson2019LaplaceLL}
\bibfield{author}{\bibinfo{person}{J.~P. Robinson}, \bibinfo{person}{Y. Li},
  \bibinfo{person}{N. Zhang}, \bibinfo{person}{Y. Fu}, {and}
  \bibinfo{person}{S.Tulyakov}.} \bibinfo{year}{2019}\natexlab{}.
\newblock \showarticletitle{Laplace Landmark Localization}. In
  \bibinfo{booktitle}{\emph{Proceedings of the IEEE International Conference on
  Computer Vision}}. \bibinfo{pages}{10102--10111}.
\newblock


\bibitem[\protect\citeauthoryear{RoyChowdhury, Yu, Sohn, Learned-Miller, and
  Chandraker}{RoyChowdhury et~al\mbox{.}}{2020}]%
        {RoyChowdhury2020ImprovingFR}
\bibfield{author}{\bibinfo{person}{A. RoyChowdhury}, \bibinfo{person}{X. Yu},
  \bibinfo{person}{K. Sohn}, \bibinfo{person}{E. Learned-Miller}, {and}
  \bibinfo{person}{M. Chandraker}.} \bibinfo{year}{2020}\natexlab{}.
\newblock \showarticletitle{Improving Face Recognition by Clustering Unlabeled
  Faces in the Wild}. In \bibinfo{booktitle}{\emph{Proceedings of the European
  Conference on Computer Vision}}. \bibinfo{pages}{119--136}.
\newblock


\bibitem[\protect\citeauthoryear{Sagonas, Antonakos, Tzimiropoulos, Zafeiriou,
  and Pantic}{Sagonas et~al\mbox{.}}{2016}]%
        {Sagonas2016300FI}
\bibfield{author}{\bibinfo{person}{C. Sagonas}, \bibinfo{person}{E. Antonakos},
  \bibinfo{person}{G. Tzimiropoulos}, \bibinfo{person}{S. Zafeiriou}, {and}
  \bibinfo{person}{M. Pantic}.} \bibinfo{year}{2016}\natexlab{}.
\newblock \showarticletitle{300 Faces In-The-Wild Challenge: database and
  results}.
\newblock \bibinfo{journal}{\emph{Image Vis. Comput.}}  \bibinfo{volume}{47}
  (\bibinfo{year}{2016}), \bibinfo{pages}{3--18}.
\newblock


\bibitem[\protect\citeauthoryear{Sagonas, Tzimiropoulos, Zafeiriou, and
  Pantic}{Sagonas et~al\mbox{.}}{2013}]%
        {Sagonas2013300FI}
\bibfield{author}{\bibinfo{person}{C. Sagonas}, \bibinfo{person}{G.
  Tzimiropoulos}, \bibinfo{person}{S. Zafeiriou}, {and} \bibinfo{person}{M.
  Pantic}.} \bibinfo{year}{2013}\natexlab{}.
\newblock \showarticletitle{300 Faces in-the-Wild Challenge: The First Facial
  Landmark Localization Challenge}. In \bibinfo{booktitle}{\emph{Proceedings of
  the IEEE International Conference on Computer Vision Workshops}}.
  \bibinfo{pages}{397--403}.
\newblock


\bibitem[\protect\citeauthoryear{Sandler, Howard, M.Zhu, Zhmoginov, and
  Chen}{Sandler et~al\mbox{.}}{2018}]%
        {Sandler2018MobileNetV2IR}
\bibfield{author}{\bibinfo{person}{M. Sandler}, \bibinfo{person}{A.~G. Howard},
  \bibinfo{person}{M.Zhu}, \bibinfo{person}{A. Zhmoginov}, {and}
  \bibinfo{person}{L. Chen}.} \bibinfo{year}{2018}\natexlab{}.
\newblock \showarticletitle{MobileNetV2: Inverted Residuals and Linear
  Bottlenecks}. In \bibinfo{booktitle}{\emph{Proceedings of the IEEE/CVF
  Conference on Computer Vision and Pattern Recognition}}.
  \bibinfo{pages}{4510--4520}.
\newblock


\bibitem[\protect\citeauthoryear{Sankaranarayanan, Alavi, Castillo, and
  Chellappa}{Sankaranarayanan et~al\mbox{.}}{2016}]%
        {Sankaranarayanan2016TripletPE}
\bibfield{author}{\bibinfo{person}{S. Sankaranarayanan}, \bibinfo{person}{A.
  Alavi}, \bibinfo{person}{C.~D. Castillo}, {and} \bibinfo{person}{R.
  Chellappa}.} \bibinfo{year}{2016}\natexlab{}.
\newblock \showarticletitle{Triplet probabilistic embedding for face
  verification and clustering}. In \bibinfo{booktitle}{\emph{Proceedings of the
  IEEE International Conference on Biometrics Theory, Applications and
  Systems}}. \bibinfo{pages}{1--8}.
\newblock


\bibitem[\protect\citeauthoryear{Saxena and Verbeek}{Saxena and
  Verbeek}{2016}]%
        {Saxena2016HeterogeneousFR}
\bibfield{author}{\bibinfo{person}{S. Saxena} {and} \bibinfo{person}{J.
  Verbeek}.} \bibinfo{year}{2016}\natexlab{}.
\newblock \showarticletitle{Heterogeneous Face Recognition with CNNs}. In
  \bibinfo{booktitle}{\emph{Proceedings of the European Conference on Computer
  Vision Workshops}}. \bibinfo{pages}{483--491}.
\newblock


\bibitem[\protect\citeauthoryear{Schroff, Kalenichenko, and Philbin}{Schroff
  et~al\mbox{.}}{2015}]%
        {Schroff2015FaceNetAU}
\bibfield{author}{\bibinfo{person}{F. Schroff}, \bibinfo{person}{D.
  Kalenichenko}, {and} \bibinfo{person}{J. Philbin}.}
  \bibinfo{year}{2015}\natexlab{}.
\newblock \showarticletitle{FaceNet: A unified embedding for face recognition
  and clustering}. In \bibinfo{booktitle}{\emph{Proceedings of the IEEE/CVF
  Conference on Computer Vision and Pattern Recognition}}.
  \bibinfo{pages}{815--823}.
\newblock


\bibitem[\protect\citeauthoryear{Schultz and Joachims}{Schultz and
  Joachims}{2004}]%
        {Schultz2003LearningAD}
\bibfield{author}{\bibinfo{person}{M. Schultz} {and} \bibinfo{person}{T.
  Joachims}.} \bibinfo{year}{2004}\natexlab{}.
\newblock \showarticletitle{Learning a distance metric from relative
  comparisons}. In \bibinfo{booktitle}{\emph{Advances in neural information
  processing systems}}. \bibinfo{pages}{41--48}.
\newblock


\bibitem[\protect\citeauthoryear{Sengupta, Chen, Castillo, Patel, Chellappa,
  and Jacobs}{Sengupta et~al\mbox{.}}{2016}]%
        {sengupta2016frontal}
\bibfield{author}{\bibinfo{person}{S. Sengupta}, \bibinfo{person}{J. Chen},
  \bibinfo{person}{C. Castillo}, \bibinfo{person}{V.~M. Patel},
  \bibinfo{person}{R. Chellappa}, {and} \bibinfo{person}{D.~W. Jacobs}.}
  \bibinfo{year}{2016}\natexlab{}.
\newblock \showarticletitle{Frontal to profile face verification in the wild}.
  In \bibinfo{booktitle}{\emph{Proceedings of the IEEE Winter Conference on
  Applications of Computer Vision}}. \bibinfo{pages}{1--9}.
\newblock


\bibitem[\protect\citeauthoryear{Shen, Zafeiriou, Chrysos, Kossaifi,
  Tzimiropoulos, and Pantic}{Shen et~al\mbox{.}}{2015}]%
        {Shen2015TheFF}
\bibfield{author}{\bibinfo{person}{J. Shen}, \bibinfo{person}{S. Zafeiriou},
  \bibinfo{person}{G.~G. Chrysos}, \bibinfo{person}{J. Kossaifi},
  \bibinfo{person}{G. Tzimiropoulos}, {and} \bibinfo{person}{M. Pantic}.}
  \bibinfo{year}{2015}\natexlab{}.
\newblock \showarticletitle{The First Facial Landmark Tracking in-the-Wild
  Challenge: Benchmark and Results}. In \bibinfo{booktitle}{\emph{Proceedings
  of the IEEE International Conference on Computer Vision Workshops}}.
  \bibinfo{pages}{1003--1011}.
\newblock


\bibitem[\protect\citeauthoryear{{Shi}, {Shan}, {Kan}, {Wu}, and {Chen}}{{Shi}
  et~al\mbox{.}}{2018}]%
        {pcn}
\bibfield{author}{\bibinfo{person}{X. {Shi}}, \bibinfo{person}{S. {Shan}},
  \bibinfo{person}{M. {Kan}}, \bibinfo{person}{S. {Wu}}, {and}
  \bibinfo{person}{X. {Chen}}.} \bibinfo{year}{2018}\natexlab{}.
\newblock \showarticletitle{Real-Time rotation-invariant face detection with
  progressive calibration networks}. In \bibinfo{booktitle}{\emph{Proceedings
  of the IEEE/CVF Conference on Computer Vision and Pattern Recognition}}.
  \bibinfo{pages}{2295--2303}.
\newblock


\bibitem[\protect\citeauthoryear{Shi, Jain, and Kalka}{Shi
  et~al\mbox{.}}{2019}]%
        {Shi2019ProbabilisticFE}
\bibfield{author}{\bibinfo{person}{Y. Shi}, \bibinfo{person}{A.~K. Jain}, {and}
  \bibinfo{person}{N.~D. Kalka}.} \bibinfo{year}{2019}\natexlab{}.
\newblock \showarticletitle{Probabilistic Face Embeddings}. In
  \bibinfo{booktitle}{\emph{Proceedings of the IEEE International Conference on
  Computer Vision}}. \bibinfo{pages}{6901--6910}.
\newblock


\bibitem[\protect\citeauthoryear{Shi, Otto, and Jain}{Shi
  et~al\mbox{.}}{2018}]%
        {shi2018face}
\bibfield{author}{\bibinfo{person}{Y. Shi}, \bibinfo{person}{C. Otto}, {and}
  \bibinfo{person}{A.~K. Jain}.} \bibinfo{year}{2018}\natexlab{}.
\newblock \showarticletitle{Face clustering: representation and pairwise
  constraints}.
\newblock \bibinfo{journal}{\emph{IEEE Transactions on Information Forensics
  and Security}} \bibinfo{volume}{13}, \bibinfo{number}{7}
  (\bibinfo{year}{2018}), \bibinfo{pages}{1626--1640}.
\newblock


\bibitem[\protect\citeauthoryear{Shi, Yu, Sohn, Chandraker, and Jain}{Shi
  et~al\mbox{.}}{2020}]%
        {shi2020universal}
\bibfield{author}{\bibinfo{person}{Y. Shi}, \bibinfo{person}{X. Yu},
  \bibinfo{person}{K. Sohn}, \bibinfo{person}{M. Chandraker}, {and}
  \bibinfo{person}{A.~K. Jain}.} \bibinfo{year}{2020}\natexlab{}.
\newblock \showarticletitle{Towards Universal Representation Learning for Deep
  Face Recognition}. In \bibinfo{booktitle}{\emph{Proceedings of the IEEE/CVF
  Conference on Computer Vision and Pattern Recognition}}.
  \bibinfo{pages}{6817--6826}.
\newblock


\bibitem[\protect\citeauthoryear{Shrivastava, Gupta, and Girshick}{Shrivastava
  et~al\mbox{.}}{2016}]%
        {shrivastava2016training}
\bibfield{author}{\bibinfo{person}{A. Shrivastava}, \bibinfo{person}{A. Gupta},
  {and} \bibinfo{person}{R. Girshick}.} \bibinfo{year}{2016}\natexlab{}.
\newblock \showarticletitle{Training region-based object detectors with online
  hard example mining}. In \bibinfo{booktitle}{\emph{Proceedings of the
  IEEE/CVF Conference on Computer Vision and Pattern Recognition}}.
  \bibinfo{pages}{761--769}.
\newblock


\bibitem[\protect\citeauthoryear{Simonyan and Zisserman}{Simonyan and
  Zisserman}{2015}]%
        {Simonyan2015VeryDC}
\bibfield{author}{\bibinfo{person}{K. Simonyan} {and} \bibinfo{person}{A.
  Zisserman}.} \bibinfo{year}{2015}\natexlab{}.
\newblock \showarticletitle{Very Deep Convolutional Networks for Large-Scale
  Image Recognition}.
\newblock  (\bibinfo{year}{2015}).
\newblock
\showeprint{1409.1556}


\bibitem[\protect\citeauthoryear{Smirnov, Melnikov, Novoselov, Luckyanets, and
  Lavrentyeva}{Smirnov et~al\mbox{.}}{2017}]%
        {smirnov2017doppelganger}
\bibfield{author}{\bibinfo{person}{E. Smirnov}, \bibinfo{person}{A. Melnikov},
  \bibinfo{person}{S. Novoselov}, \bibinfo{person}{E. Luckyanets}, {and}
  \bibinfo{person}{G. Lavrentyeva}.} \bibinfo{year}{2017}\natexlab{}.
\newblock \showarticletitle{Doppelganger mining for face representation
  learning}. In \bibinfo{booktitle}{\emph{Proceedings of the IEEE International
  Conference on Computer Vision}}. \bibinfo{pages}{1916--1923}.
\newblock


\bibitem[\protect\citeauthoryear{Sohn}{Sohn}{2016}]%
        {sohn2016improved}
\bibfield{author}{\bibinfo{person}{K. Sohn}.} \bibinfo{year}{2016}\natexlab{}.
\newblock \showarticletitle{Improved deep metric learning with multi-class
  n-pair loss objective}. In \bibinfo{booktitle}{\emph{Advances in Neural
  Information Processing Systems}}. \bibinfo{pages}{1857--1865}.
\newblock


\bibitem[\protect\citeauthoryear{Soltanpour, B.Boufama, and Wu}{Soltanpour
  et~al\mbox{.}}{2017}]%
        {Soltanpour2017ASO}
\bibfield{author}{\bibinfo{person}{S. Soltanpour}, \bibinfo{person}{B.Boufama},
  {and} \bibinfo{person}{Q.~M.~J. Wu}.} \bibinfo{year}{2017}\natexlab{}.
\newblock \showarticletitle{A survey of local feature methods for 3D face
  recognition}.
\newblock \bibinfo{journal}{\emph{Pattern Recognition}}  \bibinfo{volume}{72}
  (\bibinfo{year}{2017}), \bibinfo{pages}{391--406}.
\newblock


\bibitem[\protect\citeauthoryear{Song, Liu, Jiang, Wang, Yan, and Leng}{Song
  et~al\mbox{.}}{2018a}]%
        {Song_2018_CVPR}
\bibfield{author}{\bibinfo{person}{G. Song}, \bibinfo{person}{Y. Liu},
  \bibinfo{person}{M. Jiang}, \bibinfo{person}{Y. Wang}, \bibinfo{person}{J.
  Yan}, {and} \bibinfo{person}{B. Leng}.} \bibinfo{year}{2018}\natexlab{a}.
\newblock \showarticletitle{Beyond Trade-Off: Accelerate FCN-Based Face
  Detector With Higher Accuracy}. In \bibinfo{booktitle}{\emph{Proceedings of
  the IEEE/CVF Conference on Computer Vision and Pattern Recognition}}.
  \bibinfo{pages}{7756--7764}.
\newblock


\bibitem[\protect\citeauthoryear{Song, Xiang, Jegelka, and Savarese}{Song
  et~al\mbox{.}}{2016}]%
        {oh2016deep}
\bibfield{author}{\bibinfo{person}{H.~Oh Song}, \bibinfo{person}{Y. Xiang},
  \bibinfo{person}{S. Jegelka}, {and} \bibinfo{person}{S. Savarese}.}
  \bibinfo{year}{2016}\natexlab{}.
\newblock \showarticletitle{Deep metric learning via lifted structured feature
  embedding}. In \bibinfo{booktitle}{\emph{Proceedings of the IEEE/CVF
  Conference on Computer Vision and Pattern Recognition}}.
  \bibinfo{pages}{4004--4012}.
\newblock


\bibitem[\protect\citeauthoryear{Song, Zhang, Gao, Liu, and Shen}{Song
  et~al\mbox{.}}{2018b}]%
        {song2018dual}
\bibfield{author}{\bibinfo{person}{J. Song}, \bibinfo{person}{J. Zhang},
  \bibinfo{person}{L. Gao}, \bibinfo{person}{X. Liu}, {and} \bibinfo{person}{H
  Shen}.} \bibinfo{year}{2018}\natexlab{b}.
\newblock \showarticletitle{Dual Conditional GANs for Face Aging and
  Rejuvenation.}. In \bibinfo{booktitle}{\emph{International Joint Conference
  on Artificial Intelligence}}. \bibinfo{pages}{899--905}.
\newblock


\bibitem[\protect\citeauthoryear{Song, Zhang, Wu, and He}{Song
  et~al\mbox{.}}{2018c}]%
        {Song2018AdversarialDH}
\bibfield{author}{\bibinfo{person}{Lingxiao Song}, \bibinfo{person}{Man Zhang},
  \bibinfo{person}{Xiang Wu}, {and} \bibinfo{person}{Ran He}.}
  \bibinfo{year}{2018}\natexlab{c}.
\newblock \showarticletitle{Adversarial discriminative heterogeneous face
  recognition}. In \bibinfo{booktitle}{\emph{Proceedings of the AAAI Conference
  on Artificial Intelligence}}, Vol.~\bibinfo{volume}{32}.
\newblock


\bibitem[\protect\citeauthoryear{Song, Zhang, Liu, and Mei}{Song
  et~al\mbox{.}}{2019}]%
        {Song2019UnsupervisedPI}
\bibfield{author}{\bibinfo{person}{Sijie Song}, \bibinfo{person}{Wei Zhang},
  \bibinfo{person}{Jiaying Liu}, {and} \bibinfo{person}{Tao Mei}.}
  \bibinfo{year}{2019}\natexlab{}.
\newblock \showarticletitle{Unsupervised Person Image Generation With Semantic
  Parsing Transformation}. In \bibinfo{booktitle}{\emph{Proceedings of the
  IEEE/CVF Conference on Computer Vision and Pattern Recognition}}.
  \bibinfo{pages}{2352--2361}.
\newblock


\bibitem[\protect\citeauthoryear{Sun, Wu, Liu, Yang, Wang, Zhou, Ye, and
  Qian}{Sun et~al\mbox{.}}{2019}]%
        {Sun2019FABAR}
\bibfield{author}{\bibinfo{person}{K. Sun}, \bibinfo{person}{W. Wu},
  \bibinfo{person}{T. Liu}, \bibinfo{person}{S. Yang}, \bibinfo{person}{Q.
  Wang}, \bibinfo{person}{Q. Zhou}, \bibinfo{person}{Z. Ye}, {and}
  \bibinfo{person}{C. Qian}.} \bibinfo{year}{2019}\natexlab{}.
\newblock \showarticletitle{FAB: A Robust Facial Landmark Detection Framework
  for Motion-Blurred Videos}. In \bibinfo{booktitle}{\emph{Proceedings of the
  IEEE International Conference on Computer Vision}}.
  \bibinfo{pages}{5461--5470}.
\newblock


\bibitem[\protect\citeauthoryear{Sun, Wu, and Hoi}{Sun et~al\mbox{.}}{2018}]%
        {SUN201842}
\bibfield{author}{\bibinfo{person}{X. Sun}, \bibinfo{person}{P. Wu}, {and}
  \bibinfo{person}{S.~C.~H. Hoi}.} \bibinfo{year}{2018}\natexlab{}.
\newblock \showarticletitle{Face detection using deep learning: an improved
  faster rcnn approach}.
\newblock \bibinfo{journal}{\emph{Neurocomputing}}  \bibinfo{volume}{299}
  (\bibinfo{year}{2018}), \bibinfo{pages}{42 -- 50}.
\newblock


\bibitem[\protect\citeauthoryear{Sun, Chen, Wang, and Tang}{Sun
  et~al\mbox{.}}{2014}]%
        {sun2014deep}
\bibfield{author}{\bibinfo{person}{Y. Sun}, \bibinfo{person}{Y. Chen},
  \bibinfo{person}{X. Wang}, {and} \bibinfo{person}{X. Tang}.}
  \bibinfo{year}{2014}\natexlab{}.
\newblock \showarticletitle{Deep learning face representation by joint
  identification-verification}. In \bibinfo{booktitle}{\emph{Advances in neural
  information processing systems}}. \bibinfo{pages}{1988--1996}.
\newblock


\bibitem[\protect\citeauthoryear{Sun, Cheng, Zhang, Zhang, Zheng, Wang, and
  Wei}{Sun et~al\mbox{.}}{2020}]%
        {sun2020circle}
\bibfield{author}{\bibinfo{person}{Y. Sun}, \bibinfo{person}{C. Cheng},
  \bibinfo{person}{Y. Zhang}, \bibinfo{person}{C. Zhang}, \bibinfo{person}{L.
  Zheng}, \bibinfo{person}{Z. Wang}, {and} \bibinfo{person}{Y. Wei}.}
  \bibinfo{year}{2020}\natexlab{}.
\newblock \showarticletitle{Circle Loss: A Unified Perspective of Pair
  Similarity Optimization}. In \bibinfo{booktitle}{\emph{Proceedings of the
  IEEE/CVF Conference on Computer Vision and Pattern Recognition}}.
  \bibinfo{pages}{6398--6407}.
\newblock


\bibitem[\protect\citeauthoryear{Sun, Liang, Wang, and Tang}{Sun
  et~al\mbox{.}}{2015a}]%
        {Sun2015DeepID3FR}
\bibfield{author}{\bibinfo{person}{Y. Sun}, \bibinfo{person}{D. Liang},
  \bibinfo{person}{X. Wang}, {and} \bibinfo{person}{X. Tang}.}
  \bibinfo{year}{2015}\natexlab{a}.
\newblock \showarticletitle{DeepID3: Face Recognition with Very Deep Neural
  Networks}.
\newblock  (\bibinfo{year}{2015}).
\newblock
\showeprint{1502.00873}


\bibitem[\protect\citeauthoryear{{Sun}, {Wang}, and {Tang}}{{Sun}
  et~al\mbox{.}}{2013}]%
        {sun2013}
\bibfield{author}{\bibinfo{person}{Y. {Sun}}, \bibinfo{person}{X. {Wang}},
  {and} \bibinfo{person}{X. {Tang}}.} \bibinfo{year}{2013}\natexlab{}.
\newblock \showarticletitle{Deep Convolutional Network Cascade for Facial Point
  Detection}. In \bibinfo{booktitle}{\emph{Proceedings of the IEEE/CVF
  Conference on Computer Vision and Pattern Recognition}}.
  \bibinfo{pages}{3476--3483}.
\newblock


\bibitem[\protect\citeauthoryear{Sun, Wang, and Tang}{Sun
  et~al\mbox{.}}{2013}]%
        {Sun2013HybridDL}
\bibfield{author}{\bibinfo{person}{Y. Sun}, \bibinfo{person}{X. Wang}, {and}
  \bibinfo{person}{X. Tang}.} \bibinfo{year}{2013}\natexlab{}.
\newblock \showarticletitle{Hybrid Deep Learning for Face Verification}. In
  \bibinfo{booktitle}{\emph{Proceedings of the IEEE International Conference on
  Computer Vision}}. \bibinfo{pages}{1489--1496}.
\newblock


\bibitem[\protect\citeauthoryear{{Sun}, {Wang}, and {Tang}}{{Sun}
  et~al\mbox{.}}{2014}]%
        {Sun2014DeepID}
\bibfield{author}{\bibinfo{person}{Y. {Sun}}, \bibinfo{person}{X. {Wang}},
  {and} \bibinfo{person}{X. {Tang}}.} \bibinfo{year}{2014}\natexlab{}.
\newblock \showarticletitle{Deep Learning Face Representation from Predicting
  10,000 Classes}. In \bibinfo{booktitle}{\emph{Proceedings of the IEEE/CVF
  Conference on Computer Vision and Pattern Recognition}}.
  \bibinfo{pages}{1891--1898}.
\newblock


\bibitem[\protect\citeauthoryear{Sun, Wang, and Tang}{Sun
  et~al\mbox{.}}{2015b}]%
        {Sun2015DeeplyLF}
\bibfield{author}{\bibinfo{person}{Y. Sun}, \bibinfo{person}{X. Wang}, {and}
  \bibinfo{person}{X. Tang}.} \bibinfo{year}{2015}\natexlab{b}.
\newblock \showarticletitle{Deeply learned face representations are sparse,
  selective, and robust}. In \bibinfo{booktitle}{\emph{Proceedings of the
  IEEE/CVF Conference on Computer Vision and Pattern Recognition Workshops}}.
  \bibinfo{pages}{2892--2900}.
\newblock


\bibitem[\protect\citeauthoryear{Sun, Wang, and Tang}{Sun
  et~al\mbox{.}}{2016}]%
        {sun2015sparsifying}
\bibfield{author}{\bibinfo{person}{Y. Sun}, \bibinfo{person}{X. Wang}, {and}
  \bibinfo{person}{X. Tang}.} \bibinfo{year}{2016}\natexlab{}.
\newblock \showarticletitle{Sparsifying neural network connections for face
  recognition}. In \bibinfo{booktitle}{\emph{Proceedings of the IEEE/CVF
  Conference on Computer Vision and Pattern Recognition}}.
  \bibinfo{pages}{4856--4864}.
\newblock


\bibitem[\protect\citeauthoryear{Szegedy, Liu, Jia, Sermanet, Reed, Anguelov,
  Erhan, Vanhoucke, and Rabinovich}{Szegedy et~al\mbox{.}}{2015}]%
        {Szegedy2015GoingDW}
\bibfield{author}{\bibinfo{person}{C. Szegedy}, \bibinfo{person}{W. Liu},
  \bibinfo{person}{Y. Jia}, \bibinfo{person}{P. Sermanet}, \bibinfo{person}{S.
  Reed}, \bibinfo{person}{D. Anguelov}, \bibinfo{person}{D. Erhan},
  \bibinfo{person}{V. Vanhoucke}, {and} \bibinfo{person}{A. Rabinovich}.}
  \bibinfo{year}{2015}\natexlab{}.
\newblock \showarticletitle{Going deeper with convolutions}. In
  \bibinfo{booktitle}{\emph{Proceedings of the IEEE/CVF Conference on Computer
  Vision and Pattern Recognition}}. \bibinfo{pages}{1--9}.
\newblock


\bibitem[\protect\citeauthoryear{Tadmor, Rosenwein, Shalev-Shwartz, Wexler, and
  Shashua}{Tadmor et~al\mbox{.}}{2016}]%
        {Tadmor2016LearningAM}
\bibfield{author}{\bibinfo{person}{O. Tadmor}, \bibinfo{person}{T. Rosenwein},
  \bibinfo{person}{S. Shalev-Shwartz}, \bibinfo{person}{Y. Wexler}, {and}
  \bibinfo{person}{A. Shashua}.} \bibinfo{year}{2016}\natexlab{}.
\newblock \showarticletitle{Learning a Metric Embedding for Face Recognition
  using the Multibatch Method}. In \bibinfo{booktitle}{\emph{Advances in neural
  information processing systems}}. \bibinfo{pages}{1396–1397}.
\newblock


\bibitem[\protect\citeauthoryear{Tai, Liang, Liu, Duan, Li, Wang, Huang, and
  Chen}{Tai et~al\mbox{.}}{2019}]%
        {Tai2019TowardsHA}
\bibfield{author}{\bibinfo{person}{Y. Tai}, \bibinfo{person}{Y. Liang},
  \bibinfo{person}{X. Liu}, \bibinfo{person}{L. Duan}, \bibinfo{person}{J. Li},
  \bibinfo{person}{C. Wang}, \bibinfo{person}{F. Huang}, {and}
  \bibinfo{person}{Y. Chen}.} \bibinfo{year}{2019}\natexlab{}.
\newblock \showarticletitle{Towards Highly Accurate and Stable Face Alignment
  for High-Resolution Videos}. In \bibinfo{booktitle}{\emph{Proceedings of the
  AAAI Conference on Artificial Intelligence}}, Vol.~\bibinfo{volume}{33}.
  \bibinfo{pages}{8893--8900}.
\newblock


\bibitem[\protect\citeauthoryear{Taigman, Yang, Ranzato, and Wolf}{Taigman
  et~al\mbox{.}}{2014}]%
        {taigman2014deepface}
\bibfield{author}{\bibinfo{person}{Y. Taigman}, \bibinfo{person}{M. Yang},
  \bibinfo{person}{M. Ranzato}, {and} \bibinfo{person}{L. Wolf}.}
  \bibinfo{year}{2014}\natexlab{}.
\newblock \showarticletitle{Deepface: Closing the gap to human-level
  performance in face verification}. In \bibinfo{booktitle}{\emph{Proceedings
  of the IEEE/CVF Conference on Computer Vision and Pattern Recognition}}.
  \bibinfo{pages}{1701--1708}.
\newblock


\bibitem[\protect\citeauthoryear{Tang, Du, He, and Liu}{Tang
  et~al\mbox{.}}{2018a}]%
        {tang2018pyramidbox}
\bibfield{author}{\bibinfo{person}{X. Tang}, \bibinfo{person}{D.~K. Du},
  \bibinfo{person}{Z. He}, {and} \bibinfo{person}{J. Liu}.}
  \bibinfo{year}{2018}\natexlab{a}.
\newblock \showarticletitle{Pyramidbox: a context-assisted single shot face
  detector}. In \bibinfo{booktitle}{\emph{Proceedings of the European
  Conference on Computer Vision}}. \bibinfo{pages}{797--813}.
\newblock


\bibitem[\protect\citeauthoryear{Tang, Peng, Geng, Wu, Zhang, and Metaxas}{Tang
  et~al\mbox{.}}{2018b}]%
        {Tang2018QuantizedDC}
\bibfield{author}{\bibinfo{person}{Zhiqiang Tang}, \bibinfo{person}{Xi Peng},
  \bibinfo{person}{Shijie Geng}, \bibinfo{person}{Lingfei Wu},
  \bibinfo{person}{Shaoting Zhang}, {and} \bibinfo{person}{Dimitris~N.
  Metaxas}.} \bibinfo{year}{2018}\natexlab{b}.
\newblock \showarticletitle{Quantized Densely Connected U-Nets for Efficient
  Landmark Localization}. In \bibinfo{booktitle}{\emph{Proceedings of the
  European Conference on Computer Vision}}. \bibinfo{pages}{348--364}.
\newblock


\bibitem[\protect\citeauthoryear{Tian, Wang, Shen, Deng, Meng, Chen, Zhang,
  Zhao, and Huang}{Tian et~al\mbox{.}}{2018}]%
        {tian2018df2s2}
\bibfield{author}{\bibinfo{person}{W. Tian}, \bibinfo{person}{Z. Wang},
  \bibinfo{person}{H. Shen}, \bibinfo{person}{W. Deng}, \bibinfo{person}{Y.
  Meng}, \bibinfo{person}{B. Chen}, \bibinfo{person}{X. Zhang},
  \bibinfo{person}{Y. Zhao}, {and} \bibinfo{person}{X. Huang}.}
  \bibinfo{year}{2018}\natexlab{}.
\newblock \showarticletitle{Learning Better Features for Face Detection with
  Feature Fusion and Segmentation Supervision}.
\newblock
\showeprint{1811.08557}


\bibitem[\protect\citeauthoryear{{Tian}, {Shen}, {Chen}, and {He}}{{Tian}
  et~al\mbox{.}}{2019}]%
        {Tian2019FCOS}
\bibfield{author}{\bibinfo{person}{Z. {Tian}}, \bibinfo{person}{C. {Shen}},
  \bibinfo{person}{H. {Chen}}, {and} \bibinfo{person}{T. {He}}.}
  \bibinfo{year}{2019}\natexlab{}.
\newblock \showarticletitle{FCOS: Fully Convolutional One-Stage Object
  Detection}. In \bibinfo{booktitle}{\emph{Proceedings of the IEEE
  International Conference on Computer Vision}}. \bibinfo{pages}{9626--9635}.
\newblock


\bibitem[\protect\citeauthoryear{Tran, Yin, and Liu}{Tran
  et~al\mbox{.}}{2017}]%
        {Tran2017DisentangledRL}
\bibfield{author}{\bibinfo{person}{L. Tran}, \bibinfo{person}{X. Yin}, {and}
  \bibinfo{person}{X. Liu}.} \bibinfo{year}{2017}\natexlab{}.
\newblock \showarticletitle{Disentangled Representation Learning GAN for
  Pose-Invariant Face Recognition}. In \bibinfo{booktitle}{\emph{Proceedings of
  the IEEE/CVF Conference on Computer Vision and Pattern Recognition}}.
  \bibinfo{pages}{1283--1292}.
\newblock


\bibitem[\protect\citeauthoryear{{Trigeorgis}, {Snape}, {Nicolaou},
  {Antonakos}, and {Zafeiriou}}{{Trigeorgis} et~al\mbox{.}}{2016}]%
        {Trigeorgis2016MDM}
\bibfield{author}{\bibinfo{person}{G. {Trigeorgis}}, \bibinfo{person}{P.
  {Snape}}, \bibinfo{person}{M.~A. {Nicolaou}}, \bibinfo{person}{E.
  {Antonakos}}, {and} \bibinfo{person}{S. {Zafeiriou}}.}
  \bibinfo{year}{2016}\natexlab{}.
\newblock \showarticletitle{Mnemonic Descent Method: A Recurrent Process
  Applied for End-to-End Face Alignment}. In \bibinfo{booktitle}{\emph{IEEE
  Conference on Computer Vision and Pattern Recognition}}.
  \bibinfo{pages}{4177--4187}.
\newblock


\bibitem[\protect\citeauthoryear{Turk and Pentland}{Turk and Pentland}{1991}]%
        {1991Eigenfaces}
\bibfield{author}{\bibinfo{person}{M. Turk} {and} \bibinfo{person}{A.
  Pentland}.} \bibinfo{year}{1991}\natexlab{}.
\newblock \showarticletitle{Eigenfaces for Recognition}.
\newblock \bibinfo{journal}{\emph{Journal of Cognitive Neuroscience}}
  \bibinfo{volume}{3}, \bibinfo{number}{1} (\bibinfo{year}{1991}),
  \bibinfo{pages}{71--86}.
\newblock


\bibitem[\protect\citeauthoryear{T.Zheng, Deng, and Hu}{T.Zheng
  et~al\mbox{.}}{2017}]%
        {Zheng2017AgeEG}
\bibfield{author}{\bibinfo{person}{T.Zheng}, \bibinfo{person}{W. Deng}, {and}
  \bibinfo{person}{J. Hu}.} \bibinfo{year}{2017}\natexlab{}.
\newblock \showarticletitle{Age Estimation Guided Convolutional Neural Network
  for Age-Invariant Face Recognition}. In \bibinfo{booktitle}{\emph{Proceedings
  of the IEEE/CVF Conference on Computer Vision and Pattern RecognitionW}}.
  \bibinfo{pages}{503--511}.
\newblock


\bibitem[\protect\citeauthoryear{V and Learned-Miller}{V and
  Learned-Miller}{2010}]%
        {fddbTech}
\bibfield{author}{\bibinfo{person}{Jain V} {and} \bibinfo{person}{E.
  Learned-Miller}.} \bibinfo{year}{2010}\natexlab{}.
\newblock \bibinfo{booktitle}{\emph{FDDB: A Benchmark for Face Detection in
  Unconstrained Settings}}.
\newblock \bibinfo{type}{{T}echnical {R}eport} UM-CS-2010-009.
\newblock


\bibitem[\protect\citeauthoryear{Viola and Jones}{Viola and Jones}{2001}]%
        {viola2001rapid}
\bibfield{author}{\bibinfo{person}{Paul Viola} {and} \bibinfo{person}{Michael
  Jones}.} \bibinfo{year}{2001}\natexlab{}.
\newblock \showarticletitle{Rapid object detection using a boosted cascade of
  simple features}. In \bibinfo{booktitle}{\emph{Proceedings of the conference
  on computer vision and pattern recognition}}, Vol.~\bibinfo{volume}{1}.
  \bibinfo{pages}{I--I}.
\newblock


\bibitem[\protect\citeauthoryear{w, Chellappa, Phillips, and Rosenfeld}{w
  et~al\mbox{.}}{2003}]%
        {W2003Face}
\bibfield{author}{\bibinfo{person}{Zhao w}, \bibinfo{person}{R. Chellappa},
  \bibinfo{person}{P.~J. Phillips}, {and} \bibinfo{person}{A. Rosenfeld}.}
  \bibinfo{year}{2003}\natexlab{}.
\newblock \showarticletitle{Face recognition: A literature survey}.
\newblock \bibinfo{journal}{\emph{ACM computing surveys}} \bibinfo{volume}{35},
  \bibinfo{number}{4} (\bibinfo{year}{2003}), \bibinfo{pages}{399--458}.
\newblock


\bibitem[\protect\citeauthoryear{Wan, Chen, T.Zhang, Zhang, and Wong}{Wan
  et~al\mbox{.}}{2016}]%
        {wan2016bootstrapping}
\bibfield{author}{\bibinfo{person}{S. Wan}, \bibinfo{person}{Z. Chen},
  \bibinfo{person}{T.Zhang}, \bibinfo{person}{B. Zhang}, {and}
  \bibinfo{person}{K. Wong}.} \bibinfo{year}{2016}\natexlab{}.
\newblock \showarticletitle{Bootstrapping face detection with hard negative
  examples}.
\newblock  (\bibinfo{year}{2016}).
\newblock
\showeprint{1608.02236}


\bibitem[\protect\citeauthoryear{Wan, Gao, and Lee}{Wan et~al\mbox{.}}{2019}]%
        {Wan2019TransferDF}
\bibfield{author}{\bibinfo{person}{W. Wan}, \bibinfo{person}{Y. Gao}, {and}
  \bibinfo{person}{H.~J. Lee}.} \bibinfo{year}{2019}\natexlab{}.
\newblock \showarticletitle{Transfer deep feature learning for face sketch
  recognition}.
\newblock \bibinfo{journal}{\emph{Neural Computing and Applications}}
  (\bibinfo{year}{2019}), \bibinfo{pages}{1--10}.
\newblock


\bibitem[\protect\citeauthoryear{Wang, Chen, Li, Huang, Chen, Qian, and
  Loy}{Wang et~al\mbox{.}}{2018a}]%
        {Wang2018TheDO}
\bibfield{author}{\bibinfo{person}{F. Wang}, \bibinfo{person}{L. Chen},
  \bibinfo{person}{C. Li}, \bibinfo{person}{S. Huang}, \bibinfo{person}{Y.
  Chen}, \bibinfo{person}{C. Qian}, {and} \bibinfo{person}{C.~C. Loy}.}
  \bibinfo{year}{2018}\natexlab{a}.
\newblock \showarticletitle{The Devil of Face Recognition is in the Noise}. In
  \bibinfo{booktitle}{\emph{Proceedings of the European Conference on Computer
  Vision}}. \bibinfo{pages}{765--780}.
\newblock


\bibitem[\protect\citeauthoryear{Wang, Cheng, Liu, and Liu}{Wang
  et~al\mbox{.}}{2018b}]%
        {wang2018additive}
\bibfield{author}{\bibinfo{person}{F. Wang}, \bibinfo{person}{J. Cheng},
  \bibinfo{person}{W. Liu}, {and} \bibinfo{person}{H. Liu}.}
  \bibinfo{year}{2018}\natexlab{b}.
\newblock \showarticletitle{Additive margin softmax for face verification}.
\newblock \bibinfo{journal}{\emph{IEEE Singal processing letters}}
  \bibinfo{volume}{25}, \bibinfo{number}{7} (\bibinfo{year}{2018}),
  \bibinfo{pages}{926--930}.
\newblock


\bibitem[\protect\citeauthoryear{Wang, Jiang, Qian, Yang, Li, Zhang, Wang, and
  Tang}{Wang et~al\mbox{.}}{2017b}]%
        {Wang2017ResidualAN}
\bibfield{author}{\bibinfo{person}{F. Wang}, \bibinfo{person}{M. Jiang},
  \bibinfo{person}{C. Qian}, \bibinfo{person}{S. Yang}, \bibinfo{person}{C.
  Li}, \bibinfo{person}{H. Zhang}, \bibinfo{person}{X. Wang}, {and}
  \bibinfo{person}{X. Tang}.} \bibinfo{year}{2017}\natexlab{b}.
\newblock \showarticletitle{Residual Attention Network for Image
  Classification}. In \bibinfo{booktitle}{\emph{Proceedings of the IEEE/CVF
  Conference on Computer Vision and Pattern Recognition}}.
  \bibinfo{pages}{6450--6458}.
\newblock


\bibitem[\protect\citeauthoryear{Wang, Xiang, Cheng, and Yuille}{Wang
  et~al\mbox{.}}{2017c}]%
        {wang2017normface}
\bibfield{author}{\bibinfo{person}{F. Wang}, \bibinfo{person}{X. Xiang},
  \bibinfo{person}{J. Cheng}, {and} \bibinfo{person}{A.~L. Yuille}.}
  \bibinfo{year}{2017}\natexlab{c}.
\newblock \showarticletitle{Normface: l 2 hypersphere embedding for face
  verification}. In \bibinfo{booktitle}{\emph{Proceedings of the 25th ACM
  international conference on Multimedia}}. \bibinfo{pages}{1041--1049}.
\newblock


\bibitem[\protect\citeauthoryear{Wang, Gong, Li, and Liu}{Wang
  et~al\mbox{.}}{2019b}]%
        {Wang2019DecorrelatedAL}
\bibfield{author}{\bibinfo{person}{H. Wang}, \bibinfo{person}{D. Gong},
  \bibinfo{person}{Z. Li}, {and} \bibinfo{person}{W. Liu}.}
  \bibinfo{year}{2019}\natexlab{b}.
\newblock \showarticletitle{Decorrelated Adversarial Learning for Age-Invariant
  Face Recognition}. In \bibinfo{booktitle}{\emph{Proceedings of the IEEE/CVF
  Conference on Computer Vision and Pattern Recognition}}.
  \bibinfo{pages}{3522--3531}.
\newblock


\bibitem[\protect\citeauthoryear{Wang, Wang, Zhou, Ji, Gong, Zhou, Li, and
  Liu}{Wang et~al\mbox{.}}{2018g}]%
        {wang2018cosface}
\bibfield{author}{\bibinfo{person}{H. Wang}, \bibinfo{person}{Y. Wang},
  \bibinfo{person}{Z. Zhou}, \bibinfo{person}{X. Ji}, \bibinfo{person}{D.
  Gong}, \bibinfo{person}{J. Zhou}, \bibinfo{person}{Z. Li}, {and}
  \bibinfo{person}{W. Liu}.} \bibinfo{year}{2018}\natexlab{g}.
\newblock \showarticletitle{Cosface: Large margin cosine loss for deep face
  recognition}. In \bibinfo{booktitle}{\emph{Proceedings of the IEEE/CVF
  Conference on Computer Vision and Pattern Recognition}}.
  \bibinfo{pages}{5265--5274}.
\newblock


\bibitem[\protect\citeauthoryear{Wang, k.~Sun, Cheng, Jiang, Deng, Zhao, Liu,
  Mu, Tan, Wang, Liu, and Xiao}{Wang et~al\mbox{.}}{2021}]%
        {wang2020deep}
\bibfield{author}{\bibinfo{person}{J. Wang}, \bibinfo{person}{k. Sun},
  \bibinfo{person}{T. Cheng}, \bibinfo{person}{B. Jiang}, \bibinfo{person}{C.
  Deng}, \bibinfo{person}{Y. Zhao}, \bibinfo{person}{D. Liu},
  \bibinfo{person}{Y. Mu}, \bibinfo{person}{M. Tan}, \bibinfo{person}{X. Wang},
  \bibinfo{person}{W. Liu}, {and} \bibinfo{person}{B. Xiao}.}
  \bibinfo{year}{2021}\natexlab{}.
\newblock \showarticletitle{Deep high-resolution representation learning for
  visual recognition}.
\newblock \bibinfo{journal}{\emph{IEEE Trans. Pattern Anal. Mach. Intell.}}
  \bibinfo{volume}{43} (\bibinfo{year}{2021}), \bibinfo{pages}{3349--3364}.
\newblock


\bibitem[\protect\citeauthoryear{Wang, Yuan, Li, Yu, and Jian}{Wang
  et~al\mbox{.}}{2018h}]%
        {wang2018sface}
\bibfield{author}{\bibinfo{person}{J. Wang}, \bibinfo{person}{Y. Yuan},
  \bibinfo{person}{B. Li}, \bibinfo{person}{G. Yu}, {and} \bibinfo{person}{S.
  Jian}.} \bibinfo{year}{2018}\natexlab{h}.
\newblock \showarticletitle{Sface: An efficient network for face detection in
  large scale variations}.
\newblock  (\bibinfo{year}{2018}).
\newblock
\showeprint{1804.06559}


\bibitem[\protect\citeauthoryear{Wang, Yuan, and Yu}{Wang
  et~al\mbox{.}}{2017d}]%
        {wang2017fan}
\bibfield{author}{\bibinfo{person}{J. Wang}, \bibinfo{person}{Y. Yuan}, {and}
  \bibinfo{person}{G. Yu}.} \bibinfo{year}{2017}\natexlab{d}.
\newblock \showarticletitle{Face attention network: an effective face detector
  for the occluded faces}.
\newblock  (\bibinfo{year}{2017}).
\newblock
\showeprint{1711.07246}


\bibitem[\protect\citeauthoryear{Wang, Sindagi, and Patel}{Wang
  et~al\mbox{.}}{2018e}]%
        {Wang2018HighQualityFP}
\bibfield{author}{\bibinfo{person}{L. Wang}, \bibinfo{person}{V. Sindagi},
  {and} \bibinfo{person}{V.~M. Patel}.} \bibinfo{year}{2018}\natexlab{e}.
\newblock \showarticletitle{High-Quality Facial Photo-Sketch Synthesis Using
  Multi-Adversarial Networks}. In \bibinfo{booktitle}{\emph{Proceedings of the
  IEEE International Conference on Automatic Face $\&$ Gesture Recognition}}.
  \bibinfo{pages}{83--90}.
\newblock


\bibitem[\protect\citeauthoryear{Wang and Deng}{Wang and Deng}{2018}]%
        {wang2018deep}
\bibfield{author}{\bibinfo{person}{M. Wang} {and} \bibinfo{person}{W. Deng}.}
  \bibinfo{year}{2018}\natexlab{}.
\newblock \showarticletitle{Deep Face Recognition: A Survey}.
\newblock \bibinfo{journal}{\emph{Neurocomputing}}  \bibinfo{volume}{312}
  (\bibinfo{year}{2018}), \bibinfo{pages}{135--153}.
\newblock


\bibitem[\protect\citeauthoryear{Wang and Deng}{Wang and Deng}{2020}]%
        {Wang_2020_CVPR}
\bibfield{author}{\bibinfo{person}{M. Wang} {and} \bibinfo{person}{W. Deng}.}
  \bibinfo{year}{2020}\natexlab{}.
\newblock \showarticletitle{Mitigating Bias in Face Recognition Using
  Skewness-Aware Reinforcement Learning}. In
  \bibinfo{booktitle}{\emph{Proceedings of the IEEE/CVF Conference on Computer
  Vision and Pattern Recognition}}. \bibinfo{pages}{9322--9331}.
\newblock


\bibitem[\protect\citeauthoryear{Wang, Deng, Hu, Tao, and Huang}{Wang
  et~al\mbox{.}}{2019a}]%
        {Wang2019RacialFI}
\bibfield{author}{\bibinfo{person}{M. Wang}, \bibinfo{person}{W. Deng},
  \bibinfo{person}{J. Hu}, \bibinfo{person}{X. Tao}, {and} \bibinfo{person}{Y.
  Huang}.} \bibinfo{year}{2019}\natexlab{a}.
\newblock \showarticletitle{Racial Faces in the Wild: Reducing Racial Bias by
  Information Maximization Adaptation Network}. In
  \bibinfo{booktitle}{\emph{Proceedings of the IEEE International Conference on
  Computer Vision}}. \bibinfo{pages}{692--702}.
\newblock


\bibitem[\protect\citeauthoryear{Wang, Gao, Tao, Yang, and Li}{Wang
  et~al\mbox{.}}{2018c}]%
        {Wang2018FacialFP}
\bibfield{author}{\bibinfo{person}{N. Wang}, \bibinfo{person}{X. Gao},
  \bibinfo{person}{D. Tao}, \bibinfo{person}{H. Yang}, {and}
  \bibinfo{person}{X. Li}.} \bibinfo{year}{2018}\natexlab{c}.
\newblock \showarticletitle{Facial feature point detection: A comprehensive
  survey}.
\newblock \bibinfo{journal}{\emph{Neurocomputing}}  \bibinfo{volume}{275}
  (\bibinfo{year}{2018}), \bibinfo{pages}{50--65}.
\newblock


\bibitem[\protect\citeauthoryear{Wang, Cui, Yan, Feng, Yan, Shu, and Sebe}{Wang
  et~al\mbox{.}}{2016}]%
        {Wang2016RecurrentFA}
\bibfield{author}{\bibinfo{person}{W. Wang}, \bibinfo{person}{Z. Cui},
  \bibinfo{person}{Y. Yan}, \bibinfo{person}{J. Feng}, \bibinfo{person}{S.
  Yan}, \bibinfo{person}{X. Shu}, {and} \bibinfo{person}{N. Sebe}.}
  \bibinfo{year}{2016}\natexlab{}.
\newblock \showarticletitle{Recurrent Face Aging}. In
  \bibinfo{booktitle}{\emph{Proceedings of the IEEE/CVF Conference on Computer
  Vision and Pattern Recognition}}. \bibinfo{pages}{2378--2386}.
\newblock


\bibitem[\protect\citeauthoryear{{Wang}, {Bo}, and {Fuxin}}{{Wang}
  et~al\mbox{.}}{2019}]%
        {Wang2019adawing}
\bibfield{author}{\bibinfo{person}{X. {Wang}}, \bibinfo{person}{L. {Bo}}, {and}
  \bibinfo{person}{L. {Fuxin}}.} \bibinfo{year}{2019}\natexlab{}.
\newblock \showarticletitle{Adaptive Wing Loss for Robust Face Alignment via
  Heatmap Regression}. In \bibinfo{booktitle}{\emph{Proceedings of the IEEE
  International Conference on Computer Vision}}. \bibinfo{pages}{6970--6980}.
\newblock


\bibitem[\protect\citeauthoryear{Wang, Wang, Wang, Shi, and Mei}{Wang
  et~al\mbox{.}}{2019c}]%
        {wang2019co}
\bibfield{author}{\bibinfo{person}{X. Wang}, \bibinfo{person}{S. Wang},
  \bibinfo{person}{J. Wang}, \bibinfo{person}{H. Shi}, {and}
  \bibinfo{person}{T. Mei}.} \bibinfo{year}{2019}\natexlab{c}.
\newblock \showarticletitle{Co-Mining: Deep Face Recognition With Noisy
  Labels}. In \bibinfo{booktitle}{\emph{Proceedings of the IEEE International
  Conference on Computer Vision}}. \bibinfo{pages}{9358--9367}.
\newblock


\bibitem[\protect\citeauthoryear{Wang, Zhang, Wang, Fu, Shi, and Mei}{Wang
  et~al\mbox{.}}{2020}]%
        {Wang2019MisclassifiedVG}
\bibfield{author}{\bibinfo{person}{Xiaobo Wang}, \bibinfo{person}{Shifeng
  Zhang}, \bibinfo{person}{Shuo Wang}, \bibinfo{person}{Tianyu Fu},
  \bibinfo{person}{Hailin Shi}, {and} \bibinfo{person}{Tao Mei}.}
  \bibinfo{year}{2020}\natexlab{}.
\newblock \showarticletitle{Mis-classified vector guided softmax loss for face
  recognition}. In \bibinfo{booktitle}{\emph{Proceedings of the AAAI Conference
  on Artificial Intelligence}}, Vol.~\bibinfo{volume}{34}.
  \bibinfo{pages}{12241--12248}.
\newblock


\bibitem[\protect\citeauthoryear{Wang, Gong, Zhou, Ji, Wang, Li, Liu, and
  Zhang}{Wang et~al\mbox{.}}{2018d}]%
        {Wang2018OrthogonalDF}
\bibfield{author}{\bibinfo{person}{Y. Wang}, \bibinfo{person}{D. Gong},
  \bibinfo{person}{Z. Zhou}, \bibinfo{person}{X. Ji}, \bibinfo{person}{H.
  Wang}, \bibinfo{person}{Z. Li}, \bibinfo{person}{W. Liu}, {and}
  \bibinfo{person}{T. Zhang}.} \bibinfo{year}{2018}\natexlab{d}.
\newblock \showarticletitle{Orthogonal Deep Features Decomposition for
  Age-Invariant Face Recognition}. In \bibinfo{booktitle}{\emph{Proceedings of
  the European Conference on Computer Vision}}. \bibinfo{pages}{738--753}.
\newblock


\bibitem[\protect\citeauthoryear{Wang, Ji, Z.Zhou, Wang, and Li}{Wang
  et~al\mbox{.}}{2017a}]%
        {wang2017facerfcn}
\bibfield{author}{\bibinfo{person}{Y. Wang}, \bibinfo{person}{X. Ji},
  \bibinfo{person}{Z.Zhou}, \bibinfo{person}{H. Wang}, {and}
  \bibinfo{person}{Z. Li}.} \bibinfo{year}{2017}\natexlab{a}.
\newblock \showarticletitle{Detecting faces using region-based fully
  convolutional networks}.
\newblock  (\bibinfo{year}{2017}).
\newblock
\showeprint{1709.05256}


\bibitem[\protect\citeauthoryear{Wang, Tang, Luo, and Gao}{Wang
  et~al\mbox{.}}{2018f}]%
        {Wang2018FaceAW}
\bibfield{author}{\bibinfo{person}{Z. Wang}, \bibinfo{person}{X. Tang},
  \bibinfo{person}{W. Luo}, {and} \bibinfo{person}{S. Gao}.}
  \bibinfo{year}{2018}\natexlab{f}.
\newblock \showarticletitle{Face Aging with Identity-Preserved Conditional
  Generative Adversarial Networks}. In \bibinfo{booktitle}{\emph{Proceedings of
  the IEEE/CVF Conference on Computer Vision and Pattern Recognition}}.
  \bibinfo{pages}{7939--7947}.
\newblock


\bibitem[\protect\citeauthoryear{Wang, Zheng, Li, and Wang}{Wang
  et~al\mbox{.}}{2019d}]%
        {wang2019linkage}
\bibfield{author}{\bibinfo{person}{Z. Wang}, \bibinfo{person}{L. Zheng},
  \bibinfo{person}{Y. Li}, {and} \bibinfo{person}{S. Wang}.}
  \bibinfo{year}{2019}\natexlab{d}.
\newblock \showarticletitle{Linkage based face clustering via graph convolution
  network}. In \bibinfo{booktitle}{\emph{Proceedings of the IEEE/CVF Conference
  on Computer Vision and Pattern Recognition}}. \bibinfo{pages}{1117--1125}.
\newblock


\bibitem[\protect\citeauthoryear{Wei, Lu, and Wei}{Wei et~al\mbox{.}}{2020}]%
        {Wei2020BalancedAF}
\bibfield{author}{\bibinfo{person}{H. Wei}, \bibinfo{person}{P. Lu}, {and}
  \bibinfo{person}{Y. Wei}.} \bibinfo{year}{2020}\natexlab{}.
\newblock \showarticletitle{Balanced Alignment for Face Recognition: A Joint
  Learning Approach}.
\newblock  (\bibinfo{year}{2020}).
\newblock
\showeprint{2003.10168}


\bibitem[\protect\citeauthoryear{Weinberger and Saul}{Weinberger and
  Saul}{2006}]%
        {Weinberger2005DistanceML}
\bibfield{author}{\bibinfo{person}{K.~Q. Weinberger} {and}
  \bibinfo{person}{L.~K. Saul}.} \bibinfo{year}{2006}\natexlab{}.
\newblock \showarticletitle{Distance Metric Learning for Large Margin Nearest
  Neighbor Classification}. In \bibinfo{booktitle}{\emph{Advances in neural
  information processing systems}}. \bibinfo{pages}{1473--1480}.
\newblock


\bibitem[\protect\citeauthoryear{Wen, Li, and Qiao}{Wen et~al\mbox{.}}{2016a}]%
        {Wen2016LatentFG}
\bibfield{author}{\bibinfo{person}{Y. Wen}, \bibinfo{person}{Z. Li}, {and}
  \bibinfo{person}{Y. Qiao}.} \bibinfo{year}{2016}\natexlab{a}.
\newblock \showarticletitle{Latent Factor Guided Convolutional Neural Networks
  for Age-Invariant Face Recognition}. In \bibinfo{booktitle}{\emph{Proceedings
  of the IEEE/CVF Conference on Computer Vision and Pattern Recognition}}.
  \bibinfo{pages}{4893--4901}.
\newblock


\bibitem[\protect\citeauthoryear{Wen, Zhang, Li, and Qiao}{Wen
  et~al\mbox{.}}{2016b}]%
        {wen2016discriminative}
\bibfield{author}{\bibinfo{person}{Y. Wen}, \bibinfo{person}{K. Zhang},
  \bibinfo{person}{Z. Li}, {and} \bibinfo{person}{Y. Qiao}.}
  \bibinfo{year}{2016}\natexlab{b}.
\newblock \showarticletitle{A discriminative feature learning approach for deep
  face recognition}. In \bibinfo{booktitle}{\emph{Proceedings of the European
  Conference on Computer Vision}}. \bibinfo{pages}{499--515}.
\newblock


\bibitem[\protect\citeauthoryear{Whitelam, Taborsky, Blanton, Maze, Adams,
  Miller, Kalka, Jain, Duncan, Allen, Cheney, and Grother}{Whitelam
  et~al\mbox{.}}{2017}]%
        {Whitelam2017IARPAJB}
\bibfield{author}{\bibinfo{person}{C. Whitelam}, \bibinfo{person}{E. Taborsky},
  \bibinfo{person}{A. Blanton}, \bibinfo{person}{B. Maze},
  \bibinfo{person}{J.~C. Adams}, \bibinfo{person}{T. Miller},
  \bibinfo{person}{N.~D. Kalka}, \bibinfo{person}{A.~K. Jain},
  \bibinfo{person}{J.~A. Duncan}, \bibinfo{person}{K.~E Allen},
  \bibinfo{person}{J. Cheney}, {and} \bibinfo{person}{P. Grother}.}
  \bibinfo{year}{2017}\natexlab{}.
\newblock \showarticletitle{IARPA Janus Benchmark-B Face Dataset}. In
  \bibinfo{booktitle}{\emph{Proceedings of the IEEE/CVF Conference on Computer
  Vision and Pattern RecognitionW}}. \bibinfo{pages}{592--600}.
\newblock


\bibitem[\protect\citeauthoryear{Williford, May, and Byrne}{Williford
  et~al\mbox{.}}{2020}]%
        {williford2020explainable}
\bibfield{author}{\bibinfo{person}{Jonathan~R Williford},
  \bibinfo{person}{Brandon~B May}, {and} \bibinfo{person}{Jeffrey Byrne}.}
  \bibinfo{year}{2020}\natexlab{}.
\newblock \showarticletitle{Explainable Face Recognition}. In
  \bibinfo{booktitle}{\emph{Proceedings of the European Conference on Computer
  Vision}}. \bibinfo{pages}{248--263}.
\newblock


\bibitem[\protect\citeauthoryear{Wolf, Hassner, and Maoz}{Wolf
  et~al\mbox{.}}{2011}]%
        {Wolf2011FaceRI}
\bibfield{author}{\bibinfo{person}{L. Wolf}, \bibinfo{person}{T. Hassner},
  {and} \bibinfo{person}{I. Maoz}.} \bibinfo{year}{2011}\natexlab{}.
\newblock \showarticletitle{Face recognition in unconstrained videos with
  matched background similarity}. In \bibinfo{booktitle}{\emph{Proceedings of
  the IEEE/CVF Conference on Computer Vision and Pattern Recognition}}.
  \bibinfo{pages}{529--534}.
\newblock


\bibitem[\protect\citeauthoryear{{Wu}, {Kan}, {Liu}, {Yang}, {Shan}, and
  {Chen}}{{Wu} et~al\mbox{.}}{2017}]%
        {Wu2017ReST}
\bibfield{author}{\bibinfo{person}{W. {Wu}}, \bibinfo{person}{M. {Kan}},
  \bibinfo{person}{X. {Liu}}, \bibinfo{person}{Y. {Yang}}, \bibinfo{person}{S.
  {Shan}}, {and} \bibinfo{person}{X. {Chen}}.} \bibinfo{year}{2017}\natexlab{}.
\newblock \showarticletitle{Recursive Spatial Transformer (ReST) for
  Alignment-Free Face Recognition}. In \bibinfo{booktitle}{\emph{Proceedings of
  the IEEE International Conference on Computer Vision}}.
  \bibinfo{pages}{3792--3800}.
\newblock


\bibitem[\protect\citeauthoryear{{Wu}, {Qian}, {Yang}, {Wang}, {Cai}, and
  {Zhou}}{{Wu} et~al\mbox{.}}{2018}]%
        {Wu2018LAB}
\bibfield{author}{\bibinfo{person}{W. {Wu}}, \bibinfo{person}{C. {Qian}},
  \bibinfo{person}{S. {Yang}}, \bibinfo{person}{Q. {Wang}}, \bibinfo{person}{Y.
  {Cai}}, {and} \bibinfo{person}{Q. {Zhou}}.} \bibinfo{year}{2018}\natexlab{}.
\newblock \showarticletitle{Look at Boundary: A Boundary-Aware Face Alignment
  Algorithm}. In \bibinfo{booktitle}{\emph{Proceedings of the IEEE/CVF
  Conference on Computer Vision and Pattern Recognition}}.
  \bibinfo{pages}{2129--2138}.
\newblock


\bibitem[\protect\citeauthoryear{Wu, He, and Sun}{Wu et~al\mbox{.}}{2015}]%
        {Wu2015ALC}
\bibfield{author}{\bibinfo{person}{X. Wu}, \bibinfo{person}{R. He}, {and}
  \bibinfo{person}{Z. Sun}.} \bibinfo{year}{2015}\natexlab{}.
\newblock \showarticletitle{A Lightened CNN for Deep Face Representation}.
\newblock  (\bibinfo{year}{2015}).
\newblock
\showeprint{1511.02683}


\bibitem[\protect\citeauthoryear{Wu, He, Sun, and Tan}{Wu
  et~al\mbox{.}}{2018}]%
        {Wu2018ALC}
\bibfield{author}{\bibinfo{person}{X. Wu}, \bibinfo{person}{R. He},
  \bibinfo{person}{Z. Sun}, {and} \bibinfo{person}{T. Tan}.}
  \bibinfo{year}{2018}\natexlab{}.
\newblock \showarticletitle{A Light CNN for Deep Face Representation With Noisy
  Labels}.
\newblock \bibinfo{journal}{\emph{IEEE Transactions on Information Forensics
  and Security}}  \bibinfo{volume}{13} (\bibinfo{year}{2018}),
  \bibinfo{pages}{2884--2896}.
\newblock


\bibitem[\protect\citeauthoryear{Wu, Huang, Patel, He, and Sun}{Wu
  et~al\mbox{.}}{2019}]%
        {Wu2018DisentangledVR}
\bibfield{author}{\bibinfo{person}{X. Wu}, \bibinfo{person}{H. Huang},
  \bibinfo{person}{V.~M. Patel}, \bibinfo{person}{R. He}, {and}
  \bibinfo{person}{Z. Sun}.} \bibinfo{year}{2019}\natexlab{}.
\newblock \showarticletitle{Disentangled Variational Representation for
  Heterogeneous Face Recognition}. In \bibinfo{booktitle}{\emph{Proceedings of
  the AAAI Conference on Artificial Intelligence}}, Vol.~\bibinfo{volume}{33}.
  \bibinfo{pages}{9005--9012}.
\newblock


\bibitem[\protect\citeauthoryear{{Wu}, {Hassner}, {Kim}, {Medioni}, and
  {Natarajan}}{{Wu} et~al\mbox{.}}{2018}]%
        {Wu2018TCNN}
\bibfield{author}{\bibinfo{person}{Y. {Wu}}, \bibinfo{person}{T. {Hassner}},
  \bibinfo{person}{K. {Kim}}, \bibinfo{person}{G. {Medioni}}, {and}
  \bibinfo{person}{P. {Natarajan}}.} \bibinfo{year}{2018}\natexlab{}.
\newblock \showarticletitle{Facial Landmark Detection with Tweaked
  Convolutional Neural Networks}.
\newblock \bibinfo{journal}{\emph{IEEE Trans. Pattern Anal. Mach. Intell.}}
  \bibinfo{volume}{40}, \bibinfo{number}{12} (\bibinfo{year}{2018}),
  \bibinfo{pages}{3067--3074}.
\newblock


\bibitem[\protect\citeauthoryear{Wu, Liu, and Fu}{Wu et~al\mbox{.}}{2017}]%
        {Wu2017LowShotFR}
\bibfield{author}{\bibinfo{person}{Y. Wu}, \bibinfo{person}{H. Liu}, {and}
  \bibinfo{person}{Y. Fu}.} \bibinfo{year}{2017}\natexlab{}.
\newblock \showarticletitle{Low-Shot Face Recognition with Hybrid Classifiers}.
  In \bibinfo{booktitle}{\emph{Proceedings of the IEEE International Conference
  on Computer Vision Workshops}}. \bibinfo{pages}{1933--1939}.
\newblock


\bibitem[\protect\citeauthoryear{Wu, Wu, Gong, Lv, Chen, Liang, Hu, Liu, and
  Yan}{Wu et~al\mbox{.}}{2020}]%
        {Wu_2020_CVPR}
\bibfield{author}{\bibinfo{person}{Y. Wu}, \bibinfo{person}{Y. Wu},
  \bibinfo{person}{R. Gong}, \bibinfo{person}{Y. Lv}, \bibinfo{person}{K.
  Chen}, \bibinfo{person}{D. Liang}, \bibinfo{person}{X. Hu},
  \bibinfo{person}{X. Liu}, {and} \bibinfo{person}{J. Yan}.}
  \bibinfo{year}{2020}\natexlab{}.
\newblock \showarticletitle{Rotation Consistent Margin Loss for Efficient
  Low-Bit Face Recognition}. In \bibinfo{booktitle}{\emph{Proceedings of the
  IEEE/CVF Conference on Computer Vision and Pattern Recognition}}.
  \bibinfo{pages}{6866--6876}.
\newblock


\bibitem[\protect\citeauthoryear{X.Fan, Liu, Huyan, Feng, and Luo}{X.Fan
  et~al\mbox{.}}{2018}]%
        {Fan2018SelfReinforcedCR}
\bibfield{author}{\bibinfo{person}{X.Fan}, \bibinfo{person}{R. Liu},
  \bibinfo{person}{K. Huyan}, \bibinfo{person}{Y. Feng}, {and}
  \bibinfo{person}{Z. Luo}.} \bibinfo{year}{2018}\natexlab{}.
\newblock \showarticletitle{Self-Reinforced Cascaded Regression for Face
  Alignment}. In \bibinfo{booktitle}{\emph{Proceedings of the AAAI Conference
  on Artificial Intelligence}}, Vol.~\bibinfo{volume}{32}.
\newblock


\bibitem[\protect\citeauthoryear{{Xiao}, {Feng}, {Liu}, {Nie}, {Wang}, {Yan},
  and {Kassim}}{{Xiao} et~al\mbox{.}}{2017}]%
        {Xiao2017RDR}
\bibfield{author}{\bibinfo{person}{S. {Xiao}}, \bibinfo{person}{J. {Feng}},
  \bibinfo{person}{L. {Liu}}, \bibinfo{person}{X. {Nie}}, \bibinfo{person}{W.
  {Wang}}, \bibinfo{person}{S. {Yan}}, {and} \bibinfo{person}{A. {Kassim}}.}
  \bibinfo{year}{2017}\natexlab{}.
\newblock \showarticletitle{Recurrent 3D-2D Dual Learning for Large-Pose Facial
  Landmark Detection}. In \bibinfo{booktitle}{\emph{Proceedings of the IEEE
  International Conference on Computer Vision}}. \bibinfo{pages}{1642--1651}.
\newblock


\bibitem[\protect\citeauthoryear{Xiao, Feng, Xing, Lai, Yan, and Kassim}{Xiao
  et~al\mbox{.}}{2016}]%
        {Xiao2016Robust}
\bibfield{author}{\bibinfo{person}{S. Xiao}, \bibinfo{person}{J. Feng},
  \bibinfo{person}{J. Xing}, \bibinfo{person}{H. Lai}, \bibinfo{person}{S.
  Yan}, {and} \bibinfo{person}{A.~A. Kassim}.} \bibinfo{year}{2016}\natexlab{}.
\newblock \showarticletitle{Robust Facial Landmark Detection via Recurrent
  Attentive-Refinement Networks.}. In \bibinfo{booktitle}{\emph{Proceedings of
  the European Conference on Computer Vision}}. \bibinfo{pages}{57--72}.
\newblock


\bibitem[\protect\citeauthoryear{Xie, Shen, and Zisserman}{Xie
  et~al\mbox{.}}{2018}]%
        {Xie2018ComparatorN}
\bibfield{author}{\bibinfo{person}{W. Xie}, \bibinfo{person}{L. Shen}, {and}
  \bibinfo{person}{A. Zisserman}.} \bibinfo{year}{2018}\natexlab{}.
\newblock \showarticletitle{Comparator Networks}. In
  \bibinfo{booktitle}{\emph{Proceedings of the European Conference on Computer
  Vision}}. \bibinfo{pages}{782--797}.
\newblock


\bibitem[\protect\citeauthoryear{Xiong, Zhou, Dou, and Su}{Xiong
  et~al\mbox{.}}{2020}]%
        {Xiong2020GaussianVA}
\bibfield{author}{\bibinfo{person}{Yilin Xiong}, \bibinfo{person}{Zijian Zhou},
  \bibinfo{person}{Yuhao Dou}, {and} \bibinfo{person}{Zhizhong Su}.}
  \bibinfo{year}{2020}\natexlab{}.
\newblock \showarticletitle{Gaussian vector: An efficient solution for facial
  landmark detection}. In \bibinfo{booktitle}{\emph{Proceedings of the Asian
  Conference on Computer Vision}}. \bibinfo{pages}{70--87}.
\newblock


\bibitem[\protect\citeauthoryear{Xu, Liu, and Ye}{Xu et~al\mbox{.}}{2017}]%
        {Xu2017AgeIF}
\bibfield{author}{\bibinfo{person}{C. Xu}, \bibinfo{person}{Q. Liu}, {and}
  \bibinfo{person}{M. Ye}.} \bibinfo{year}{2017}\natexlab{}.
\newblock \showarticletitle{Age invariant face recognition and retrieval by
  coupled auto-encoder networks}.
\newblock \bibinfo{journal}{\emph{Neurocomputing}}  \bibinfo{volume}{222}
  (\bibinfo{year}{2017}), \bibinfo{pages}{62--71}.
\newblock


\bibitem[\protect\citeauthoryear{{Xu} and {Kakadiaris}}{{Xu} and
  {Kakadiaris}}{2017}]%
        {Xu2017JFA}
\bibfield{author}{\bibinfo{person}{X. {Xu}} {and} \bibinfo{person}{I.~A.
  {Kakadiaris}}.} \bibinfo{year}{2017}\natexlab{}.
\newblock \showarticletitle{Joint head pose estimation and face alignment
  framework using global and local cnn features}. In
  \bibinfo{booktitle}{\emph{Proceedings of the IEEE International Conference on
  Automatic Face $\&$ Gesture Recognition}}. \bibinfo{pages}{642--649}.
\newblock


\bibitem[\protect\citeauthoryear{Xu, Meng, Qin, Guo, Zhao, Zhou, and Lei}{Xu
  et~al\mbox{.}}{2021}]%
        {Xu2021SearchingFA}
\bibfield{author}{\bibinfo{person}{Xiaqing Xu}, \bibinfo{person}{Qiang Meng},
  \bibinfo{person}{Yunxiao Qin}, \bibinfo{person}{Jianzhu Guo},
  \bibinfo{person}{Chenxu Zhao}, \bibinfo{person}{Feng Zhou}, {and}
  \bibinfo{person}{Zhen Lei}.} \bibinfo{year}{2021}\natexlab{}.
\newblock \showarticletitle{Searching for Alignment in Face Recognition}. In
  \bibinfo{booktitle}{\emph{Proceedings of the AAAI Conference on Artificial
  Intelligence}}, Vol.~\bibinfo{volume}{35}. \bibinfo{pages}{3065--3073}.
\newblock


\bibitem[\protect\citeauthoryear{Xu, Yan, Sun, Yang, and Luo}{Xu
  et~al\mbox{.}}{2019}]%
        {xu2019centerface}
\bibfield{author}{\bibinfo{person}{Y. Xu}, \bibinfo{person}{W. Yan},
  \bibinfo{person}{H. Sun}, \bibinfo{person}{G. Yang}, {and}
  \bibinfo{person}{J. Luo}.} \bibinfo{year}{2019}\natexlab{}.
\newblock \showarticletitle{CenterFace: Joint Face Detection and Alignment
  Using Face as Point}.
\newblock
\showeprint{1911.03599}


\bibitem[\protect\citeauthoryear{Yan, Zhang, Lei, and Li}{Yan
  et~al\mbox{.}}{2014}]%
        {YAN2014790}
\bibfield{author}{\bibinfo{person}{J. Yan}, \bibinfo{person}{X. Zhang},
  \bibinfo{person}{Z. Lei}, {and} \bibinfo{person}{S.~Z. Li}.}
  \bibinfo{year}{2014}\natexlab{}.
\newblock \showarticletitle{Face detection by structural models}.
\newblock \bibinfo{journal}{\emph{Image and Vision Computing}}
  \bibinfo{volume}{32}, \bibinfo{number}{10} (\bibinfo{year}{2014}),
  \bibinfo{pages}{790 -- 799}.
\newblock


\bibitem[\protect\citeauthoryear{{Yang}, {Yan}, {Lei}, and {Li}}{{Yang}
  et~al\mbox{.}}{2015}]%
        {faceevaluation15}
\bibfield{author}{\bibinfo{person}{B. {Yang}}, \bibinfo{person}{J. {Yan}},
  \bibinfo{person}{Z. {Lei}}, {and} \bibinfo{person}{S.~Z. {Li}}.}
  \bibinfo{year}{2015}\natexlab{}.
\newblock \showarticletitle{Fine-grained evaluation on face detection in the
  wild}. In \bibinfo{booktitle}{\emph{Proceedings of the IEEE International
  Conference on Automatic Face $\&$ Gesture Recognition}},
  Vol.~\bibinfo{volume}{1}. \bibinfo{pages}{1--7}.
\newblock


\bibitem[\protect\citeauthoryear{Yang, Bulat, and Tzimiropoulos}{Yang
  et~al\mbox{.}}{2020a}]%
        {Yang2020FANFaceAS}
\bibfield{author}{\bibinfo{person}{Jing Yang}, \bibinfo{person}{Adrian Bulat},
  {and} \bibinfo{person}{Georgios Tzimiropoulos}.}
  \bibinfo{year}{2020}\natexlab{a}.
\newblock \showarticletitle{Fan-face: a simple orthogonal improvement to deep
  face recognition}. In \bibinfo{booktitle}{\emph{Proceedings of the AAAI
  Conference on Artificial Intelligence}}, Vol.~\bibinfo{volume}{34}.
  \bibinfo{pages}{12621--12628}.
\newblock


\bibitem[\protect\citeauthoryear{Yang, Ren, Zhang, Chen, Wen, Li, and Hua}{Yang
  et~al\mbox{.}}{2017}]%
        {Yang2017NeuralAN}
\bibfield{author}{\bibinfo{person}{J. Yang}, \bibinfo{person}{P. Ren},
  \bibinfo{person}{D. Zhang}, \bibinfo{person}{D. Chen}, \bibinfo{person}{F.
  Wen}, \bibinfo{person}{H. Li}, {and} \bibinfo{person}{G. Hua}.}
  \bibinfo{year}{2017}\natexlab{}.
\newblock \showarticletitle{Neural Aggregation Network for Video Face
  Recognition}. In \bibinfo{booktitle}{\emph{Proceedings of the IEEE/CVF
  Conference on Computer Vision and Pattern Recognition}}.
  \bibinfo{pages}{5216--5225}.
\newblock


\bibitem[\protect\citeauthoryear{Yang, Chen, Zhan, Zhao, Loy, and Lin}{Yang
  et~al\mbox{.}}{2020b}]%
        {Yang2020LearningTC}
\bibfield{author}{\bibinfo{person}{L. Yang}, \bibinfo{person}{D. Chen},
  \bibinfo{person}{X. Zhan}, \bibinfo{person}{R. Zhao}, \bibinfo{person}{C.~C.
  Loy}, {and} \bibinfo{person}{D. Lin}.} \bibinfo{year}{2020}\natexlab{b}.
\newblock \showarticletitle{Learning to Cluster Faces via Confidence and
  Connectivity Estimation}. In \bibinfo{booktitle}{\emph{Proceedings of the
  IEEE/CVF Conference on Computer Vision and Pattern Recognition}}.
  \bibinfo{pages}{13369--13378}.
\newblock


\bibitem[\protect\citeauthoryear{Yang, Zhan, Chen, Yan, Loy, and Lin}{Yang
  et~al\mbox{.}}{2019}]%
        {yang2019learning}
\bibfield{author}{\bibinfo{person}{L. Yang}, \bibinfo{person}{X. Zhan},
  \bibinfo{person}{D. Chen}, \bibinfo{person}{J. Yan}, \bibinfo{person}{C.~C.
  Loy}, {and} \bibinfo{person}{D. Lin}.} \bibinfo{year}{2019}\natexlab{}.
\newblock \showarticletitle{Learning to cluster faces on an affinity graph}. In
  \bibinfo{booktitle}{\emph{Proceedings of the IEEE/CVF Conference on Computer
  Vision and Pattern Recognition}}. \bibinfo{pages}{2298--2306}.
\newblock


\bibitem[\protect\citeauthoryear{Yang, Kriegman, and Ahuja}{Yang
  et~al\mbox{.}}{2002}]%
        {Yang2002DetectingFI}
\bibfield{author}{\bibinfo{person}{Ming-Hsuan Yang}, \bibinfo{person}{D.
  Kriegman}, {and} \bibinfo{person}{N. Ahuja}.}
  \bibinfo{year}{2002}\natexlab{}.
\newblock \showarticletitle{Detecting Faces in Images: A Survey}.
\newblock \bibinfo{journal}{\emph{IEEE Trans. Pattern Anal. Mach. Intell.}}
  \bibinfo{volume}{24} (\bibinfo{year}{2002}), \bibinfo{pages}{34--58}.
\newblock


\bibitem[\protect\citeauthoryear{Yang, Luo, Loy, and Tang}{Yang
  et~al\mbox{.}}{2015}]%
        {2015Faceness}
\bibfield{author}{\bibinfo{person}{S. Yang}, \bibinfo{person}{P. Luo},
  \bibinfo{person}{C.~C. Loy}, {and} \bibinfo{person}{X. Tang}.}
  \bibinfo{year}{2015}\natexlab{}.
\newblock \showarticletitle{From facial parts responses to face detection: a
  deep learning approach}. In \bibinfo{booktitle}{\emph{Proceedings of the IEEE
  International Conference on Computer Vision}}. \bibinfo{pages}{3676--3684}.
\newblock


\bibitem[\protect\citeauthoryear{Yang, Luo, Loy, and Tang}{Yang
  et~al\mbox{.}}{2016}]%
        {Yang_2016_CVPR}
\bibfield{author}{\bibinfo{person}{S. Yang}, \bibinfo{person}{P. Luo},
  \bibinfo{person}{C.~C. Loy}, {and} \bibinfo{person}{X. Tang}.}
  \bibinfo{year}{2016}\natexlab{}.
\newblock \showarticletitle{WIDER FACE: A Face Detection Benchmark}. In
  \bibinfo{booktitle}{\emph{Proceedings of the IEEE/CVF Conference on Computer
  Vision and Pattern Recognition}}. \bibinfo{pages}{5525--5533}.
\newblock


\bibitem[\protect\citeauthoryear{Yi, Lei, Liao, and Li}{Yi
  et~al\mbox{.}}{2014}]%
        {yi2014learning}
\bibfield{author}{\bibinfo{person}{D. Yi}, \bibinfo{person}{Z. Lei},
  \bibinfo{person}{S. Liao}, {and} \bibinfo{person}{S.~Z. Li}.}
  \bibinfo{year}{2014}\natexlab{}.
\newblock \showarticletitle{Learning Face Representation from Scratch}.
\newblock  (\bibinfo{year}{2014}).
\newblock
\showeprint{1411.7923}


\bibitem[\protect\citeauthoryear{Yi, Zhang, Tan, and Gong}{Yi
  et~al\mbox{.}}{2017}]%
        {Yi2017DualGANUD}
\bibfield{author}{\bibinfo{person}{Z. Yi}, \bibinfo{person}{H. Zhang},
  \bibinfo{person}{P. Tan}, {and} \bibinfo{person}{M. Gong}.}
  \bibinfo{year}{2017}\natexlab{}.
\newblock \showarticletitle{DualGAN: Unsupervised Dual Learning for
  Image-to-Image Translation}. In \bibinfo{booktitle}{\emph{Proceedings of the
  IEEE International Conference on Computer Vision}}.
  \bibinfo{pages}{2868--2876}.
\newblock


\bibitem[\protect\citeauthoryear{Yin, Tran, Li, Shen, and Liu}{Yin
  et~al\mbox{.}}{2019a}]%
        {yin2019towards}
\bibfield{author}{\bibinfo{person}{Bangjie Yin}, \bibinfo{person}{Luan Tran},
  \bibinfo{person}{Haoxiang Li}, \bibinfo{person}{Xiaohui Shen}, {and}
  \bibinfo{person}{Xiaoming Liu}.} \bibinfo{year}{2019}\natexlab{a}.
\newblock \showarticletitle{Towards interpretable face recognition}. In
  \bibinfo{booktitle}{\emph{Proceedings of the IEEE International Conference on
  Computer Vision}}. \bibinfo{pages}{9348--9357}.
\newblock


\bibitem[\protect\citeauthoryear{Yin, Yu, Sohn, Liu, and Chandraker}{Yin
  et~al\mbox{.}}{2019b}]%
        {yin2019feature}
\bibfield{author}{\bibinfo{person}{X. Yin}, \bibinfo{person}{X. Yu},
  \bibinfo{person}{K. Sohn}, \bibinfo{person}{X. Liu}, {and}
  \bibinfo{person}{M. Chandraker}.} \bibinfo{year}{2019}\natexlab{b}.
\newblock \showarticletitle{Feature Transfer Learning for Face Recognition with
  Under-Represented Data}. In \bibinfo{booktitle}{\emph{Proceedings of the
  IEEE/CVF Conference on Computer Vision and Pattern Recognition}}.
  \bibinfo{pages}{5704--5713}.
\newblock


\bibitem[\protect\citeauthoryear{Yu, Fan, Chen, Yan, Lu, Liu, and Xie}{Yu
  et~al\mbox{.}}{2019a}]%
        {Yu2019UnknownIR}
\bibfield{author}{\bibinfo{person}{H. Yu}, \bibinfo{person}{Y. Fan},
  \bibinfo{person}{K. Chen}, \bibinfo{person}{H. Yan}, \bibinfo{person}{X. Lu},
  \bibinfo{person}{J. Liu}, {and} \bibinfo{person}{D. Xie}.}
  \bibinfo{year}{2019}\natexlab{a}.
\newblock \showarticletitle{Unknown Identity Rejection Loss: Utilizing
  Unlabeled Data for Face Recognition}. In
  \bibinfo{booktitle}{\emph{Proceedings of the IEEE International Conference on
  Computer Vision Workshops}}. \bibinfo{pages}{2662--2669}.
\newblock


\bibitem[\protect\citeauthoryear{Yu, Jiang, Wang, Cao, and Huang}{Yu
  et~al\mbox{.}}{2016}]%
        {UnitBox}
\bibfield{author}{\bibinfo{person}{J. Yu}, \bibinfo{person}{Y. Jiang},
  \bibinfo{person}{Z. Wang}, \bibinfo{person}{Z. Cao}, {and}
  \bibinfo{person}{T. Huang}.} \bibinfo{year}{2016}\natexlab{}.
\newblock \showarticletitle{Unitbox: An advanced object detection network}. In
  \bibinfo{booktitle}{\emph{Proceedings of the 24th ACM international
  conference on Multimedia}}. \bibinfo{pages}{516--520}.
\newblock


\bibitem[\protect\citeauthoryear{Yu, Song, Zhang, Liu, Zhou, and Yan}{Yu
  et~al\mbox{.}}{2019b}]%
        {liu2019towards}
\bibfield{author}{\bibinfo{person}{Y. Yu}, \bibinfo{person}{G. Song},
  \bibinfo{person}{M. Zhang}, \bibinfo{person}{J. Liu}, \bibinfo{person}{Y.
  Zhou}, {and} \bibinfo{person}{J. Yan}.} \bibinfo{year}{2019}\natexlab{b}.
\newblock \showarticletitle{Towards flops-constrained face recognition}. In
  \bibinfo{booktitle}{\emph{Proceedings of the IEEE International Conference on
  Computer Vision Workshops}}. \bibinfo{pages}{2698--2702}.
\newblock


\bibitem[\protect\citeauthoryear{Yue, Miao, Wang, Zhang, Zhen, and Cao}{Yue
  et~al\mbox{.}}{2018}]%
        {Yue2018AttentionalAN}
\bibfield{author}{\bibinfo{person}{L. Yue}, \bibinfo{person}{X. Miao},
  \bibinfo{person}{P. Wang}, \bibinfo{person}{B. Zhang}, \bibinfo{person}{X.
  Zhen}, {and} \bibinfo{person}{X. Cao}.} \bibinfo{year}{2018}\natexlab{}.
\newblock \showarticletitle{Attentional Alignment Networks}. In
  \bibinfo{booktitle}{\emph{Proceedings of the British Machine Vision
  Conference}}, Vol.~\bibinfo{volume}{2}. \bibinfo{pages}{6--13}.
\newblock


\bibitem[\protect\citeauthoryear{z.~Liu, p.~Luo, x.~Wang, and Tang}{z.~Liu
  et~al\mbox{.}}{2015}]%
        {liu2015faceattributes}
\bibfield{author}{\bibinfo{person}{z. Liu}, \bibinfo{person}{p. Luo},
  \bibinfo{person}{x. Wang}, {and} \bibinfo{person}{X. Tang}.}
  \bibinfo{year}{2015}\natexlab{}.
\newblock \showarticletitle{Deep learning face attributes in the wild}. In
  \bibinfo{booktitle}{\emph{Proceedings of the IEEE International Conference on
  Computer Vision}}. \bibinfo{pages}{3730--3738}.
\newblock


\bibitem[\protect\citeauthoryear{Zafeiriou, Trigeorgis, Chrysos, Deng, and
  Shen}{Zafeiriou et~al\mbox{.}}{2017}]%
        {Zafeiriou2017TheMF}
\bibfield{author}{\bibinfo{person}{S. Zafeiriou}, \bibinfo{person}{G.
  Trigeorgis}, \bibinfo{person}{G. Chrysos}, \bibinfo{person}{J. Deng}, {and}
  \bibinfo{person}{Ji. Shen}.} \bibinfo{year}{2017}\natexlab{}.
\newblock \showarticletitle{The Menpo Facial Landmark Localisation Challenge: A
  Step Towards the Solution}. In \bibinfo{booktitle}{\emph{Proceedings of the
  IEEE/CVF Conference on Computer Vision and Pattern RecognitionW}}.
  \bibinfo{pages}{2116--2125}.
\newblock


\bibitem[\protect\citeauthoryear{Zafeiriou, Zhang, and Zhang}{Zafeiriou
  et~al\mbox{.}}{2015}]%
        {Stefanos2015A}
\bibfield{author}{\bibinfo{person}{S. Zafeiriou}, \bibinfo{person}{C. Zhang},
  {and} \bibinfo{person}{Z. Zhang}.} \bibinfo{year}{2015}\natexlab{}.
\newblock \showarticletitle{A survey on face detection in the wild: Past,
  present and future}.
\newblock \bibinfo{journal}{\emph{Computer Vision and Image Understanding}}
  \bibinfo{volume}{138} (\bibinfo{year}{2015}), \bibinfo{pages}{1--24}.
\newblock


\bibitem[\protect\citeauthoryear{Zee, Gali, and Nwogu}{Zee
  et~al\mbox{.}}{2019}]%
        {zee2019enhancing}
\bibfield{author}{\bibinfo{person}{Timothy Zee}, \bibinfo{person}{Geeta Gali},
  {and} \bibinfo{person}{Ifeoma Nwogu}.} \bibinfo{year}{2019}\natexlab{}.
\newblock \showarticletitle{Enhancing human face recognition with an
  interpretable neural network}. In \bibinfo{booktitle}{\emph{Proceedings of
  the IEEE International Conference on Computer Vision Workshops}}.
  \bibinfo{pages}{514--522}.
\newblock


\bibitem[\protect\citeauthoryear{Zeng, Liu, Zhao, Ge, Shen, and Zhang}{Zeng
  et~al\mbox{.}}{2019}]%
        {ZENG2019PPN}
\bibfield{author}{\bibinfo{person}{D. Zeng}, \bibinfo{person}{H. Liu},
  \bibinfo{person}{F. Zhao}, \bibinfo{person}{S. Ge}, \bibinfo{person}{W.
  Shen}, {and} \bibinfo{person}{Z. Zhang}.} \bibinfo{year}{2019}\natexlab{}.
\newblock \showarticletitle{Proposal pyramid networks for fast face detection}.
\newblock \bibinfo{journal}{\emph{Information Sciences}}  \bibinfo{volume}{495}
  (\bibinfo{year}{2019}), \bibinfo{pages}{136 -- 149}.
\newblock


\bibitem[\protect\citeauthoryear{Zhan, Liu, Yan, Lin, and Loy}{Zhan
  et~al\mbox{.}}{2018}]%
        {Zhan2018ConsensusDrivenPI}
\bibfield{author}{\bibinfo{person}{X. Zhan}, \bibinfo{person}{Z. Liu},
  \bibinfo{person}{J. Yan}, \bibinfo{person}{D. Lin}, {and}
  \bibinfo{person}{C.~C. Loy}.} \bibinfo{year}{2018}\natexlab{}.
\newblock \showarticletitle{Consensus-Driven Propagation in Massive Unlabeled
  Data for Face Recognition}. In \bibinfo{booktitle}{\emph{Proceedings of the
  European Conference on Computer Vision}}. \bibinfo{pages}{568--583}.
\newblock


\bibitem[\protect\citeauthoryear{Zhang, Li, Wang, Tai, Wang, Li, Huang, Xia,
  Pei, and Ji}{Zhang et~al\mbox{.}}{2020d}]%
        {Zhang2020ASFDAA}
\bibfield{author}{\bibinfo{person}{B. Zhang}, \bibinfo{person}{J. Li},
  \bibinfo{person}{Y. Wang}, \bibinfo{person}{Y. Tai}, \bibinfo{person}{C.
  Wang}, \bibinfo{person}{J. Li}, \bibinfo{person}{F. Huang},
  \bibinfo{person}{Y. Xia}, \bibinfo{person}{W. Pei}, {and} \bibinfo{person}{R.
  Ji}.} \bibinfo{year}{2020}\natexlab{d}.
\newblock \showarticletitle{ASFD: Automatic and Scalable Face Detector}.
\newblock
\showeprint{2003.11228}


\bibitem[\protect\citeauthoryear{Zhang, Xu, and Tu}{Zhang
  et~al\mbox{.}}{2018c}]%
        {zhang2018face}
\bibfield{author}{\bibinfo{person}{C. Zhang}, \bibinfo{person}{X. Xu}, {and}
  \bibinfo{person}{D. Tu}.} \bibinfo{year}{2018}\natexlab{c}.
\newblock \showarticletitle{Face detection using improved faster rcnn}.
\newblock  (\bibinfo{year}{2018}).
\newblock
\showeprint{1802.02142}


\bibitem[\protect\citeauthoryear{{Zhang} and {Zhang}}{{Zhang} and
  {Zhang}}{2014}]%
        {zhang2014mt}
\bibfield{author}{\bibinfo{person}{C. {Zhang}} {and} \bibinfo{person}{Z.
  {Zhang}}.} \bibinfo{year}{2014}\natexlab{}.
\newblock \showarticletitle{Improving multiview face detection with multi-task
  deep convolutional neural networks}. In \bibinfo{booktitle}{\emph{IEEE Winter
  Conference on Applications of Computer Vision}}. \bibinfo{pages}{1036--1041}.
\newblock


\bibitem[\protect\citeauthoryear{Zhang, Lin, Chen, Wu, Tan, and
  Izquierdo}{Zhang et~al\mbox{.}}{2017b}]%
        {Zhang2017ContentAdaptiveSP}
\bibfield{author}{\bibinfo{person}{D. Zhang}, \bibinfo{person}{L. Lin},
  \bibinfo{person}{T. Chen}, \bibinfo{person}{X. Wu}, \bibinfo{person}{W. Tan},
  {and} \bibinfo{person}{E. Izquierdo}.} \bibinfo{year}{2017}\natexlab{b}.
\newblock \showarticletitle{Content-Adaptive Sketch Portrait Generation by
  Decompositional Representation Learning}.
\newblock \bibinfo{journal}{\emph{Trans. Image Process.}}  \bibinfo{volume}{26}
  (\bibinfo{year}{2017}), \bibinfo{pages}{328--339}.
\newblock


\bibitem[\protect\citeauthoryear{Zhang}{Zhang}{2006}]%
        {2006fr}
\bibfield{author}{\bibinfo{person}{F. Zhang}.} \bibinfo{year}{2006}\natexlab{}.
\newblock \showarticletitle{Face recognition from a single image per person: A
  survey}.
\newblock \bibinfo{journal}{\emph{Pattern Recognition}} \bibinfo{volume}{39},
  \bibinfo{number}{9} (\bibinfo{year}{2006}), \bibinfo{pages}{1725--1745}.
\newblock


\bibitem[\protect\citeauthoryear{Zhang, Fan, Ai, Song, Qin, and Wu}{Zhang
  et~al\mbox{.}}{2019a}]%
        {zhang2019accurate}
\bibfield{author}{\bibinfo{person}{F. Zhang}, \bibinfo{person}{X. Fan},
  \bibinfo{person}{G. Ai}, \bibinfo{person}{J. Song}, \bibinfo{person}{Y. Qin},
  {and} \bibinfo{person}{J. Wu}.} \bibinfo{year}{2019}\natexlab{a}.
\newblock \showarticletitle{Accurate Face Detection for High Performance}.
\newblock  (\bibinfo{year}{2019}).
\newblock
\showeprint{1905.01585}


\bibitem[\protect\citeauthoryear{Zhang, Han, Shan, Song, and Chen}{Zhang
  et~al\mbox{.}}{2018a}]%
        {Zhang2018FaceAA}
\bibfield{author}{\bibinfo{person}{G. Zhang}, \bibinfo{person}{H. Han},
  \bibinfo{person}{S. Shan}, \bibinfo{person}{X. Song}, {and}
  \bibinfo{person}{X. Chen}.} \bibinfo{year}{2018}\natexlab{a}.
\newblock \showarticletitle{Face Alignment across Large Pose via MT-CNN Based
  3D Shape Reconstruction}. In \bibinfo{booktitle}{\emph{Proceedings of the
  IEEE International Conference on Automatic Face $\&$ Gesture Recognition}}.
  \bibinfo{pages}{210--217}.
\newblock


\bibitem[\protect\citeauthoryear{Zhang, Riggan, Hu, Short, and Patel}{Zhang
  et~al\mbox{.}}{2019b}]%
        {Zhang2019SynthesisOH}
\bibfield{author}{\bibinfo{person}{H. Zhang}, \bibinfo{person}{B.~S. Riggan},
  \bibinfo{person}{S. Hu}, \bibinfo{person}{N.~J. Short}, {and}
  \bibinfo{person}{V.~M. Patel}.} \bibinfo{year}{2019}\natexlab{b}.
\newblock \showarticletitle{Synthesis of High-Quality Visible Faces from
  Polarimetric Thermal Faces using Generative Adversarial Networks}.
\newblock \bibinfo{journal}{\emph{International Journal of Computer Vision}}
  \bibinfo{volume}{127} (\bibinfo{year}{2019}), \bibinfo{pages}{845--862}.
\newblock


\bibitem[\protect\citeauthoryear{Zhang and Hu}{Zhang and Hu}{2019}]%
        {Zhang2019StackedHN}
\bibfield{author}{\bibinfo{person}{J. Zhang} {and} \bibinfo{person}{H. Hu}.}
  \bibinfo{year}{2019}\natexlab{}.
\newblock \showarticletitle{Stacked Hourglass Network Joint with Salient Region
  Attention Refinement for Face Alignment}. In
  \bibinfo{booktitle}{\emph{Proceedings of the IEEE International Conference on
  Automatic Face $\&$ Gesture Recognition}}. \bibinfo{pages}{1--7}.
\newblock


\bibitem[\protect\citeauthoryear{Zhang, Shan, Kan, and Chen}{Zhang
  et~al\mbox{.}}{2014b}]%
        {Jie2014Coarse}
\bibfield{author}{\bibinfo{person}{J. Zhang}, \bibinfo{person}{S. Shan},
  \bibinfo{person}{M. Kan}, {and} \bibinfo{person}{X. Chen}.}
  \bibinfo{year}{2014}\natexlab{b}.
\newblock \showarticletitle{Coarse-to-fine auto-encoder networks (cfan) for
  real-time face alignment}. In \bibinfo{booktitle}{\emph{Proceedings of the
  European Conference on Computer Vision}}. \bibinfo{pages}{1--16}.
\newblock


\bibitem[\protect\citeauthoryear{Zhang, Wu, Hoi, and Zhu}{Zhang
  et~al\mbox{.}}{2020e}]%
        {zhang2017fanet}
\bibfield{author}{\bibinfo{person}{J. Zhang}, \bibinfo{person}{X. Wu},
  \bibinfo{person}{S.~C. Hoi}, {and} \bibinfo{person}{J. Zhu}.}
  \bibinfo{year}{2020}\natexlab{e}.
\newblock \showarticletitle{Feature agglomeration networks for single stage
  face detection}.
\newblock \bibinfo{journal}{\emph{Neurocomputing}}  \bibinfo{volume}{380}
  (\bibinfo{year}{2020}), \bibinfo{pages}{180--189}.
\newblock


\bibitem[\protect\citeauthoryear{{Zhang}, {Zhang}, {Li}, and {Qiao}}{{Zhang}
  et~al\mbox{.}}{2016}]%
        {mtcnn}
\bibfield{author}{\bibinfo{person}{K. {Zhang}}, \bibinfo{person}{Z. {Zhang}},
  \bibinfo{person}{Z. {Li}}, {and} \bibinfo{person}{Y. {Qiao}}.}
  \bibinfo{year}{2016}\natexlab{}.
\newblock \showarticletitle{Joint face detection and alignment using multitask
  cascaded convolutional networks}.
\newblock \bibinfo{journal}{\emph{IEEE Signal Processing Letters}}
  \bibinfo{volume}{23}, \bibinfo{number}{10} (\bibinfo{year}{2016}),
  \bibinfo{pages}{1499--1503}.
\newblock


\bibitem[\protect\citeauthoryear{Zhang, Lin, Wu, Ding, and Zhang}{Zhang
  et~al\mbox{.}}{2015}]%
        {Zhang2015EndtoEndPG}
\bibfield{author}{\bibinfo{person}{L. Zhang}, \bibinfo{person}{L. Lin},
  \bibinfo{person}{X. Wu}, \bibinfo{person}{S. Ding}, {and} \bibinfo{person}{L.
  Zhang}.} \bibinfo{year}{2015}\natexlab{}.
\newblock \showarticletitle{End-to-End Photo-Sketch Generation via Fully
  Convolutional Representation Learning}. In
  \bibinfo{booktitle}{\emph{Proceedings of the ACM on International Conference
  on Multimedia Retrieval}}. \bibinfo{pages}{627–634}.
\newblock


\bibitem[\protect\citeauthoryear{Zhang, Li, Wang, Chi, and Gao}{Zhang
  et~al\mbox{.}}{2020c}]%
        {Zhang2020CascadedFS}
\bibfield{author}{\bibinfo{person}{M. Zhang}, \bibinfo{person}{Y. Li},
  \bibinfo{person}{N. Wang}, \bibinfo{person}{Y. Chi}, {and}
  \bibinfo{person}{X. Gao}.} \bibinfo{year}{2020}\natexlab{c}.
\newblock \showarticletitle{Cascaded Face Sketch Synthesis Under Various
  Illuminations}.
\newblock \bibinfo{journal}{\emph{IEEE Transactions on Image Processing}}
  \bibinfo{volume}{29} (\bibinfo{year}{2020}), \bibinfo{pages}{1507--1521}.
\newblock


\bibitem[\protect\citeauthoryear{Zhang, Wang, Gao, Li, and Tao}{Zhang
  et~al\mbox{.}}{2019c}]%
        {Zhang2019DualTransferFS}
\bibfield{author}{\bibinfo{person}{M. Zhang}, \bibinfo{person}{R. Wang},
  \bibinfo{person}{X. Gao}, \bibinfo{person}{J. Li}, {and} \bibinfo{person}{D.
  Tao}.} \bibinfo{year}{2019}\natexlab{c}.
\newblock \showarticletitle{Dual-Transfer Face Sketch–Photo Synthesis}.
\newblock \bibinfo{journal}{\emph{Trans. Image Process.}}  \bibinfo{volume}{28}
  (\bibinfo{year}{2019}), \bibinfo{pages}{642--657}.
\newblock


\bibitem[\protect\citeauthoryear{Zhang, Chi, Lei, and Li}{Zhang
  et~al\mbox{.}}{2020a}]%
        {zhang2019refineface}
\bibfield{author}{\bibinfo{person}{S. Zhang}, \bibinfo{person}{C. Chi},
  \bibinfo{person}{Z. Lei}, {and} \bibinfo{person}{S.~Z. Li}.}
  \bibinfo{year}{2020}\natexlab{a}.
\newblock \showarticletitle{Refineface: Refinement neural network for high
  performance face detection}.
\newblock \bibinfo{journal}{\emph{IEEE Trans. Pattern Anal. Mach. Intell.}}
  (\bibinfo{year}{2020}).
\newblock


\bibitem[\protect\citeauthoryear{Zhang, Wen, Bian, Lei, and Li}{Zhang
  et~al\mbox{.}}{2018b}]%
        {Zhang_2018_refinedet}
\bibfield{author}{\bibinfo{person}{S. Zhang}, \bibinfo{person}{L. Wen},
  \bibinfo{person}{X. Bian}, \bibinfo{person}{Z. Lei}, {and}
  \bibinfo{person}{S.~Z. Li}.} \bibinfo{year}{2018}\natexlab{b}.
\newblock \showarticletitle{Single-Shot Refinement Neural Network for Object
  Detection}. In \bibinfo{booktitle}{\emph{Proceedings of the IEEE/CVF
  Conference on Computer Vision and Pattern Recognition}}.
  \bibinfo{pages}{4203--4212}.
\newblock


\bibitem[\protect\citeauthoryear{Zhang, Zhu, Wang, Shi, Fu, Wang, Mei, and
  Li}{Zhang et~al\mbox{.}}{2019g}]%
        {zhang2019improved}
\bibfield{author}{\bibinfo{person}{S. Zhang}, \bibinfo{person}{R. Zhu},
  \bibinfo{person}{X. Wang}, \bibinfo{person}{H. Shi}, \bibinfo{person}{F. Fu},
  \bibinfo{person}{S. Wang}, \bibinfo{person}{T. Mei}, {and}
  \bibinfo{person}{S.~Z. Li}.} \bibinfo{year}{2019}\natexlab{g}.
\newblock \showarticletitle{Improved Selective Refinement Network for Face
  Detection}.
\newblock  (\bibinfo{year}{2019}).
\newblock
\showeprint{1901.06651}


\bibitem[\protect\citeauthoryear{{Zhang}, {Zhu}, {Lei}, {Shi}, {Wang}, and
  {Li}}{{Zhang} et~al\mbox{.}}{2017a}]%
        {Zhang2017Faceboxes}
\bibfield{author}{\bibinfo{person}{S. {Zhang}}, \bibinfo{person}{X. {Zhu}},
  \bibinfo{person}{Z. {Lei}}, \bibinfo{person}{H. {Shi}}, \bibinfo{person}{X.
  {Wang}}, {and} \bibinfo{person}{S.~Z. {Li}}.}
  \bibinfo{year}{2017}\natexlab{a}.
\newblock \showarticletitle{FaceBoxes: a CPU real-time face detector with high
  accuracy}. In \bibinfo{booktitle}{\emph{Proceedings of the IEEE International
  Joint Conference on Biometrics}}. \bibinfo{pages}{1--9}.
\newblock


\bibitem[\protect\citeauthoryear{{Zhang}, {Zhu}, {Lei}, {Shi}, {Wang}, and
  {Li}}{{Zhang} et~al\mbox{.}}{2017b}]%
        {Zhang2017S3FD}
\bibfield{author}{\bibinfo{person}{S. {Zhang}}, \bibinfo{person}{X. {Zhu}},
  \bibinfo{person}{Z. {Lei}}, \bibinfo{person}{H. {Shi}}, \bibinfo{person}{X.
  {Wang}}, {and} \bibinfo{person}{S.~Z. {Li}}.}
  \bibinfo{year}{2017}\natexlab{b}.
\newblock \showarticletitle{S$^3$FD: single shot scale-Invariant face
  detector}. In \bibinfo{booktitle}{\emph{Proceedings of the IEEE International
  Conference on Computer Vision}}. \bibinfo{pages}{192--201}.
\newblock


\bibitem[\protect\citeauthoryear{Zhang, Zhu, Lei, Wang, and Li}{Zhang
  et~al\mbox{.}}{2018e}]%
        {Zhang2018DCFPN}
\bibfield{author}{\bibinfo{person}{S. Zhang}, \bibinfo{person}{X. Zhu},
  \bibinfo{person}{Z. Lei}, \bibinfo{person}{X. Wang}, {and}
  \bibinfo{person}{Stan~Z Li}.} \bibinfo{year}{2018}\natexlab{e}.
\newblock \showarticletitle{Detecting face with densely connected face proposal
  network}.
\newblock \bibinfo{journal}{\emph{Neurocomputing}}  \bibinfo{volume}{284}
  (\bibinfo{year}{2018}), \bibinfo{pages}{119--127}.
\newblock


\bibitem[\protect\citeauthoryear{Zhang, Wang, and Tang}{Zhang
  et~al\mbox{.}}{2011}]%
        {Zhang2011CoupledIE}
\bibfield{author}{\bibinfo{person}{W. Zhang}, \bibinfo{person}{X. Wang}, {and}
  \bibinfo{person}{X. Tang}.} \bibinfo{year}{2011}\natexlab{}.
\newblock \showarticletitle{Coupled information-theoretic encoding for face
  photo-sketch recognition}. In \bibinfo{booktitle}{\emph{Proceedings of the
  IEEE/CVF Conference on Computer Vision and Pattern Recognition}}.
  \bibinfo{pages}{513--520}.
\newblock


\bibitem[\protect\citeauthoryear{Zhang, Fang, Wen, Li, and Qiao}{Zhang
  et~al\mbox{.}}{2017a}]%
        {Zhang2017RangeLF}
\bibfield{author}{\bibinfo{person}{X. Zhang}, \bibinfo{person}{Z. Fang},
  \bibinfo{person}{Y. Wen}, \bibinfo{person}{Z. Li}, {and} \bibinfo{person}{Y.
  Qiao}.} \bibinfo{year}{2017}\natexlab{a}.
\newblock \showarticletitle{Range Loss for Deep Face Recognition with
  Long-Tailed Training Data}. In \bibinfo{booktitle}{\emph{Proceedings of the
  IEEE International Conference on Computer Vision}}.
  \bibinfo{pages}{5419--5428}.
\newblock


\bibitem[\protect\citeauthoryear{Zhang, Zhao, Qiao, Wang, and Li}{Zhang
  et~al\mbox{.}}{2019e}]%
        {zhang2019adacos}
\bibfield{author}{\bibinfo{person}{X. Zhang}, \bibinfo{person}{R. Zhao},
  \bibinfo{person}{Y. Qiao}, \bibinfo{person}{X. Wang}, {and}
  \bibinfo{person}{H. Li}.} \bibinfo{year}{2019}\natexlab{e}.
\newblock \showarticletitle{AdaCos: Adaptively Scaling Cosine Logits for
  Effectively Learning Deep Face Representations}. In
  \bibinfo{booktitle}{\emph{Proceedings of the IEEE/CVF Conference on Computer
  Vision and Pattern Recognition}}. \bibinfo{pages}{10823--10832}.
\newblock


\bibitem[\protect\citeauthoryear{Zhang, Zhao, Yan, Gao, Qiao, Wang, and
  Li}{Zhang et~al\mbox{.}}{2019f}]%
        {Zhang2019P2SGradRG}
\bibfield{author}{\bibinfo{person}{X. Zhang}, \bibinfo{person}{R. Zhao},
  \bibinfo{person}{J. Yan}, \bibinfo{person}{M. Gao}, \bibinfo{person}{Y.
  Qiao}, \bibinfo{person}{X. Wang}, {and} \bibinfo{person}{H. Li}.}
  \bibinfo{year}{2019}\natexlab{f}.
\newblock \showarticletitle{P2SGrad: Refined Gradients for Optimizing Deep Face
  Models}. In \bibinfo{booktitle}{\emph{Proceedings of the IEEE/CVF Conference
  on Computer Vision and Pattern Recognition}}. \bibinfo{pages}{9898--9906}.
\newblock


\bibitem[\protect\citeauthoryear{Zhang, Zhou, Lin, and Sun}{Zhang
  et~al\mbox{.}}{2018d}]%
        {Zhang2018ShuffleNetAE}
\bibfield{author}{\bibinfo{person}{X. Zhang}, \bibinfo{person}{X. Zhou},
  \bibinfo{person}{M. Lin}, {and} \bibinfo{person}{J. Sun}.}
  \bibinfo{year}{2018}\natexlab{d}.
\newblock \showarticletitle{ShuffleNet: An Extremely Efficient Convolutional
  Neural Network for Mobile Devices}. In \bibinfo{booktitle}{\emph{Proceedings
  of the IEEE/CVF Conference on Computer Vision and Pattern Recognition}}.
  \bibinfo{pages}{6848--6856}.
\newblock


\bibitem[\protect\citeauthoryear{Zhang, Deng, Wang, Hu, Li, Zhao, and
  Wen}{Zhang et~al\mbox{.}}{2020b}]%
        {Zhang_2020_FaceGraph}
\bibfield{author}{\bibinfo{person}{Y. Zhang}, \bibinfo{person}{W. Deng},
  \bibinfo{person}{M. Wang}, \bibinfo{person}{J. Hu}, \bibinfo{person}{X. Li},
  \bibinfo{person}{D. Zhao}, {and} \bibinfo{person}{D. Wen}.}
  \bibinfo{year}{2020}\natexlab{b}.
\newblock \showarticletitle{Global-Local GCN: Large-Scale Label Noise Cleansing
  for Face Recognition}. In \bibinfo{booktitle}{\emph{Proceedings of the
  IEEE/CVF Conference on Computer Vision and Pattern Recognition}}.
  \bibinfo{pages}{7728--7737}.
\newblock


\bibitem[\protect\citeauthoryear{Zhang, Xu, and Liu}{Zhang
  et~al\mbox{.}}{2019d}]%
        {zhang2019robust}
\bibfield{author}{\bibinfo{person}{Y. Zhang}, \bibinfo{person}{X. Xu}, {and}
  \bibinfo{person}{X. Liu}.} \bibinfo{year}{2019}\natexlab{d}.
\newblock \showarticletitle{Robust and High Performance Face Detector}.
\newblock
\showeprint{1901.02350}


\bibitem[\protect\citeauthoryear{Zhang, Ping, Chen, and Tang}{Zhang
  et~al\mbox{.}}{2014a}]%
        {Zhang2014Facial}
\bibfield{author}{\bibinfo{person}{Z. Zhang}, \bibinfo{person}{L. Ping},
  \bibinfo{person}{C.~L. Chen}, {and} \bibinfo{person}{X. Tang}.}
  \bibinfo{year}{2014}\natexlab{a}.
\newblock \showarticletitle{Facial Landmark Detection by Deep Multi-task
  Learning}. In \bibinfo{booktitle}{\emph{Proceedings of the European
  Conference on Computer Vision}}. \bibinfo{pages}{94--108}.
\newblock


\bibitem[\protect\citeauthoryear{Zhao, Ying, Shi, Tong, Wen, and Zha}{Zhao
  et~al\mbox{.}}{2020}]%
        {Zhao_2020_CVPR}
\bibfield{author}{\bibinfo{person}{H. Zhao}, \bibinfo{person}{X. Ying},
  \bibinfo{person}{Y. Shi}, \bibinfo{person}{X. Tong}, \bibinfo{person}{J.
  Wen}, {and} \bibinfo{person}{H. Zha}.} \bibinfo{year}{2020}\natexlab{}.
\newblock \showarticletitle{RDCFace: Radial Distortion Correction for Face
  Recognition}. In \bibinfo{booktitle}{\emph{Proceedings of the IEEE/CVF
  Conference on Computer Vision and Pattern Recognition}}.
  \bibinfo{pages}{7721--7730}.
\newblock


\bibitem[\protect\citeauthoryear{Zhao, Cheng, Cheng, Yang, Lan, Zhao, Xiong,
  Xu, Li, Pranata, Shen, Xing, Liu, Yan, and Feng}{Zhao et~al\mbox{.}}{2019a}]%
        {Zhao2019LookAE}
\bibfield{author}{\bibinfo{person}{J. Zhao}, \bibinfo{person}{Y. Cheng},
  \bibinfo{person}{Y.~P. Cheng}, \bibinfo{person}{Y. Yang}, \bibinfo{person}{H.
  Lan}, \bibinfo{person}{F. Zhao}, \bibinfo{person}{L. Xiong},
  \bibinfo{person}{Y. Xu}, \bibinfo{person}{J. Li}, \bibinfo{person}{S.
  Pranata}, \bibinfo{person}{S. Shen}, \bibinfo{person}{J. Xing},
  \bibinfo{person}{H. Liu}, \bibinfo{person}{S. Yan}, {and} \bibinfo{person}{J.
  Feng}.} \bibinfo{year}{2019}\natexlab{a}.
\newblock \showarticletitle{Look Across Elapse: Disentangled Representation
  Learning and Photorealistic Cross-Age Face Synthesis for Age-Invariant Face
  Recognition}. In \bibinfo{booktitle}{\emph{Proceedings of the AAAI Conference
  on Artificial Intelligence}}, Vol.~\bibinfo{volume}{33}.
  \bibinfo{pages}{9251--9258}.
\newblock


\bibitem[\protect\citeauthoryear{{Zhao}, {Cheng}, {Xu}, {Xiong}, {Li}, {Zhao},
  {Jayashree}, {Pranata}, {Shen}, {Xing}, {Yan}, and {Feng}}{{Zhao}
  et~al\mbox{.}}{2018}]%
        {Zhao2018pim}
\bibfield{author}{\bibinfo{person}{J. {Zhao}}, \bibinfo{person}{Y. {Cheng}},
  \bibinfo{person}{Y. {Xu}}, \bibinfo{person}{L. {Xiong}}, \bibinfo{person}{J.
  {Li}}, \bibinfo{person}{F. {Zhao}}, \bibinfo{person}{K. {Jayashree}},
  \bibinfo{person}{S. {Pranata}}, \bibinfo{person}{S. {Shen}},
  \bibinfo{person}{J. {Xing}}, \bibinfo{person}{S. {Yan}}, {and}
  \bibinfo{person}{J. {Feng}}.} \bibinfo{year}{2018}\natexlab{}.
\newblock \showarticletitle{Towards Pose Invariant Face Recognition in the
  Wild}. In \bibinfo{booktitle}{\emph{Proceedings of the IEEE/CVF Conference on
  Computer Vision and Pattern Recognition}}. \bibinfo{pages}{2207--2216}.
\newblock


\bibitem[\protect\citeauthoryear{Zhao, Xiong, Karlekar, Li, Zhao, Wang,
  Pranata, Shen, Yan, and Feng}{Zhao et~al\mbox{.}}{2017}]%
        {Zhao2017DualAgentGF}
\bibfield{author}{\bibinfo{person}{J. Zhao}, \bibinfo{person}{L. Xiong},
  \bibinfo{person}{J. Karlekar}, \bibinfo{person}{J. Li}, \bibinfo{person}{F.
  Zhao}, \bibinfo{person}{Z. Wang}, \bibinfo{person}{S. Pranata},
  \bibinfo{person}{S. Shen}, \bibinfo{person}{S. Yan}, {and}
  \bibinfo{person}{J. Feng}.} \bibinfo{year}{2017}\natexlab{}.
\newblock \showarticletitle{Dual-Agent GANs for Photorealistic and Identity
  Preserving Profile Face Synthesis}. In \bibinfo{booktitle}{\emph{Advances in
  neural information processing systems}}. \bibinfo{pages}{65–75}.
\newblock


\bibitem[\protect\citeauthoryear{Zhao, Xu, and Cheng}{Zhao
  et~al\mbox{.}}{2019b}]%
        {zhao2019regularface}
\bibfield{author}{\bibinfo{person}{K. Zhao}, \bibinfo{person}{K. Xu}, {and}
  \bibinfo{person}{M. Cheng}.} \bibinfo{year}{2019}\natexlab{b}.
\newblock \showarticletitle{RegularFace: Deep Face Recognition via Exclusive
  Regularization}. In \bibinfo{booktitle}{\emph{Proceedings of the IEEE/CVF
  Conference on Computer Vision and Pattern Recognition}}.
  \bibinfo{pages}{1136--1144}.
\newblock


\bibitem[\protect\citeauthoryear{Zheng and Deng}{Zheng and Deng}{2018}]%
        {zheng2018cross}
\bibfield{author}{\bibinfo{person}{T. Zheng} {and} \bibinfo{person}{W. Deng}.}
  \bibinfo{year}{2018}\natexlab{}.
\newblock \showarticletitle{Cross-pose lfw: A database for studying crosspose
  face recognition in unconstrained environments}.
\newblock \bibinfo{journal}{\emph{Beijing University of Posts and
  Telecommunications, Tech. Rep}} (\bibinfo{year}{2018}),
  \bibinfo{pages}{18--01}.
\newblock


\bibitem[\protect\citeauthoryear{Zheng, Deng, and Hu}{Zheng
  et~al\mbox{.}}{2017}]%
        {zheng2017cross}
\bibfield{author}{\bibinfo{person}{T. Zheng}, \bibinfo{person}{W. Deng}, {and}
  \bibinfo{person}{J. Hu}.} \bibinfo{year}{2017}\natexlab{}.
\newblock \showarticletitle{Cross-age lfw: A database for studying cross-age
  face recognition in unconstrained environments}.
\newblock  (\bibinfo{year}{2017}).
\newblock
\showeprint{1708.08197}


\bibitem[\protect\citeauthoryear{Zheng, Pal, and Savvides}{Zheng
  et~al\mbox{.}}{2018}]%
        {zheng2018ring}
\bibfield{author}{\bibinfo{person}{Y. Zheng}, \bibinfo{person}{D.~K. Pal},
  {and} \bibinfo{person}{M. Savvides}.} \bibinfo{year}{2018}\natexlab{}.
\newblock \showarticletitle{Ring loss: Convex feature normalization for face
  recognition}. In \bibinfo{booktitle}{\emph{Proceedings of the IEEE/CVF
  Conference on Computer Vision and Pattern Recognition}}.
  \bibinfo{pages}{5089--5097}.
\newblock


\bibitem[\protect\citeauthoryear{{Zhong}, {Chen}, and {Huang}}{{Zhong}
  et~al\mbox{.}}{2017}]%
        {Zhong2017e2e}
\bibfield{author}{\bibinfo{person}{Y. {Zhong}}, \bibinfo{person}{J. {Chen}},
  {and} \bibinfo{person}{B. {Huang}}.} \bibinfo{year}{2017}\natexlab{}.
\newblock \showarticletitle{Toward End-to-End Face Recognition Through
  Alignment Learning}.
\newblock \bibinfo{journal}{\emph{IEEE Singal processing letters}}
  \bibinfo{volume}{24}, \bibinfo{number}{8} (\bibinfo{year}{2017}),
  \bibinfo{pages}{1213--1217}.
\newblock


\bibitem[\protect\citeauthoryear{Zhong and Deng}{Zhong and Deng}{2018}]%
        {zhong2018deep}
\bibfield{author}{\bibinfo{person}{Yaoyao Zhong} {and} \bibinfo{person}{Weihong
  Deng}.} \bibinfo{year}{2018}\natexlab{}.
\newblock \showarticletitle{Deep Difference Analysis in Similar-looking Face
  recognition}. In \bibinfo{booktitle}{\emph{Proceedings of the International
  Conference on Pattern Recognition (ICPR)}}. IEEE,
  \bibinfo{pages}{3353--3358}.
\newblock


\bibitem[\protect\citeauthoryear{Zhong, Deng, Wang, Hu, Peng, Tao, and
  Huang}{Zhong et~al\mbox{.}}{2019}]%
        {Zhong_2019_CVPR}
\bibfield{author}{\bibinfo{person}{Y. Zhong}, \bibinfo{person}{W. Deng},
  \bibinfo{person}{M. Wang}, \bibinfo{person}{J. Hu}, \bibinfo{person}{J.
  Peng}, \bibinfo{person}{X. Tao}, {and} \bibinfo{person}{Y. Huang}.}
  \bibinfo{year}{2019}\natexlab{}.
\newblock \showarticletitle{Unequal-Training for Deep Face Recognition With
  Long-Tailed Noisy Data}. In \bibinfo{booktitle}{\emph{Proceedings of the
  IEEE/CVF Conference on Computer Vision and Pattern Recognition}}.
  \bibinfo{pages}{7804--7813}.
\newblock


\bibitem[\protect\citeauthoryear{Zhou, Cao, and Sun}{Zhou
  et~al\mbox{.}}{2018}]%
        {Zhou2018GridFaceFR}
\bibfield{author}{\bibinfo{person}{E. Zhou}, \bibinfo{person}{Z. Cao}, {and}
  \bibinfo{person}{J. Sun}.} \bibinfo{year}{2018}\natexlab{}.
\newblock \showarticletitle{Gridface: Face rectification via learning local
  homography transformations}. In \bibinfo{booktitle}{\emph{Proceedings of the
  European Conference on Computer Vision}}. \bibinfo{pages}{3--19}.
\newblock


\bibitem[\protect\citeauthoryear{Zhou, Fan, Cao, Jiang, and Yin}{Zhou
  et~al\mbox{.}}{2013}]%
        {Zhou2013ExtensiveFL}
\bibfield{author}{\bibinfo{person}{E. Zhou}, \bibinfo{person}{H. Fan},
  \bibinfo{person}{Z. Cao}, \bibinfo{person}{Y. Jiang}, {and}
  \bibinfo{person}{Q. Yin}.} \bibinfo{year}{2013}\natexlab{}.
\newblock \showarticletitle{Extensive Facial Landmark Localization with
  Coarse-to-Fine Convolutional Network Cascade}. In
  \bibinfo{booktitle}{\emph{Proceedings of the IEEE International Conference on
  Computer Vision Workshops}}. \bibinfo{pages}{386--391}.
\newblock


\bibitem[\protect\citeauthoryear{Zhou and Comaniciu}{Zhou and
  Comaniciu}{2007}]%
        {Zhou2007ShapeRM}
\bibfield{author}{\bibinfo{person}{S.~K. Zhou} {and} \bibinfo{person}{D.
  Comaniciu}.} \bibinfo{year}{2007}\natexlab{}.
\newblock \showarticletitle{Shape regression machine}. In
  \bibinfo{booktitle}{\emph{Biennial International Conference on Information
  Processing in Medical Imaging}}. \bibinfo{pages}{13--25}.
\newblock


\bibitem[\protect\citeauthoryear{{Zhu}, {He}, and {Savvides}}{{Zhu}
  et~al\mbox{.}}{2019}]%
        {zhu2019fs}
\bibfield{author}{\bibinfo{person}{C. {Zhu}}, \bibinfo{person}{Y. {He}}, {and}
  \bibinfo{person}{M. {Savvides}}.} \bibinfo{year}{2019}\natexlab{}.
\newblock \showarticletitle{Feature Selective Anchor-Free Module for
  Single-Shot Object Detection}. In \bibinfo{booktitle}{\emph{Proceedings of
  the IEEE/CVF Conference on Computer Vision and Pattern Recognition}}.
  \bibinfo{pages}{840--849}.
\newblock


\bibitem[\protect\citeauthoryear{{Zhu}, {Tao}, {Luu}, and {Savvides}}{{Zhu}
  et~al\mbox{.}}{2018}]%
        {Zhu2018}
\bibfield{author}{\bibinfo{person}{C. {Zhu}}, \bibinfo{person}{R. {Tao}},
  \bibinfo{person}{K. {Luu}}, {and} \bibinfo{person}{M. {Savvides}}.}
  \bibinfo{year}{2018}\natexlab{}.
\newblock \showarticletitle{Seeing small faces from robust anchor's
  perspective}. In \bibinfo{booktitle}{\emph{Proceedings of the IEEE/CVF
  Conference on Computer Vision and Pattern Recognition}}.
  \bibinfo{pages}{5127--5136}.
\newblock


\bibitem[\protect\citeauthoryear{Zhu, Zheng, Luu, and Savvides}{Zhu
  et~al\mbox{.}}{2017b}]%
        {Zhu2017CMS-RCNN}
\bibfield{author}{\bibinfo{person}{C. Zhu}, \bibinfo{person}{Y. Zheng},
  \bibinfo{person}{K. Luu}, {and} \bibinfo{person}{M. Savvides}.}
  \bibinfo{year}{2017}\natexlab{b}.
\newblock \showarticletitle{Cms-rcnn: contextual multi-scale region-based cnn
  for unconstrained face detection}.
\newblock In \bibinfo{booktitle}{\emph{Deep learning for biometrics}}.
  \bibinfo{pages}{57--79}.
\newblock


\bibitem[\protect\citeauthoryear{Zhu, Park, Isola, and Efros}{Zhu
  et~al\mbox{.}}{2017a}]%
        {Zhu2017UnpairedIT}
\bibfield{author}{\bibinfo{person}{J. Zhu}, \bibinfo{person}{T. Park},
  \bibinfo{person}{P. Isola}, {and} \bibinfo{person}{A.~A. Efros}.}
  \bibinfo{year}{2017}\natexlab{a}.
\newblock \showarticletitle{Unpaired Image-to-Image Translation Using
  Cycle-Consistent Adversarial Networks}. In
  \bibinfo{booktitle}{\emph{Proceedings of the IEEE International Conference on
  Computer Vision}}. \bibinfo{pages}{2242--2251}.
\newblock


\bibitem[\protect\citeauthoryear{Zhu, Shi, Zheng, and Sadiq}{Zhu
  et~al\mbox{.}}{2019b}]%
        {Zhu_2019_CVPR}
\bibfield{author}{\bibinfo{person}{M. Zhu}, \bibinfo{person}{F. Shi},
  \bibinfo{person}{M. Zheng}, {and} \bibinfo{person}{M. Sadiq}.}
  \bibinfo{year}{2019}\natexlab{b}.
\newblock \showarticletitle{Robust facial landmark detection via
  occlusion-adaptive deep networks}. In \bibinfo{booktitle}{\emph{Proceedings
  of the IEEE/CVF Conference on Computer Vision and Pattern Recognition}}.
  \bibinfo{pages}{3486--3496}.
\newblock


\bibitem[\protect\citeauthoryear{{Zhu}, {Lei}, {Liu}, {Shi}, and {Li}}{{Zhu}
  et~al\mbox{.}}{2016}]%
        {Zhu2016_3DDFA}
\bibfield{author}{\bibinfo{person}{X. {Zhu}}, \bibinfo{person}{Z. {Lei}},
  \bibinfo{person}{X. {Liu}}, \bibinfo{person}{H. {Shi}}, {and}
  \bibinfo{person}{S.~Z. {Li}}.} \bibinfo{year}{2016}\natexlab{}.
\newblock \showarticletitle{Face Alignment Across Large Poses: A 3D Solution}.
  In \bibinfo{booktitle}{\emph{Proceedings of the IEEE/CVF Conference on
  Computer Vision and Pattern Recognition}}. \bibinfo{pages}{146--155}.
\newblock


\bibitem[\protect\citeauthoryear{Zhu, Liu, Lei, Shi, Yang, Yi, Qi, and Li}{Zhu
  et~al\mbox{.}}{2019a}]%
        {zhu2019large}
\bibfield{author}{\bibinfo{person}{X. Zhu}, \bibinfo{person}{H. Liu},
  \bibinfo{person}{Z. Lei}, \bibinfo{person}{H. Shi}, \bibinfo{person}{F.
  Yang}, \bibinfo{person}{D. Yi}, \bibinfo{person}{G. Qi}, {and}
  \bibinfo{person}{S.~Z. Li}.} \bibinfo{year}{2019}\natexlab{a}.
\newblock \showarticletitle{Large-scale bisample learning on id versus spot
  face recognition}.
\newblock \bibinfo{journal}{\emph{International Journal of Computer Vision}}
  \bibinfo{volume}{127}, \bibinfo{number}{6-7} (\bibinfo{year}{2019}),
  \bibinfo{pages}{684--700}.
\newblock


\bibitem[\protect\citeauthoryear{{Zhu} and {Ramanan}}{{Zhu} and
  {Ramanan}}{2012}]%
        {Zhu2012Face}
\bibfield{author}{\bibinfo{person}{X. {Zhu}} {and} \bibinfo{person}{D.
  {Ramanan}}.} \bibinfo{year}{2012}\natexlab{}.
\newblock \showarticletitle{Face detection, pose estimation, and landmark
  localization in the wild}. In \bibinfo{booktitle}{\emph{Proceedings of the
  IEEE/CVF Conference on Computer Vision and Pattern Recognition}}.
  \bibinfo{pages}{2879--2886}.
\newblock


\bibitem[\protect\citeauthoryear{Zhu, Huang, Deng, Ye, Huang, Chen, Zhu, Yang,
  Lu, Du, and Zhou}{Zhu et~al\mbox{.}}{2021}]%
        {Zhu2021WebFace260MAB}
\bibfield{author}{\bibinfo{person}{Zheng Zhu}, \bibinfo{person}{Guan Huang},
  \bibinfo{person}{Jiankang Deng}, \bibinfo{person}{Yun Ye},
  \bibinfo{person}{Junjie Huang}, \bibinfo{person}{Xinze Chen},
  \bibinfo{person}{Jiagang Zhu}, \bibinfo{person}{Tian Yang},
  \bibinfo{person}{Jiwen Lu}, \bibinfo{person}{Dalong Du}, {and}
  \bibinfo{person}{Jie Zhou}.} \bibinfo{year}{2021}\natexlab{}.
\newblock \showarticletitle{WebFace260M: A Benchmark Unveiling the Power of
  Million-Scale Deep Face Recognition}.
\newblock \bibinfo{journal}{\emph{Proceedings of the IEEE/CVF Conference on
  Computer Vision and Pattern Recognition}} (\bibinfo{year}{2021}),
  \bibinfo{pages}{10487--10497}.
\newblock


\bibitem[\protect\citeauthoryear{{Zhuang}, {Zhang}, {Zhu}, {Lei}, {Wang}, and
  {Li}}{{Zhuang} et~al\mbox{.}}{2019}]%
        {2019fldet}
\bibfield{author}{\bibinfo{person}{C. {Zhuang}}, \bibinfo{person}{S. {Zhang}},
  \bibinfo{person}{X. {Zhu}}, \bibinfo{person}{Z. {Lei}}, \bibinfo{person}{J.
  {Wang}}, {and} \bibinfo{person}{S.~Z. {Li}}.}
  \bibinfo{year}{2019}\natexlab{}.
\newblock \showarticletitle{FLDet: A CPU Real-time Joint Face and Landmark
  Detector}. In \bibinfo{booktitle}{\emph{Proceedings of the International
  Conference on Biometrics}}. \bibinfo{pages}{1--8}.
\newblock


\bibitem[\protect\citeauthoryear{Zou, Kittler, and Messer}{Zou
  et~al\mbox{.}}{2007}]%
        {Zou2007IlluminationIF}
\bibfield{author}{\bibinfo{person}{X. Zou}, \bibinfo{person}{J. Kittler}, {and}
  \bibinfo{person}{K. Messer}.} \bibinfo{year}{2007}\natexlab{}.
\newblock \showarticletitle{Illumination Invariant Face Recognition: A Survey}.
  In \bibinfo{booktitle}{\emph{Proceedings of the IEEE International Conference
  on Biometrics: Theory, Applications, and Systems}}. \bibinfo{pages}{1--8}.
\newblock


\bibitem[\protect\citeauthoryear{Zou, Zhong, Yan, Zhao, Zhou, and Wu}{Zou
  et~al\mbox{.}}{2019}]%
        {Zou2019LearningRF}
\bibfield{author}{\bibinfo{person}{X. Zou}, \bibinfo{person}{S. Zhong},
  \bibinfo{person}{L. Yan}, \bibinfo{person}{X. Zhao}, \bibinfo{person}{J.
  Zhou}, {and} \bibinfo{person}{Y. Wu}.} \bibinfo{year}{2019}\natexlab{}.
\newblock \showarticletitle{Learning Robust Facial Landmark Detection via
  Hierarchical Structured Ensemble}. In \bibinfo{booktitle}{\emph{Proceedings
  of the IEEE International Conference on Computer Vision}}.
  \bibinfo{pages}{141--150}.
\newblock


\bibitem[\protect\citeauthoryear{Çeliktutan, S.Ulukaya, and
  Sankur}{Çeliktutan et~al\mbox{.}}{2013}]%
        {eliktutan2013ACS}
\bibfield{author}{\bibinfo{person}{O. Çeliktutan},
  \bibinfo{person}{S.Ulukaya}, {and} \bibinfo{person}{B. Sankur}.}
  \bibinfo{year}{2013}\natexlab{}.
\newblock \showarticletitle{A comparative study of face landmarking
  techniques}.
\newblock \bibinfo{journal}{\emph{EURASIP Journal on Image and Video
  Processing}}  \bibinfo{volume}{2013} (\bibinfo{year}{2013}),
  \bibinfo{pages}{1--27}.
\newblock


\end{thebibliography}

\newpage
\appendix

\section{Representative surveys of face recognition}

A number of face recognition surveys have been published in the past twenty years. We summarize them in Table~\ref{fr_surveys}.

\begin{table*}[!h]
\begin{center}
\caption{Representative surveys of face recognition}
\label{fr_surveys}
\resizebox{\linewidth}{!}{
\begin{tabular}{|p{10cm}|c|p{8cm}|}
\hline
{Title}&{Year}&{Description}\\
\hline\hline
Face Recognition: A Literature Survey~\cite{W2003Face}&2003& Traditional image- and video-based methods in face recognition. Not covering deep face recognition.\\
\hline
Face Recognition from a Single Image per Person: A Survey~\cite{2006fr}&2006& The methods to address the single sample problem in face recognition, not covering deep face recognition. \\
\hline
A Survey of Approaches and Challenges in 3D and Multi-modal 3D+2D Face Recognition~\cite{bowyer2006survey}&2006& A survey of 3D and multi-modal face recognition, not covering deep face recognition. \\
\hline
Illumination Invariant Face Recognition: A Survey~\cite{Zou2007IlluminationIF}&2007& Focus on illumination-invariant face recognition task, not covering deep face recognition. \\

\hline
A Survey of Face Recognition Techniques~\cite{2009fr}&2009& Traditional face recognition methods on different modal face data, not covering deep face recognition. \\

\hline
A Comprehensive Survey on Pose-Invariant Face Recognition~\cite{Ding2016ACS}&2016& Focus on pose-invariant face recognition task. 
\\

\hline
A Survey of Local Feature Methods for 3D Face Recognition~\cite{Soltanpour2017ASO}&2017& A review of feature extraction based methods for 3D face recognition. \\
\hline
Deep Learning for Understanding Faces~\cite{Ranjan2018deep}&2018& Provide a brief overview of the end-to-end deep face recognition, not covering the recent works. \\
\hline
Deep Face Recognition: A Survey~\cite{wang2018deep}&2018& Focus on the deep face representation learning.\\
\hline
Past, Present, and Future of Face Recognition: A Review
~\cite{electronics9081188}&2020& A review of 2D and 3D face recognition, not covering end-to-end deep face recognition.   \\
\hline
\end{tabular}}
\end{center}
\end{table*}

\section{Face detection}

\subsection{Single-stage and multi-stage face detectors} 
\begin{figure}[ht]
\centering
\includegraphics[scale=0.25]{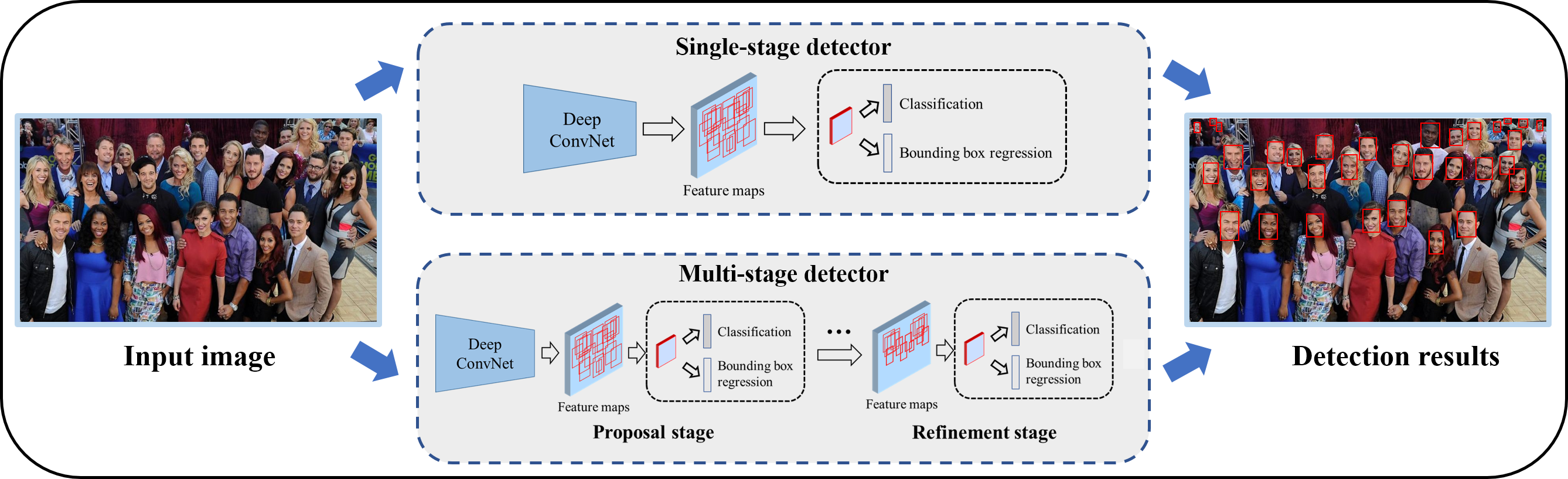}
\caption{The illustration of single-stage and multi-stage face detectors. The single-stage detector accomplishes the face detection directly from the feature maps, whereas the multi-stage detector adopts a proposal stage to generate candidates and one or more stages to refine these candidates.}
\label{face_detectors}
\end{figure}

Fig.~\ref{face_detectors} illustrates the difference between single-stage and multi-stage face detectors. For comparison, the single-stage face detector accomplishes the detection processing directly from the feature maps, whereas the multi-stage face detector adopts a proposal stage to generate candidates and one or more stages to refine these candidates.

\subsection{Performance comparison of CPU real-time face detection methods} 

Table~\ref{cpu_detection} shows the running efficiency of CPU real-time face detection methods, among which the lightweight backbone~\cite{xu2019centerface,deng2020retinaface}, rapidly digested convolutional layer~\cite{Zhang2017Faceboxes,Zhang2018DCFPN}, knowledge distillation~\cite{2019fbi} and region-of-interest (RoI) convolution~\cite{2016stn} are the common practices.

\begin{table}[ht]
\begin{center}
\caption{Running efficiency of CPU real-time face detectors. ``Accuracy (\%)'' denotes the true positive rate at 1000 false positives on FDDB. }
\label{cpu_detection}
\resizebox{0.7\linewidth}{!}{
\begin{tabular}{|c|c|c|c|c|}
\hline
{Method}&{Publication}&{CPU-model}&{Speed (FPS)}&{Accuracy ($\%$)}\\
\hline\hline
Faceboxes~\cite{Zhang2017Faceboxes} &{IJCB’17}&E5-2660v3@2.60GHz&20&96.0\\
\hline
STN~\cite{2016stn}&{ECCV’16} &i7-4770K&30&-\\
\hline
DCFPN~\cite{Zhang2018DCFPN}  &{Neurocomputing'18}&2.60GHz&30&-\\
\hline
FBI~\cite{2019fbi} &{ICB'19}&E5-2660v3@2.60GHz&20&96.8\\
\hline
PCN~\cite{pcn}&{CVPR'18}&3.40GHz&29&-\\
\hline
PPN~\cite{ZENG2019PPN} &{Information Sciences'19}&i5&60&-\\
\hline
RetinaFace~\cite{deng2020retinaface}&{CVPR'19}&i7-6700K&60&-\\
\hline
CenterFace~\cite{xu2019centerface}&{arXiv'19}&i7-6700@2.60GHz&30&98.0\\
\hline
\end{tabular}}
\end{center}
\end{table}

\section{Face alignment}

\subsection{Hourglass network for facial landmark localization}
\begin{figure}[ht]
\centering
\includegraphics[height=3.5cm]{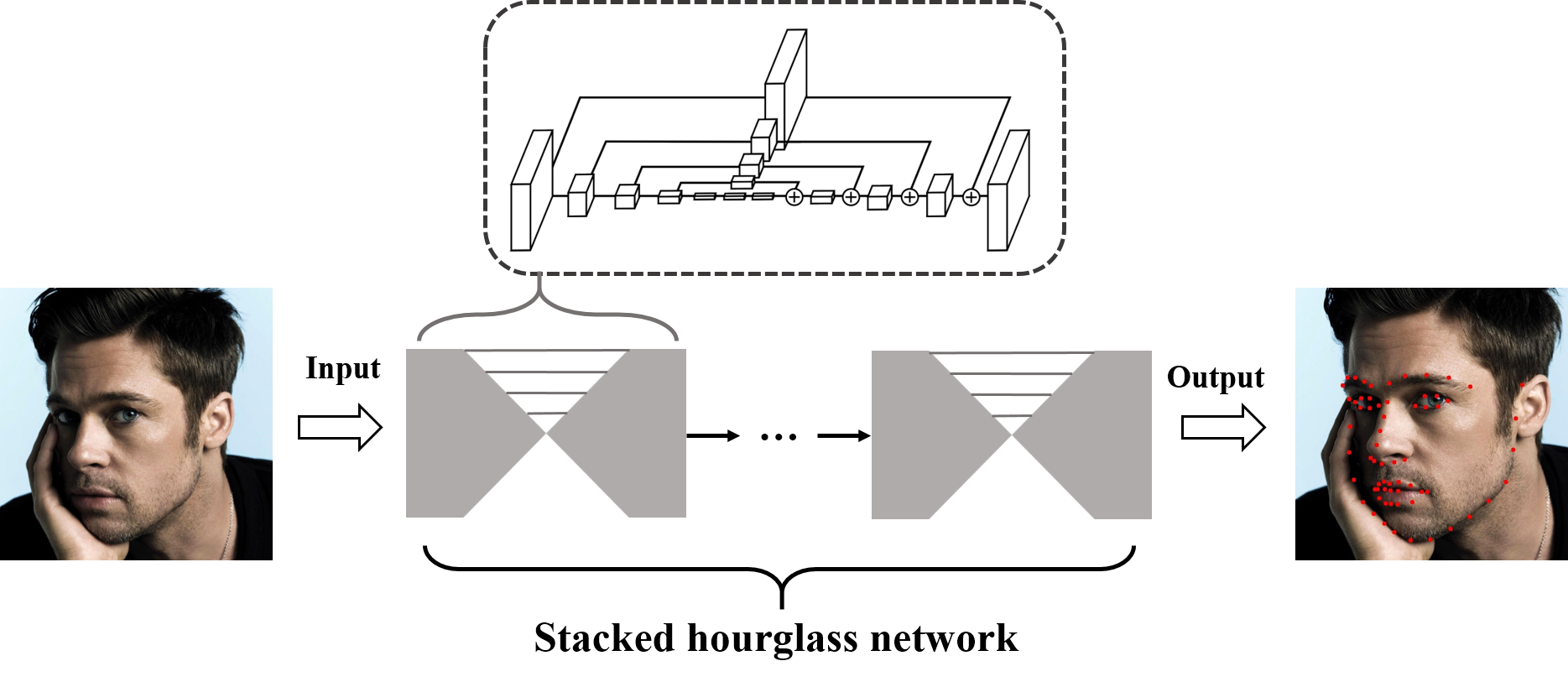}
\caption{An illustration of stacked hourglass network~\cite{Newell2016StackedHN} for facial landmark localization. 
In each hourglass structure, the width (\textit{i.e.,} feature channels) is consistent, and the boxes represent the residual modules.}
\label{hourglass}
\end{figure}

Hourglass~\cite{Newell2016StackedHN} is a bottom-up and top-down architecture, playing an important role in the deep stack of bottleneck blocks along with intermediate supervision. Fig.~\ref{hourglass} is an illustration of stacked hourglass network.

\subsection{3D model fitting for facial landmark localization}
\begin{figure}[t]
\centering
\includegraphics[height=2.2cm]{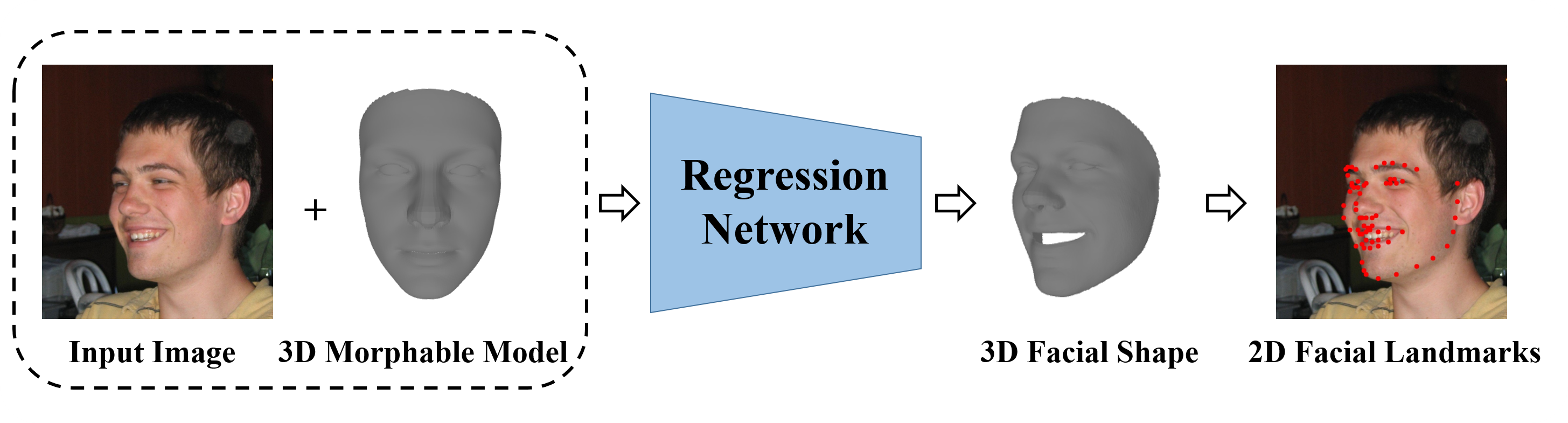}
\caption{The process of 3D model fitting for face alignment. A dense 3D Morphable Model is used to model a 2D face to 3D mesh. The regression network estimates the parameters of 3D shape and projection matrix, and then the 3D shape is projected onto the image plane to obtain the 2D landmarks. }
\label{3D_model_fitting}
\end{figure}

As illustrated in Fig.~\ref{3D_model_fitting}, some 3D model fitting based methods employ cascaded regression manner with a dense 3D Morphable Model (3DMM)~\cite{BlanzVolker2003FaceRB} to estimate the 3D face shape.

\subsection{Landmark-free face alignment}
\begin{figure}[ht]
\centering
\includegraphics[height=2cm]{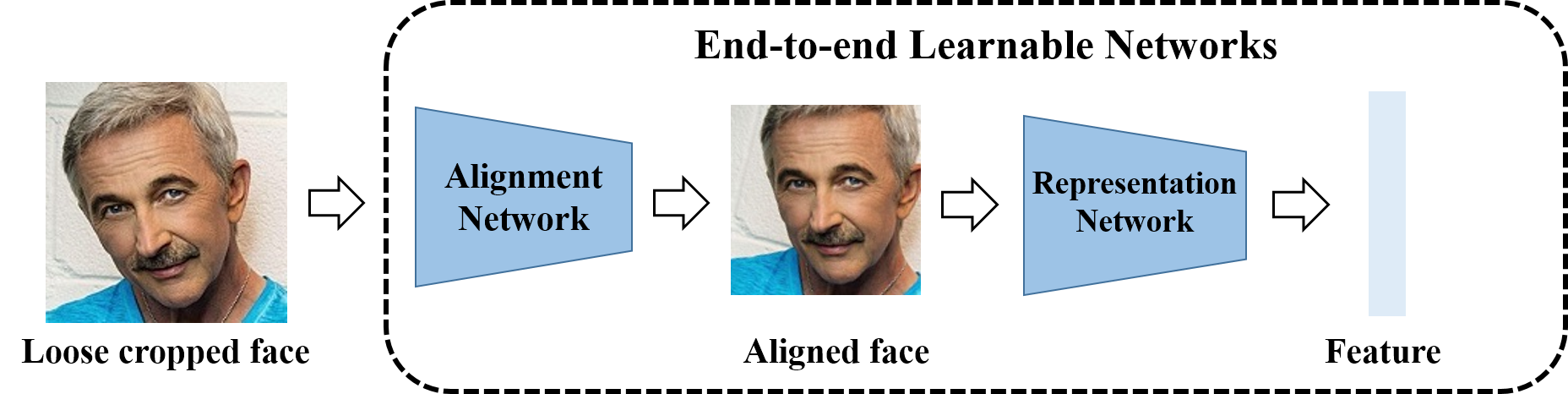}
\caption{
An illustration of integrated framework that accomplishes landmark-free face alignment and representation computation.
}
\label{e2e}
\end{figure}

Landmark-free face alignment methods integrate the alignment transformation processing into DCNNs and output aligned face without relying on facial landmarks (Fig.~\ref{e2e}). 

\section{Face representation}

\begin{figure}[ht]
\centering
\includegraphics[height=5cm]{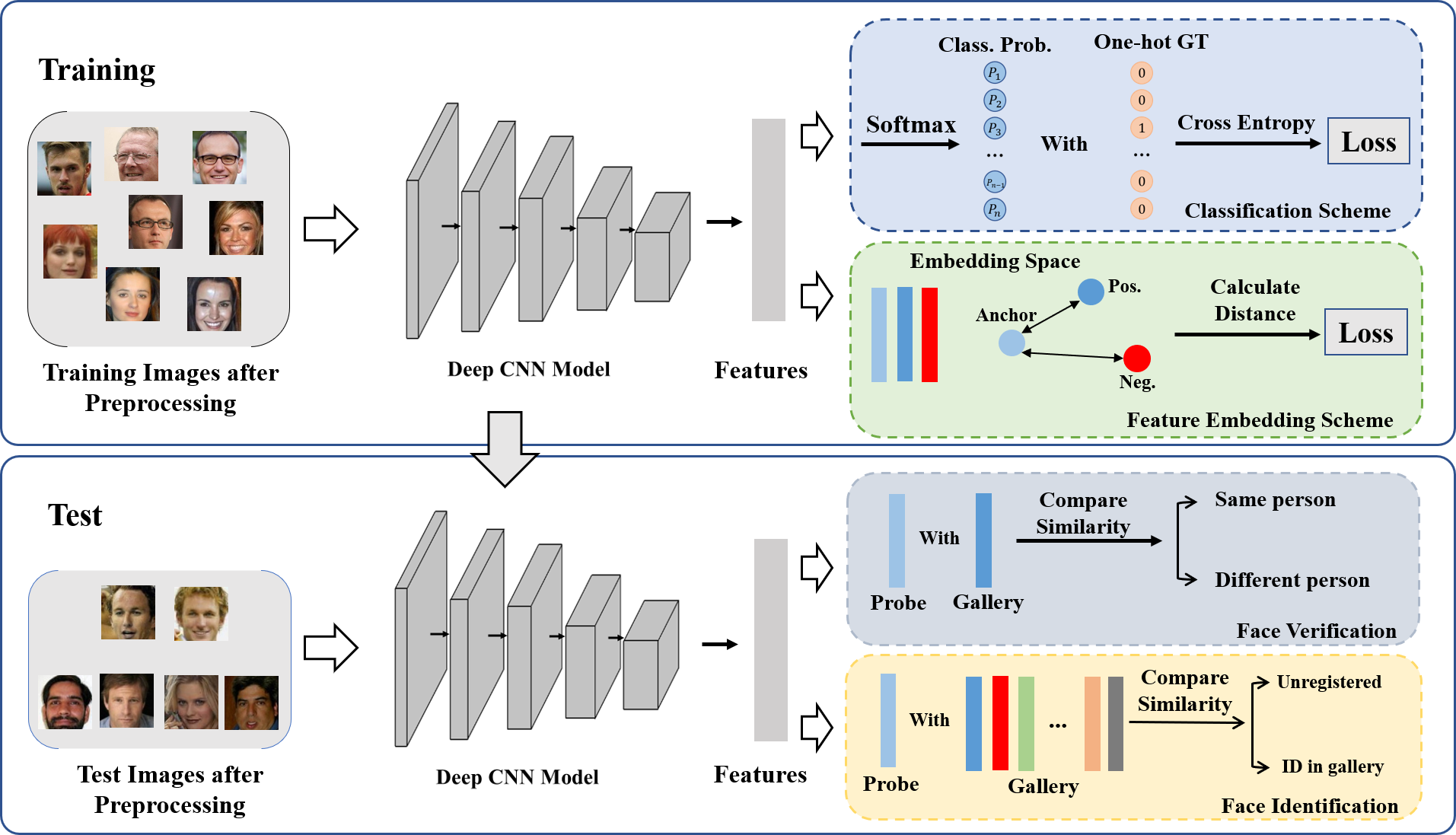}
\caption{The pipeline of face representation training phase and test phase. In the training phase, two schemes, \textit{i.e.,} classification and feature embedding, are often used for learning face representation. In the test phase, face verification and face identification are the major tasks.}
\label{fig:face_representaion}
\end{figure}

Fig.~\ref{fig:face_representaion} shows the pipeline of face representation training phase and test phase. In the training phase, two types of training supervision are widely used, \textit{i.e.,} classification and feature embedding. 
As for test phase, there are two major tasks, \textit{i.e.,} face verification or face identification.

\begin{figure}[ht]
\centering
\includegraphics[height=4cm]{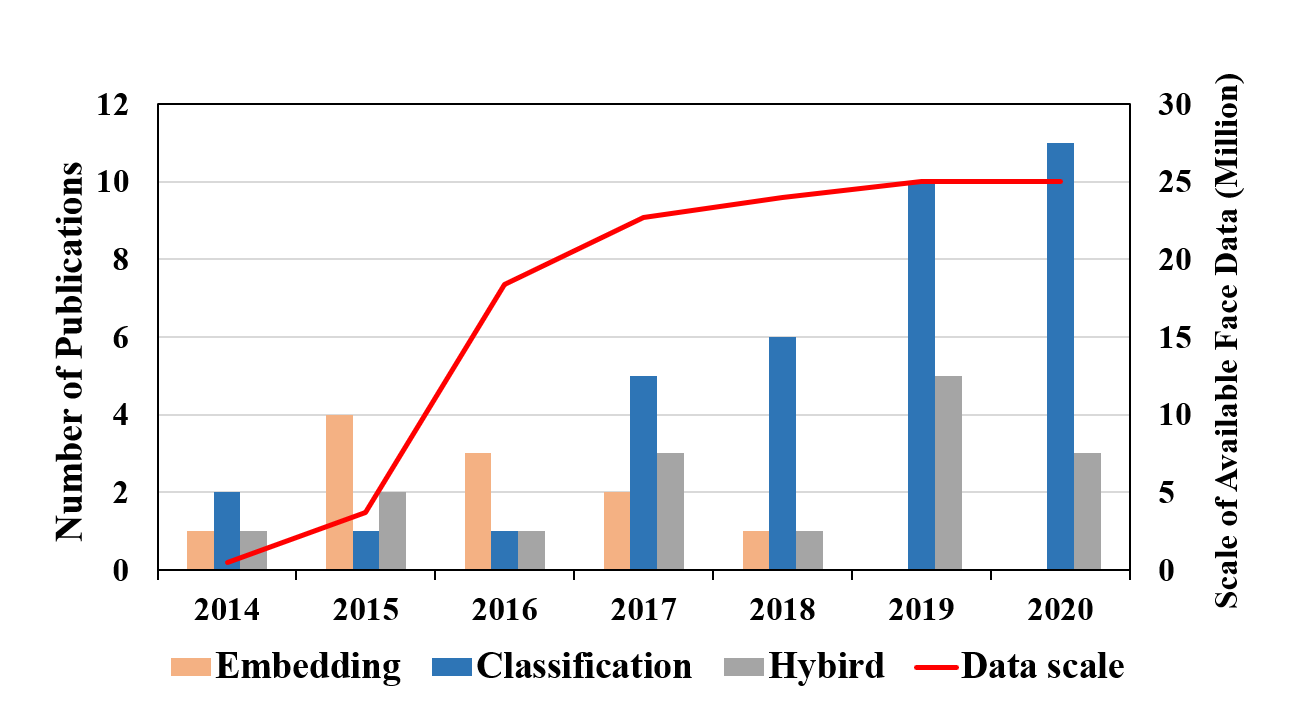}
\caption{The publication trend of three supervised face representation learning schemes with the growing scale of available face datasets from 2014 - 2020.  }
\label{fig:face_representaion_data_scale}
\end{figure}

In addition, as shown Fig.~\ref{fig:face_representaion_data_scale}, we can observe that the publications of classification based training supervision exceed those of the feature embedding and hybrid methods with the growing scale of available face data. The reason is that the closed-set classification training on the large-scale datasets enables to approach open-set face recognition scenario.

\clearpage

\end{document}